%% file: main.tex
\documentclass[a4paper,12pt,openany,oneside]{memoir}
\usepackage{graphicx} % Required for inserting images
\usepackage[backend=biber,bibencoding=ascii,texencoding=ascii]{biblatex}
\usepackage{tikz}
\usetikzlibrary{shapes.arrows, fadings, arrows.meta}
\usepackage{adjustbox}
\usepackage[Bjornstrup]{fncychap}
\usepackage{xurl}
\usepackage[hidelinks,colorlinks=true,linkcolor=black,citecolor=blue,hypertexnames=false]{hyperref}
\usepackage{pythonhighlight}
\usepackage{subcaption}
\usepackage{arydshln}
\usepackage{pdfpages}
\usepackage{mwe}
\usepackage{amsmath}
\usepackage{listings}
\usepackage{amsfonts} 
\usepackage{booktabs} % For prettier tables
\usepackage{diagbox} % Include this in your document's preamble
\usepackage{multirow}
\usepackage{listings}
\usepackage{xcolor}

\definecolor{pink}{rgb}{1.0, 0.0, 0.5}
\definecolor{darkblue}{rgb}{0.0, 0.0, 0.55}
\definecolor{blue}{rgb}{0.0, 0.0, 1.0}
\lstdefinestyle{pythonstyle}{
    language=Python,
    basicstyle=\ttfamily\footnotesize,
    keywordstyle=\color{blue},
    commentstyle=\color{green},
    stringstyle=\color{red},
    breakatwhitespace=false,
    breaklines=true,
    captionpos=b,
    keepspaces=true,
    numbers=left,
    numbersep=5pt,
    showspaces=false,
    showstringspaces=false,
    showtabs=false,
    tabsize=2,
    frame=single,
    frameround=ffff,
    framexleftmargin=8mm,
    rulecolor=\color{black},
    backgroundcolor=\color{white},
    morekeywords={True, False, self}, % Extend with actual Python keywords or built-ins
    keywordstyle=\color{darkblue}, % Style for extended keywords
}

\newenvironment{acknowledgements}
	{\chapter*{Acknowledgements}}												%kai
	{\addcontentsline{toc}{chapter}{Acknowledgements}}	%kai
\input{eth-template/extrapackages}
\input{eth-template/layoutsetup}

\input{eth-template/theoremsetup}

\input{eth-template/macrosetup}

\nonzeroparskip
\defaultlists

\title{Solving the Elastic Wave Equation with Physics-Informed Neural Networks: A Robust and Critical Assessment}
\author{Davide Staub\and Ben Moseley}
\thesistype{ArXiv Version}
\advisors{Based on the Master Thesis by Davide Staub (March 2024)\\
Original thesis advisor: Prof.\ Dr.\ Siddhartha Mishra\\
Original thesis co-supervisor: Dr.\ Ben Moseley\\
\href{https://doi.org/10.3929/ethz-b-000668359}{doi:10.3929/ethz-b-000668359}}
\department{Department of Mathematics}
\date{September 2026}
\hypersetup{
  pdftitle={Solving the Elastic Wave Equation with Physics-Informed Neural Networks: A Robust and Critical Assessment},
  pdfauthor={Davide Staub and Ben Moseley},
  pdfsubject={Physics-Informed Neural Networks for the elastic wave equation},
  pdfkeywords={physics-informed neural networks, elastic wave equation, scientific machine learning, seismology}
}
\begin{document}
\begin{titlingpage}
  \calccentering{\unitlength}
  \begin{adjustwidth*}{\unitlength-24pt}{-\unitlength-24pt}
    \maketitle
\end{adjustwidth*}
\end{titlingpage}
\newpage
\chapter*{Note on this arXiv version}
This document substantially reproduces the Master Thesis of Davide Staub,
submitted to the Department of Mathematics at ETH Z\"urich on 5 March 2024 and
available at
\href{https://doi.org/10.3929/ethz-b-000668359}{doi:10.3929/ethz-b-000668359}.
The original thesis was supervised by Prof.\ Dr.\ Siddhartha Mishra and
co-supervised by Dr.\ Ben Moseley. The author list for this arXiv version is
Davide Staub and Ben Moseley. Apart from the updated front matter and minor
typographical corrections, the scientific content is unchanged from the
deposited thesis.

\newpage
\begin{abstract}
Physics-Informed Neural Networks (PINNs) have recently emerged as a promising approach for solving Partial Differential Equations (PDEs), offering a meshfree alternative that integrates physical principles into the learning process. This presents a new paradigm compared to traditional discretization methods and purely data-driven machine learning techniques. While promising, PINNs are not a panacea; they inherit challenges such as spectral bias and unstable convergence. Moreover, their potential in seismology remains largely unexplored. In this work, we provide a robust and critical assessment of PINNs for solving the elastic wave equation in seismology. We investigate the performance of PINNs on problems with varying degrees of complexity across various seismic sources and parameter models, from constant to highly heterogeneous settings. A pivotal aspect of our work involves investigating whether embedding physical principles directly into the network architecture enhances convergence and accuracy. We test an extensive range of neural architecture designs, from unrestricted, uninformed PINNs to highly specialized ones. We find that integrating an understanding of wave physics into the network design significantly improves accuracy. For instance, introducing a custom wavelet or plane wave layer, coupled with encoder and decoder layers, consistently yields a relative $L_2$ error approximately half that of the standard PINN, as evidenced across numerous experiments. We further demonstrate that this novel architecture enhances accuracy when applied to the acoustic wave equation, underlying the versatility of our network. Another key contribution of our research is the successful conditioning of PINNs on seismic source locations. For seismology applications, this advancement is crucial as it traditionally requires running thousands of seismic simulations to assess various potential source locations. By conditioning PINNs on the source location, the network learns solutions for a class of problems, not just a single setting. Once trained, this method significantly outperforms traditional numerical methods, such as Finite Difference methods, in terms of simulation speed. The capability of conditioned PINNs to infer the wavefield for countless source locations in a single forward pass signifies a considerable advancement towards rapid seismic hazard detection and seismic analysis.

\end{abstract}
\newpage
\begin{acknowledgements}
\emph{The following acknowledgements are reproduced from the original thesis.}

First and foremost, I extend my sincerest gratitude to my supervisor, Prof. Dr. Siddhartha Mishra, for allowing me to pursue a thesis topic that truly captivated me. His openness to my passion, even when it diverged from his initial suggestion, has been incredibly inspiring. I am profoundly thankful for his invaluable insights and keen identification of the shortcomings in my approach, which significantly enhanced my work and led to more robust outcomes.

Secondly, I express my deep appreciation to my co-supervisor, Dr. Ben Moseley, for his immense support, insightful contributions, and engaging discussions throughout this journey. His pivotal role in sparking my interest in scientific machine learning through previous collaborative projects has indubitably and positively shaped my academic journey and aspirations.

Lastly, my gratitude extends to my colleagues---Theo Smertnig, Matthias Vogel, and Daniel Widmer---who, despite their demanding schedules, proofread my thesis and provided critical feedback. Their contributions were invaluable to the refinement of my work.
\end{acknowledgements}
\newpage
{\small\tableofcontents}
\newpage
\pagenumbering{arabic}
\chapter{Introduction}
% PDEs are important
\section{Background and Motivation}
Partial Differential Equations (PDEs) are the cornerstone of many scientific and engineering disciplines. Their role is paramount in modelling the complex dynamics of various physical phenomena, ranging from fluid dynamics to electromagnetism. Achieving accurate solutions to PDEs presents considerable challenges, particularly in fields where precision or performance is crucial.

% traditional methods are great but computationally expensive
Traditional methods for solving PDEs primarily involve numerical methods. Prominent among these are finite difference (FDM), finite element (FEM), and finite volume (FVM) techniques. These methods were established decades ago \cite{FD} and, as a result of continuing refinement and improvement, offer robust solutions, validated by their strong convergence properties and accuracy. In some cases, arbitrarily high-order accuracy can be achieved \cite{arbitrarly_high_order}. 
Despite their widespread usage and proven track record, traditional numerical methods come with certain limitations. They typically require discretizing the problem domain using a mesh, which can become computationally prohibitive, particularly when dealing with multi-scale or high-dimensional problems.

%What if I make it shorter: 
Meshfree methods like Smoothed Particle Hydrodynamics (SPH) \cite{SPH} and the Material Point Method (MPM) \cite{MPM} leverage a Lagrangian framework for enhanced flexibility in navigating complex geometries and dynamic interfaces \cite{SPH}. Despite their advantages, challenges such as managing boundary conditions and high computational demands persist. These methods and traditional, mesh-based ones conform to causality principles, necessitating sequential time-stepping that becomes a bottleneck for simulations of complex or highly non-linear systems.

% ML is exploding
Meanwhile, the advent of machine learning (ML) and deep learning (DL) technologies, propelled by advancements in computational hardware, such as GPUs, and an increasing volume of data, has ushered in a new era in computational science. These technologies have led to groundbreaking discoveries across various fields. In biology, AlphaFold by DeepMind \cite{AlphaFold} has revolutionised the prediction of protein structures with unparalleled atomic accuracy. In the realm of chemistry, Sch{\"u}tt et al. introduced a deep learning methodology that offers profound insights into the quantum-mechanical observables of molecular systems \cite{Schuett}. Moreover, the development of Transformer Networks \cite{Attention_is_all_you_need} has significantly transformed natural language processing (NLP), facilitating the emergence of large language models like GPT-4 \cite{GPT4}, which have made remarkable impacts globally. Traditionally, ML and DL have predominantly focused on data-driven approaches, excelling in creating models that map input data to outputs in a supervised manner or in identifying patterns within an unsupervised fashion.

Amidst these advancements, a new approach for solving problems related to differential equations has emerged in the form of Physics-Informed Neural Networks (PINNs) (\cite{Raissi},\cite{Psichogios}) in the broader concept of scientific machine learning (SciML) (\cite{PINNwithdata},\cite{FF2},\cite{PINO},\cite{DeepONets}). They reduce the reliance on extensive datasets characteristic of traditional ML techniques by directly incorporating physical laws into the learning algorithm, enabling learning even without pre-gathered data. This is especially beneficial if the data generation process is difficult or impossible altogether.

% we use PINNs to solve EWE
In this thesis, we investigate the use of PINNs within seismology, a field with broad applications such as earthquake hazard assessment and the exploration of subsurface resources. This exploration is pivotal for identifying deep mineral deposits, vital for conducting efficient and safe mining operations \cite{mineral}. At the heart of our study is the PDE that is central to seismology---the elastic wave equation. This equation models the propagation of seismic waves through Earth's solid materials, encompassing compressional (P) and shear (S) waves. It provides a depiction of seismic phenomena that more accurately reflects the dynamics of Earth's solid structure, compared to simpler models like the acoustic wave equation.\cite{elastic_FWI}. 

% Traditional methods exist but are expensive
Traditional methods like FD and SEM have long been employed to solve seismic equations accurately. Both methods, with SEM introduced by Komatitsch and Tromp in 1999 \cite{SEM}, have evolved significantly, enhancing both performance and accuracy. This progress has facilitated the creation of popular software applications such as SPECFEM \cite{SPECFEM} and Seismic CPML \cite{seismic_CPML}. Despite their proven effectiveness and widespread adoption, the accurate modelling of seismic activities continues to pose challenges, primarily due to the high computational demands of these methods. This issue becomes particularly acute in scenarios involving three-dimensional domains or the simulation of numerous seismic sources.

% Potential of PINNs
The potential of PINNs in this context is substantial. As a meshfree approach, PINNs offer the prospect of significantly reduced computational demands, particularly for large-scale or complex seismic problems. Furthermore, they can be trained as surrogate models, conditioned on the initial conditions of the simulation. Once trained, they can infer responses at any given point in the space-time continuum without executing a full-scale simulation. This characteristic is especially beneficial for analyzing the amplitude responses generated by earthquakes. Traditional mesh-based methods would necessitate running a complete simulation for each source location, a time-consuming and resource-intensive task.

While PINNs show promise, their application in seismology, especially in solving the elastic wave equation, remains limited. The optimal network architecture for this task and the performance of different networks under various parameter settings are still undetermined. PINNs face inherent challenges, such as spectral bias, which hampers their ability to accurately resolve high-frequency components and address multi-scale problems. These issues have been partially addressed in the context of various PDEs but not for the elastic wave equation. Further, the potential of PINNs to scale from toy problems to practical applications in seismology, moving beyond simple test cases to tackle more complex scenarios, has yet to be explored.

% Our contributions
The primary objective of this thesis is to assess the effectiveness of PINNs in mitigating the computational burden associated with traditional numerical simulations and to ascertain whether they offer a more efficient and feasible approach for seismic wavefield analysis. To this end, we apply PINNs to the elastic wave equation across various parameter models and conditions. This examination encompasses scenarios featuring single seismic sources and conditioning the model based on the source location, thereby enabling the inference of seismic waves for any specified source location. We will rigorously assess the accuracy, efficiency, and scalability of PINNs compared to conventional numerical methods. Moreover, we undertake a comprehensive study of different novel neural network architectures to identify those most suited for addressing the challenges inherent in modelling seismic phenomena. Through this detailed exploration, this thesis aims to contribute significantly to the expanding body of knowledge on how modern SciML techniques, particularly PINNs, can advance the field of seismic wavefield simulation.

\section{Thesis Structure Overview}
This thesis is structured as follows. Initially, we provide background information, detailing the elastic wave equation mathematically and introducing the general formulation of PINNs. The body of the thesis is organized into three main chapters. In the first chapter, we demonstrate our application of PINNs to solve the elastic wave equation for various seismic settings. We establish a baseline PINN model, highlighting its promising applications and identifying three primary limitations: accuracy, hyperparameter tuning intensity, and efficiency. The subsequent chapter addresses the first two limitations by exploring innovative architectural adjustments to deliver more accurate and robust solutions. In the final chapter, we delve into strategies aimed at augmenting the efficiency of PINNs by conditioning these networks on the source location. 

\newpage
\chapter{Background}
\section{The Elastic Wave Equation}
\label{section:ELASTIC_WAVE_EQUATION}
\subsection{Introduction}

Seismology, stemming from the Greek words \textit{seismos} (earthquake) and \textit{logos} (science), has expanded its scope of being the mere study of earthquakes \cite{intro_to_seismology}. While studying earthquakes remains the primary aspect, it also encompasses the broader study of Earth's interior physics. Through analyzing seismic waves --- originating from natural earthquakes or human-made explosions --- and their journey through Earth's interior, we gain insights into its internal structure \cite{intro_to_seismology} \cite{seismic_source_theory}. Advancements in physics theory and technology have significantly propelled the evolution of seismology. The first introduction of seismographs in the late $19^{th}$ century marked a pivotal milestone, enabling the capture and recording of ground vibrations over time, including remote ones, thus fostering a more quantitative approach in the field \cite{history_of_seismology} \cite{quantitative_seismology}. 
In recent years, the development of seismology has been characterised by the integration of increasingly high-quality and high-quantity data, sophisticated models of seismic sources, enhancements in wave propagation theory and recent improvements in computational methods, all contributing to a deeper understanding of seismic phenomena \cite{recent_advances} \cite{seismic_source_theory} \cite{quantitative_seismology}. 
Central to all subfields of seismology are seismic waves, which, for most applications, can be well approximated by elastic waves. These waves cause reversible deformations in the Earth's material \cite{eng_seismology}. To be more specific, the term \textit{elastic} refers to the assumption that deformations of the Earth's material induced by seismic waves are temporary and will disappear upon the removal of the wave-induced stress \cite{eng_seismology}.
The elastic wave equation governs the propagation of these waves and their interaction with solid materials, including their reflection, refraction, and diffraction. This PDE is fundamental to this thesis and will be the subject of a more detailed examination in the subsequent section.

\subsection{Formulation}
\label{Section:Elastic_wave_equation}
The elastic wave equation governs the propagation of seismic or elastic waves through heterogeneous, viscoelastic solid media. This PDE is expressed as follows:

\begin{equation}
\begin{aligned}
\rho^s \partial_t^2 \mathbf{u} &= \nabla \cdot \Sigma^s + F^s, \\
\Sigma^s &= C^U : \varepsilon - \sum_{l=1}^{L} \mathbf{R}^l, \\
\varepsilon &= \frac{1}{2} (\nabla \mathbf{u} + (\nabla \mathbf{u})^T), \\
\forall 1 \leq l \leq L, \quad \partial_t \mathbf{R}^l &= \frac{1}{{\tau^{S^l}}} (-\mathbf{R}^l + \delta c^l : \varepsilon).
\end{aligned}
\end{equation}
In this equation, $\rho^s \in \mathbb{R} [kg/m^3]$ represents the density of the solid, $\mathbf{u} \in \mathbb{R}^d [m]$ denotes the displacement of the solid, $\Sigma^s \in \mathbb{R}^{d \times d} [Pa]$ is the stress tensor of the solid, $C^U \in \mathbb{R}^{d \times d \times d \times d} [Pa]$ refers to the unrelaxed modulus, $F^s \in \mathbb{R}^d [N/m^3]$ signifies the external forces acting on the solid, and $\varepsilon \in \mathbb{R}^{d \times d} [-]$ is the strain tensor. The term $L$ indicates the number of relaxation mechanisms, the relaxation auxiliary variables are represented by $(\mathbf{R}^l)_{1\leq l \leq L} \in \mathbb{R}^{d \times d} [Pa]$, the stress relaxation times are denoted by $(\tau^{S^l})_{1\leq l \leq L} \in \mathbb{R} [s]$, and the modulus defect is given by $\delta c^l \in \mathbb{R}^{d \times d \times d \times d} [Pa]$. For detailed derivations of the parameters $\mathbf{R}^l$, $\tau^{S^l}$, and $\delta c^l$, refer to \cite{SPECFEM2D-DG} and \cite{SEM}.

For the simulations carried out in this thesis, the external forces $F^s$ are not directly included in the governing equations but are represented by initial conditions. Furthermore, considering the medium as purely elastic, we eliminate the need for the auxiliary variables $\mathbf{R}^l$, thereby simplifying our equations to:

\begin{equation}
\begin{aligned}
\rho^s \partial_t^2 \mathbf{u} &= \nabla \cdot \Sigma^s, \\
\Sigma^s &= C^U : \varepsilon, \\
\varepsilon &= \frac{1}{2} (\nabla \mathbf{u} + (\nabla \mathbf{u})^T). \\
\end{aligned}
\end{equation}
In this context, the stress tensor $\Sigma^s$ is linearly dependent on the 21 independent constants represented by the unrelated modulus $C^U$ \cite{21}.

\begin{figure}
\begin{center}
  \includegraphics[width=0.6\linewidth]{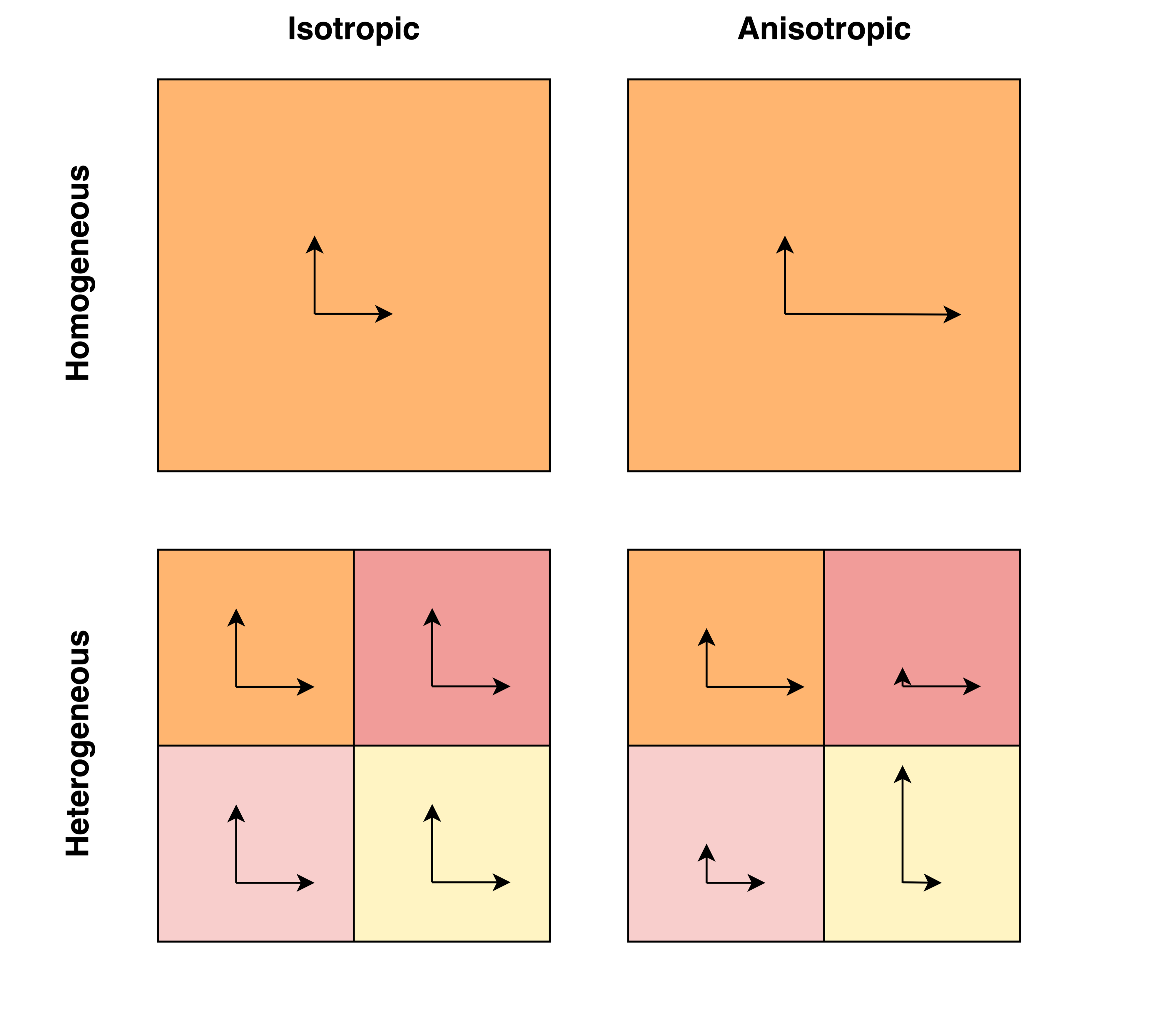}
  \caption{An illustrative depiction of homogeneous/heterogeneous and isotropic/anisotropic materials is presented, where the colours denote the type of material, and the vectors signify the directional dependence of certain medium properties, such as the shear modulus $\mu$. The term \textit{homogeneous} describes a domain consisting of a single material type, whereas \textit{heterogeneous} refers to a composition of at least two materials with differing properties. The concept of \textit{anisotropy} is introduced to describe the directional dependence of a medium's property, as indicated by the non-uniform lengths of the vectors in the $x$ and $y$ directions (assuming a two-dimensional domain). Conversely, \textit{isotropic} materials are characterised by uniform behaviour across different directions; this is depicted through vectors of equal length, demonstrating that an isotropic material undergoes identical deformation along the $x$ and $y$ axes when subjected to an external force. Regarding material parameters, this thesis focuses on the first and second Lam{\'e} parameters, $\lambda$ and $\mu$ and the density $\rho$.}
  \label{fig:isotropic}
  \end{center}
\end{figure}

By assuming isotropy,\footnote{Isotropy signifies that the properties of a medium are direction agnostic. When pressure is exerted on a perfectly isotropic cube, the cube's elastic or inelastic deformations remain consistent irrespective of the pressure's directional application. For an illustrative example, see Figure (\ref{fig:isotropic}).} the complexity of the material's behaviours is significantly decreased, resulting in reducing the number of constants to two: The first and second Lam{\'e} constants $\lambda$ and $\mu$. $\lambda$ indicates a material's compressibility, reflecting its response to uniform pressure application. Conversely, $\mu$, known as the shear modulus, quantifies the material's resistance against shearing deformations. These constants are pivotal in determining the velocities of pressure waves ($v_p$) and shear waves ($v_s$) propagating through a material. Specifically, these velocities are calculated from the Lam{\'e} constants as
\begin{equation} 
v_p = \sqrt{\frac{\lambda + 2\mu}{\rho}},
v_s = \sqrt{\frac{\mu}{\rho}}.
\label{eq:lame}
\end{equation}
The assumption of isotropy significantly simplifies the computation of the stress tensor, enabling it to be succinctly expressed as:
\begin{equation} \label{eq:sigma}
\begin{aligned}
\Sigma^s_{ij} &= C_{ijkl} \varepsilon_{kl}, \\
C_{ijkl} &= \lambda \delta_{ij}\delta_{kl} + \mu \delta_{ik}\delta_{jl} + \delta_{il}\delta_{jk}.
\end{aligned}
\end{equation}

In this formulation, $\delta_{jl}$ represents the Kronecker-delta. By adopting matrix notation, the equation for $\Sigma^s$ is transformed into:
\begin{equation} \label{eq:stress_1}
\begin{aligned}
\Sigma^s = \lambda \begin{bmatrix}
\epsilon_{xx} + \epsilon_{yy} & 0 \\
0 & \epsilon_{xx} + \epsilon_{yy}
\end{bmatrix} + 2\mu \begin{bmatrix}
\epsilon_{xx} & \epsilon_{xy} \\
\epsilon_{xy} & \epsilon_{yy}
\end{bmatrix},
\end{aligned}  
\end{equation}
which can be further simplified to:
\begin{equation} \label{eq:stress}
\begin{aligned}
\Sigma^s = \lambda \, \text{tr}(\epsilon) \, I + 2\mu \, \epsilon.
\end{aligned}  
\end{equation}
In the context of this thesis, where both homogeneous and heterogeneous domains are analyzed, a general assumption of a heterogeneous domain is made. Consequently, the parameters $\lambda$ and $\mu$ are considered as spatial functions, denoted by $\lambda(\mathbf{x})$ and $\mu(\mathbf{x})$. The comprehensive governing equations, encapsulating our assumptions of elasticity, heterogeneity, and isotropy, are thus represented as:
\begin{equation} \label{eq:solid}
\begin{aligned}
\rho^s \partial_t^2 \mathbf{u} &= \nabla \cdot \Sigma^s, \\
\Sigma^s &= \lambda(\mathbf{x}) \, \text{tr}(\epsilon) \, I + 2\mu(\mathbf{x}) \, \epsilon, \\
\varepsilon &= \frac{1}{2} (\nabla \mathbf{u} + (\nabla \mathbf{u})^T). \\
\end{aligned}
\end{equation}
The equations in this section are presented using the so-called displacement formulation, where the unknown function $\textbf{u}$ that we solve for in the presented PDE is the displacement field. However, other formulations exist, such as the velocity-stress or displacement-stress formulations \cite{formulations}.

\section{Traditional Numerical Methods}
\subsection{Introduction}

In this thesis, we diverge from the conventional numerical methods typically utilised for solving the wave equation, focusing instead on PINNs. This section offers a concise yet comprehensive overview of traditional methodologies, aiming to position PINNs within the broader context of computational seismology and to establish a comparative baseline. This is crucial, given the lack of a universal analytical solution for the elastic wave equation, with closed-form solutions being limited to particular initial conditions \cite{analytic}. Therefore, all evaluations and comparisons of PINNs in this study are benchmarked against solutions derived from numerical simulations of the elastic wave equation.

\subsection{Brief Overview of Different Numerical Methods}
The array of numerical methods within computational seismology is extensive and continuously evolving, making a comprehensive discussion impractical for this thesis. However, we will concisely introduce three primary categories, emphasizing their significance in seismology: Finite Difference Methods (FDMs), Finite Element Methods (FEMs), and Spectral Element Methods (SEMs).

\paragraph{FDMs} employ finite differences to discretise spatial and temporal gradients in PDEs. For instance, the first-order forward temporal derivative of a continuous one-dimensional function $u(t)$ is approximated as $\frac{du}{dt} \approx \frac{u(t+h) - u(t)}{h}$, where $h$ is a sufficiently small temporal discretization interval. By using a suitable stencil that dictates the directions of the discretization for all derivatives present in a PDE, an algebraic system of equations is formed, which can be efficiently solved. Despite their long-standing applications, FDMs remain popular due to their simplicity and quick convergence. However, their reliance on regular meshes limits their capability in modelling complex, irregular geological structures encountered in seismic analyses \cite{Joly}.

\paragraph{FEMs} involve partitioning the domain into finite elements (lines in one dimension, triangles/quadrangles in two dimensions, tetrahedra in three dimensions), as opposed to discrete points like in FDMs. Each element is defined by shape functions, representing how physical quantities such as displacement or temperature vary within the element. These values are typically stored at the nodes of the elements. The equations for each element, either differential or integral, compile into a global system that is numerically solvable. FEM's significant advantage over FDMs is their ability to accommodate an irregular "grid", making them highly suitable for seismology.

\paragraph{SEMs} were initially introduced for fluid dynamics by Patera et al. in 1983 \cite{Patera} and later adopted for seismology by Komatitsch, Tsuboi, and Tromp \cite{SEM}. SEMs combine the geometric flexibility of FEM with the accuracy of spectral methods, making them a preferred choice in seismology for computing accurate wavefield simulations in highly heterogeneous media. SEMs employ hexahedral elements and discretise the wavefield within these elements using high-degree Lagrangian interpolants. Integration is performed using the Gauss-Lobatto-Legendre rule, resulting in a diagonal mass matrix. This approach allows using explicit methods without inverting linear systems, significantly reducing computational costs \cite{SEM}.
 
\subsection{General Limitations and Applicability to Seismology}
\label{sectoin:traditional_limitatoins}
This section provides a consolidated overview of the previously discussed numerical methods, elucidating their commonalities, differences and inherent limitations while acknowledging that many of these could be mitigated through specialised techniques. This discussion is intended as a general overview rather than a detailed exposition of state-of-the-art numerical methods.

%DISCUSS WITH BEN
\begin{itemize}
\item A critical aspect of numerical simulations is the adherence to the principle of causality.
Specifically, methods employing an explicit or forward scheme must stringently comply with this principle. This compliance is governed by the Courant-Freidrichs-Lewy (CFL) condition, which restricts the time step size for advancing the simulation in time. Exceeding this limit leads to numerical instabilities, hence these schemes are classified as conditionally stable \cite{conditionally_stable}. The SEM exemplifies such a scheme with its explicit approach \cite{SEM_conditionally}.  In contrast, implicit schemes are considered unconditionally stable, as they are not subject to the CFL condition. However, they require solving complex, nonlinear equations, increasing computational demands, complexity in implementation, and the possibility of introducing numerical damping.
 
\item Discretization of the domain is a requirement for all the methods discussed, whether through a regular grid, as for FDMs or through variously shaped, potentially tedious to create, finite elements for FEMs and SEMs. The meshing process is intricate and time-consuming, especially when dealing with highly heterogeneous domains that demand fine resolutions. This complexity poses computational challenges, particularly when coupled with explicit time-stepping schemes, where a finer spatial resolution necessitates smaller temporal discretization to adhere to the CFL condition. 

\item Boundary conditions represent a significant challenge, especially in seismology, where non-reflecting, absorbing boundaries are often preferred to simulate an effectively infinite domain. This concept is crucial in seismology, as the Earth's interior does not provide a natural reflective boundary for seismic waves. Implementing such conditions is complex, with the state-of-the-art approach being the Perfectly Matched Layer (PML), an artificially introduced absorbing layer \cite{PML}.
 \end{itemize}
 
Significant advancements have been made in enhancing the capabilities of traditional numerical methods, particularly in capturing complex behaviours. For instance, the integration of elastic and acoustic domains \cite{coupled} \cite{SPECFEM2D-DG}, as well as the simulation of wavefields in anisotropic media \cite{axissymmetric}, represent noteworthy developments. Although these methods deliver highly accurate solutions for seismic wave propagation, even in complex domains, they incur substantial computational costs. This is especially true when high-resolution solutions are required to model seismic waves precisely with high-frequency components. Such precision demands exceptionally fine meshing in FEM and SEM and even finer grids, alongside minuscule time steps, in FDMs. The discerning reader may observe that the limitations highlighted here are intentionally selected to emphasise the contrast with PINNs, a differentiation that will be further explored in the following section.
%DISCUSS WITH BEN: FD VS FDM
\subsection{Numerical Baseline Used in This Thesis}
This section elaborates on the numerical FDM implementation that serves as the benchmark in our study. The chosen method is DEVITO \cite{Devito}, a high-performance FDM Python package that capitalises on symbolic computation and the optimisation of FDM stencils. DEVITO is distinguished by its user-friendliness and rapid development capabilities, making it a favoured tool for wave propagation simulations within the seismic research community. The selection of DEVITO as our benchmark stems from its ease of use, flexibility, and the specific challenges in comparing it with other popular FDM simulations for the elastic wave equation, such as seismic CPML, which employs a different equation formulation. Established solvers like 
SPECFEM could not be chosen due to integration issues with our approach to handling initial conditions.  Consequently, we developed our own source code for a displacement-based, elastic wave equation FDM solver using DEVITO, affording us comprehensive control over the implementation process. Refer to Appendix (\ref{DEVITO_IMPLEMENTATION}) for a detailed pseudo-code representation.
Unless specified otherwise, we utilized a spatial grid of $512 \times 512$ points and a time step of $dt = 1e-3$.

\section{Physics-Informed Neural Networks}
\label{section:PINNS}

\subsection{Introduction}
Our objective is to provide a comprehensive insight into PINNs. To achieve this, we embark on a scenic route, beginning with traditional numerical methods before swiftly advancing towards ML and DL. We aim to showcase the success of ML/DL across various scientific disciplines, always with a particular focus on seismology. Our journey ultimately narrows down to SciML, highlighting PINNs as one of its most promising areas.

\paragraph{Traditional Numerical Methods}
The progress of scientific discovery owes much to the ongoing evolution and refinement of numerical methods. These traditional techniques have served as a crucial foundation for scientific inquiry, and their impressive development over time has empowered scientists to tackle intricate equations in geophysics and seismology, even within vast, heterogeneous domains. Through this progress, our comprehension of the physics of the Earth's interior and the causes, transmissions, and consequences of earthquakes has dramatically expanded, leading to more precise risk assessment and damage evaluation.

\paragraph{Deep Learning in Seismology}
The emergence and proliferation of ML and DL techniques have brought forth a new epoch in scientific discoveries. These methods have gained considerable traction and have led to an ever-expanding corpus of research on DL, rendering it one of the most extensively studied domains. In recent times, seismology has been increasingly recognised for its potential in applying ML, reflected in the continuously increasing body of research papers \cite{70years}. An exponential surge in computing power, marked primarily by the development of more powerful and affordable GPUs, coupled with the enrichment in the quality and volume of labelled data and continuous enhancements in ML algorithms, have empowered ML techniques to tackle tasks previously considered unattainable. Such tasks include identifying and classifying previously unseen signals and patterns in seismic or geological data. For instance, Leduc et al. used ML to identify acoustic signals previously dismissed as low-amplitude noise, thereby suggesting an imminent fault failure with high accuracy \cite{Leduc}. Wiszniowski et al. successfully utilised real-time Recurrent Neural Networks to predict small-scale natural earthquakes in noisy data, a task at which conventional numerical methods failed \cite{Wiszniowski}. More recently, Moseley, Nissen-Meyer, and Markham introduced a DL-based approach for efficient simulation of seismic responses in layered and faulted acoustic media  \cite{WaveNET}. At a high level, these data-driven methodologies share a common strategy: they compile and process extensive datasets and select from a continuously evolving array of ML techniques, including linear regression, support vector machines, random forests, and various neural network-based approaches. Subsequently, they train the chosen model on the data, either in a supervised manner, with labelled 'key-value' pairs where the model learns the mapping function from input to output, or in an unsupervised manner, where the model learns solely from unlabeled data, performing tasks like clustering. Irrespective of the data type or method employed, these approaches' shared characteristic, whether in seismology or other scientific fields, is that they are entirely data-driven. The model remains impartial to underlying physical principles that may or may not provide significant insights. However, this is not to detract from the value of ML techniques in seismology, where they have demonstrated remarkable versatility. Frequently, the objectives can be accomplished effectively within a purely data-driven environment.
Nonetheless, these techniques have limitations that warrant attention. 
Relying solely on a black-box\footnote{ML and DL techniques are commonly known as black-box models due to their non-transparent internal mechanisms. The workings of individual neurons in a deep neural network and their specific purposes are not fully comprehensible. Nonetheless, in most cases, the quality of the model's output is all that is sought, and understanding these internal mechanisms is not essential.} model that lacks physical interpretability is not without concern. In scientific applications, it's highly desirable for the method used and the results generated to be interpretable \cite{Ben_thesis}. This enables us to comprehend and reason about their effectiveness and limitations. For example, while the accuracy of simulating seismic wave propagation using FDMs is well-understood in terms of temporal and spatial resolution, such clarity is absent when training a deep neural network for earthquake simulation based on seismic data.  Moreover, although black-box ML methods excel in data interpolation, their capacity for generalization beyond the training dataset is often limited. Therefore, achieving excellent results with these methods does not always guarantee successful extrapolation.
This issue is especially pertinent in the Earth sciences, where the scarcity of observational data emphasises the necessity for approaches that transcend a purely data-driven framework. 
Kong et al. have highlighted the substantial benefits that ML techniques offer in seismology \cite{Kong}. Their study encompasses extensive scientific research that extends beyond our current discussion. Fascinatingly, they conclude with a vital proposition for a hybrid approach that uses ML/DL techniques but also leverages long-standing physical expertise in the field. This approach is critical for further advancement in computational seismology.

\paragraph{PINNs as an Example of SciML}
Recently, SciML has been gaining significant momentum. This emerging discipline aims to harmonise and leverage the power of ML techniques with the rigour of scientific inquiry and domain-specific expertise. It is more than simply implementing ML methods in a scientific domain; rather, it is the process of incorporating physical laws and domain-specific scientific knowledge into the ML model learning process. This methodology aims to ensure that the model fits the data well and observes the fundamental physical principles that oversee the underlying phenomena, leading to a more holistic and interpretable framework for scientific inquiry compared to purely data-driven models. With the increasing success and application of SciML, the field continues to evolve rapidly. Brunton et al.'s research showcased how sparsity-promoting techniques and ML can be used in conjunction with nonlinear dynamical systems to uncover physical equations governing measurement data \cite{Brunton}. Anker et al. introduced Exp2SimGAN, an unsupervised ML model, which can transform an artificial, simulated dataset into an experimental one containing typical experimental artefacts utilised in inelastic neutron scattering \cite{Anker}. Moseley et al. presented a novel denoising approach for low-light lunar images, incorporating a physical noise model of the camera that surpasses conventional camera-calibration methods \cite{Lowlight}. These studies only provide a glimpse of the vast range of SciML's capabilities.
One promising development in SciML is the advent of PINNs. These are neural networks which are designed to obey the fundamental physical laws of a system, usually expressed as ODEs or PDEs, by minimising the violation of these laws in their loss function. The benefits of PINNs are numerous. In a supervised learning context, such as simulating a physical entity, utilizing PINNs ensures that the neural network's proposed solution is physically sound and data-consistent. PINNs are also valuable in low-data situations, such as seismology, where data is scarce or noisy. In such cases, relying on a purely data-driven network is impractical. However, a network that adheres to the underlying physical principles allows for work with minimal or no data. This makes PINNs a powerful tool, akin to traditional numerical methods for solving complex PDEs across various scientific domains.
In the following sections, we will delve into the evolution and formulation of PINNs in greater detail.

Within the domain of SciML, a variety of techniques extend beyond PINNs. Notable among these are Fourier Neural Operators (\cite{FNO},\cite{FNO_elastic}), Deep Operator Networks \cite{DeepONets}, Physics-Informed Neural Operators (\cite{PINO},\cite{PINO_acoustic}), and Neural Differential Equations (\cite{NDE}). These methods are elaborated upon in Section (\ref{cond_related_work}).

\subsection{Previous Work}
\label{section:PINNs_previous_work}
The concept of PINNs can be traced back to developments in the early 1990s, well before the mainstream adoption of ML techniques. For instance, in 1992, Psichogios and Ungar pioneered the integration of physical laws into neural network models for process modelling. They formulated a custom loss function based on differential equations that represent the governing dynamics of a bioreactor, marking one of the earliest instances of a PINN-like approach \cite{Psichogios}. Similarly, in 1994, Dissanayake et al. introduced a numerical method for solving PDEs using neural networks as 'universal approximators' in conjunction with point collocation, effectively transforming the problem into an unconstrained minimization task \cite{even_older_PINN}.
Furthermore, in 1998, Lagaris et al. developed a method where neural networks were utilised to solve PDEs. This approach involved decomposing the solution into two components: one satisfying the initial and boundary conditions and another incorporating a trainable feedforward neural network to ensure compliance with the differential equations \cite{ogLagaris}.
Although these methods resemble modern PINNs, they may have been ahead of their time, limited by the computational resources available. The term PINNs gained prominence in 2019 through a seminal paper by Raissi, Perdikaris, and Karniadakis. They made groundbreaking contributions to the field by proposing novel methodologies for addressing both forward and inverse problems involving PDEs, which is often regarded as the definitive emergence of PINNs as known today \cite{Raissi}.

Over the years, PINNs have evolved to address increasingly complex physical problems. For instance, significant advancements have been made in computational fluid dynamics, as demonstrated by Jin et al., who effectively employed PINNs to simulate complex incompressible flows \cite{Jin}. Sirignano and Spiliopoulos demonstrated the efficacy of PINNs in solving PDEs, specifically the Burgers equation, in very high-dimensional settings using their Deep Galerkin Method \cite{DGM}. With each new contribution, the methodology continues to evolve and improve, as evidenced by the work of Wang et al. on addressing fundamental modes of failure in PINNs. These failures, related to numerical stiffness and unbalanced back-propagation, were overcome through the use of a learning rate annealing algorithm that balanced different terms in the loss function \cite{Wang}. As optimization algorithms become more suited and tailor-made \cite{optimtheoptim}, network architectures more refined \cite{Sinusoidal}, and training schemes more advanced \cite{causality} \cite{CPINNs}, the applications of PINNs in various scientific fields continue to grow. In fluid dynamics and weather modelling, Kashinath et al. successfully implemented PINNs to predict complex weather and climate patterns accurately \cite{Kashinath}. Jagtap et al. employed Extended-PINNs (XPINNs) to solve the inverse supersonic-compressible flow problem, utilizing domain decomposition to deploy local neural networks in each subdomain for more efficient problem-solving \cite{supersonic}. Furthermore, Moseley et al. proposed Finite-Basis PINNs (FBPINNs), a FEM-inspired approach that utilises individual subdomain normalization and a flexible subdomain training schedule to address spectral bias and achieve rapid convergence of multi-scale physics problems \cite{FBPINNS}. Beyond these examples, PINNs have been applied in biomedical engineering \cite{Bio}, in addressing the ongoing power problem \cite{power}, and in material science for non-homogeneous material identification \cite{material}.

\subsection{General Formulation}
This section provides an overview of the general formulation of PINNs. Subsequently, we present a detailed account of the formulation, adaptation, and application of PINNs employed in our study.

\subsubsection{Underlying PDE}
We consider an arbitrary PDE, presented in its residual form, which characterizes a physical phenomenon:
\begin{equation}
\begin{aligned}
\label{eg:PINNS1}
\mathcal{N}[\textbf{u}(t,\textbf{x}),\lambda] = 0, \quad \textbf{x} \in \Omega, \quad t \in [0,T],
\end{aligned}
\end{equation}
subject to the subsequent initial and boundary conditions:
\begin{equation}
\begin{aligned}
\label{eg:PINNS2}
\textbf{u}(0,\textbf{x}) = \textbf{g}(\textbf{x}), \
\mathcal{B}[\textbf{u}(t,\textbf{x}),\lambda] = 0, \quad t \in [0,T], \quad \textbf{x} \in \partial \Omega. \
\end{aligned}
\end{equation}
The differential operator, represented by $ \mathcal{N}[\cdot]$, is parameterized by $\lambda$, while $\textbf{u}(t,\textbf{x}): \Bar{\Omega} \times [0,T] \rightarrow \mathbb{R}^D$ denotes the potentially vectorial, unknown latent solution dictated by the PDE. The spatial and temporal variables, $\textbf{x}$ and $t$, respectively, are confined to their respective domains $\Bar{\Omega} = \Omega \cup \partial \Omega$ and $[0,T]$. The initial condition is signified by $\textbf{u}(0,\textbf{x})$, parameterized by a function $\mathbf{g}(\textbf{x}): \Bar{\Omega} \rightarrow \mathbb{R}^D$. The boundary conditions are encapsulated by $\mathcal{B}[\cdot]$, which may assume various forms, including Dirichlet, Neumann, periodic, or Robin boundary conditions

\subsubsection{Structuring the PINN Loss Function}
In the general formulation, PINNs approximate the unknown latent solution $\textbf{u}(t,\textbf{x})$ using a neural network $\Lambda(t,\textbf{x},\theta)$, where $\theta$ denotes the trainable parameters, namely weights and biases, of the network. The network is trained by directly incorporating the underlying equations (\ref{eg:PINNS1}) - (\ref{eg:PINNS2}) through a suitable loss function, typically represented as:
\begin{equation}
\label{eq:Loss}
\mathcal{L} =
\lambda_{\mathcal{P}}\mathcal{L}_{\mathcal{P}}(\mathbf{\theta}) + \lambda_{\mathcal{IC}} \mathcal{L}_{\mathcal{IC}}(\mathbf{\theta}) + \lambda_{\mathcal{BC}}\mathcal{L}_{\mathcal{BC}}(\mathbf{\theta}) + \lambda_{\mathcal{D}} \mathcal{L}_{\mathcal{D}}(\mathbf{\theta}),
\end{equation}
with $\lambda_{\mathcal{P}}$, $\lambda_{\mathcal{IC}}$, $\lambda_{\mathcal{BC}}$, and $\lambda_{\mathcal{D}}$ being potentially trainable parameters that balance the importance of each loss term. The physics loss, $\mathcal{L}_{\mathcal{P}}(\mathbf{\theta})$, enforces the governing Equation (\ref{eg:PINNS1}) on the network's solution. The initial and boundary losses, $\mathcal{L}_{\mathcal{IC}}(\mathbf{\theta})$ and $\mathcal{L}_{\mathcal{BC}}(\mathbf{\theta})$, respectively, ensure the uniqueness of the approximation solution by enforcing the initial and boundary conditions from Equation (\ref{eg:PINNS2}). The data loss, $\mathcal{L}_{\mathcal{D}}(\mathbf{\theta})$, measures the fit of the approximation solution to the given measurement data. It is crucial to note that the data loss is not necessary for forward problems, where the objective is to simulate the solution of a PDE. However, it becomes essential in inverse problems aimed at estimating parameters from data measurements. While there are various approaches to incorporate initial and boundary conditions, the formulation presented here is only one possible method. The individual loss terms are derived as follows:
\begin{equation}
\begin{aligned}
\mathcal{L}_{\mathcal{P}}(\mathbf{\theta}) &= \frac{1}{N_{\mathcal{P}}}\sum_{i=0}^{N_\mathcal{P}} || \mathcal{N}[\Lambda(t_i,\textbf{x}_i,\theta),\lambda] ||^2, \\
\mathcal{L}_{\mathcal{IC}}(\mathbf{\theta}) &= \frac{1}{N_\mathcal{IC}}\sum_{j=0}^{N_\mathcal{IC}} || \textbf{g}(\textbf{x}_j) - \Lambda(0,\textbf{x}_j,\theta) ||^2, \\
\mathcal{L}_{\mathcal{BC}}(\mathbf{\theta}) &= \frac{1}{N_{\mathcal{BC}}}\sum_{k=0}^{N_\mathcal{BC}} || \mathcal{B}[\Lambda(t_k,\textbf{x}_k,\theta),\lambda] ||^2, \\
\mathcal{L}_{\mathcal{D}}(\mathbf{\theta}) &= \frac{1}{N_\mathcal{D}}\sum_{h=0}^{N_\mathcal{D}} || \textbf{u}^\ast(t_h,\textbf{x}_h) - \Lambda(t_h,\textbf{x}_h,\theta) ||^2, 
\end{aligned}
\end{equation}
where $\textbf{u}^\ast(t,\textbf{x})$ represents known solution points provided by labeled measurement data. The loss terms in a PINN are designed to constrain the network solution to adhere to the governing PDE, initial conditions, boundary conditions, and any available measurement data. Typically, the training points are either randomly sampled or Sobol sequence-generated. They include $\{t_i,\textbf{x}_i\}_{i=0}^{N_\mathcal{P}}$ on the spatial and temporal domain $\Bar{\Omega} \times [0,T]$, $\{\textbf{x}_j\}_{j=0}^{N_{\mathcal{IC}}}$ on the domain $\Bar{\Omega}$ for $t=0$, $\{t_k,\textbf{x}_k\}_{k=0}^{N_\mathcal{BC}}$ on the boundary domain $\partial \Omega \times [0,T]$, and $\{\textbf{u}^\ast(t_h,\textbf{x}_h),t_h,\textbf{x}_h\}_{h=0}^{N_\mathcal{D}}$ for the explicitly defined measurement data points.

\subsubsection{Advantages of Unsupervised Learning in PINNs}
A distinct feature of PINNs that differentiates them from traditional numerical methods is their independence from explicit discretization of space and time. Instead, they allow for the random sampling of training and evaluation points. As with any neural network, PINNs act as function approximators, capable of mapping any point in space-time upon training. PINNs are particularly advantageous in domains where data is scarce, difficult to generate, or unattainable. Notably, PINNs, except the optional data loss term, do not depend on labelled data and can learn solely from a PDE presented in its residual form. In this context, PINNs are categorized as unsupervised learning models for forward problems and as supervised learning models for inverse problems. Henceforth, our discussion will concentrate exclusively on forward problems in this thesis.

\subsubsection{Designing the Neural Network Architecture}
The objective is to develop an approximation function that closely mirrors the true underlying function. This involves minimizing the difference between the true solution $\textbf{u}(t,\textbf{x})$ and the approximation function $\Lambda(t,\textbf{x},\theta)$ to near-zero as the number of training iterations increases. To achieve this, selecting an appropriate neural network architecture and determining a suitable number of trainable parameters $\theta$ are crucial. In practice, the true solution $\textbf{u}(t,\textbf{x})$ to which PINNs are compared is typically either an analytical solution, experimentally obtained, or derived from FDM, FEM or SEM-type simulations.

\begin{figure}
\begin{center}
\includegraphics[width=1.0\linewidth]{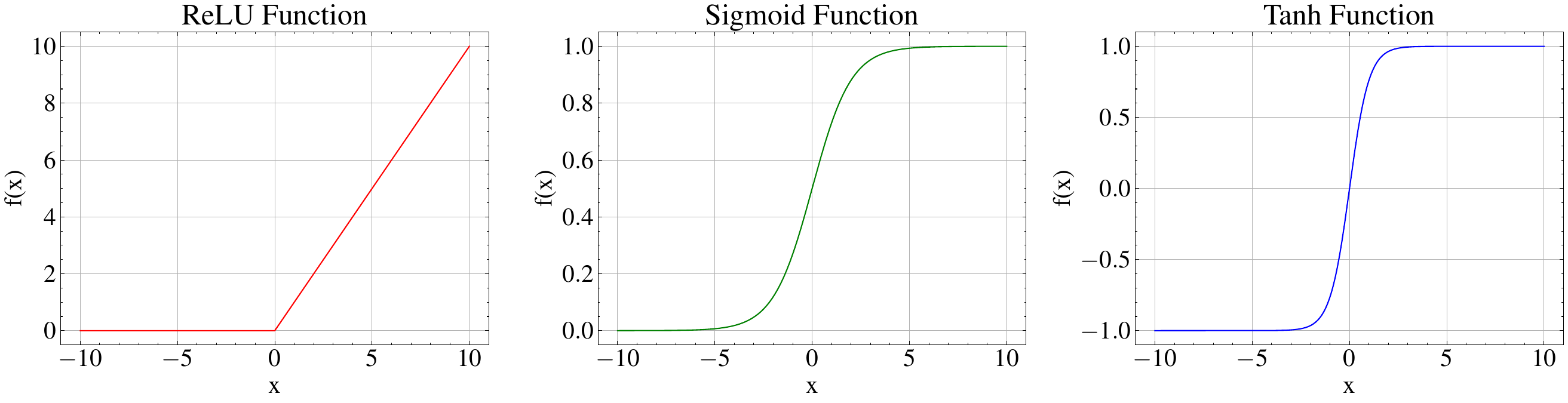}
\caption{Illustration of the three most commonly used activation functions in neural networks: ReLU, sigmoid, and tanh. Note the distinctive kink at $x=0$ for ReLU, which sets it apart from sigmoid and tanh by making it non-differentiable at $x=0$.}
\label{fig:activations}
\end{center}
\end{figure}

A variety of network architectures have been explored for PINNs, including Long-Short Term Memory (LSTM) Networks \cite{LSTM1} \cite{LSTM2}, Generative Adversarial Networks (GAN) \cite{GAN} \cite{GAN2}, and Recurrent Neural Networks (RNN) \cite{RNN}, among others. However, the most commonly adopted architecture is the Feed-Forward Fully Connected Neural Network (FFFCN), hereafter abbreviated as FCN. In an FCN, signals move strictly forward, without any backward or feedback connections, unlike in an RNN. This architecture consists entirely of fully connected layers, where each neuron in one layer is connected to every neuron in the subsequent layer. The output of the network, $\Lambda(t,\textbf{x},\theta)$, is obtained through the forward pass as follows:
\begin{equation}
\label{eq:forward}
\begin{aligned}
\textbf{z}^{0} &= W^{0}(t,\textbf{x}) + b^{0}, \\
\textbf{a}^{1} &= \sigma^{1}(\textbf{z}^{0}), \\
\textbf{z}^{1} &= W^{1}\textbf{a}^{1} + b^{1}, \\
&\vdots \nonumber \\
\textbf{a}^{L-1} &= \sigma^{L-1}(\textbf{z}^{L-2}), \\
\textbf{z}^{L-1} &= W^{L-1}\textbf{a}^{L-1} + b^{L-1}, \\
\textbf{a}^{L} &= \sigma^{L}(\textbf{z}^{L-1}), \\
\Lambda(t,\textbf{x},\theta) &= W^{L}\textbf{a}^{L} + b^{L}.
\end{aligned}
\end{equation}
The input $(t,\textbf{x})$ undergoes a linear transformation via the weight matrix $W^0$ and bias vector $b^0$ at the input layer. Subsequently, for each of the $L-1$ hidden layers, the output from the preceding layer is processed through the activation function $\sigma^l$ and then linearly transformed by the current layer's weight and bias. This sequence continues up to the final output layer, where the output undergoes one last linear transformation by $W^L$ and $b^L$ to yield the final output $\Lambda(t,\textbf{x},\theta)$. Activation functions $\sigma$ introduce non-linearities, enabling the modelling of complex nonlinear functions. Popular activation functions include ReLU, sigmoid, and tanh, each with distinct advantages (see Figure (\ref{fig:activations})).

For PINN training, a continuous function such as tanh is often preferred over ReLU as an activation function, a choice that will be further elucidated later in the discussion.

\subsubsection{Training Process and Backpropagation}
Like any neural network, a PINN is trained using the backpropagation algorithm, also referred to as the backward pass. This process encompasses several steps:
Initially, the input $(t,\textbf{x})$ is propagated through the network (forward pass), as delineated in Equation (\ref{eq:forward}), to procure the approximate solution $\Lambda(t,\textbf{x},\theta)$. Subsequently, with the current approximation, we compute the loss function's value as specified in Equation (\ref{eq:Loss}). The gradient of the loss function with respect to the network parameters $\theta$ is then determined by applying the chain rule to each component of the loss function. Ultimately, the network parameters are updated towards minimizing the loss, with the direction informed by the computed gradient.
Theoretically, one could employ the gradient descent algorithm with a constant learning rate $\eta$ for this purpose. The update for each parameter $\theta_i$ at every iteration would involve moving in the opposite direction of the gradient, with the step size dictated by the learning rate:
\begin{equation}
\theta_i^{n+1} = \theta_i^{n} - \eta \frac{\partial\mathcal{L}}{\partial\theta_i^{n}}.
\end{equation}
\subsubsection{Optimisation Algorithms in PINN Training} \label{Optimisation_algorithms_in_PINN_training}
The utilization of optimisation algorithms is pivotal in training deep neural networks. These algorithms, commonly referred to as optimisers, are responsible for updating the model parameters throughout each training iteration. While the standard gradient descent approach may not always be suitable for training a deep neural network, various other optimisation algorithms are employed to achieve optimal results. There are a plethora of optimisers to choose from, such as stochastic gradient descent (SGD) \cite{SGD_paper}, adaptive moment estimation (ADAM) \cite{ADAM_paper}, adaptive gradient algorithm (Adagrad) \cite{Adagrad}, root mean squared propagation (RMSprop) \cite{RMSprop}, Broyden--Fletcher--Goldfarb--Shanno algorithm (BFGS) \cite{BFGS}, and limited memory BFGS algorithm (LBFGS) \cite{LBFGS_paper}.

For the training of PINNs, ADAM and LBFGS are the two most popular optimisers. ADAM, a first-order optimisation method, optimises the model parameters using the first derivative of the loss function. It calculates adaptive learning rates for each parameter by estimating the first and second moments of the gradient. ADAM is known for its rapid convergence properties, though it may not always converge to an optimal solution, as it can get stuck in local minima or saddle points.

On the other hand, LBFGS is a quasi-Newton method that approximates the second-order information, precisely the inverse of the Hessian matrix, which represents the second-order partial derivatives of the loss function. This approximation allows the algorithm to consider the curvature of the loss landscape. LBFGS uses past updates of positions and gradients to implicitly approximate the inverse Hessian, offering a more efficient alternative to directly computing the full inverse Hessian (BFGS), which is computationally and memory-intensive. Although LBFGS can achieve convergence in fewer steps than ADAM, it has a higher memory requirement and is not well-suited for batched or mini-batch training. The dependency of LBFGS on past inverse Hessian estimates can introduce instability when there is a change in the data points used for gradient computation. While there are novel approaches to enable batched training with LBFGS, such as \cite{batched_LBFGS}, these are not yet widely adopted in tools like PyTorch. Conversely, ADAM efficiently supports mini-batch training.

The specific characteristics of the problem influence the selection of an optimiser in PINN training. Different underlying PDEs generate distinct loss landscapes that may favour one optimiser over another. Several studies utilizing PINNs have reported successful results with the ADAM optimiser \cite{ADAM1} \cite{physical_activation_functions}, while others have preferred the LBFGS optimiser \cite{LBFGS1} \cite{LBFGS2}. It is also a common strategy to initially employ ADAM to achieve a good preliminary approximation and subsequently fine-tune the model using LBFGS \cite{AandL1} \cite{optimtheoptim}. This approach is contingent on achieving an accurate initial approximation with ADAM, ensuring it does not become entrapped in a suboptimal local minimum.

\begin{figure}
\begin{center}
  \includegraphics[width=1.0\linewidth]{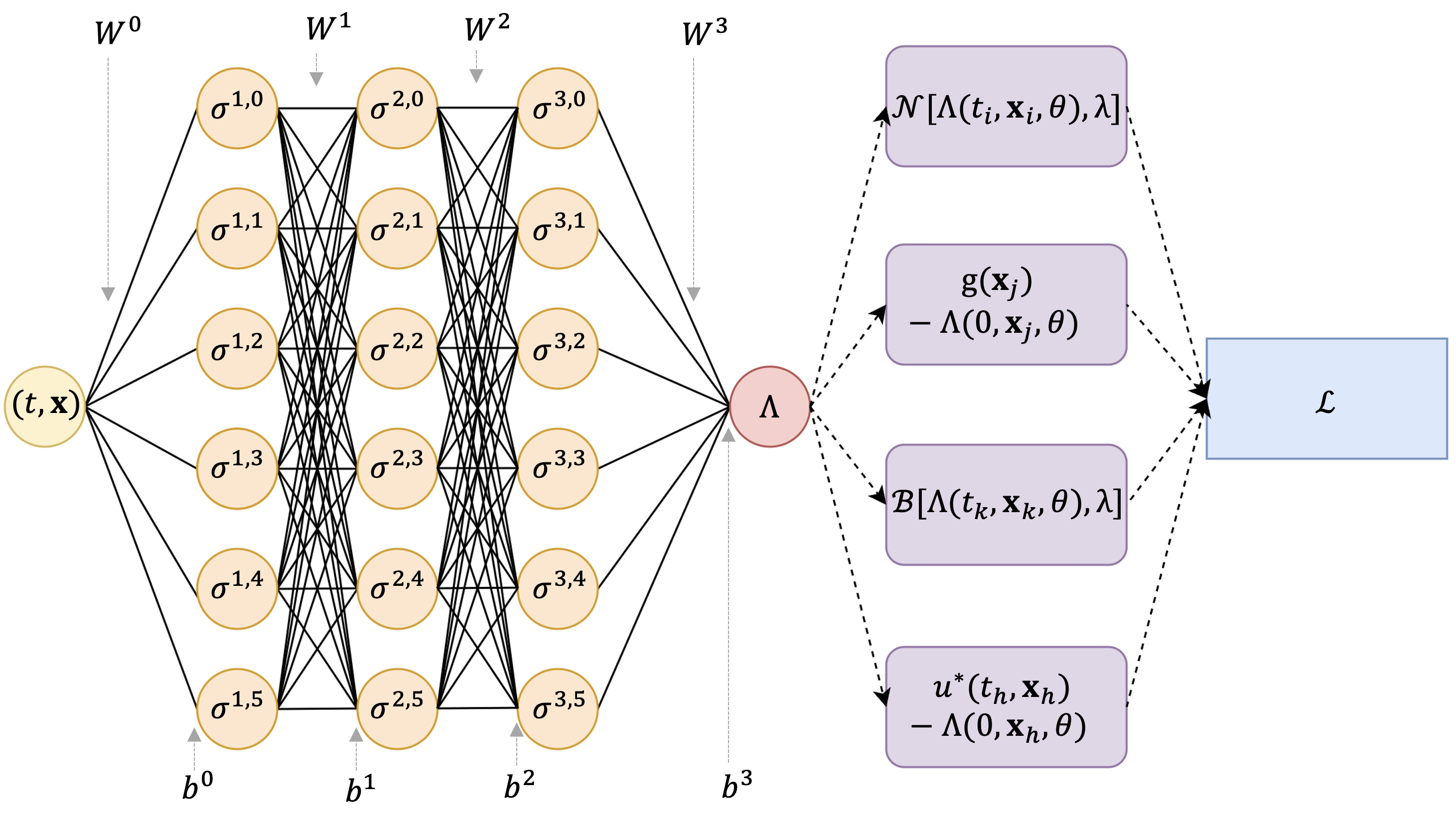}
  \caption{Depiction of a PINN workflow with an FCN illustrated by yellow, orange, and red nodes. The yellow node symbolizes the network input, incorporating time, space, and potentially other dimensions. The input connects to an input layer of orange nodes through weights $W^0$ and biases $b^0$, followed by several hidden layers. Each hidden layer applies activation functions ($\sigma^{1,n}$, $\sigma^{2,n}$, etc.) and transforms inputs via weights ($W^1$, $W^2$, etc.) and biases ($b^1$, $b^2$, etc.). The output layer linearly transforms the final hidden layer's output to match the desired dimension (one-dimensional for scalar fields, two-dimensional for vector fields, etc.), resulting in the output $\Lambda(t_i,\textbf{x}_i)$. The overall loss $\mathcal{L}$ is determined by evaluating the PDE residual, the initial conditions, the boundary conditions, and any available data measurements, as depicted by the purple boxes from top to bottom.  $\mathcal{L}$ is then utilised during the backpropagation (not depicted), leading to parameter updates. The depicted network features three hidden layers, each with six neurons.}
  \label{fig:nn_sketch}
  \end{center}
\end{figure}

\subsubsection{Derivative Calculation and Implications for Activation Functions}
The process of evaluating the loss function during training necessitates the computation of all partial derivatives of the network output $\Lambda(t,\textbf{x},\theta)$ with respect to time and space, as dictated by the specified PDE. While derivatives can be computed analytically \cite{Ben}, the advent of automatic differentiation in widely used machine learning frameworks such as PyTorch, JAX, and TensorFlow simplifies this task considerably. This capability proves particularly beneficial in the development of PINNs. It is important to note that PINNs often employ continuous activation functions, including tanh, sigmoid, or sin. The rationale behind this choice is that for a PDE involving k-th order derivatives, the activation function must belong to the class $C^k$ to ensure the computation of derivatives with respect to the input variables at least k times while maintaining continuity. The RELU function is classified as $C^0$ because, despite its continuity, it is not differentiable at zero (see Figure (\ref{fig:activations})). This characteristic introduces a challenge for PINNs, as it renders the gradients with respect to the input variables not well-defined. Conversely, activation functions like tanh are classified as $C^\infty$, indicating their capability for smooth differentiation.
\newpage

\chapter{Solving the Elastic Wave Equation with PINNs}
\label{section:Elastic PINNs}
\section{Introduction}
The exploration of PINNs in relation to the elastic wave equation constitutes the cornerstone of this thesis. This investigation is motivated by the urgent necessity to understand and enhance the deployment of PINNs within the specific context of seismology. The rationale behind this motivation is multifaceted.

\paragraph{Limitations of Traditional Numerical Methods} 
As Section (\ref{sectoin:traditional_limitatoins}) elaborates, simulating seismic wave propagation or solving the elastic wave equation via traditional numerical methods presents significant computational challenges in large-scale, highly heterogeneous media. This challenge becomes particularly pronounced when numerous simulations are required, such as in assessing seismic hazards, where it is necessary to simulate seismic wave propagation for various seismic sources to evaluate potential surface impacts. Such tasks demand substantial computational resources, to the extent of engaging the world's most powerful supercomputers \cite{Ben_thesis}. Consequently, there is a pressing need for alternative methodologies that circumvent the limitations of expensive meshing or the use of very fine grids.

\paragraph{Advantages of PINNs}
PINNs offer a promising solution to computational challenges by being meshfree. Furthermore, PINNs learn continuous functions, differentiating them from traditional discretization techniques. This attribute ensures that, upon training, the solution for any point within the defined spatial and temporal domain is \textit{instantaneously}\footnote{Here, \textit{instantaneously} implies the elimination of computing previous time steps. However, it is acknowledged that neural network inference times, while quick, are not zero.} accessible. Additionally, PINNs possess the capability to learn the solution to a PDE for an entire class of problem parameters (\cite{conditional_PINNs2},\cite{conditional_PINNs2},\cite{Ben}). This endeavour of learning a class of functions simultaneously is known as conditioning and, once trained, provides a fast surrogate model, which is often orders of magnitude faster than numerical simulation. Chapter (\ref{chap:conditioning}) will extensively discuss this approach.

\paragraph{Gap in Existing Literature}
Despite the promising capabilities of PINNs across various scientific domains, their application within seismology remains surprisingly underexplored, with notable exceptions highlighted in \cite{elasto_PINNS} and \cite{scattered}. This significant gap in the literature underscores the need for a thorough qualitative and quantitative evaluation of the effectiveness of PINNs under diverse parameter settings in the modelling of seismic wave propagation.

\paragraph{Research Questions}
In this chapter, we aim to assess the efficacy of PINNs within seismology, particularly whether they offer a viable alternative to traditional numerical methods in specific problem settings. Specifically, we examine the performance of PINNs across a range of seismic source sizes, which induce varying frequencies under several parameter configurations of increasing complexity. The investigation begins with models using constant Lam{\'e} parameters (as explored in \cite{elasto_PINNS}) and progresses to more complex, heterogeneous models. The objective is to determine whether PINNs, in their standard form, can achieve sufficient accuracy and efficiency to be considered feasible substitutes for conventional numerical methods in seismology.

\paragraph{Chapter Overview} 
This chapter is dedicated to the development and training of a baseline PINN model tailored for the elastic wave equation under a variety of experimental conditions, with an emphasis on identifying the optimal hyperparameter configuration.

The chapter unfolds as follows: Initially, we review existing literature to set the context and highlight the novel contribution this chapter aims to make. The methods section follows, where we elaborate on configuring our baseline PINN. This includes a detailed description of the network architecture, the formulation of the loss function, the approach to initial and boundary conditions, the strategies for sampling and normalization and the training regimen.
We then present our findings, starting with results from experiments using a constant Lam{\'e} parameter model and extending to tests on two heterogeneous parameter models. A critical analysis of the $L_2$ error variations in response to different seismic source sizes is provided, alongside an examination of the model's sensitivity to various hyperparameters, such as the number of neurons and layers.

The discussion of results, while acknowledging the limitations of our baseline model---particularly its underwhelming accuracy in challenging experimental setups and its inefficiency stemming from the necessity for retraining for each new scenario---serves as a foundation for the subsequent chapters. These chapters aim to address these issues individually, proposing innovative solutions to enhance both the precision and efficiency of PINNs in seismology. Among these solutions are novel network architectures designed to improve accuracy and strategies for conditioning the PINN model on seismic source locations, thereby enhancing efficiency.

\section{Contributions}
This chapter is dedicated to exploring the application of PINNs to model the elastic wave equation. This topic has seen relatively limited research, especially when compared to its application to the acoustic wave equation.

In this chapter, we focus on deploying PINNs to solve the elastic wave equation for a single source scenario. We examine the effectiveness of a traditional Feedforward Neural Network in modelling the elastic wave equation and evaluate its accuracy across a range of parameter models. These models vary from a simple constant parameter setting to a highly heterogeneous but smoothly varying setting and finally to a layered setting characterized by sharp transitions. Furthermore, we investigate the performance of PINNs with different sizes of the initial seismic source, shedding light on their behaviour in scenarios involving higher frequencies and multi-scale solution components.

The exploration of PINNs across various seismic settings lays the foundation for identifying and articulating the primary limitations of our standard PINN approach. These limitations will be addressed in the subsequent chapters, guiding our efforts towards proposing and examining solutions to overcome these challenges.

\section{Related Work}
\label{Section_ELASTic_relatedwork}
Section (\ref{section:PINNs_previous_work}) presents an extensive examination of the development and evolution of PINNs, chronicling the significant advancements facilitated by various research efforts. These efforts span a wide array of learning and sampling methodologies, optimization strategies, and network architectures, paving the way for the effective deployment of PINNs across many scientific disciplines. In this section, the emphasis is placed on the existing literature concerning the application of PINNs within seismology.

While the elastic wave equation more accurately depicts seismic activities, the acoustic wave equation, which simulates scalar sound or pressure waves, is frequently employed as a simplified model. Despite the relatively few instances of PINNs being applied to the elastic wave equation, there has been considerable success with their application to the acoustic wave equation. The challenges inherent to the acoustic wave equation, such as solutions being characterized by components of varying scales that propagate and exhibit oscillatory behaviour, were effectively addressed by Moseley et al. through the use of PINNs \cite{Ben}. Karimpouli et al. conducted a comparative analysis of PINNs and Gaussian processes (GP) to solve the forward and inverse problems associated with the 1D acoustic wave equation \cite{1Dacoustic}. Rasht-Behesht et al. showcased promising outcomes in full waveform inversion for the two-dimensional acoustic wave equation using PINNs \cite{2DFWI}. Alkhadhr et al. investigated the implications of applying hard versus soft constraints on the initial and boundary conditions in the two-dimensional acoustic wave equation \cite{2Dacoustic_IC}. Additionally, Ding et al. introduced self-adaptive PINNs (SA-PINNs) aimed at improving the scalability and precision of PINNs when tackling the acoustic wave equation in areas of complex topography \cite{SA-PINNS}.

These studies have shown that, with appropriate extensions, PINNs can successfully solve the acoustic wave equation, even in complex domains. However, the application of PINNs to the elastic wave equation is not as thoroughly explored. Currently, only two studies have been identified that apply PINNs to the elastic wave equation. Rao et al. successfully applied PINNs to this equation within homogeneous parameter settings for infinite and bounded domains, utilizing a mixed-variable output including the displacement vector and stress components \cite{elasto_PINNS}. The other recent study by Song et al. applied PINNs to the scattered form of the frequency domain elastic wave equation \cite{scattered}, marking a pivotal exploration into the utility of PINNs in simulating elastic wave phenomena.

Our research diverges in several key aspects. The first study (\cite{elasto_PINNS}) primarily addresses the constant parameter model in both bounded and infinite domains. In contrast, we extend our investigation to include multiple parameter models, notably highly heterogeneous ones, to more accurately reflect the complex reality of seismological environments, which often deal with parameter models derived from the Earth's varied internal structure. Additionally, we explore how the accuracy of PINNs varies with the size of the initial seismic source, an aspect not examined in the cited works. The second study (\cite{scattered}) delves into the scattered form of the frequency domain elastic wave equation, incorporating non-constant parameter cases. Contrary to their methodology, our study opts for the time domain approach to the elastic wave equation. This choice is driven by the relative ease and the direct relevance of time domain modelling in effectively capturing the dynamics of seismic wave propagation, offering a clearer insight into the transient nature of seismic events.

\section{Methods}
This section delineates our application of the principle of PINNs, as discussed in Section (\ref{section:PINNS}), in addressing the complexities of solving the elastic wave equation, outlined in Section (\ref{section:ELASTIC_WAVE_EQUATION}). It delves into the specifics of the neural network architecture employed and the formulation of the loss function to ensure compliance with the underlying physical laws. Moreover, this section elaborates on implementing boundary and initial conditions, the chosen sampling strategy, and the training methodologies, including network initialization and the selected optimizer. To ensure transparency and clarity, we also detail the process for calculating the test loss and the chosen metrics for evaluation.

\subsection{Physical Loss}
It is imperative to construct an appropriate loss function to enable the model to learn the physics encapsulated in Equation (\ref{eq:solid}). This is achieved by transforming the PDE into its residual form and subsequently utilizing it as a loss function:
\begin{equation}
\begin{aligned}
L_\mathcal{P} = \frac{1}{N_\mathcal{P}}\sum_{i=0}^{N_\mathcal{P}} \left| \rho \frac{\partial ^2 \Lambda(t_i,\mathbf{x}_i;\theta)}{\partial t^2} - \nabla \cdot \Sigma^s_{\Lambda}\right|^2.
\end{aligned}
\end{equation}
Here, $\Sigma^s_{\Lambda}$ represents the neural network's approximation of the stress tensor. The computation of its values follows Equation (\ref{eq:stress}), wherein the approximation of the stress tensor, $\varepsilon_{\Lambda}$, is obtained by Equation (\ref{eq:sigma}). This approximation replaces the true solution $\mathbf{u}$ with the neural network's prediction $\Lambda(t_i,\mathbf{x}i, \theta)$, articulated as:
\begin{equation}
\varepsilon_{\Lambda} = \frac{1}{2} (\nabla \Lambda(t_i,\mathbf{x}_i;\theta) + (\nabla \Lambda(t_i,\mathbf{x}_i;\theta))^T). \
\end{equation}
Here, $\Lambda(t_i,\mathbf{x}i;\theta)$ denotes the network's approximation of the solution or the prediction for a given sample $(t_i,\mathbf{x}i)$, based on the current network parameters $\theta$.
In summary, the physical loss, $L_\mathcal{P}$, is formulated as:
\begin{equation}
L_{\mathcal{P}} = \frac{1}{N_\mathcal{P}}\sum_{i=0}^{N_\mathcal{P}} \left| \rho \frac{\partial ^2 \Lambda(t_i,\mathbf{x}_i;\theta)}{\partial t^2} - \nabla \cdot \left[\lambda(\mathbf{x}) \text{tr}(\varepsilon_{\Lambda}) \
I + 2\mu(\mathbf{x}) \varepsilon_{\Lambda}\right]\right|^2.
\end{equation}
The calculation involves second-order derivatives with respect to both space and time. However, computing the loss, including the derivatives of the neural network's approximate solution, is facilitated by automatic differentiation. Refer to Algorithm (\ref{lst:loss}) for a practical demonstration of this process using PyTorch.

\subsection{Network Architecture}
The neural network employed to approximate the solution to the specified PDE is an FCN, referred to here as the approximation network. The input to this network includes spatial and temporal dimensions, specifically represented as $(t_i, x_i, y_i)$ for each sample in the training dataset. The network's output is two-dimensional, corresponding to the components of the displacement field components ($u_x, u_y$), thus establishing a mapping $\textbf{u}: \mathbb{R}^3 \rightarrow \mathbb{R}^2$. The architecture comprises an input layer, multiple hidden layers, and a linear output layer, forming a deep neural network.

The exact configuration of hidden layers and neurons varies across experiments, precluding a definitive specification at this juncture. Unless specified otherwise, the hyperbolic tangent (tanh) function was utilized as the activation function $\sigma$, selected for its smoothness as a $C^\infty$ function. Although alternative activation functions, including sigmoid and swish, were evaluated, tanh consistently demonstrated superior performance in convergence and accuracy. This preference is supported by numerous studies on PINNs, which recommend tanh as an optimal activation function \cite{activation_functions_for_PINNs} \cite{old_and_new}.

The trainable parameters of the network, specifically the weights and biases, were initialized using Xavier initialization \cite{Xavier}. Proper initialization is critical to prevent the vanishing or exploding gradient phenomena, which is potentially detrimental to network training. Xavier initialization addresses these issues by ensuring that the weights are centred around zero and that each layer maintains consistent variance. This is achieved by drawing weights from a Gaussian distribution with mean zero and a variance defined as:
\begin{equation}
\frac{2}{f_{in}+f_{out}},
\end{equation}
where $f_{in}$ and $f_{out}$ represent the number of neurons in the preceding and subsequent layers, respectively. Employing Xavier initialization when using the tanh activation function is advantageous for several reasons. The tanh function is susceptible to gradient saturation as it constrains output values between $-1$ and $1$. Xavier initialization addresses this challenge by ensuring uniform variance across layers, thereby minimizing the risk of saturation due to excessively large or small gradients. This uniformity is particularly beneficial during the initial stages of training, preventing the initialization of excessively large weights that could lead to saturation at the limits of the limits of the tanh function. Moreover, Xavier initialization positions the weights around zero, harmonizing with the tanh function's symmetric nature around zero. Empirical research further supports the claim that Xavier initialization paired with the tanh function yields positive results \cite{xavier_and_tanh}.

\subsection{Initial Condition}
\begin{figure}
\begin{center}
\includegraphics[width=1.0\linewidth]{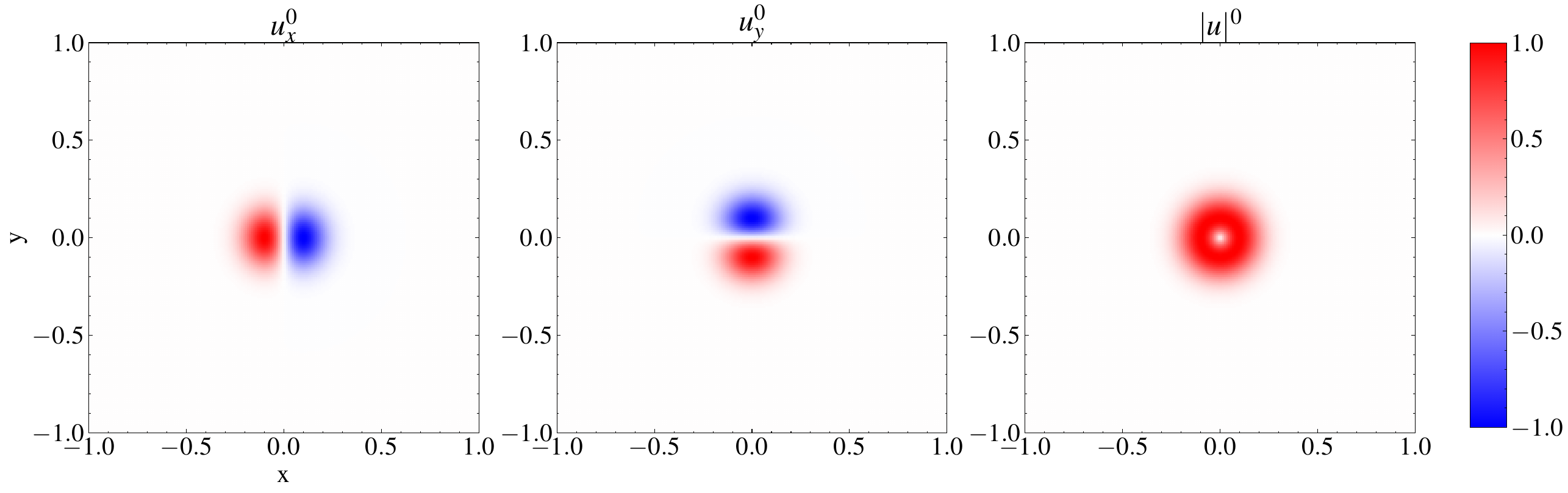}
\caption{Visualization of the initial conditions as the initial displacement field. From left to right: $u_x^0$, the x-component of the initial displacement field; $u_y^0$, the y-component of the initial displacement field; and $|u|^0 =\sqrt{(u_x^0)^2 +(u_y^0)^2}$, the magnitude of the initial displacement field.}
\label{fig:init}
\end{center}
\end{figure}

Initial conditions are required to ensure a unique solution for a PDE. These conditions are articulated as follows:
\begin{equation} \label{eq:init}
\begin{aligned}
\mathbf{u}(0,\mathbf{x}) = g(\mathbf{x}), \
\partial_t\mathbf{u}(0,\mathbf{x}) = 0.
\end{aligned}
\end{equation}
We designed the function $g(\mathbf{x})$ to emulate an explosion, instigating the propagation of elastic waves. Let $f(\mathbf{x}): \mathbb{R}^2 \rightarrow \mathbb{R}^2$ denote the Gaussian function, expressed as
\begin{equation}
f(\mathbf{x}) = e^{-\frac{|\mathbf{x} - \boldsymbol{\mu} |}{2\sigma^2}},
\end{equation}
where $\sigma$ represents the standard deviation, and $\boldsymbol{\mu}$ indicates the explosion's epicenter. The derivative of $f(\mathbf{x})$ can be analytically determined as
\begin{equation}
\frac{d f(\mathbf{x})}{d\mathbf{x}} = -\frac{\mathbf{x} f(\mathbf{x})}{\sigma^2}.
\end{equation}
Accordingly, $g(\mathbf{x})$ is defined by
\begin{equation}
\begin{aligned}
g(\mathbf{x}) = \frac{d f(\mathbf{x})}{d\mathbf{x}} / {\left\Vert \frac{d f(\mathbf{x})}{d\mathbf{x}}\right\Vert_{\text{max}}}.
\end{aligned} \label{eq:initial_condition}
\end{equation}
This formulation yields a normalized depiction of an explosion that catalyzes the propagation of elastic waves.
Unless stated otherwise, we assume a standard deviation of $\sigma = 0.1$, with the explosion's centre located at the origin, $\boldsymbol{\mu} = [0,0]$. Figure (\ref{fig:init}) depicts the initial condition.

\paragraph{Hard Initial Conditions}
\label{section:Hard_Initial_Conditions} In conventional PINN frameworks, initial conditions are incorporated by adding supplementary terms to the loss function, as illustrated in Equation (\ref{eq:Loss}), to enforce these constraints. This method, known as soft-constraining, however, has its limitations. The initial condition is only approximatively enforced, potentially leading to inconsistent solutions, as the exact initial condition is achieved only if the loss converges to zero \cite{FBPINNS}. Moreover, incorporating additional loss terms may complicate the optimization process, as competing loss components can decelerate convergence \cite{SA-PINNs-OG}.

A preferable approach is to embed the neural network within a solution \textit{Ansatz} that strictly enforces the initial condition, referred to as a hard constraint. This methodology was first proposed by Lagaris et al. \cite{ogLagaris}, with Moseley et al. providing a comprehensive examination of its application to the acoustic wave equation \cite{FBPINNS}. We adhere closely to their methodology.

\begin{figure} 
\begin{center}
\includegraphics[width=1.0\linewidth]{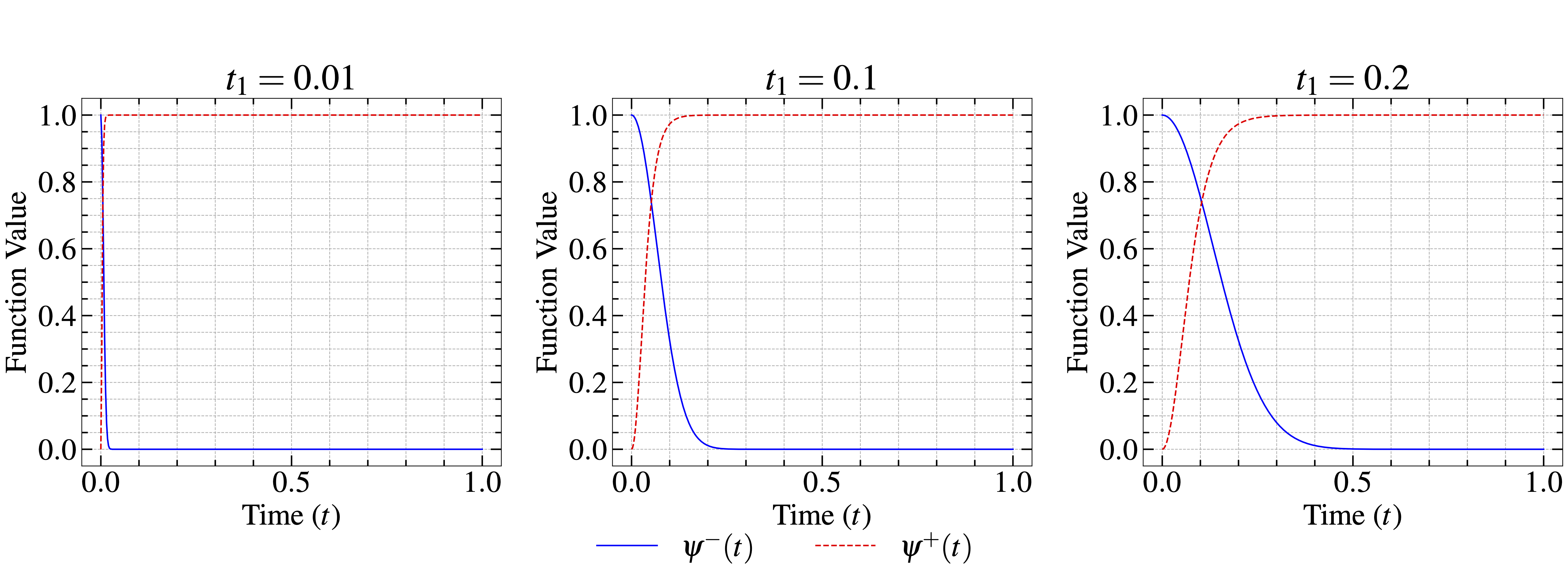}
\caption{Behavioral depiction of $\psi^{-}(t)$ and $\psi^{+}(t)$ across varying $t_1$ values. From left to right: $t_1$ values are 0.01, 0.1, and 0.2. A smaller $t_1$ value results in a sharper transition from enforcing initial conditions to the neural network approximation.}
\label{fig:Ansatz}
\end{center}
\end{figure}

The essence of this \textit{Ansatz} is to guarantee the fulfilment of both conditions in Equation (\ref{eq:init}) inherently. This objective is achieved by redefining the approximate solution as:
\begin{equation}
\hat{\Lambda}(t,\mathbf{x};\theta) = \psi^{-}(t) g(\mathbf{x}) + \psi^{+}(t) \Lambda(t,\mathbf{x};\theta).
\end{equation}
The functions $\psi^{-}(t)$ and $\psi^{+}(t)$ must satisfy specific criteria to render $\hat{\Lambda}(t,\mathbf{x};\theta)$ a suitable solution \textit{Ansatz}:
\[
\begin{aligned}
    & \psi^{-}(0) = 1, \\
    & \psi^{+}(0) = 0, \\
    & \frac{d \psi^{-}(t)}{dt}\bigg|_{t=0} = 0, \\
    & \forall T \gg 0, \psi^{-}(T) = 0.
\end{aligned}
\]
The functions are defined as follows:
\begin{equation}
\begin{aligned}
\psi^{-}(t) = e^{-\frac{1}{2} (\frac{2}{3} t/t_1)^2)}, \
\psi^{+}(t) = \tanh^2(2.5 t/t_1).
\end{aligned}
\end{equation}
These functions satisfy all prerequisite conditions. Figure (\ref{fig:Ansatz}) portrays the behaviour of these functions.

At $t=0$, the network prediction $\Lambda(t,\mathbf{x};\theta)$ is suppressed, aligning the solution $\hat{\Lambda}(t,\mathbf{x};\theta)$ strictly with the initial condition $g(\mathbf{x})$. As time progresses, $\psi^{-}(t)$ diminishes, and $\psi^{+}(t)$ escalates, gradually blending the initial condition with the neural network's solution. This mechanism allows the network to learn the residual, which is the discrepancy between the actual solution and $g(\mathbf{x})$. The parameter $t_1$ governs the rate at which these functions attenuate or amplify. Empirically, we find $t_1 = 0.1$ to be optimal. A too-rapid decline of $\psi^{-}(t)$, associated with a small $t_1$, can slow convergence by introducing an artificial high-frequency change.
Section (\ref{Section:t1}) delves into how the accuracy of the PINN solution is contingent upon the value of $t_1$.

\subsection{Boundary Conditions}
In seismology, the modelling of absorbing boundary conditions is often preferred to simulate an infinite domain, as discussed in Section (\ref{section:ELASTIC_WAVE_EQUATION}). Traditionally, imposing initial and boundary conditions is essential to secure a unique and accurate solution. However, PINNs exhibit a distinctive inherent behaviour that naturally emulates absorbing boundary conditions without the necessity to define boundary conditions in their loss function explicitly.

This characteristic arises from the continuous nature of PINNs. Specifically, the solution from a PINN employing an infinitely continuous activation function, such as the hyperbolic tangent (tanh), will inherently be smooth (i.e., a $C^\infty$ function). This smoothness results from its output being a linear combination of infinitely continuous functions. Given that the composition of two differentiable functions remains differentiable, this principle also applies to the output of a PINN using a differentiable activation function. Therefore, with the output function being smooth, the only conceivable effect at the boundary, in the absence of any other imposed boundary condition, is the absorption of a wavefront. This implies that absorbing boundary conditions are effectively utilized by foregoing the specification of any boundary conditions.

\subsection{Overall Loss}
We streamline the training process by seamlessly integrating boundary and initial conditions, thus negating the need for supplementary loss functions. Additionally, our approach does not incorporate measurement data, which results in excluding the loss term $L_{\mathcal{D}}$. Consequently, the overall loss function is reduced to $L = L_{\mathcal{P}}$, adhering to the conventional PINN loss function framework as specified in Equation (\ref{eq:Loss}). This reduction transforms the optimization challenge from a constrained to an unconstrained format, significantly benefiting the training process by enhancing convergence rates due to the absence of conflicting loss components.

The decision against using measurement data stems from our focus on forward problems, where the necessity for labelled data is inherently minimal. Furthermore, the difficulties of obtaining measurement data in real-world scenarios reinforce our preference for minimizing dependency on labelled data. In contrast to studies that utilize sequential time steps from FDM or SEM simulations to facilitate network learning from the initial state to future states \cite{Ben}, our model strictly adheres to an analytically defined initial condition. This strategy ensures that our network's training is exclusively influenced by the physics-based loss, eliminating the need for additional labelled data.

\subsection{Sampling}
\label{Section:Sampling}
As previously discussed, our model does not require labelled data for training. Nevertheless, to enable the PINN to learn the mapping from input to output space, it is essential to provide input samples ($t_i,x_i,y_i$), known as collocation points. The quantity of these points must be sufficiently large to guarantee accurate approximations, particularly in heterogeneous domains where the wavefield might display rapid, high-frequency variations both spatially and temporally. The Nyquist-Shannon sampling theorem, which stipulates sampling at twice the frequency of the highest frequency present in the true solution \cite{Nyquist}, theoretically dictates the sampling rate. However, predetermining the exact Nyquist frequency is challenging in heterogeneous domains. Our empirical evidence suggests that for a homogeneous domain of our specified size and with the chosen Lam{\'e} parameters, approximately $100,000$ collocation points are sufficient. However, more complex geometries may require an increased number of collocation points. The exact number of collocation points employed in each experiment will be detailed.

\begin{figure} 
\begin{center}
  \includegraphics[width=0.8\linewidth]{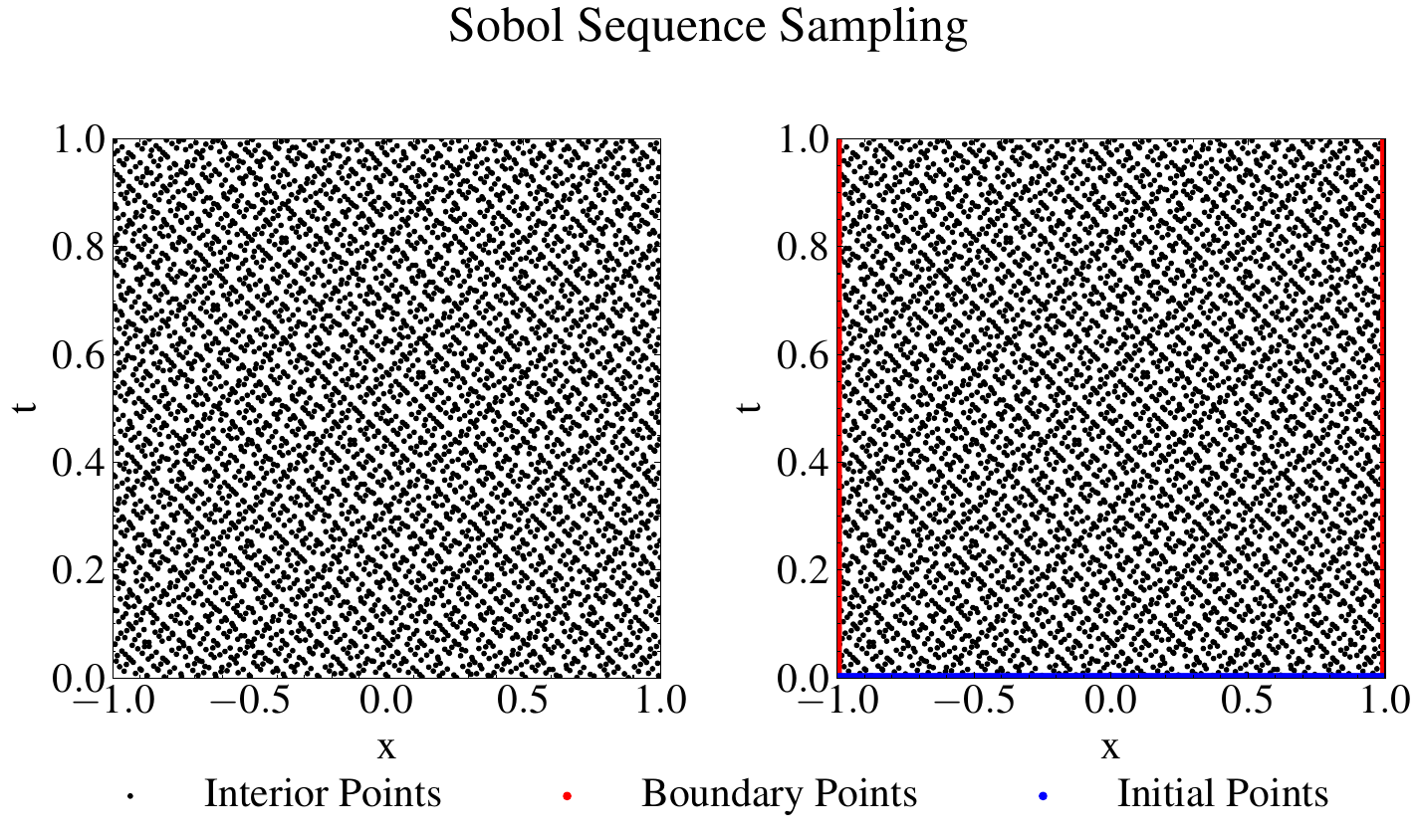}
  \caption{Comparison of sampling strategies in two dimensions (space and time): The left figure illustrates the sampling method we employ, utilizing a Sobol sequence generator for $5000$ points, excluding boundary points in space and time due to hard constraint initial conditions and automatic absorbing boundary conditions. The right figure showcases a traditional sampling approach, which requires boundary points to address additional initial and boundary condition loss terms.}
  \label{fig:sampling}
\end{center}
\end{figure}

For collocation point generation, we opt for the Sobol sequence, a quasi-random number generator, to distribute points within a given domain efficiently. The Sobol sequence is part of the low-discrepancy sequence family, which means that the sequence fills the underlying domain effectively. Mishra et al. have shown that low-discrepancy sequences, compared to simple random point generation, yield superior results \cite{low_discrepancy}. Figure (\ref{fig:sampling}) illustrates the distribution of collocation points generated by a Sobol sequence in a two-dimensional domain.

\subsection{Normalization}
To ensure neural network convergence, input data normalization is crucial, especially since the data can span various scales. In our case, absent any training data constraints, we have the liberty to define our domain. Spatial points are confined within $[-1,1]$, while temporal points are set within $[0,1]$. This approach is tantamount to selecting an arbitrary domain for $x$ and $y$ and then rescaling these to $[-1,1]$, a range that is compatible with the $\tanh$ activation function, whose output values also fall within $[-1,1]$. Despite the common practice of normalizing all inputs to a uniform range, our findings indicate that maintaining $t$ within $[0,1]$ neither hinders convergence nor detracts from accuracy.

\subsection{Training Setup}
This section outlines the configuration of our training regimen. As detailed in Section~(\ref{Optimisation_algorithms_in_PINN_training}), ADAM and LBFGS, either used separately or in combination, are the preferred optimizers for PINNs. In our experiments, we selected LBFGS as the sole optimizer based on a comprehensive hyperparameter exploration. This choice was underpinned by observations that LBFGS outperformed both standalone ADAM and its sequential application with LBFGS in terms of convergence.

We implemented LBFGS with an initial learning rate set to $1.0$. It's important to note that LBFGS inherently adjusts its step sizes during optimization, guided by the geometry of the loss surface and satisfying the Wolfe conditions, rather than relying on predetermined learning rates \cite{LBFGS_paper} \cite{LBFGS_step}. A history size of $100$ was employed, defining the number of previous iterations considered in approximating the inverse Hessian matrix. This specific history size was identified as optimal; increasing it did not further improve convergence efficiency or accuracy, yet it did increase memory requirements. The \textit{strong-wolfe} was selected as our line search algorithm. The epoch count was experiment-specific and will be detailed individually for each experiment. It is essential to note the inability to precisely predict the total number of training steps due to various influencing factors, including $N_{epochs}$, a user-specified value and not a parameter of LBFGS. Furthermore, LBFGS has two built-in parameters, $max_{iter}$ and $max_{eval}$, dictating the maximum number of iterations and function evaluations per optimization step, respectively. Theoretically, the maximum step count is $N_{epochs} \times max_{iter} \times max_{eval}$, serving as an upper estimate. In practice, the actual number of iterations more closely approximates $ 2 \times N_{epochs} \times max_{iter} $ due to internal tolerance change criteria.
Section(\ref{Optimisation_algorithms_in_PINN_training}) also notes that LBFGS necessitates full-batch optimization, which is advantageous for quicker convergence when compared to mini-batch strategies, albeit at the expense of greater GPU memory consumption. Our access to ETH's Euler supercomputer, equipped with high-end GPUs capable of up to 80GB of GPU memory, facilitated the execution of experiments with more than one million collocation points, circumventing the potential limitations of available GPU memory.

\subsection{Workflow and Test Loss Evaluation}
\begin{figure} 
\begin{center}
\includegraphics[width=1.0\linewidth]{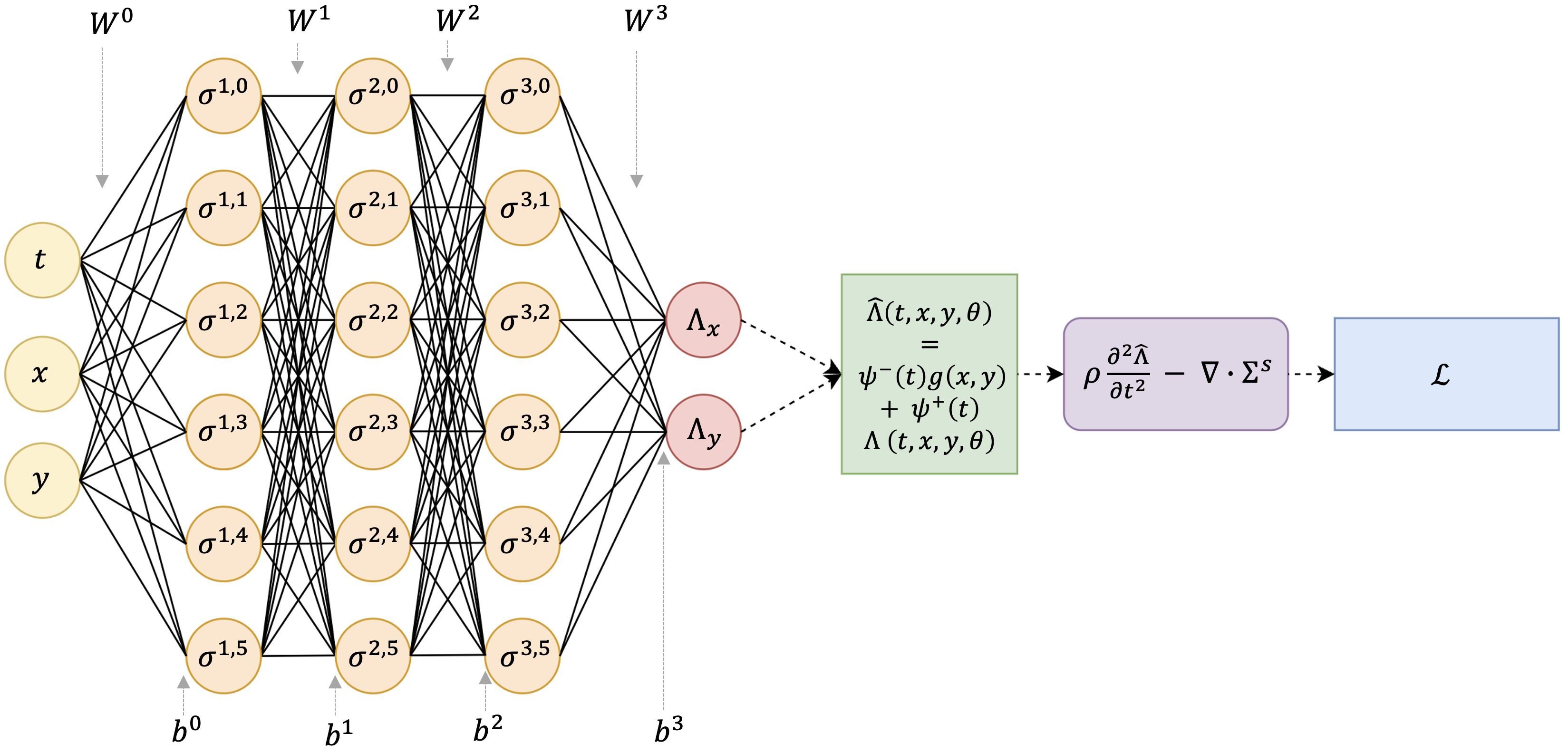}
\caption{Illustration of the workflow of the PINN model implemented in our research. On the left, it shows the approximation network, an FCN, starting with an input layer that transforms the three-dimensional input ($t, x, y$) through the weight matrix $W^0$ and bias vector $b^0$. The input then moves through $N_{layers}$ hidden layers, each with $N_{neurons}$ using the tanh activation function $\sigma$, and concludes at a linear output layer with the weight matrix $W^3$ and bias vector $b^3$, yielding a two-dimensional output ($\Lambda_x, \Lambda_y$). This output advances through the solution \textit{Ansatz} (in green) to determine the final predicted current $\hat{\Lambda}$. Subsequent steps involve computing necessary gradients for the current physics residual (in purple), leading to the current physics loss (in blue). Note that the backpropagation process is not illustrated.}
\label{fig:final_nn}
  \end{center}
\end{figure}

The preceding section detailed the components required to assemble the final PINN model, with Figure (\ref{fig:final_nn}) illustrating the comprehensive, streamlined workflow. Establishing a test loss metric is vital for evaluating the predictive accuracy of a PINN model, as relying solely on the training loss is insufficient. To this end, we employed DEVITO to generate a second-order in time and fourth-order in space FDM simulation of elastic wave propagation. This simulation serves as a benchmark for calculating the test loss at $100$ uniformly distributed intervals between $t=0$ and $t=1$. At each interval, the relative $L_2$ error is determined using the equation:

\begin{equation}
\text{Relative $L_2$ Error} = \frac{|\mathbf{u_{FD}} - \mathbf{\hat{\Lambda}}|_2}{|\mathbf{u_{FD}}|_2}.
\end{equation}

In this formula, $|\cdot|_2$ denotes the $L_2$ norm, $\mathbf{u_{FD}}$ represents the reference solution generated by DEVITO, and $\mathbf{\hat{\Lambda}}$ signifies the approximate PINN solution. The relative $L_2$ error is computed at each of the 100 time steps, and the average of these errors is taken. Unless explicitly stated otherwise, this average relative $L_2$ error constitutes our primary metric for assessing the accuracy of a PINN model.

\section{Results}
This section presents the outcomes of our baseline PINN in accurately solving the elastic wave equation with constant Lam{\'e} parameters. It encompasses an examination of the PINN model's sensitivity to particular hyperparameters and different source sizes, alongside challenges associated with heterogeneous Lam{\'e} parameter scenarios. Additionally, we will address efficiency considerations.
\subsection{Constant Parameters}
Our initial experiments were conducted within a simplified framework, employing constant Lam{\'e} coefficients with $\lambda$ set to $20$ and $\mu$ set to $30$, alongside a fixed density, $\rho$, of $100$. While these values may initially appear arbitrary, they were chosen to produce reasonable wave velocities for the purpose of our simulation. In comparison, the Lam{\'e} parameters of rocks within the Earth's interior typically vary from approximately $20$ to $140$ GPa, and the density averages around $1000 kg/m^3$ \cite{typical_lame}. However, it is essential to note that seismic simulations often operate on a much larger scale.

Calculating the primary (P-wave) and secondary (S-wave) velocities from the Lam{\'e} parameter and the density is straightforward, as illustrated in Equation (\ref{eq:lame}). We determine that $V_p = 0.89$ and $V_s = 0.56$ for the given parameters. Within our specified spatial domain of $x,y \in [-1,1]$ and temporal domain of $t \in [0,1]$, these velocity values are sensible. They roughly ensure that the primary wavefronts reach the spatial boundary by the end of the simulation, assuming the seismic source is centrally located.

Although the values we used are not representative of real-world scenarios, adjusting them to reflect actual geological conditions through proper rescaling can easily be done.

\begin{figure}
\begin{center}
  \includegraphics[width=0.65\linewidth]{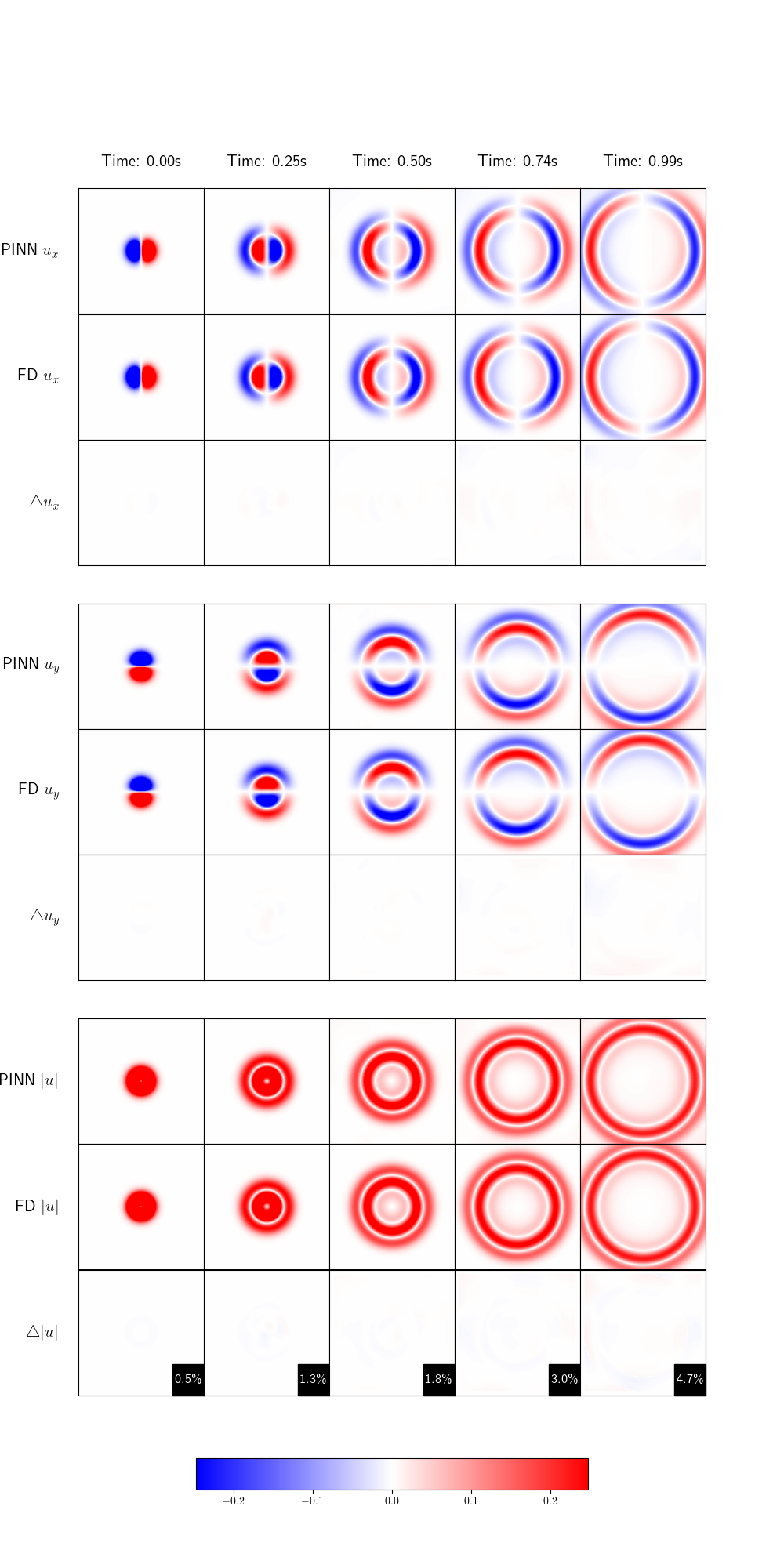}
  \caption{Comparison of elastic wavefield predictions from our baseline PINN model, featuring optimal hyperparameter configurations, to the FDM solution computed using DEVITO for constant underlying Lam{\'e} parameters.}
  \label{fig:vanilla_constant}
  \end{center}
\end{figure}
The comparison between the PINN model results and the FDM solution produced by Devito is illustrated in Figure (\ref{fig:vanilla_constant}). This comparison showcases the optimal performance in terms of average relative $L_2$ error, achieved after an exhaustive series of tests and hyperparameter optimization. For this purpose, a neural network architecture with five hidden layers, each layer consisting of $128$ neurons, was utilized. The training process involved $100000$ collocation points and was conducted over $200$ epochs, setting both $max_{iter}$ and $max_{eval}$ to $100$. With the $t_1$ value fixed at $0.1$, we achieved an average relative $L_2$ error of $2.22\%$, acknowledging some variance potentially due to discretization errors in the FDM solution
\footnote{This variance does not reflect the $L_2$ error variance observed across different simulation time steps but originates from the inherent uncertainty when comparing our PINN model to a non-analytical reference model. The primary source of the observed variance is the discretization error within the FDM simulation, which decreases as the number of discretization points, or resolution, increases. Moreover, by examining the discrepancy between the PINN and FD solutions at the initial time $t=0$---where the strictly enforced initial conditions should theoretically align for both methods---we note a non-zero difference. This discrepancy arises because including second-order derivatives requires using a second-order difference scheme in time, necessitating initial conditions for both the first and second-time steps. Such a requirement can slightly alter the velocity scale, though this effect can be mitigated by selecting an appropriate timestep for DEVITO, aligning the temporal comparison between DEVITO and the PINN. Nevertheless, achieving a precise comparison remains challenging due to DEVITO's intrinsic characteristics. Therefore, the solution provided by DEVITO should be considered a close approximation rather than an absolute truth. It is plausible that the actual $L_2$ error of the PINN is somewhat overstated and could be lower when compared to an analytical benchmark. However, in the absence of such a benchmark, accurately estimating the error margin is challenging.}.

This result, while promising, underscores the potential for further enhancement. A detailed analysis of the error distribution at various time steps reveals an expected trend: the relative error tends to increase as the simulation progresses. This phenomenon, attributable to the cumulative nature of errors in forward-time solutions, becomes particularly evident at later simulation stages. Notably, as the primary wavefront nears or intersects with the boundary, an escalation in error is observed in proximity to the physical boundary. This issue stems from the fact that training points outside the given domain are not considered in the training process, which makes it challenging to constrain the solution outside the modelling domain. Such a scenario highlights one limitation of employing an automatically applied absorbing boundary condition.

Introducing an additional loss term to define boundary conditions could potentially mitigate ambiguity regarding the solution's behaviour beyond the domain. However, this approach introduces its own set of challenges, including the risk of competing loss terms that may hinder convergence.

Moreover, the error is predominantly most pronounced at the location of the primary wavefront, where the solution undergoes the most significant temporal and spatial variations---a consistent observation across all conducted experiments.

Figure (\ref{fig:vanilla_constant}) further illustrates an anomalously higher error in the domain's top-left corner compared to other areas, a discrepancy that eludes straightforward explanation. Although PINNs offer more transparency than purely data-driven models, forecasting the optimization process during training remains a complex challenge. Another noted source of error involves the sharp delineation between wavefronts of differing polarizations. The PINN model exhibits minor difficulties in replicating the exact straight-line trajectory characteristic of the true solution, though this discrepancy is subtle and not easily perceived.

\paragraph{Hyperparameter Sensitivity}
As mentioned earlier, developing this solution entailed a rigorous process of hyperparameter exploration. Here, we aim to briefly highlight some of these hyperparameters and demonstrate their impact on the convergence of PINNs.

\begin{figure}
\begin{subfigure}{.45\textwidth}
\centering
\includegraphics[width=\linewidth]{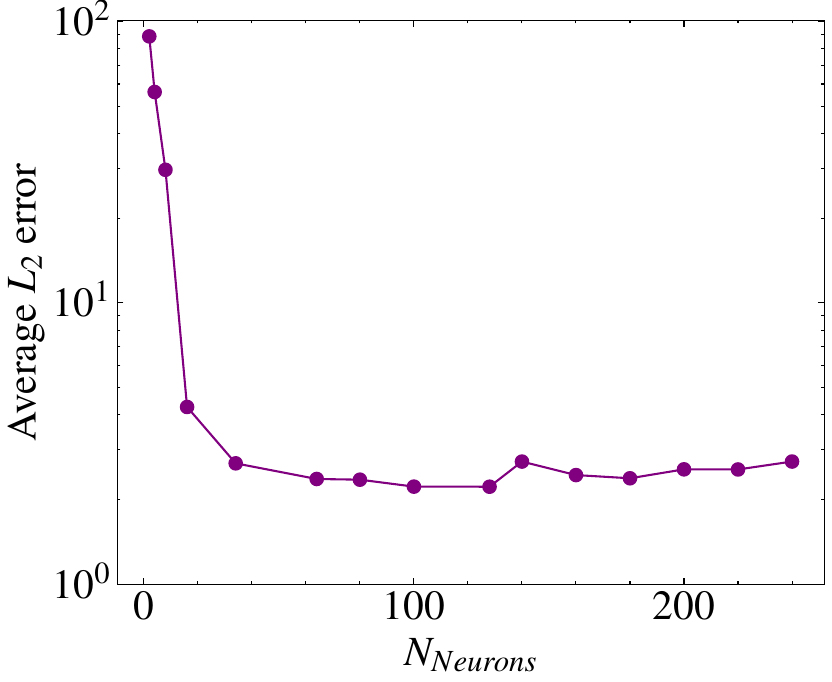}
\caption{Average relative $L_2$ error as the number of neurons per hidden layers increases, with the number of hidden layers ($N_{Layers}$) fixed at five.}
\label{fig:neurons}
\end{subfigure} \hfill
\begin{subfigure}{.45\textwidth}
\centering
\includegraphics[width=\linewidth]{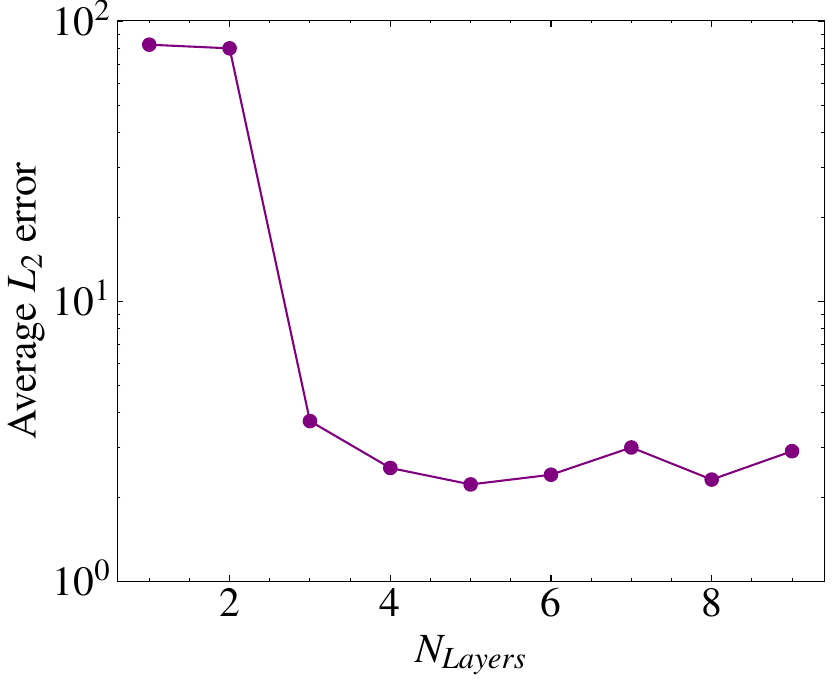}
\caption{Average relative $L_2$ error as the number of hidden layers increases, with the number of neurons per layer ($N_{Neurons}$) fixed at $128$.}
\label{fig:layers}
\end{subfigure}
\caption{Illustration of how the average relative $L_2$ error varies with changes in the number of neurons and hidden layers within our approximation network.}
\label{fig:n_and_l}
\end{figure}
In both experiments described below and illustrated in Figure (\ref{fig:n_and_l}), the network was trained on $100000$ collocation points for $200$ epochs, using the LBFGS optimizer with a starting learning rate of $1.0$ and a maximum number of iterations and evaluations of $100$. The $t_1$ value was kept constant at $0.1$.
The graph in Figure (\ref{fig:neurons}) reveals the relationship between the $L_2$ error of the PINN model and the number of neurons per layer, with the number of hidden layers fixed at five. The neurons tested varied from two to $240$, constrained by GPU memory limits. As expected, a lower neuron count (below $32$ per layer) led to inadequate convergence. However, beyond this threshold, the $L_2$ error declined until reaching about $128$ neurons, which emerged as the optimal count for minimizing error while efficiently utilizing GPU memory. Increasing the neuron count further did not result in a significant reduction in error.
Figure (\ref{fig:layers}) depicts how PINN accuracy changes with an increase in the number of hidden layers, maintaining a steady count of $128$ neurons per layer. Analogous to the prior experiment, having too few layers (two or fewer) led to unsatisfactory or absent convergence. The most favourable outcomes were observed with five hidden layers, beyond which no further improvements in accuracy were noted.
These observations align with our expectations and corroborate findings from the existing literature, affirming that PINNs typically exhibit superior performance with architectures that are wide (a high number of neurons per layer) yet shallow (a minimal number of hidden layers)\cite{shortwide1}\cite{shortwide2}\cite{shortbutwide3}. Empirical analysis helped us identify a practical hyperparameter range of $N_{Layers} \in [4,6]$ and $N_{Neurons} \in [64,200]$. Subsequent experiments utilized these ranges to fine-tune the model towards optimal performance.

\begin{figure}
\centering
\includegraphics[width=0.8\linewidth]{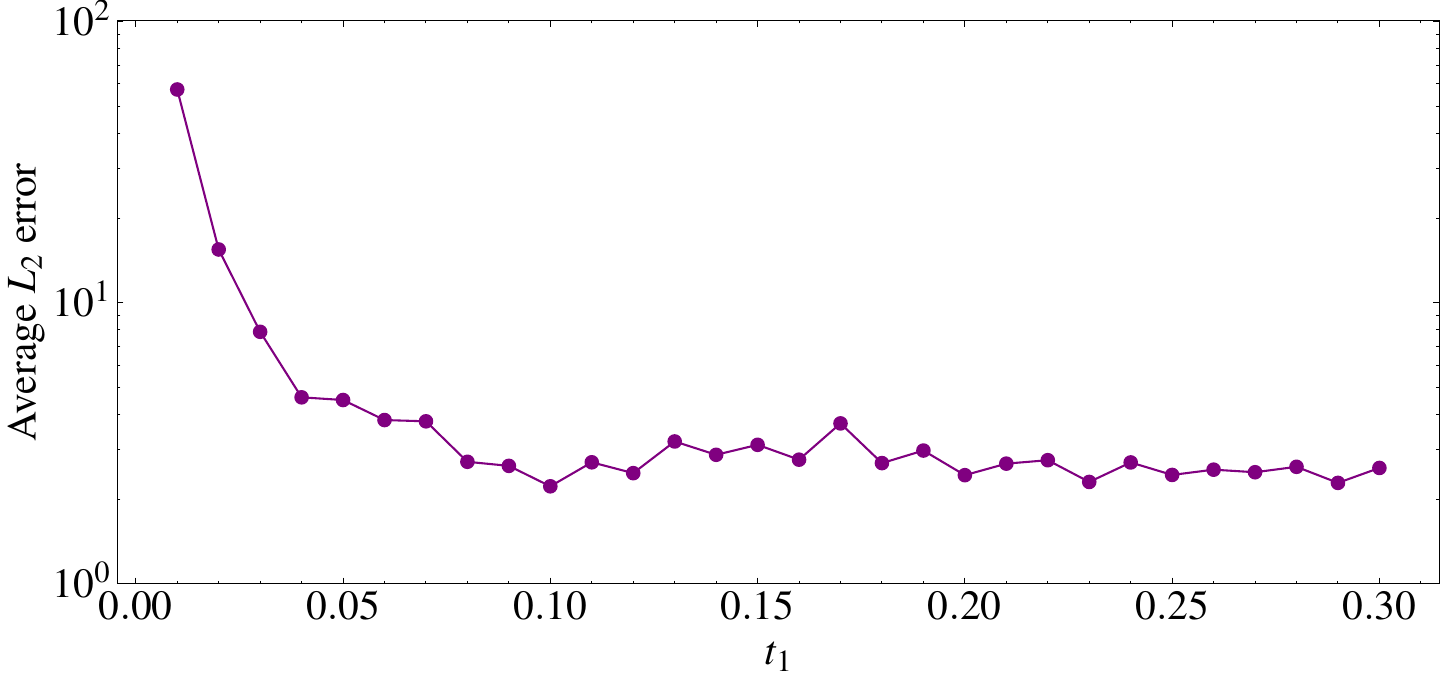}
\caption{Average relative $L_2$ error as a function of increasing $t_1$ value, with $N_{layer}$ and $N_{Neurons}$ maintained at five and 128, respectively, under constant Lam{\'e} parameters.}
\label{fig:t1}
\end{figure}

\paragraph{$t_1$-Sensitivity}
\label{Section:t1}
In Section (\ref{section:Hard_Initial_Conditions}), we introduced a pivotal parameter influencing the convergence of a PINN - the $t_1$ parameter. This parameter governs the rate at which the strict enforcement of initial conditions is relaxed, and the network's approximation begins to take precedence. Its impact on the model's convergence is significant, as illustrated in Figure (\ref{fig:t1}).

The graph shows the average $L_2$ error at a constant number of layers and neurons $(5,128)$ across varying $t_1$ values. It reveals that achieving reasonable convergence necessitates a sufficiently large $t_1$ value. After surpassing a specific threshold (approximately $t_1 \geq 0.7$ in this context), the $L_2$ error stabilizes, though some variance is observed.
It is critical to understand that the $t_1$ value delineates the timeframe over which the initial condition is strictly applied. A diminutive $t_1$ value can lead to the abrupt cessation of the initial condition, introducing artificial high-frequency components that hinder the network's ability to efficiently traverse the loss landscape, culminating in suboptimal convergence. The optimal configuration, under constant parameters, was identified with a $t_1$ value set at $0.1$.

\paragraph{Evaluating Model Performance with Varying Source Sizes}
\label{strong_scaling}
To assess our model's performance under varying complexities while maintaining constant parameters, we adjust the size of the explosive source. This size is defined by the standard deviation $\sigma$ of the associated Gaussian function. By reducing the source size, we aim to understand how our model copes with increased difficulty. This increase in difficulty arises as smaller sources produce solutions with higher frequencies, which generally necessitate a greater number of mesh or collocation points for accurate resolution.
\begin{figure}
\begin{subfigure}{.45\textwidth}
  \centering
  \includegraphics[width=\linewidth]{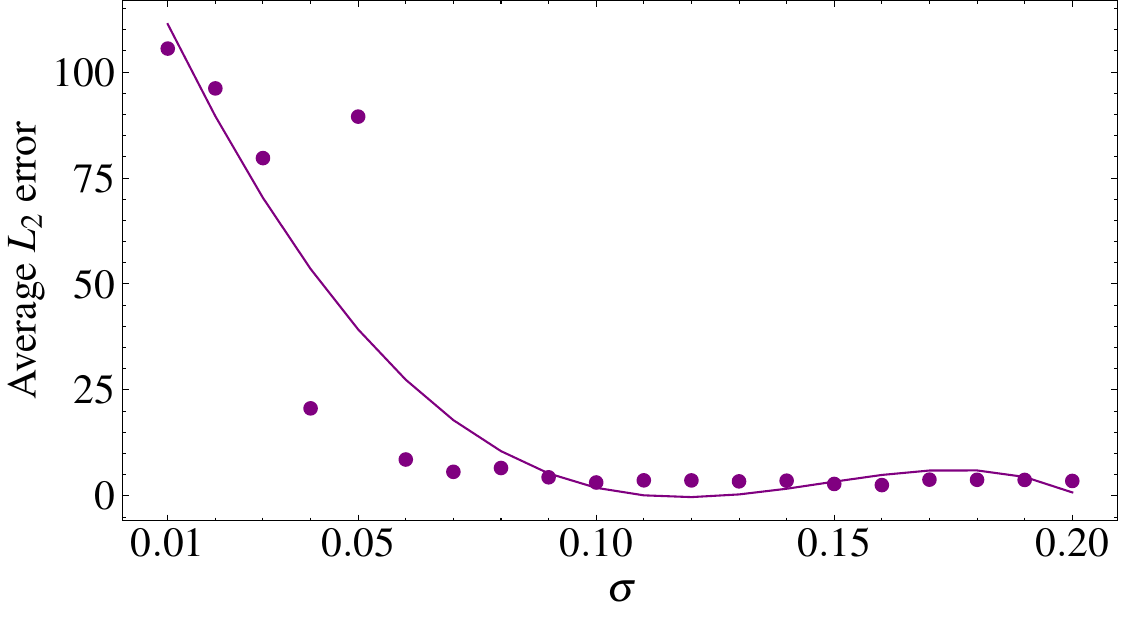}
  \caption{Relative $L_2$ error across full  Range of source sizes $\sigma$}
  \label{fig:strong_scaling_full}
\end{subfigure} \hfill
\begin{subfigure}{.45\textwidth}
  \centering
  \includegraphics[width=\linewidth]{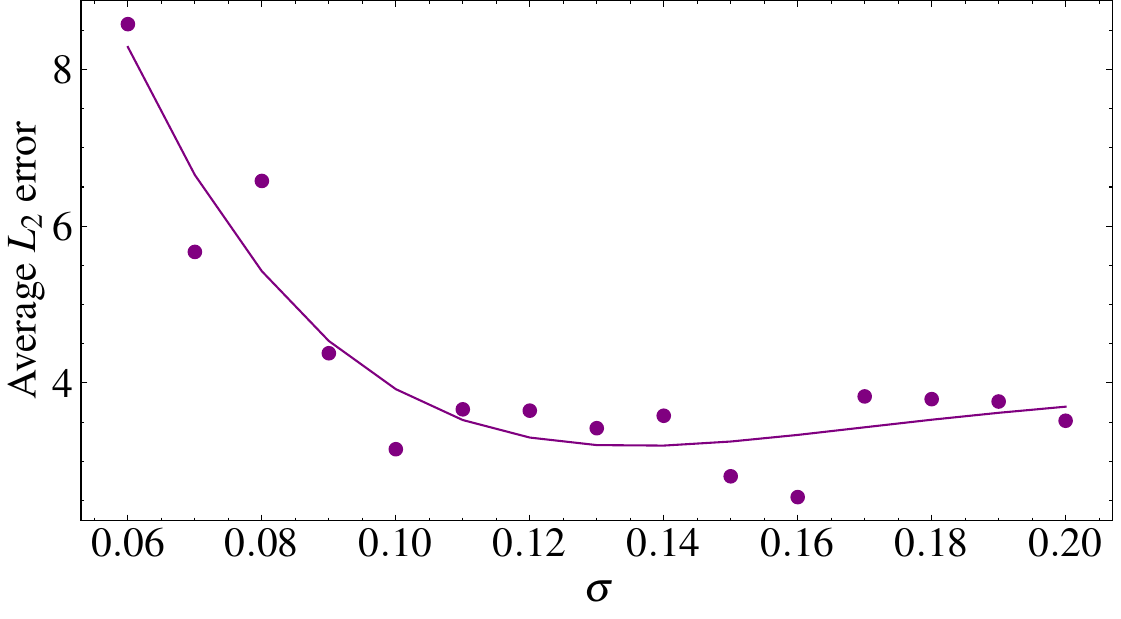}
  \caption{Relative $L_2$ error with Nyquist-compliant source sizes $\sigma$}
  \label{fig:strong_scaling_cut}
\end{subfigure}
\caption{Illustration of the relative $L_2$ error as a function of increasing standard deviation $\sigma$ of the initial source; on the left: the full tested range of $\sigma \in [0.01,0.2]$, and on the right: only source sizes guaranteed to be large enough to comply with the Nyquist-Shannon sampling theorem ($\sigma \in [0.06,0.2]$).}
\label{fig:strong_scaling}
\end{figure}

Figure (\ref{fig:strong_scaling_full}) presents an analysis of the average relative $L_2$ error as the source size increases, with a fixed model comprising five hidden layers, each with $128$ neurons, and a constant collocation point count of $100000$. The training setup adheres to the methodology detailed in Section (\ref{Section:t1}). A smaller source size leads to a tighter wavefront, entailing higher frequency components in the solution, as dictated by the Fourier principle.

According to Figure (\ref{fig:strong_scaling_full}), an overly small source term (specifically, a standard deviation $\sigma \le 0.06$) hampers network convergence. This issue can be traced back to two main factors. The first is known as spectral bias, a typical limitation of PINNs, highlighting that PINNs, like other neural networks, initially approximate low-frequency solution components and struggle with high-frequency and multi-scale features \cite{SB1}\cite{SB2}\cite{SB3}. The second factor relates to violating the Nyquist-Shannon sampling theorem, as elaborated in Section (\ref{Section:Sampling}). We determined that an underlying standard deviation of $\sigma = 0.06$ conservatively guarantees compliance with the Nyquist criterion, using a total of $100000$ collocation points. Detailed derivations are provided in Appendix (\ref{Nyquist_approximation}).

Consequently, in Figure (\ref{fig:strong_scaling_cut}), the analysis excludes the smallest $\sigma$ values below 0.06 for a detailed examination of the $L_2$ error's dependence on $\sigma$. A quasi-quadratic decay is observed up to $\sigma = 0.16$. Surprisingly, beyond this point, increasing $\sigma$ does not further reduce the $L_2$ error, contrary to expectations as the solution simplifies with larger $\sigma$. No definitive physical explanation currently describes the phenomenon observed. Asserting that PINNs face difficulties in learning the lower-frequency components of the elastic wave equation based solely on these results would be speculative, given that our findings are constrained to a single hyperparameter configuration and its effect on source size convergence. Nevertheless, it is important to highlight this anomaly, as it suggests that PINNs have not achieved complete transparency, and a degree of ambiguity in their behaviour persists.

\subsection{Heterogeneous Parameter Models}
\label{heterogeneous_models}
The preceding section provided an extensive discussion of the constant parameter scenario, representing the simplest form of parameter configuration. However, the real-world scenario is inherently heterogeneous, not homogeneous. This heterogeneity stems from the Earth's crust being composed of a diverse array of materials arranged non-uniformly. To address this complexity, we introduce two additional parameter settings: a mixture model and a layered model, both evaluated using our PINN framework.

\begin{figure}
\begin{subfigure}{.5\textwidth}
\centering
\includegraphics[width=\linewidth]{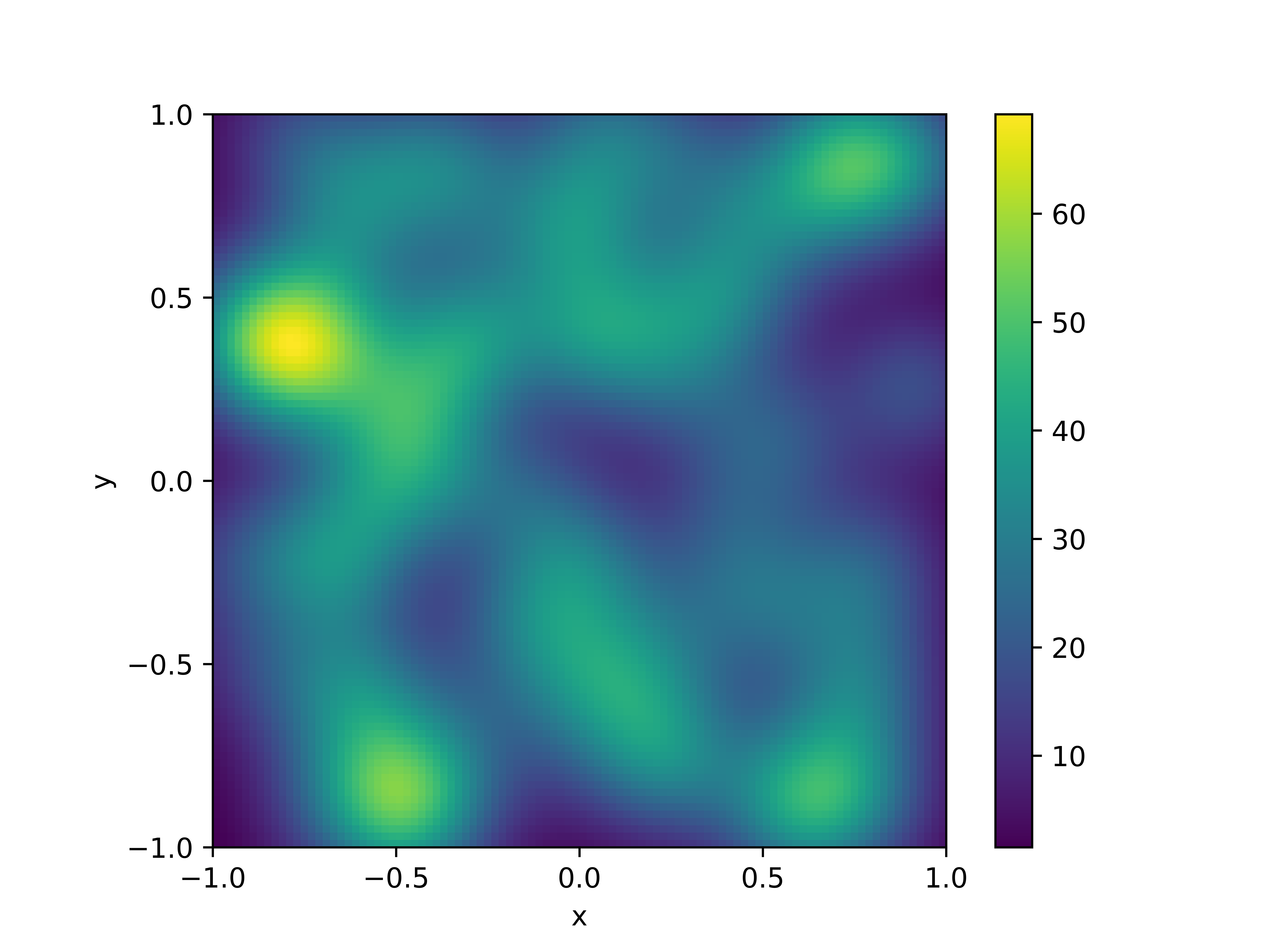}
\caption{Mixture model}
\label{fig:mixture_model}
\end{subfigure}%
\begin{subfigure}{.5\textwidth}
\centering
\includegraphics[width=\linewidth]{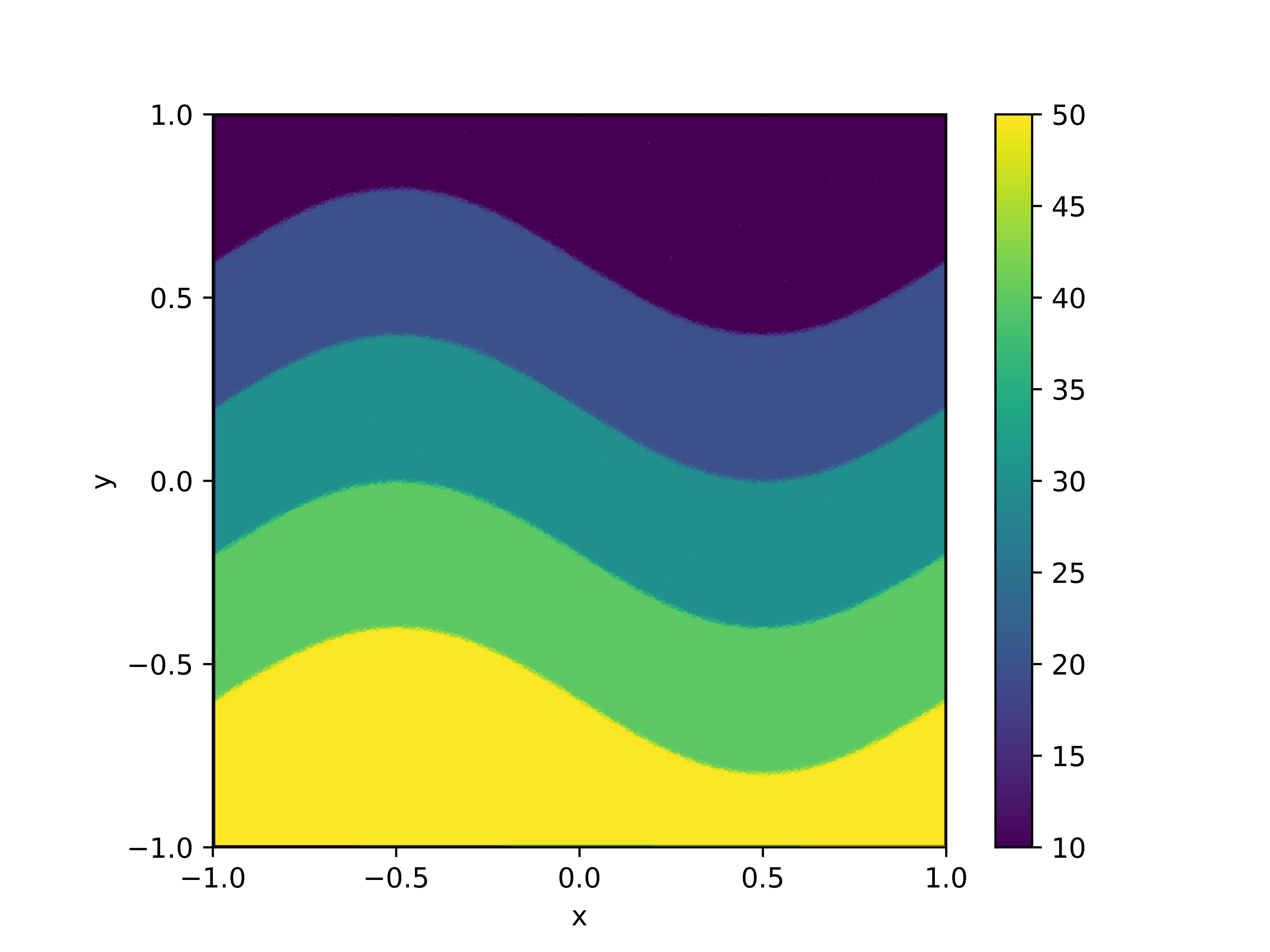}
\caption{Layered model}
\label{fig:layered_model}
\end{subfigure}
\caption{Two heterogeneous parameter models for the Lam{\'e} parameters $\lambda$ and $\mu$. In each model, $\lambda = \mu$ at every point, while the density $\rho$ is consistently set to $100$ across all experiments.}
\label{fig:heterogeneous_models}
\end{figure}

\paragraph{Mixture Model}
\label{section:mixture_model_setup}
The mixture model, characterized by its high degree of heterogeneity, is constructed using a Gaussian mixture model. This approach involves the superposition of $100$ Gaussian distributions, each with varying frequency and amplitude. Figure (\ref{fig:mixture_model}) illustrates the spatial distribution of the first and second Lam{\'e} parameters $(\lambda,\mu)$, as generated by the mixture model. Although the condition $\lambda = \mu$ is consistent across the spatial domain, this does not undermine the model's heterogeneity. This is because the calculated $P$-wave and $S$-wave velocities ($V_P$ and $V_S$, respectively) do not coincide, as derived in equation (\ref{eq:lame}), despite a constant density $\rho$ of $100$ throughout the spatial domain.

\begin{figure}
\begin{subfigure}{.5\textwidth}
  \centering
  \includegraphics[width=\linewidth]{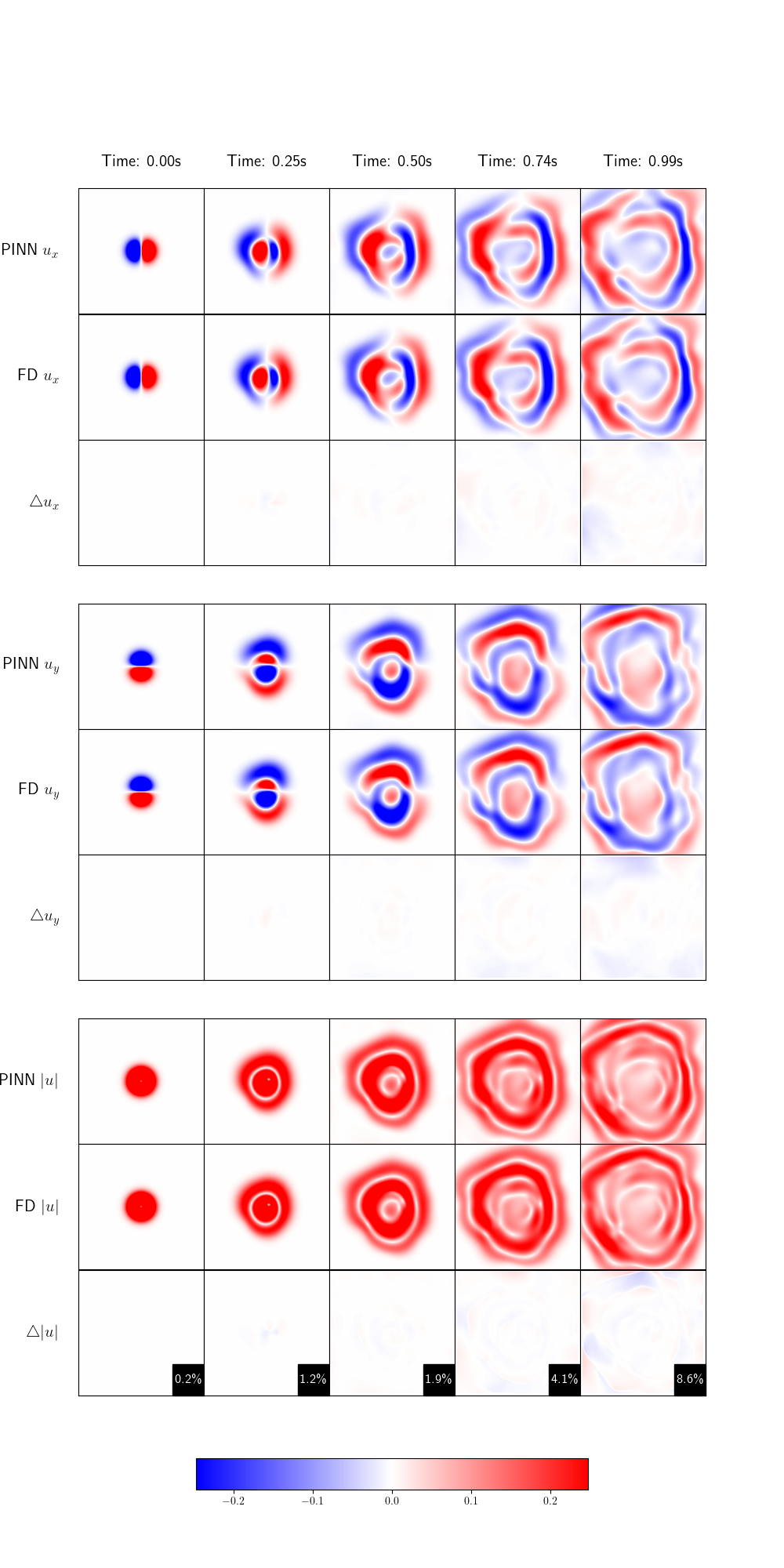}
  \caption{Mixture model}
  \label{fig:vanilla_mixture}
\end{subfigure}%
\begin{subfigure}{.5\textwidth}
  \centering
  \includegraphics[width=\linewidth]{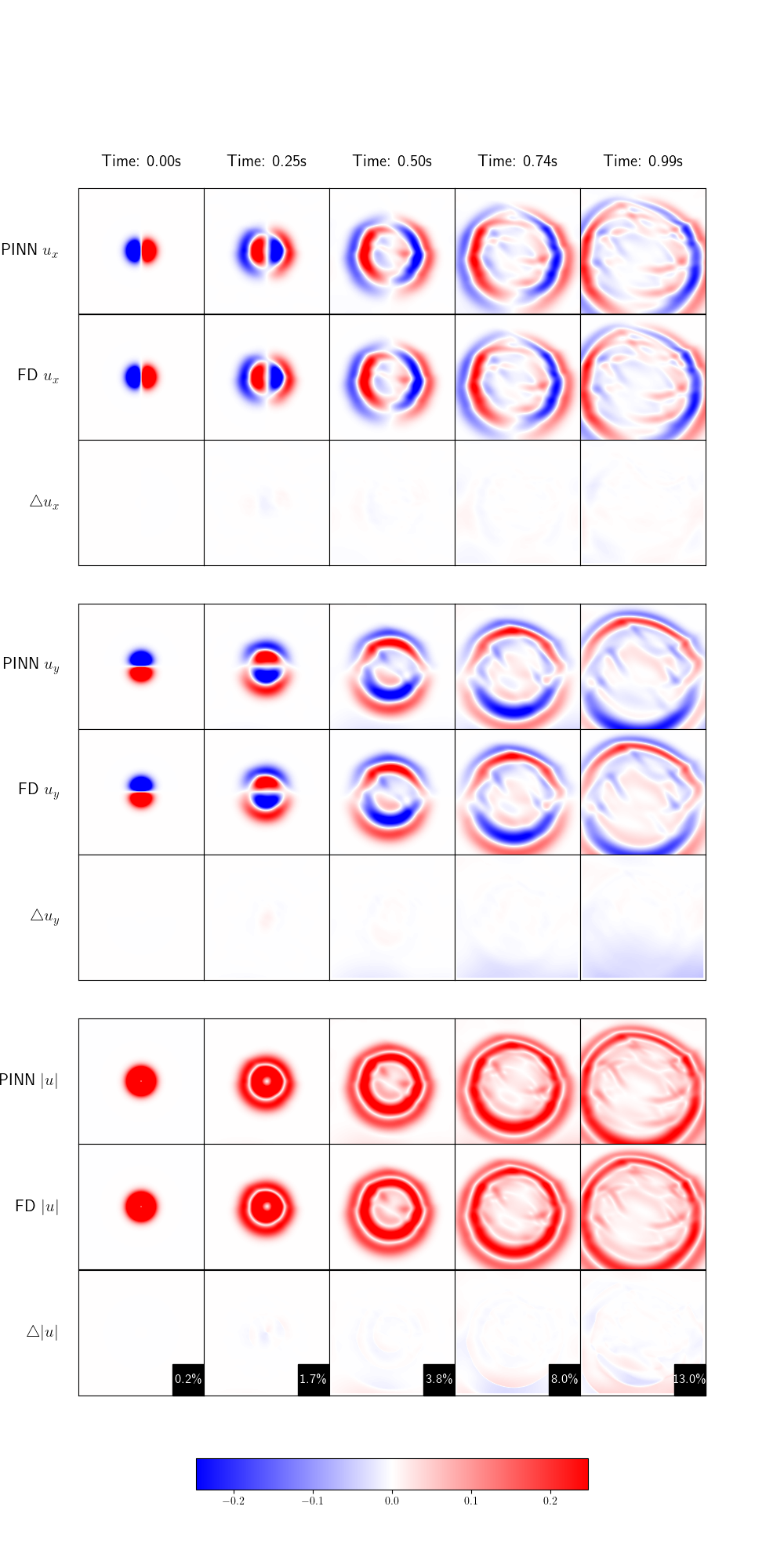}
  \caption{Layered model}
  \label{fig:vanilla_layered}

\end{subfigure}
\caption{Illustration of the outcomes achieved using optimal hyperparameter configurations for both the mixture model and the layered model. The initial trio of rows in each figure presents the PINN predictions for the x-component of the displacement field ($u_x$), compared to the FDM solution generated with Devito. The subsequent group of rows depict the $u_y$ displacement field, while the final segment showcases the magnitude of displacement, denoted as $|u|$.}
\end{figure}

Figure (\ref{fig:vanilla_mixture}) depicts the PINN's performance. The optimal hyperparameters included a learning rate of $2.0$, six hidden layers with 128 neurons each, and a $t_1$ value of 0.2, alongside $100000$ collocation points, trained for $200$ epochs, with $100$ maximum evaluations and iterations. The resultant average relative $L_2$ error was 2.88\%. This error margin, significantly larger than that observed in the constant parameter case (Figure (\ref{fig:vanilla_constant})), underscores the increase in difficulty due to the heterogeneity of the Lam{\'e} parameters. As time progresses, both the error and the solution complexity increase, with high error regions correlating with sharp solution features, dramatic changes in the wavefront, and proximity to boundaries.

\paragraph{Layered Model}
\label{section:layered_model_setup}
The layered model divides the domain into five sinusoidally shaped layers, chosen for their added complexity and non-uniformity over straight layers. This configuration aims to mimic the stratified structure of the Earth's interior, composed of various material layers. Figure (\ref{fig:layered_model}) displays the spatial distribution of the Lam{\'e} parameters $(\lambda,\mu)$ within the layered model, maintaining $\lambda = \mu$ across the domain and a constant density $\rho$ of $100$, mirroring the argument for heterogeneity presented in the mixture model Section (\ref{section:mixture_model_setup}). Though less heterogeneous than the mixture model, the layered model introduces sharp changes in Lam{\'e} parameters, potentially causing reflections and refractions.
The accuracy of the PINN in this layered configuration is shown in Figure (\ref{fig:vanilla_layered}), with hyperparameters set to $400$ epochs, $200$ maximum iterations and evaluations, a learning rate of $2.0$, five hidden layers of 128 neurons each, and a $t_1$ value of $0.2$. This parameter setting presented a more significant challenge for the PINN, reflected by an average $L_2$ error of 5.02\%. Notably, at $t=0.99$, the error escalated to 13\%. While the PINN successfully captures the primary reflections and refractions, its accuracy is compromised, particularly in approximating the low-frequency aspects of the wavefront in the final stage of the $u_x$ displacement field. This highlights the spectral bias inherent in PINNs, especially evident in layered settings where the model struggles with higher frequency and multi-scale components.

\begin{figure}
\centering
\includegraphics[width=0.6\linewidth]{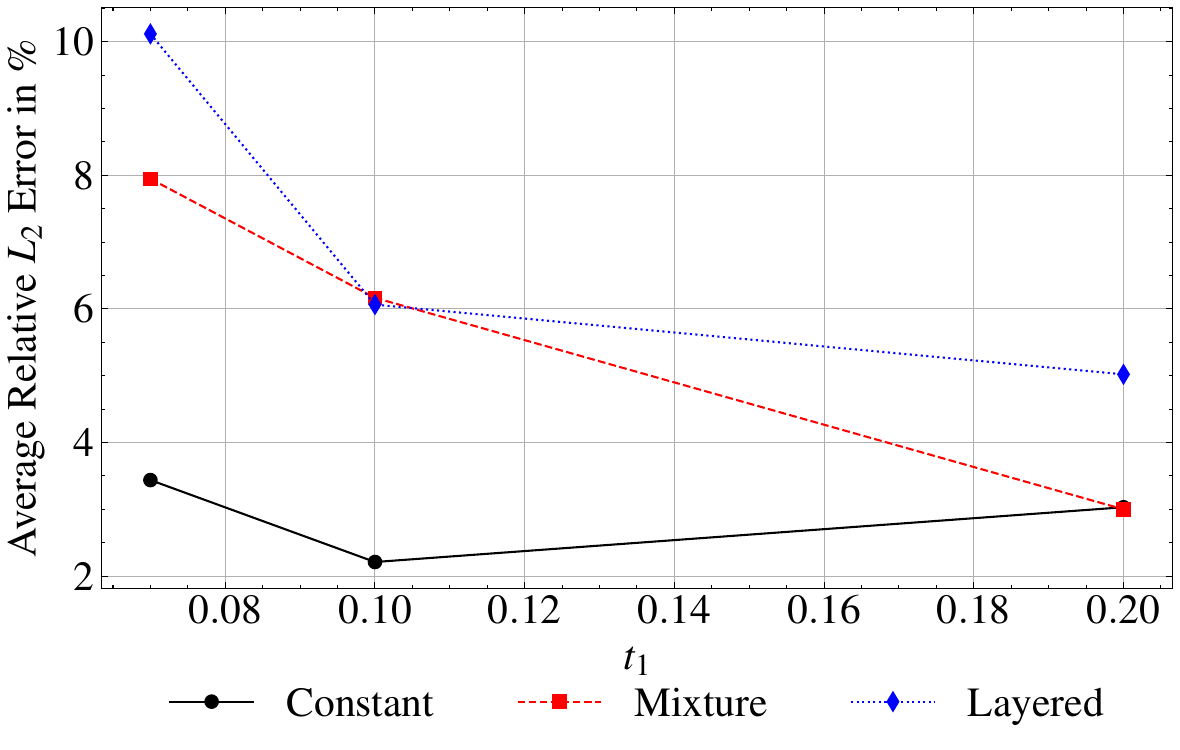}
\caption{Illustration of the PINNs sensitivity on the $t_1$ parameter across multiple underlying Lam{\'e} parameter models.}
\label{fig:t1_sensitivity_cross}
\end{figure}

\paragraph{$t_1$ Sensitivity Across Multiple Parameter Settings}
\label{section:t1_vanilla}
In Section (\ref{Section:t1}), the importance of selecting an apt $t_1$ parameter to facilitate optimal convergence in scenarios with constant parameter settings was underscored. This section delves into the variation of this relationship across diverse parameter settings.
Figure (\ref{fig:t1_sensitivity_cross}) elucidates the relative $L_2$ error for three distinct $t_1$ values, namely $t_1 \in [0.07,0.1,0.2]$, across various parameter settings. These specific values were selected to represent pivotal points of interest: a minimal $t_1$ value of $0.07$, ensuring convergence in the constant parameter scenario; an optimal $t_1$ value of $0.1$ for the same scenario; and a comparatively larger $t_1$ value of $0.2$, which showed to be adequate in more complex settings.

The graphical representation in Figure (\ref{fig:t1_sensitivity_cross}) demonstrates that $t_1$'s influence is inconsistent across the three experimental setups. For the constant parameter scenario, the optimal value is established at $t_1 = 0.1$. In contrast, the two more complex scenarios exhibit enhanced performance with a higher $t_1$ value ($t_1 = 0.2$), each presenting unique variations in their sensitivity to changes in $t_1$.
This observation underscores the absence of a universally optimal $t_1$ value; instead, it necessitates tailored determination for each specific experiment. Notably, even minor adjustments in $t_1$ values can lead to substantial disparities in $L_2$ error rates, emphasizing the necessity for a thorough hyperparameter optimization process for PINNs.

\subsection{Efficiency}
%CHECK WITH BEN
The prior section elucidated the accuracy metrics of our baseline PINN model under diverse scenarios. This section aims to assess the efficiency of PINNs compared to traditional numerical methods.

\begin{table}[ht]
\centering
\begin{tabular}{lcccc}
\toprule
 & FD-512 & PINN Single & PINN Full & PINN Training \\
\midrule
Constant & 20.5s & 1.91s & 219s & 4h 31m 11s \\
Mixture & 20.5s & 1.91s & 219s & 4h 35m 13s \\
Layered & 20.5s & 1.91s & 219s & 10h 51m 4s \\
\bottomrule
\end{tabular}
\caption{Comparison of inference and training times for PINN models against the execution time of the FDM solution across various Lam{\'e} parameter models. Inference for PINNs and the execution of the FDM were carried out on a single CPU, whereas the training of PINNs was conducted on a single NVIDIA TITAN RTX GPU.}
\label{tab:execution_times}
\end{table}

Table (\ref{tab:execution_times}) showcases the inference durations of PINNs alongside the execution times of our FDM reference solution, in addition to the training durations for PINNs. Given the uniform architecture of the approximation network employed (comprising five layers, each with $128$ neurons), the inference durations remain constant across different Lam{\'e} parameter models. This constancy extends to the execution time of our FDM solution, generated utilizing DEVITO, which consistently employs a spatial resolution of $512\times512$ points over $100$ time intervals with a time step of $dt=1e-3$. The predominant advantage of PINNs is observed in scenarios requiring a single inference (PINN-single), such as the evaluation of the final displacement field at $t=1$, which, by virtue of the nature of PINNs, negates the need for computing preceding time steps. Nonetheless, when a high-resolution displacement field is reauired at each of the $100$ time steps (as denoted in the PINN-full column), the efficiency of DEVITOs execution markedly surpasses that of PINNs. A notable challenge is the extensive training duration of PINNs, which spans several hours. It is pertinent to acknowledge that strategies such as early stopping---upon reaching an acceptable test loss during training---or the application of sophisticated training methodologies, such as domain decomposition \cite{FBPINNS}, could significantly ameliorate this duration.

\section{Discussion}
This chapter delves into an extensive study of how PINNS can be applied to solve the elastic wave equation - a critical area of study in seismology.

This research is motivated by the necessity of surmounting the substantial computational demands imposed by conventional numerical methods for simulating elastic wave propagation in large, heterogeneous mediums. 
The research presented in this chapter addresses these challenges by demonstrating the potential of PINNs as a meshfree approach that adeptly captures physical behaviour without the need for labelled data and offers efficient solutions across space-time, thereby addressing the constraints of traditional methodologies.

As highlighted in this chapter, the utilization of PINNs in seismology has demonstrated promising potential. Our experiments have yielded high levels of accuracy, particularly in straightforward scenarios with constant underlying Lam{\'e} parameters. Nonetheless, it is crucial to acknowledge the limitations observed in our results. These constraints are articulated in the subsequent three paragraphs.

\paragraph{Accuracy} Achieving high accuracy in complex scenarios, such as within layered parameter models where solutions include reflections and transmissions at media interfaces or when addressing high-frequency features induced by small source sizes, presents a significant challenge. This difficulty is attributed to the inherent challenges PINNs face with high-frequency components - their spectral bias. Despite PINNs offering an innovative approach and demonstrating commendable performance in certain experimental setups, their accuracy has yet to surpass that of traditional numerical methods.

\paragraph{Hyperparameters} Optimizing PINNs involves a costly and complex hyperparameter tuning process, which is necessary before each experiment to tailor the model to each new parameter setting. Finding the optimal network configuration for specific challenges is both resource-intensive and time-consuming, with outcomes that lack consistency. This complexity presents a significant obstacle to the broader adoption and practical application of PINNs.

\paragraph{Efficiency} The efficiency of PINNs, particularly in comparison to traditional numerical methods, remains a principal concern. For PINNs to be considered practically viable, their training and inference time must be competitive with the simulation times of conventional methods. Despite fast inference capabilities, the training phase, which can last hours, does not meet this criterion. The necessity for retraining PINNs with each new source location and parameter adjustment significantly hampers their adaptability to diverse seismological conditions, limiting their practical use in real-world scenarios.

\section{Conclusion}
We showcase the application of PINNs across multiple parameter settings that gradually increase in difficulty and realism. These settings include a constant setting, a highly heterogeneous mixture setting, and a custom layered setting. The exploration of these different parameter settings reveals the adaptability of PINNs to different complexities inherent in seismological models. Furthermore, we explore the impact of decreasing the source size on the convergence of PINN models. We demonstrate that while PINNs provide relatively low error in simple settings and can learn the primary components of the underlying solution, they still face significant challenges in terms of accuracy, particularly in complex settings like layered models where reflections and refractions are prevalent. This observation underscores the need for further refinement and development of PINNs to enhance their applicability in seismology.
One of the main limitations of PINNs is the need to retrain the model for every new source location, making them practically unviable. In the upcoming sections, our aim is to address these challenges by focusing on improving accuracy and developing a more versatile PINN model that does not require retraining for every new source location. These improvements are crucial for establishing PINNs as a valuable tool in seismology.

\newpage

\chapter{Neural Network Architecture Search}
\section{Introduction}
\label{section:Arch_Motivation}
%Beating the baseline
In Section (\ref{section:Elastic PINNs}), we introduced a standard feed-forward neural network and evaluated its accuracy within the framework of PINNs for modelling the elastic wave equation. This chapter extends our investigation into the network architecture by exploring various designs and their performance implications. Pursuing this line of inquiry is motivated by several factors. Primarily, the accuracy of the standard PINN model frequently falls short of expectations, as illustrated, for instance, in Figure (\ref{fig:vanilla_layered}). Our goal is to develop innovative network designs that not only meet but exceed our established benchmarks, thereby achieving superior accuracy in wavefield simulations across diverse parameter settings. Nonetheless, surpassing the baseline constitutes a secondary objective in our broader research agenda.

%Incorporating wave knowledge into the design
At the heart of our exploration is the question: Can we successfully incorporate prior knowledge about the solution into the network design, and does such integration generally enhance accuracy and convergence? Of course, prior knowledge of the solution depends on the underlying PDE. In the context of the elastic wave equation, we identify several key solution characteristics that inform our network design enhancements. For instance, wavefields typically exhibit oscillatory behaviour and propagate outward from the seismic source in scenarios involving a singular source term. Additionally, seismic waves traversing non-homogeneous media display distinct polarizations, including longitudinally polarized P-waves and transversely polarized S-waves. We also anticipate that the solution will feature varying frequency components due to dispersion---a phenomenon that causes different frequency components of a wave to travel at varying velocities. Moreover, over distances and time, waves undergo attenuation, characterized by a decrease in amplitude and intensity, a consequence of the medium's intrinsic properties and the geometrical spreading of the waves. These considerations represent just a subset of the high-level properties pertinent to (seismic) wave behaviour.

\paragraph{The Spectrum}
A pivotal aspect of our exploration involves determining the optimal extent to which prior knowledge about wave phenomena should influence network design. We conceptualize this as a spectrum of information utilization within neural network architectures. At one end of this spectrum lies the conventional PINN model employing a fully connected architecture characterized by its lack of integration of solution-specific prior knowledge. Conversely, the opposite end of the spectrum is occupied by models meticulously designed to leverage all conceivable prior information, thereby, to the best of our knowledge, closely approximating the underlying solution from the outset. Such models potentially require fewer learnable parameters compared to their less informed counterparts. In an idealized scenario, the network might even incorporate the exact analytical solution as part of its structure or an \textit{Ansatz}, though this approach is impractical in our context due to the absence of a universally applicable analytical solution.
%CHECK WITH BEN: NOW OKAY WHEN TALKING ABOUT RESTRICTION?

The spectrum not only delineates a network's reliance on pre-existing knowledge but also its inherent constraints. All neural networks, PINNs included, are subject to intrinsic limitations due to their inductive bias \cite{inductive_bias}, which is influenced by various architectural decisions, such as the selection of activation functions or the configuration of layers and neurons. Models incorporating physical insights into their architecture, however, impose additional constraints on the solution formulation process. While the term "restriction" might imply a drawback, in this context, it signifies a strategic narrowing of the function space that the network explores. A standard PINN, without additional constraints, can approximate any continuous function across compact subsets of real numbers according to the Universal Approximation Theorem \cite{Universalapproximationtheorem}. This capability requires at least one hidden layer and a sufficient number of neurons. In contrast, a model situated towards the more informed end of our spectrum offers a more focused solution space, albeit with greater specificity than that of a standard PINN model.

This chapter introduces various network designs positioned across this spectrum and evaluates their effectiveness. Our aim is to ascertain whether a discernible correlation exists between a model's placement on this spectrum and its accuracy and convergence rates and to determine if a universally optimal position emerges across diverse problem settings. 

%Summary and outlook
\paragraph{Chapter Overview}
This chapter is organized as follows: Initially, we explore the examination of related work. Subsequently, we outline our main contributions. In the Methods section, the discussion begins with a description of the three primary basis functions employed in constructing our diverse network architectures, followed by an elaboration on four principal design templates. The placement of these models within our predefined spectrum is then discussed for further analysis. The subsequent evaluation includes the performance assessment of all models across the spectrum, tested on the three Lam{\'e} parameter settings as previously delineated in Section (\ref{heterogeneous_models}). This comprehensive analysis leads to the identification of a singular model type that outperforms all other network types. Following this, the best model is compared to the baseline PINN under a wide range of novel experimental conditions, including varied seismic source functions, different source sizes, a higher frequency layered case, and its sensitivity to changes in hyperparameters. This sets the stage for the next chapter, where this optimal model is utilized to solve the conditioned elastic wave equation.

\section{Contributions}
In this chapter, we extend the standard Physics-Informed Neural Network (PINN) approach by developing a broad array of neural network designs---some entirely novel, others significantly inspired by existing work---to assess how and to what extent the incorporation of prior wave physics knowledge enhances accuracy. Instead of evaluating a single new method, we propose a spectrum along which these models are positioned, illustrating the correlation between accuracy/performance and their placement on this spectrum. We rigorously test these models across multiple parameter settings of increasing complexity and a range of hyperparameter configurations to address the question posed. Furthermore, we pinpoint an optimal neural network architecture that outperforms all other architectures, including the baseline PINN model, across various Lam{\'e} parameter settings and source sizes in solving the elastic wave equation. This architecture's superiority is further validated by its successful application to an additional PDE, the acoustic wave equation, thereby strongly affirming its generalizability.

\section{Related Work}
In this section, we delve into the existing literature pertinent to the concepts introduced in this chapter. The term "related" is apt, considering the absence of studies that directly mirror our approach---specifically, the generation of multiple novel networks that integrate prior knowledge of wave physics in diverse manners, particularly within the context of seismology. While other studies have introduced specialized, domain-specific functions to enhance performance, they often do not explicitly aim to embed prior domain knowledge within the network architecture. Instead, they primarily focus on developing a singular, specialized neural network optimized for a distinct task (\cite{WaveletLSTM}, \cite{planewave_super}, \cite{PINNsformer}.
Nonetheless, instances exist where the explicit goal was to incorporate prior domain knowledge into network designs, and these endeavours have been successful. For example, Alkahlifah et al. enhanced PINNs by modelling them as a linear combination of Gabor basis functions, specifically targeting the Helmholtz equation \cite{Gabor}.
Moreover, the basis functions we introduce---sinusoidal functions, wavelets, and plane waves---have been previously employed in conjunction with SciML. For instance, Li et al. developed deep wavelet neural networks (DWNN) by integrating wavelets into deep neural network architectures, achieving refined feature description and extraction across various PDEs, including the shock-wave problem \cite{DWNN}.
Fourier Feature Neural Networks, proposed by Tancik et al., enable FCNs to learn high-frequency functions in low-dimensional problem spaces by transforming input points through a layer composed of sine and cosine functions \cite{Fourier_Features}. Wang et al. expanded on this concept by introducing spatiotemporal and multi-scale random Fourier features, broadening the spectrum of Fourier features available \cite{FF2}.
Cui et al. suggested the use of plane wave activation-based neural networks (PWNN), replacing traditional tanh activation functions with plane waves to address the Helmholtz equation using deep neural networks, markedly enhancing both speed and accuracy \cite{planewaveactivations}.
Tripura et al. proposed the Wavelet Neural Operator (WNO), which combines integral kernels with wavelet transformations. This approach leverages the wavelets' superior time-frequency localization capabilities, enabling precise pattern tracking in spatial domains and efficient learning of functional mappings \cite{WNO}.
Where relevant, these studies will be revisited in the method Section (\ref{Sectoin:Arch_Methods}) to detail how we were inspired by existing research and how our innovative network architectures distinguish themselves from those previously proposed.

\section{Methods}
\label{Sectoin:Arch_Methods}
In this section, we explore a variety of network architectures. A broad range of designs incorporating different underlying basis functions is examined. Specifically, we detail four distinct design templates utilized alongside various basis functions, including sinusoidal functions, wavelets, and plane waves. Additionally, we illustrate their respective positions on the spectrum. Toward the end of this section, we briefly discuss some designs that, despite their theoretical promise, do not yield successful outcomes in practice.

\subsection{Basis Functions}
For clarity, we first define several basis functions and abbreviations that will be employed throughout this chapter. Certain functions inherently satisfy high-level solution properties, including sinusoidal functions, wavelets, plane waves, and Bessel functions. These functions are instrumental in incorporating prior wave knowledge into our network design. The subsequent sections will explore the specific network designs that utilize these basis functions.

\paragraph{Sinusoidal Functions}
The sinusoidal basis function is defined as follows:
\begin{equation}
\gamma(\mathbf{X}) = \sin(2 \pi (\mathbf{X} \mathbf{W} + \mathbf{b})),
\end{equation}
where $\mathbf{X} \in \mathbb{R}^{n \times 3}$ represents the input tensor consisting of $n$, three-dimensional $(t,x,y)$ samples, $\mathbf{W} \in \mathbb{R}^{3 \times d}$ denotes a trainable weight tensor, and $\mathbf{b} \in \mathbb{R}^{d}$ is a trainable bias vector.

\begin{figure}
\centering
  \includegraphics[width=0.5\linewidth]{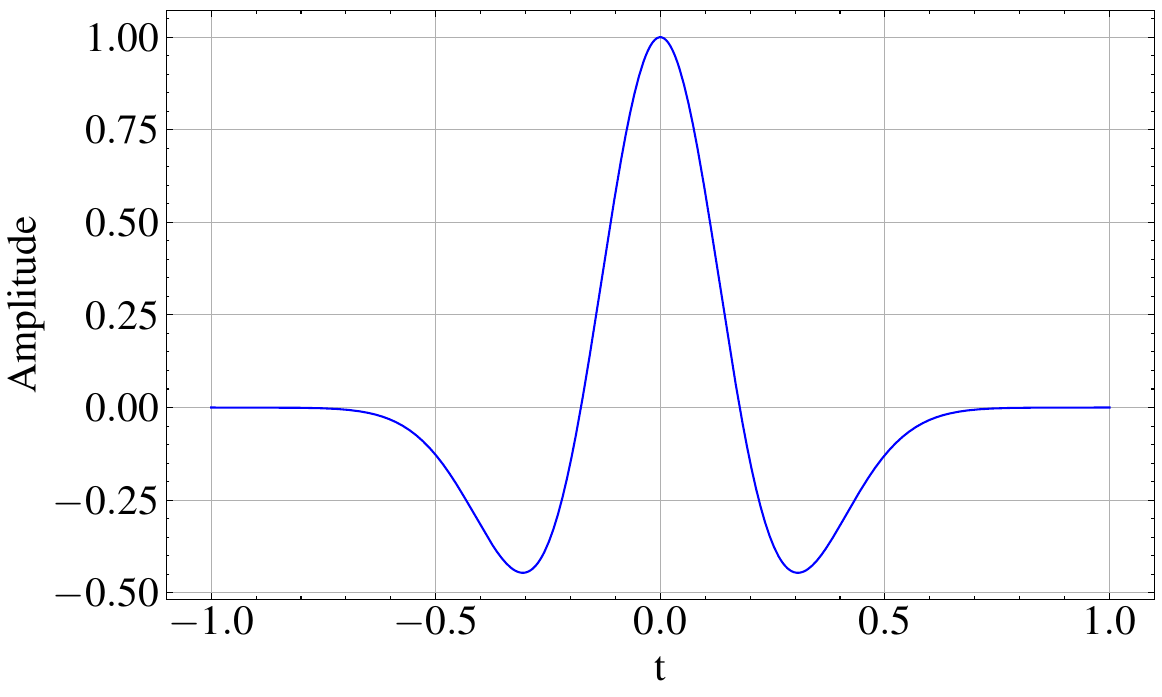}
  \caption{Example of Ricker Wavelet with $a=4$.}
  \label{fig:ricker}
\end{figure}

\paragraph{Wavelets}
The Ricker wavelet is selected as the mother wavelet:
\begin{equation}
\psi(\mathbf{t}) = \left(1 - 2 a^2 \mathbf{t}^2\right) e^{-a^2 \mathbf{t}^2},
\end{equation}
with $a$ set to $4$. A depiction of the Ricker Wavelet is provided in Figure (\ref{fig:ricker}).
To incorporate spatial information, the wavelet is adapted as follows:
\begin{equation}
\tilde{\psi}(\mathbf{t},\mathbf{x},\mathbf{y}) = \psi\left(\boldsymbol{\alpha} \cdot \left(\mathbf{t} - \boldsymbol{\beta} - \mathbf{x} \cdot \mathbf{w}_x - \mathbf{y} \cdot \mathbf{w}_y\right)\right),
\end{equation}
where $\boldsymbol{\alpha} \in \mathbb{R}^d$ signifies a vector of trainable parameters for scaling, $\boldsymbol{\beta} \in \mathbb{R}^d$ for temporal translation, and $\mathbf{w}_x, \mathbf{w}_y \in \mathbb{R}^d$ are trainable weight vectors for spatial dimensions.

\begin{figure}
\centering
  \includegraphics[width=0.5\linewidth]{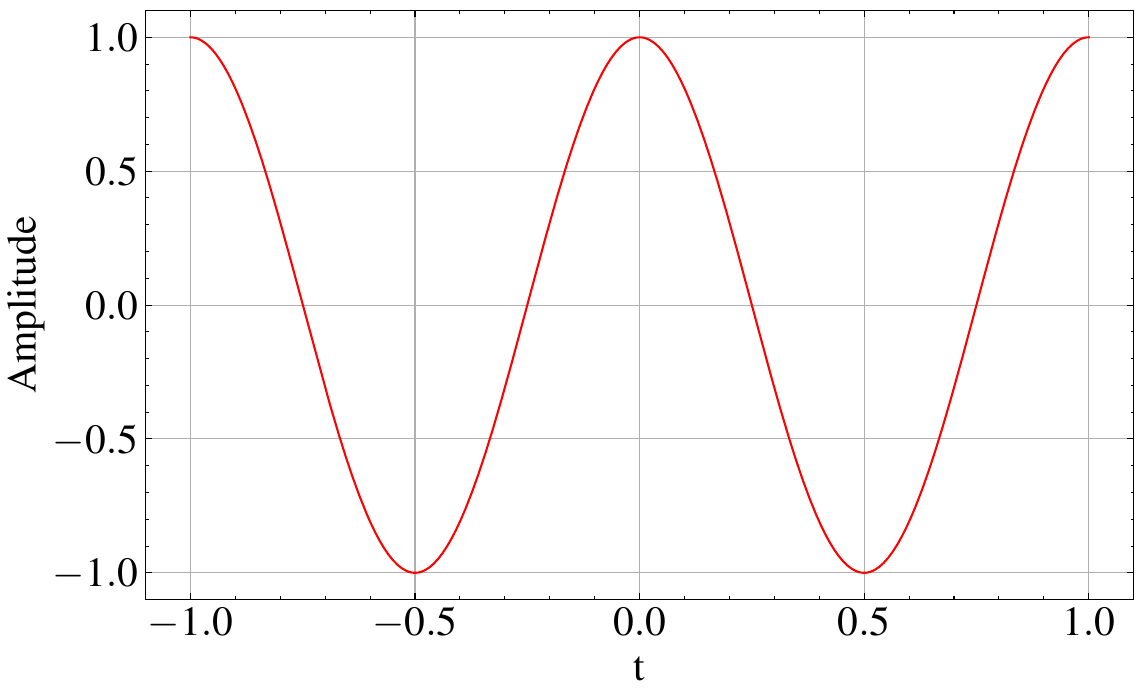}
  \caption{Example of one-dimensional plane wave, parameterized by: $x=0, y=0; A=1, k_x=2, k_y=2, \omega=2\pi, \phi=0$.}
  \label{fig:planewave}
\end{figure}

\paragraph{Plane Waves}
Plane waves are represented as:
\begin{equation}
\Psi(\mathbf{t},\mathbf{x},\mathbf{y})= \text{Re}(\mathbf{A} \cdot e^{i (\mathbf{k_x} \mathbf{x} + \mathbf{k_y} \mathbf{y} - \mathbf{\omega} \mathbf{t} + \mathbf{\phi} \sqrt{\mathbf{x}^2 + \mathbf{y}^2})}).
\end{equation}
Here, $\mathbf{A} \in \mathbb{R}^d$ corresponds to the amplitude, $\mathbf{k_x}, \mathbf{k_y} \in \mathbb{R}^d$ are the wave numbers in the $x$ and $y$ directions, respectively, $\boldsymbol{\omega} \in \mathbb{R}^d$ denotes the angular frequency, and $\mathbf{\Phi} \in \mathbb{R}^d$ represents the phase shift parameter. All five parameters ($\mathbf{A},\mathbf{k_x},\mathbf{k_y},\boldsymbol{\omega},\mathbf{\Phi}$) are potentially trainable. An example of a one-dimensional plane wave is illustrated in Figure (\ref{fig:planewave}).
While both of the above-described basis functions intrinsically satisfy some of the mentioned characteristics of wavefield solutions discussed in Section (\ref{section:Arch_Motivation}), the use of plane waves as a basis function is particularly compelling. This is because a harmonic spherical elastic wave, as encountered in a homogeneous domain, can be solely represented by a superposition of both homogeneous and inhomogeneous plane waves \cite{quantitative_seismology}.

\subsection{Design Templates}
We utilize four design templates in conjunction with the aforementioned basis function to construct neural network architectures. Each template is detailed subsequently. Additionally, we outline the default model, serving as a benchmark for comparing these architectures. Finally, we discuss several designs that, despite not producing satisfactory outcomes, offer valuable perspectives on the architectural spectrum.

\subsection{PINNs- tanh}
The default model, extensively reviewed in Section (\ref{section:Elastic PINNs}), epitomizes the foundational approach among the spectrum of models examined in this chapter. Positioned at the spectrum's extreme left, this model is characterized by its independence from prior wave knowledge and the absence of constraints. It thus serves as a baseline, enabling subsequent models to diverge by integrating increased complexity or domain-specific insights.

\subsection{Input Transformation PINNs}
\label{input_transforrmation}
\begin{figure}
\centering
\includegraphics[width=.8\linewidth]{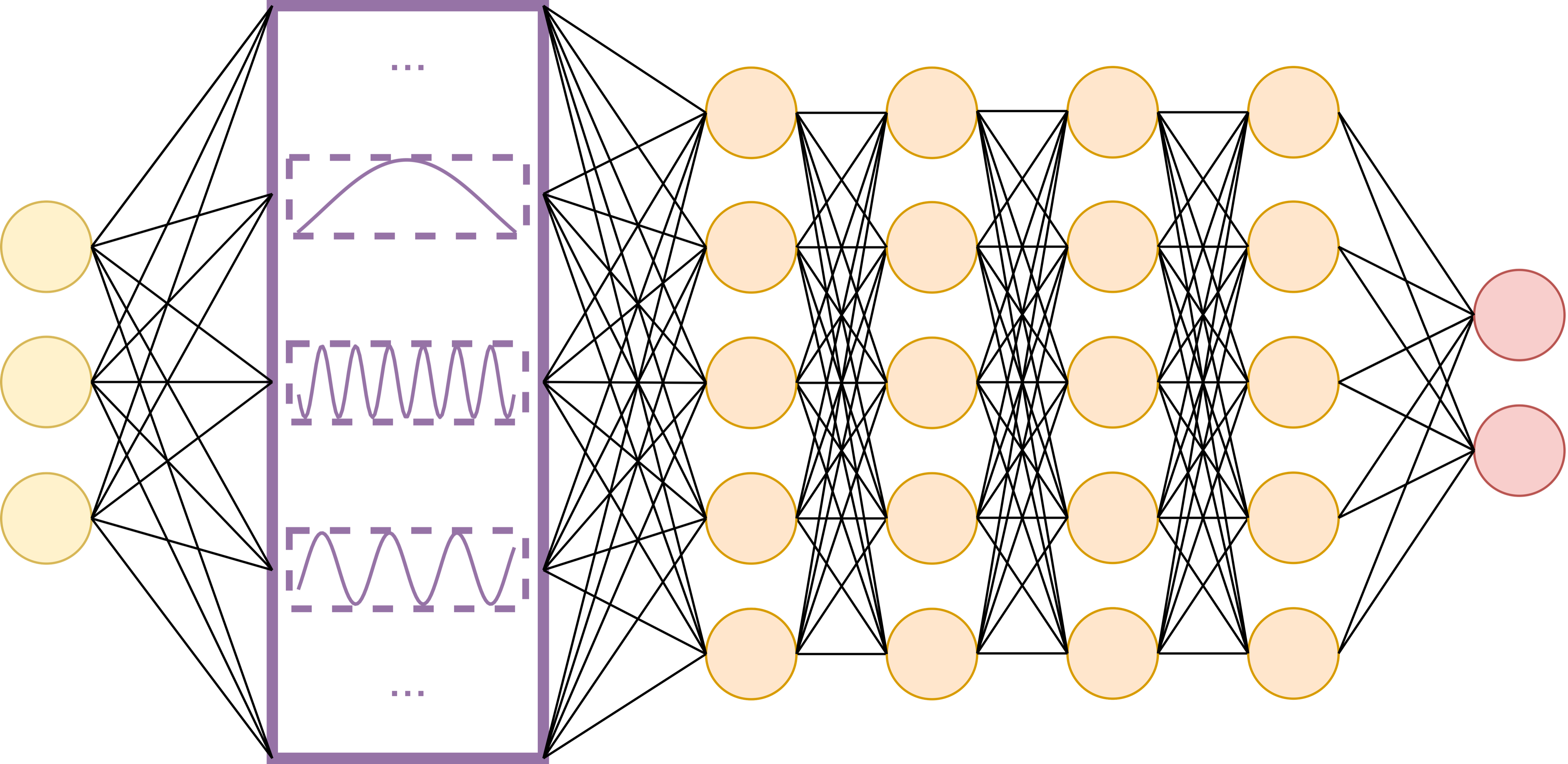}
\caption{A graphical representation of an input transformation PINN, capable of utilizing wavelets, plane waves, or sinusoidal functions, is shown. Yellow nodes represent the network's temporospatial inputs. In this example, the input transformation layer utilizes a plane wave approach, depicted by three distinct purple plane waves. This layer elevates the original input into a higher-dimensional space. Following this, the transformed input undergoes processing through an FCN featuring nonlinear tanh activation functions, illustrated by the four layers of orange nodes. The process culminates in generating the final two-dimensional output, indicated by two pink nodes.}\label{fig:Morlet}
\end{figure}

The first design template we introduce features an input transformation strategy. Specifically, this strategy involves transforming the original input space, defined by $t, x, y$ or $\Omega \times [0, T]$, into a higher-dimensional space, denoted as $\mathcal{H} \in \mathbb{R}^d$, via a mapping $\mathcal{T}: \Omega \times [0, T] \rightarrow \mathcal{H}$. This transformation, $\mathcal{T}$, is realised  by integrating a custom hidden layer equipped with $d$ neurons, which transforms the input space by employing, for instance, $d$ distinct trainable wavelets:
\[
\mathcal{T}(t,\mathbf{x},\theta) = \left[ \tilde{\psi}(t,\mathbf{x};\theta_1), \ldots, \tilde{\psi}(t,\mathbf{x};\theta_d) \right].
\]
We denote this specialized layer as a wavelet/sinusoidal/plane Wave mapping layer, where each $\theta_k$ signifies the trainable parameters of the $k$-th neuron. In the specific instance of a wavelet mapping layer, these parameters would be $(\alpha_k,\beta_k,w_{xk},w_{yk})$. We employed this design template in conjunction with wavelets, plane waves, and sinusoidal functions. Following the initial mapping layer, several fully connected layers are incorporated, mirroring those in the conventional PINN framework (PINN-tanh). A schematic of an architecture embodying this design template is depicted in Figure (\ref{fig:Morlet}).
  
Mathematically, this transformation is represented in the novel forward pass equation:
\begin{equation}
\label{eq:sinusoidal_forward}
\begin{aligned}
\textbf{z}^{0} &= W^{0} \mathcal{T}(t,\textbf{x},\theta) + b^{0}, \\
\textbf{a}^{1} &= \sigma(\textbf{z}^{0}), \\
\textbf{z}^{1} &= W^{1}\textbf{a}^{1} + b^{1}, \\
&\vdots \\
\textbf{a}^{L-1} &= \sigma(\textbf{z}^{L-2}), \\
\textbf{z}^{L-1} &= W^{L-1}\textbf{a}^{L-1} + b^{L-1}, \\
\textbf{a}^{L} &= \sigma(\textbf{z}^{L-1}), \\
\Lambda(t,\textbf{x},\theta) &= W^{L}\textbf{a}^{L} + b^{L}.
\end{aligned}
\end{equation}
where $\mathcal{T}(t,\mathbf{x},\theta)$ varies depending on the mapping layer used.
  
This concept of transforming the input space indicates that the neural network learns not in Cartesian space but rather in a wavelet/plane Wave/sinusoidal space. This is intuitive, as wavelets, for instance, possess intrinsic properties that align with known properties of the wavefield, such as local oscillations and outward propagations. 

\paragraph{Naming}
Henceforth, the three custom mapping layer networks will be referred to as sinusoidal-transformation-PINN (\textit{sT}-PINN), wavelet-transformation-PINN (\textit{wT}-PINN), and plane wave-transformation-PINN (\textit{pT-PINN}).

\paragraph{Initialization}
The fully connected layers within the network follow the configuration and initialization guidelines of a standard PINN, incorporating the tanh activation function and Xavier initialization method. The transformation or mapping layer, however, demands specific attention due to each neuron's multiple weights, designed to encompass a wide array of wave phenomena. Through extensive experimentation, we identified optimal initialization ranges for our problem settings: For plane waves, the amplitude $A$ is initialized within $[-1,1]$, wavenumbers $k_x$ and $k_y$ within $[-2,2]$, velocity $v$ within $[0.5,1.5]$, and the phase term $\phi$ within $[0,\frac{\pi}{4}]$. For wavelets, we set the scaling factor $\alpha$ to initialize within $[0.5,1.5]$, the temporal translation factor $\beta$ within $[-1,1]$, and spatial weights $w_x$ and $w_y$ within $[-0.1,0.1]$. In the case of the sinusoidal mapping layer, the weight matrix $\mathbf{W}$ begins with weights drawn from a normal distribution $\mathcal{N}(0,1)$, and the bias vector is initially zero.

\paragraph{Related Work}
The design template we present closely aligns with the work of Wong et al. \cite{Sinusoidal}, who demonstrated that employing a sinusoidal mapping space can significantly enhance convergence for a wide array of forward and inverse problems by effectively mitigating spectral bias. We build upon this foundation by extending the approach to include wavelet and plane wave mappings, thus broadening the scope and applicability of the method.
Additionally, our proposed technique draws parallels with the field of Fourier Feature Neural Networks, as investigated by Tancik et al. \cite{Fourier_Features}. Their approach introduces a transformation layer consisting of sine and cosine functions, articulated through the random Fourier mapping as
\begin{equation}
\gamma(\mathbf{v}) = \begin{bmatrix}
\cos(\mathbf{Bv}) \
\sin(\mathbf{Bv})
\end{bmatrix},
\end{equation}
where each element of the weight matrix $\mathbf{B}$ is sampled from a normal distribution with a specified standard deviation $\sigma$. This concept shares similarities with ours, with a notable distinction in the necessity to choose an appropriate $\sigma$ for each specific application to ensure optimal convergence. To address this limitation, Wang et al. proposed the introduction of spatiotemporal and multi-scale random Fourier features, where each neuron's weight matrix $\mathbf{B}^i$ is independently sampled from a normal distribution $\mathcal{N}(0,\sigma^i)$, facilitating a wider array of Fourier features \cite{FF2}. However, determining $\sigma^i$ values is still problem-dependent and requires prior specification. In contrast, our method (and the one in \cite{Sinusoidal}) incorporates trainable parameters within the feature mapping, eliminating the need for predetermining optimal $\sigma$ values, which can be challenging for the diverse, multi-scale solutions encountered in our investigations. While our approach does not claim superiority, it offers an alternative strategy.

The motivation behind employing Fourier feature methods stems from their ability to counteract the spectral bias intrinsic to PINNs, which can hinder convergence for high-frequency components of the solution. Elevating the input space into a higher-dimensional domain enriched with various frequency components can greatly alleviate spectral bias. Our goal in developing this network architecture goes beyond merely reducing spectral bias; it includes incorporating physical insights by transforming the input space into, for instance, a wavelet space. This transformation aims to enhance the network's learning effectiveness by implicitly adhering to the underlying solution's properties.
In conclusion, while our work extends the concept of input space transformation, it is not entirely novel. Nonetheless, it is imperative to discuss it in this context, as it forms a fundamental aspect of the spectrum of models we have delineated.

\subsection{Encoder-PINN}
Similar to input transformation PINNs such as \textit{sT}-PINN, \textit{wT}-PINN, and \textit{pT}-PINN, this network design incorporates a custom wavelet/plane wave layer. However, a key distinction lies in the placement of this layer, which is situated at the end of the network rather than at the beginning. Instead of elevating the original Cartesian input space into a higher dimensional space, this design posits that the final wavefield is a superposition of wavelets/plane waves. Furthermore, the intrinsic parameters of the custom layer are not directly learnable; they are derived as functions of the input $(t,x,y)$, obtained by processing the input through an FCN whose output dimension is $d \times k$. Here, $d$ represents the number of neurons in the wavelet/plane wave layer, and $k$ signifies the number of dynamic parameters for each neuron (five for plane waves, four for wavelets). The process entails processing the original temporal and spatial input through several fully connected layers to generate the parameters for each plane wave/wavelet neuron. These parameters, along with the original input (as plane waves and wavelets also depend on $t, x, y$), are then utilized by the custom plane wave/wavelet layer. The final wavefield is a \textbf{linear} combination of the $d$ wavelets/plane waves, where the term \textit{linear} implies that the output layer consists of weights and biases without an additional activation function.

\begin{figure}
\centering
    \includegraphics[width=0.8\linewidth]{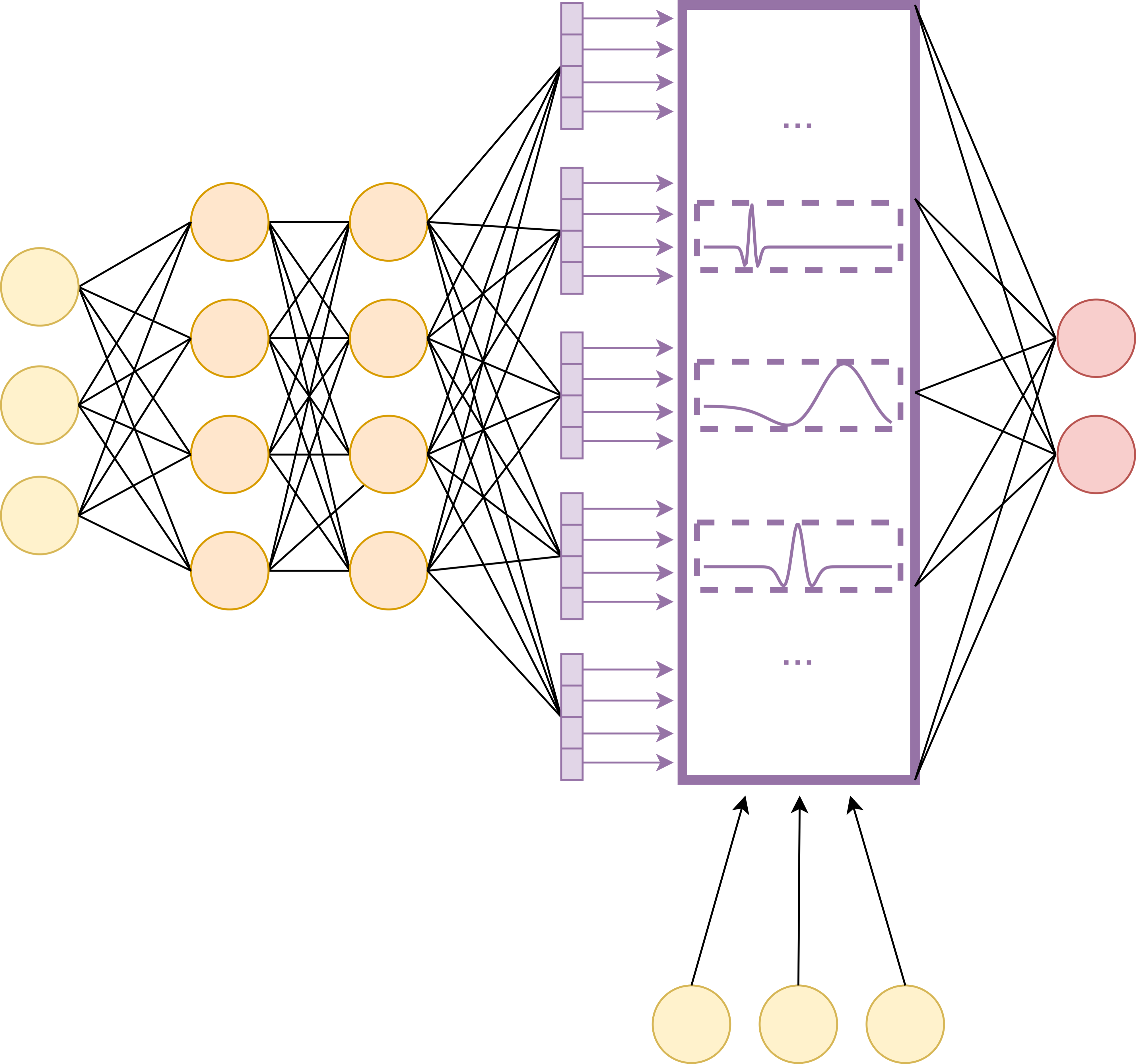}
    \caption{Graphical representation of an Encoder-type PINN, utilizing wavelets or plane waves. Yellow nodes indicate temporospatial inputs. Orange nodes, comprising an FCN utilizing the tanh activation function, serve as the encoder tasked with generating parameters for the wavelet/plane wave layer; wavelets are used for illustration. The encoder outputs $d=4$ parameters for each of the five wavelet neurons, depicted as four squares per neuron. The wavelet layer, illustrated with three wavelets for clarity, receives both the encoder-generated parameters and the original temporospatial inputs (yellow nodes at the bottom). The network's two-dimensional output emerges from a linear combination of the wavelet layer outputs, represented by two pink nodes.}
\label{fig:FCN_ALL_PARALMS}
  \end{figure}

The forward pass of encoder-type PINNs can be modeled as:
\begin{equation}
\begin{aligned}
\textbf{z}^{0}_E &= W^{0}_E(t,\mathbf{x}) + b^{0}_E, \\
\textbf{a}^{1}_E &= \sigma^{1}_E(\textbf{z}^{0}_E), \\
\textbf{z}^{1}_E &= W^{1}_E\textbf{a}^{1}_E + b^{1}_E, \\
&\vdots  \\
\textbf{a}^{L-1}_E &= \sigma^{L-1}_E(\textbf{z}^{L-2}_E), \\
\textbf{z}^{L-1}_E &= W^{L-1}_E\textbf{a}^{L-1}_E + b^{L-1}_E, \\
\textbf{a}^{L}_E &= \sigma^{L}_E(\textbf{z}^{L-1}_E) + b^{L}_E, \\
\theta(\Lambda_E) &= W^{L}_E\textbf{a}^{L}_E, \\
\Lambda(\mathbf{x}) &= W^{L} \mathcal{T}(t,\mathbf{x}, \theta(\Lambda_E)) + b^{L}.
\end{aligned}
\label{eq:encoder_pinn_forward}
\end{equation}
where $\mathcal{T}$ denotes the wavelet or plane wave functions, for instance,  $\mathcal{T}(t,\mathbf{x},\theta) = \tilde{\psi}(t,\mathbf{x})$ when utilizing wavelets. $\theta(\Lambda_E)$ represents the parameter layer, which is input-dependent and determined by the encoder network, characterized by the encoder's weights $W^{1}_E, \ldots, W^{L-1}_E, W^{L}_E$ and biases $b^{0}_E, \ldots, b^{L}_E$. $W^{L}$ and $b^{L}$ are the weights and bias of the final output layer, mapping the $d$-dimensional wavelet/plane wave space to the two-dimensional output space.
Figure (\ref{fig:FCN_ALL_PARALMS}) illustrates a simplified setup of this architecture.

\paragraph{Naming}
The networks are designated as plane wave-encoder-PINNs (\textit{pE}-PINN) and wavelet-encoder-PINNs (\textit{wE}-PINN). The designation "encoder" is appropriately selected, reflecting the role of the FCN that computes the wavelet/plane wave parameters as an encoder network. This encoder network effectively maps from the input space to the plane wave/wavelet parameter space.

\paragraph{Initialization}
Distinct from the input transformation networks, the initialization of wavelet/plane wave parameters is not required in this context because these parameters are generated by an FCN. Consequently, the initialization and configuration of the FCN are directly aligned with those of a standard PINN employing the tanh activation function (PINN-tanh).

%CHECK WITH BEN: is this correct and okay?
\paragraph{Related Work}
Existing literature, such as \cite{Gabor}, demonstrates instances where the PINN solution is articulated as a linear combination of Gabor basis functions. However, to the best of our knowledge, there is no precedent for modelling the solution as a linear combination of custom basis functions, where the basis functions themselves are learned and dependent on input parameters. For a more detailed comparison of these innovative architectural approaches to established models, refer to the related work section in the Encode-Decoder-PINN segment (\ref{section:encoder-decoder}).

\subsection{Amplitude-PINN}
\begin{figure}\centering
    \includegraphics[width=.8\linewidth]{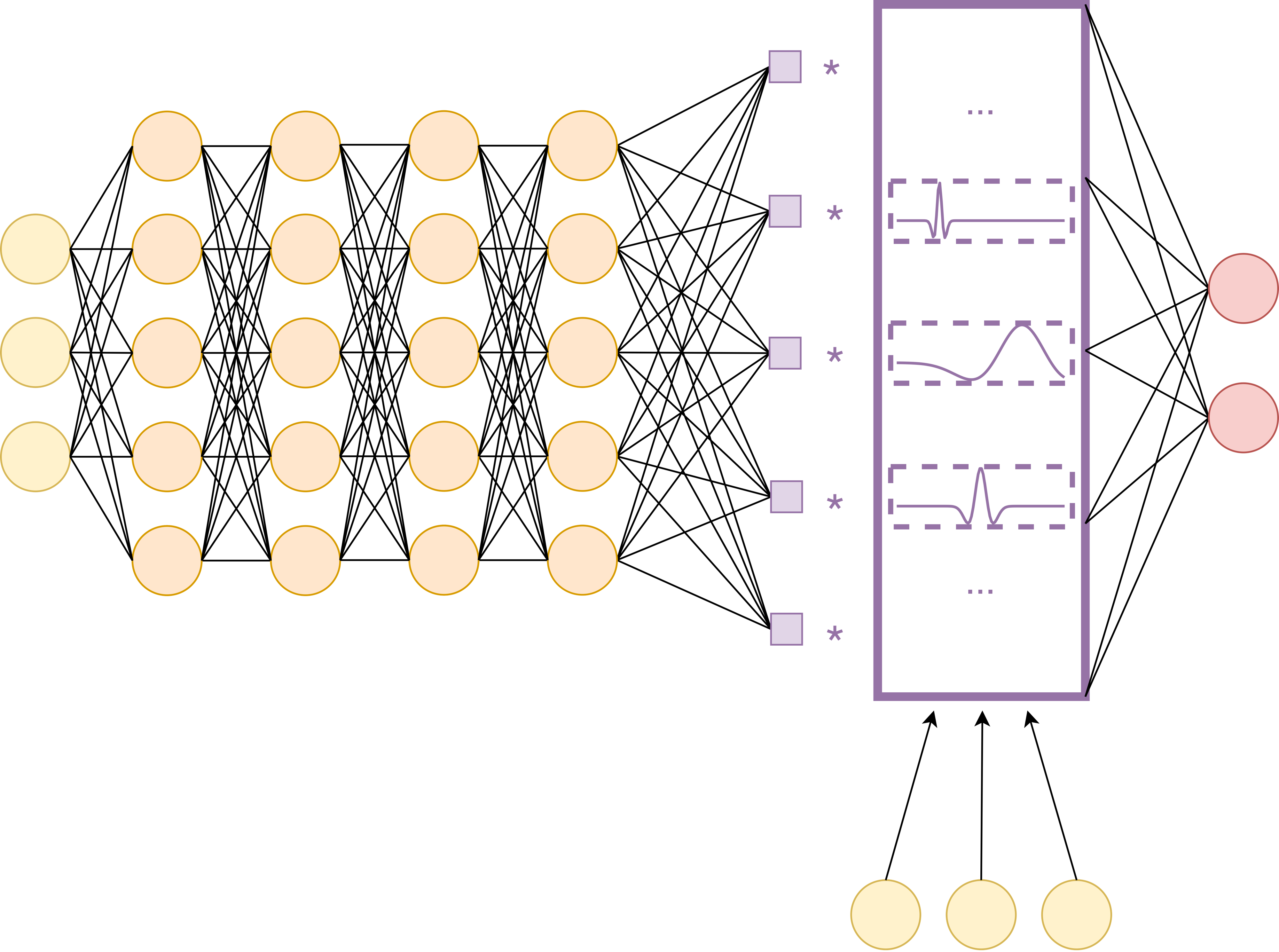}
    \caption{This graphical representation outlines an Amplitude-PINN, adaptable for either wavelets or plane waves. Yellow nodes signify the network's temporospatial inputs. Encoded as orange nodes, the hidden layers employ tanh activation functions to form an Encoder-type FCN dedicated to calculating Amplitudes, which are represented by purple squares. These Amplitudes are subsequently multiplied by the outputs from the custom wavelet/plane wave layer, exemplified here by three distinct wavelets. Importantly, the wavelet/plane wave layer, with its trainable but input-independent parameters, operates autonomously from the preceding FCN. The temporospatial inputs are also directly channelled into this wavelet layer, as illustrated by the yellow nodes at the bottom. The culmination of this process is the network's two-dimensional output, denoted by pink nodes, achieved without additional non-linear layers.}
    \label{fig:Amplitude_FCN}
  \end{figure}

The amplitude-PINN embodies a hybrid model, integrating concepts from both the input transformation PINN and the Encoder PINN frameworks. Similar to the Encoder PINN, it theorizes that the solution can be represented as a linear combination of plane waves or wavelets. Nonetheless, a distinctive feature is that only the amplitudes of these plane waves or wavelets are input-dependent, determined by a primary FCN. The remaining parameters of the plane waves or wavelets remain trainable but not influenced by the input, akin to the methodology employed in the input transformation PINN.
The forward pass mechanism of the Amplitude-PINN is formulated as:
\begin{equation}
\begin{aligned}
\textbf{z}^{0}_E &= W^{0}_E(t,\mathbf{x}) + b^{0}_E, \\
\textbf{a}^{1}_E &= \sigma^{1}_E(\textbf{z}^{0}_E), \\
\textbf{z}^{1}_E &= W^{1}_E\textbf{a}^{1}_E + b^{1}_E, \\
&\vdots\\
\textbf{a}^{L-1}_E &= \sigma^{L-1}_E(\textbf{z}^{L-2}_E), \\
\textbf{z}^{L-1}_E &= W^{L-1}_E\textbf{a}^{L-1}_E + b^{L-1}_E, \\
\textbf{a}^{L}_E &= \sigma^{L}_E(\textbf{z}^{L-1}_E) + b^{L}_E, \\
A &= W^{L}_E\textbf{a}^{L}_E, \\
\Lambda(\mathbf{x}) &= W^{L} (A \cdot \mathcal{T}(t,\mathbf{x}, \theta) + b^{L}).
\end{aligned}
\label{eq:amplitude_pinn_forward}
\end{equation}
where $A$ signifies the amplitude, which is input-dependent and computed by the encoder network via the weights and biases $W^{1}_E, \ldots, W^{L}_E$ and $b^{0}_E, \ldots, b^{L}_E$. Subsequently, in the penultimate step before the linear output layer, this amplitude $A$ is multiplied by the output of each neuron's custom wavelet or plane wave function. A simplified depiction of this network configuration is illustrated in Figure (\ref{fig:Amplitude_FCN}).

\paragraph{Naming}
The networks introduced are collectively known as amplitude-PINNs, with specific variations including the wavelet-amplitude-PINNs (\textit{wA}-PINN) and plane wave-amplitude-PINNs (\textit{pA}-PINN). 
\paragraph{Initialization}
Except for the amplitude, all parameters adhere to the initialization protocols specified in the Section on input transformation PINNs (\ref{input_transforrmation}). The amplitude parameters are uniquely exempt from manual initialization, as they are dynamically generated by a preceding FCN. 

\paragraph{Related Work}
The conceptualization of the amplitude-PINN architecture was significantly influenced by the pioneering efforts of Alkahlifah et al. \cite{Gabor}, who integrated Gabor basis functions with PINNs to tackle the Helmholtz equation. Their methodology involved the multiplication of an FCN's output---akin to the amplitude in our context---with a custom layer employing a Gabor activation function. 

\subsection{Encoder-Decoder-PINN}
\label{section:encoder-decoder}
\begin{figure}\centering
    \includegraphics[width=.8\linewidth]{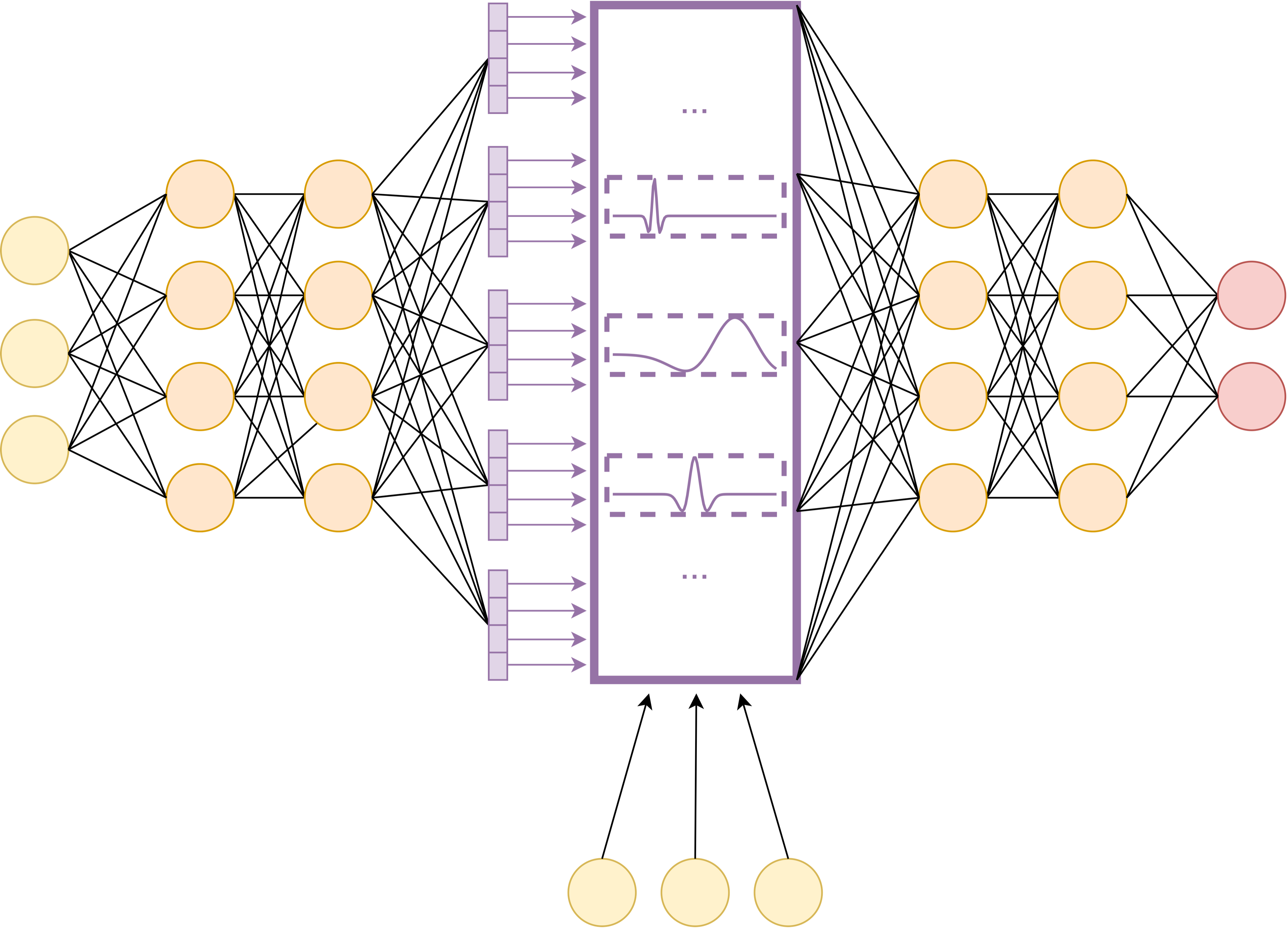}
    \caption{This illustration showcases an Encoder-Decoder PINN (\textit{ED}-PINN), capable of incorporating either wavelets or plane waves. Yellow nodes represent the network's temporospatial inputs. Orange nodes indicate the hidden layers equipped with tanh activation functions, functioning in both the encoding (initial) and decoding (final) phases of the network. The encoder's output, depicted by purple squares, serves as the parameter layer, which also acts as the input to the custom wavelet/plane wave layer, illustrated here by three distinct wavelet examples for clarity. This layer receives inputs from both the input-dependent parameters generated by the encoder and the original temporospatial inputs, as shown by the yellow nodes at the bottom. The information then passes through the decoder layer, leading to the network's final two-dimensional output, indicated by the two pink nodes.}
    \label{fig:FCN_ALL_PARALMS_FCN}
\end{figure}

Encoder-decoder networks share a foundational architecture with Encoder-PINN types but introduce a critical modification. Instead of conceptualizing the final wavefield as a linear combination of wavelets or plane waves, these networks incorporate a secondary FCN after the wavelet/plane wave layer. This secondary FCN consists of multiple fully connected hidden layers, offering enhanced flexibility by allowing for the construction of the final wavefield as a nonlinear superposition of the wavelets/plane waves. This adjustment not only increases the model's adaptability but also enables the network to potentially model complex solution features, such as reflections and refractions in layered media, more effectively.
The forward pass in Encoder-Decoder PINNs, while bearing similarity to that of Encoder PINNs, is defined as follows:
\begin{equation}
\label{eq:encoder_decoder_pinn_forward}
\begin{aligned}
&\left.\begin{aligned}
\mathbf{z}^{0}_E &= W^{0}_E(t,\mathbf{x}) + b^{0}_E, \\
\mathbf{a}^{1}_E &= \sigma^{1}_E(\mathbf{z}^{0}_E), \\
\mathbf{z}^{1}_E &= W^{1}_E\mathbf{a}^{1}_E + b^{1}_E, \\
&\vdots \\
\mathbf{a}^{L-1}_E &= \sigma^{L-1}_E(\mathbf{z}^{L-2}_E), \\
\mathbf{z}^{L-1}_E &= W^{L-1}_E\mathbf{a}^{L-1}_E + b^{L-1}_E, \\
\mathbf{a}^{L}_E &= \sigma^{L}_E(\mathbf{z}^{L-1}_E) + b^{L}_E, \\
\end{aligned} \right\} \text{Encoder Network} \\
&\theta(\Lambda_E) = W^{L}_E\mathbf{a}^{L}_E, \\
&\left.\begin{aligned}
\mathbf{z}^{0}_D &= W^{0}_D \mathcal{T}(t,\mathbf{x}, \theta(\Lambda_E)) + b^{0}_D, \\
\mathbf{a}^{1}_D &= \sigma^{1}_D(\mathbf{z}^{0}_D), \\
\mathbf{z}^{1}_D &= W^{1}_D\mathbf{a}^{1}_D + b^{1}_D, \\
&\vdots \\
\mathbf{a}^{L-1}_D &= \sigma^{L-1}_D(\mathbf{z}^{L-2}_D), \\
\mathbf{z}^{L-1}_D &= W^{L-1}_D\mathbf{a}^{L-1}_D + b^{L-1}_D, \\
\mathbf{a}^{L}_D &= \sigma^{L}_D(\mathbf{z}^{L-1}_D), \\
\end{aligned} \right\} \text{Decoder Network} \\
&\Lambda(\mathbf{x}) = W^{L}_D\mathbf{a}^{L}_D + b^{L}_D.
\end{aligned}
\end{equation}
Here, $\mathcal{T}$ signifies the wavelet or plane wave functions, for instance, $\mathcal{T}(t,\mathbf{x},\theta) = \tilde{\psi}(t,\mathbf{x})$ when utilizing wavelets. $\theta(\Lambda_E)$ represents the parameter layer, which is the output from the encoder network, indicated by the weights $W^{1}_E, \ldots, W^{L}_E$ and biases $b^{0}_E, \ldots, b^{L}_E$. The final output, $\Lambda(\mathbf{x})$, is produced by passing the output from the wavelet/plane wave layer through the decoder network, identified by its weights $W^{1}_D, \ldots, W^{L}_D$ and biases $b^{0}_D, \ldots, b^{L}_D$.
Figure (\ref{fig:FCN_ALL_PARALMS_FCN}) illustrates a simplified model of this architecture template.

\paragraph{Naming}
This network design, characterized by an initial encoder FCN and a secondary FCN that serves as a decoder, is designated as encoder-decoder-PINN. The specific configurations are referred to as plane wave-encoder-decoder-PINNs (\textit{pED}-PINN) and wavelet-encoder-decoder-PINNs (\textit{wED}-PINN).

\paragraph{Related Work}
A key distinction of encoder-decoder and encoder-PINNs, compared to previously discussed input transformation-PINNs, as well as Fourier Feature and Multi-Scale Fourier Features Neural Networks \cite{Fourier_Features} \cite{FF2}, is their adaptability in the parameterization of the transformation, wavelet, or plane wave layers. This design eliminates the need for predefined or randomly chosen parameter values. Such an approach significantly lessens the reliance on prior knowledge of the solution's frequency components, which can be complex and challenging to predict. This flexibility allows the model to autonomously learn and adapt to the spectral characteristics of the solution, providing a more dynamic and potentially more precise modelling approach.

\paragraph{Initialization}
Consistent with the encoder network paradigm, the wavelet/plane wave parameters do not require explicit initialization since they are derived from learned, input-dependent functions. Both the encoder and decoder networks employ Xavier initialization, mirroring the approach used in standard PINN models (PINN-tanh) that utilize tanh activation functions. 

\subsection{Failure Cases}
The models elaborated upon earlier were chosen for deeper comparative analysis as they exhibited convergence in some contexts. Nevertheless, we evaluated a broad array of models spanning the entire spectrum, which did not display convergence even after rigorous hyperparameter optimization. In the following section, we detail two network architectures that notably failed to converge.

\paragraph{Modified-Radial-Bessel-PINNs}
This network design initiated our neural network architecture search; the core idea was to create an \textit{Ansatz} solution closely approximating the true solution by incorporating Bessel functions. These functions satisfy most of the known solution properties, arguably more so than plane waves and wavelets. We define $\mathcal{B}(\textbf{x})$, the output of a single Modified-radial-bessel Neuron (MRB) as follows:
\begin{equation}
\begin{aligned}
&\mathcal{B}(t,x,y) = g(x,y) \cdot \cos\left(\arctan2\left(y + \varepsilon, x + \varepsilon\right) - \alpha + (\beta \cdot \arctan2\left(y + \varepsilon, x + \varepsilon\right)^2)\right) \\
&\cdot A \cdot e^{-t} \cdot \tilde{j0}(k \cdot \sqrt{x^2 + y^2 + \varepsilon} - \omega \cdot t + \varepsilon), 
\end{aligned}
\label{eq:bessel}
\end{equation}
where $g(x,y)$ produces a modulated radial-Gaussian envelope, which propagates over time with a given speed $\omega$, controlled by the parameters $\gamma$ and $\kappa$. This design aims to limit the Bessel functions' continuous oscillatory nature to model the typically observed limited number of outward propagating wavefronts in wave physics. The oscillatory nature is introduced through the Bessel function $\tilde{j0}(x)$, using an approximation of the Bessel function of the first kind. Due to the lack of a derivative for PyTorch's built-in Bessel function, we opted for an algebraic approximation, whose implementation and derivatives can be computed readily (see Appendix (\ref{Appendix:Architecture})). $A$ determines the amplitude, and the term $e^{-t}$ accounts for attenuation. The cosine term, including $arctan2$ terms, models the direction and polarity of the primary wavefront propagation. Each parameter, excluding constants in the Bessel approximation, is learnable, totalling eight learnable parameters.

\begin{figure}
\begin{center}
  \includegraphics[width=1.0\linewidth]{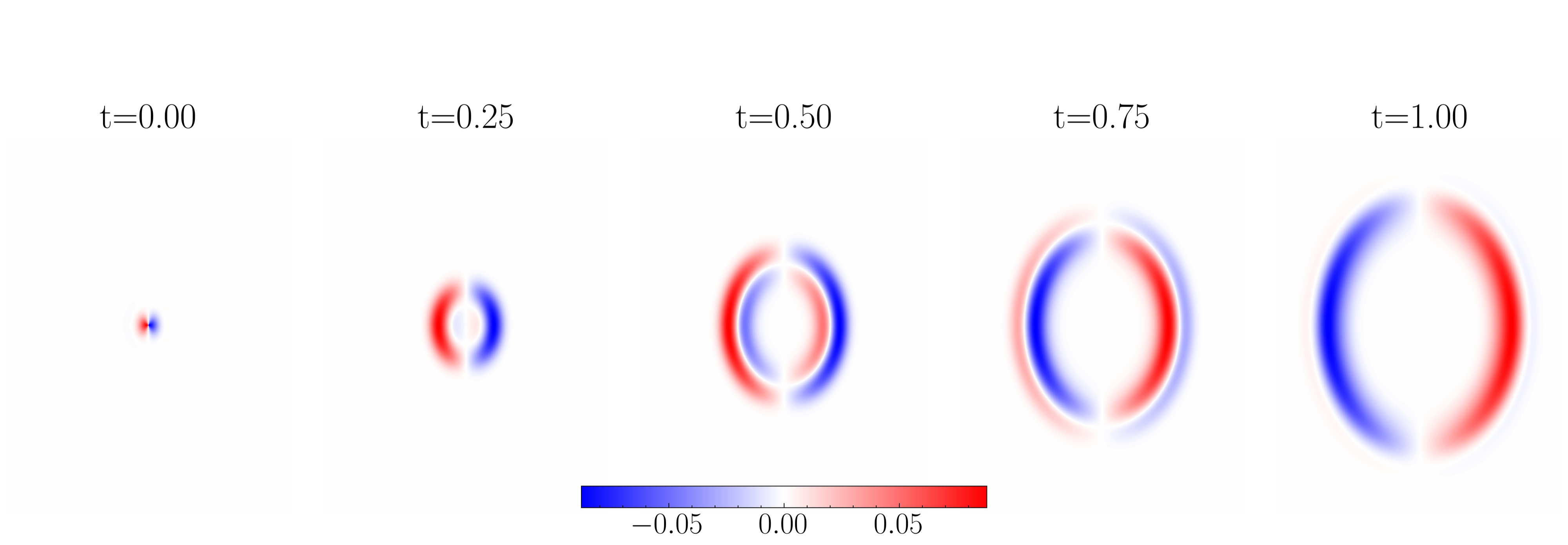}
  \caption{Visualization of the x-component of the wavefield generated by a single MRB neuron. The parameters used can be found in Appendix (\ref{APPENDIX:MRB})}
  \label{fig:bessel}
\end{center}
\end{figure}

Figure (\ref{fig:bessel}) shows the output generated by a \textbf{single} MRB-neuron. With the correct parameters, a single MRB neuron can generate a wavefield very close to the true solution for a constant parameter case. However, this Figure was produced by extensive tweaking of parameters, not by NN-learned parameters. We attempted to incorporate MRB layers in various configurations (input transformation, encoder layer, encoder+decoder layer), but none showed any convergence signs, even after exhaustive hyperparameter searches. We argue that the MRB basis functions create a loss landscape that is too complex for the optimiser to navigate effectively.

\paragraph{Modified-Far-Field-PINNs}
Similar to MRB-PINNs, Modified-Far-Field-PINNs (MFF-PINNs) employ a sophisticated solution \textit{Ansatz} that, in theory, should closely approximate the true wavefield solution. This model draws upon an analytical approach by Keiiti Aki and Paul G. Richards, as outlined in their seminal work, "Quantitative Seismology" \cite{quantitative_seismology}. We have adapted their equation to suit our specific needs (refer to Appendix (\ref{Appendix:Architecture}) for the equation details). Like MRB-PINNs, MFF-PINNs feature a high number of trainable parameters per neuron---specifically, seven---thus presenting a considerable challenge to the optimization process due to the complexity of the employed basis function. Despite the theoretical proximity of this approximation to the actual modelling of elastic wave propagation, the model consistently failed to converge.

\paragraph{Additional Models}
For the sake of completeness, we list other models that were part of our investigation but will not be discussed in detail due to their inadequate convergence or the absence of novel insights:
\begin{itemize}
\item The PINN-sin and PINN-swish models, employing sine and swish activation functions, respectively, instead of the tanh activation, exhibited inferior convergence performance and did not provide new insights for integrating wave-physics knowledge into the network architecture.
\item Both Encoder-PINNs and encoder-decoder PINNs with sinusoidal functions struggled with convergence issues.
\item Various dual-network structures, designed with separate $u_x$ and $u_y$ component wavelet/plane wave layers, underperformed compared to their counterparts featuring unified custom layers.
\item Traditional FCN networks utilizing wavelets or plane wave activation functions failed to converge. This failure is likely due to the physical incongruity introduced by nesting these functions; specifically, using the output of one wavelet as the input to another contradicts physical principles. In wave physics, the input to a wavelet should be in (modified) Cartesian space, and deviations from this principle result in non-convergence.
\end{itemize}
These explorations into various network designs underscore the necessity of adherence to the physical laws governing the phenomena being modelled and the intricate requirements for developing effective PINNs tailored to wave physics applications. Given the significant time and resources necessary to configure and test over ten different network architectures with various initialization schemes and hyperparameter settings, our focus has shifted to presenting a subset of functional models for a qualitative comparison.

\subsection{Placement on the Spectrum}
\label{spectrum_placement}
\begin{figure}
    \centering
    \includegraphics[width=1.0\linewidth]{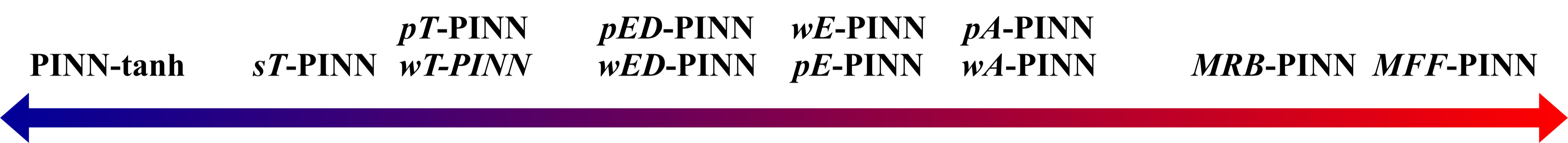}
    \caption{Approximate placement of the discussed PINN network architectures on the Spectrum. From left to right: Increasing restrictiveness and increasing utilization of prior wave-physics knowledge.}
    \label{fig:spectrum_placement}
\end{figure}

This section discusses the placement of the network designs we've explored on our previously constructed spectrum. It's crucial to state upfront that our placements are subjective and not absolute. The exact placement of each model on the spectrum cannot be definitively proven.
Starting from the very left of the spectrum, where the most unrestricted and uninformed architectures are found, we move rightwards towards models that incorporate more prior knowledge but may also be more restricted in their capacity to generalise.
The starting point on this spectrum is occupied by the standard PINN (PINN-tanh), which employs fully connected layers with the tanh activation function and utilises no prior knowledge of the wave function. This model is the least restricted in its function approximation process.
Next, we place the \textit{sT}-PINN model. A minimal level of restriction and information is introduced by transforming the input into sinusoidal space, suggesting that some wave physics knowledge is incorporated by transforming the input space into a domain that implicitly satisfies oscillatory features. Despite this, the network retains significant flexibility due to the standard FCN network that follows the input-mapping layer.
Following this reasoning, both the \textit{wT}-PINN and \textit{pT}-PINN models are positioned adjacent to the \textit{sT}-PINN. While one could argue that these three models should occupy the same position on the spectrum, we believe that using wavelets and plane waves in the input transformation layer conveys more prior wave knowledge than more straightforward sinusoidal functions. This is attributed to wavelets and plane waves including more intrinsic learnable parameters compared to the sinusoidal functions, which feature only a trainable weight matrix (corresponding to wave frequency) and a bias vector (corresponding to phase shift). Moreover, plane waves and wavelets inherently capture more nuanced representations of wave behaviours like attenuation, reflection, or varying frequency components.
The \textit{pED}-PINN and \textit{wED}-PINN models are positioned next. These models are considered to utilise more prior knowledge due to their architectural differences from the \textit{wT}-type-PINNs. The main difference lies in the fact that the wavelet/plane wave parameters are now directly dependent on the input, allowing the model to encode temporally varying parameters like frequency directly. This architecture better reflects the reality that characteristics of wave phenomena are often directly dependent on environmental conditions. By having wavelet/plane wave parameters that are functions directly dependent on the original input space (t,x,y), the model can more accurately mimic this aspect of wave physics. Real-world seismic wavefields often exhibit non-uniformities and complex behaviours that vary spatially and temporally. Input-dependent wavelet/plane wave parameters enable the model to capture these variations more effectively than "independent" learnable parameters, which provide a broader, more general approach to modelling waves but might not be as closely tied to specific wave physics. A more technical detail that further solidifies our choice of placement is the design of \textit{sT}-type-PINNs, where the entire FCN follows after the custom wavelets layer. To ensure fair cross-comparison, we modelled the encoder and decoder FCN in \textit{wED}-type-PINNs to have roughly the same amount of parameters together as the subsequent FCN in \textit{sT}-type-PINNs. This means the wavelet/plane wave layer of \textit{wED}-type-PINNs is roughly in the "middle" of the overall network, making it more closely situated to the final output. Thus, the wavefield solution is a non-linear superposition of wavelets/plane waves with less flexibility compared to \textit{sT}-type-PINNs.
Following this, the \textit{wE}-PINNs and \textit{pE}-PINNs are placed to the right of \textit{wED}-type-PINNs, as they do not use a decoder network. This indicates that the overall wavefield solution is a linear combination of wavelets, introducing less flexibility and thus positioning these models further to the right on the spectrum.
Finally, the \textit{wA}-PINNs and \textit{pA}-PINNs are positioned as being more restrictive than the \textit{wE}-type-PINNs, as they utilise a single parameter (amplitude) learned from the encoder network, which is then multiplied with the output of the wavelet layer. This model type is considered the most restrictive among the four discussed design templates.
Failure cases like MRB-PINNs and MFF-PINNs are placed at the very end of the right-hand side of the spectrum due to their highly specialised and restrictive nature. They aim to model the output function directly.
In summary, while the placement of these models on the spectrum is based on our subjective assessment, certain conclusions are clear:
\begin{itemize}
\item The PINN-tanh model stands out as the least informed and least restrictive among the models considered.
\item Models lacking a decoder network (\textit{wA}-PINNs, \textit{pA}-PINNs, \textit{wE}-PINNs, \textit{pE}-PINNs) impose more restrictions compared to those incorporating one (\textit{wED}-PINNs, \textit{pED}-PINNs, and, to a certain extent, \textit{sT}-PINNs, \textit{pT}-PINNs, and \textit{wT}-PINNs).
\item Highly specialised models like the MRB-PINN and the MFF-PINN, which attempt to directly model the output function, are the most restrictive models.
\end{itemize}
Figure (\ref{fig:spectrum_placement}) visualises the approximate placement of all discussed network architectures on the spectrum.

\section{Results}
\label{section:Arch_res}
In this section, we present the results, which are divided into two subsections. Firstly, we examine the performance of various models across the spectrum under different Lam{\'e} parameter settings to evaluate the correlation between their placement on the spectrum (i.e., the extent to which prior wave physics is incorporated into the network design) and the models' accuracy.
In the subsequent subsection, we delve deeper into a specific network architecture, the encoder-decoder-PINN, which has been identified as the most effective model overall. We test its capabilities under a broader range of experimental settings, comparing it against the baseline PINN model (PINN-tanh).

\subsection{Evaluation along the Spectrum}
The previously presented network designs were evaluated against each other using the three Lam{\'e} parameter settings as detailed in Section (\ref{heterogeneous_models}). These models ranged from the simplest, with constant Lam{\'e} parameters, to more complex models, such as the mixture and layered models.
\paragraph{Experimental Framework}
\label{section:exp_setup}
To ensure equitable comparisons among various network architectures and Lam{\'e} parameter models, certain hyperparameters were consistently held constant as follows:
\begin{itemize}
\item \textbf{For tests involving constant Lam{\'e} parameter models:} The number of collocation points was set to $100000$, epochs were fixed at $200$, and both maximum iterations and evaluations were limited to $100$. The LBFGS optimizer was utilized with a learning rate set to $1.0$.
\item \textbf{For tests involving the mixture model:} Settings remained identical to those for the constant parameter model mentioned above.
\item \textbf{For tests involving the layered model:} The number of collocation points was kept at $100000$, with epochs increased to $400$, and both maximum iterations and evaluations expanded to $200$. The LBFGS optimizer was applied, with the learning rate adjusted to $2.0$.
\end{itemize}
Other hyperparameters were deliberately varied to explore their influence on model efficacy. Each architecture underwent evaluation using three specific $t_1$ values: $[0.07, 0.1, 0.2]$, as justified in Section (\ref{section:t1_vanilla}). The selection process for the number of hidden layers and neurons, including the arrangement of distinct encoder and decoder networks for architectures necessitating such configurations, warrants further elucidation. Given the constraints imposed by resources and time, it was impractical to undertake exhaustive hyperparameter optimization, such as a complete grid search for each network architecture and subtype. Consequently, a pragmatic approach was adopted, where, based on numerous experiments, hyperparameter configurations that appeared effective for each model type were identified through random selection and intuitive judgement. Such \textit{good points} were subsequently utilized as starting points for scanning around these regions in the hyperparameter space. Within these more confined intervals, a grid search was conducted for each model to identify the optimal configuration for each network architecture. Unless explicitly indicated otherwise, the results and comparative analyses discussed in subsequent sections and tables are derived from the most effective configurations identified for each network design. For detailed information on the specific hyperparameter configurations employed for each model across the three tested Lam{\'e} parameter scenarios, refer to Tables (\ref{tab:hyperparameters_constant} - \ref{tab:hyperparameters_layered}) in Appendix (\ref{Appendix:Architecture}).
A critical aspect was to ensure that each experiment included a PINN-tanh model with a comparable number of trainable parameters, thereby mitigating any potential assertions that the enhanced performance of novel, wave-physics-informed networks could be ascribed merely to their having a more substantial approximation network base.
 
\begin{figure}
\centering
% First subfigure
\begin{subfigure}[b]{.8\textwidth}
  \includegraphics[width=\linewidth]{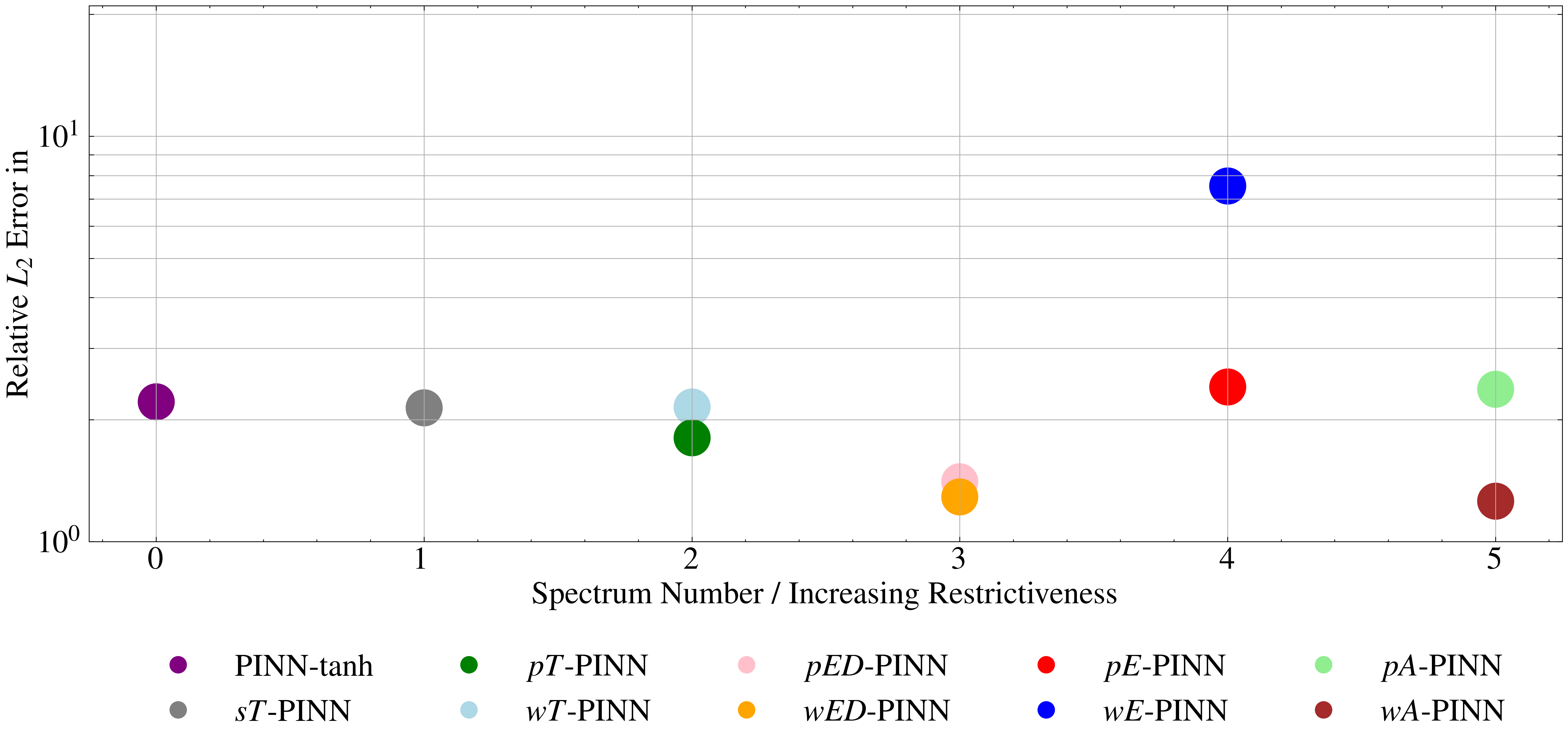}
  \caption{Constant Lam{\'e} parameters}
  \label{fig:sub1}
\end{subfigure}
% Second subfigure
\begin{subfigure}[b]{.8\textwidth}
  \includegraphics[width=\linewidth]{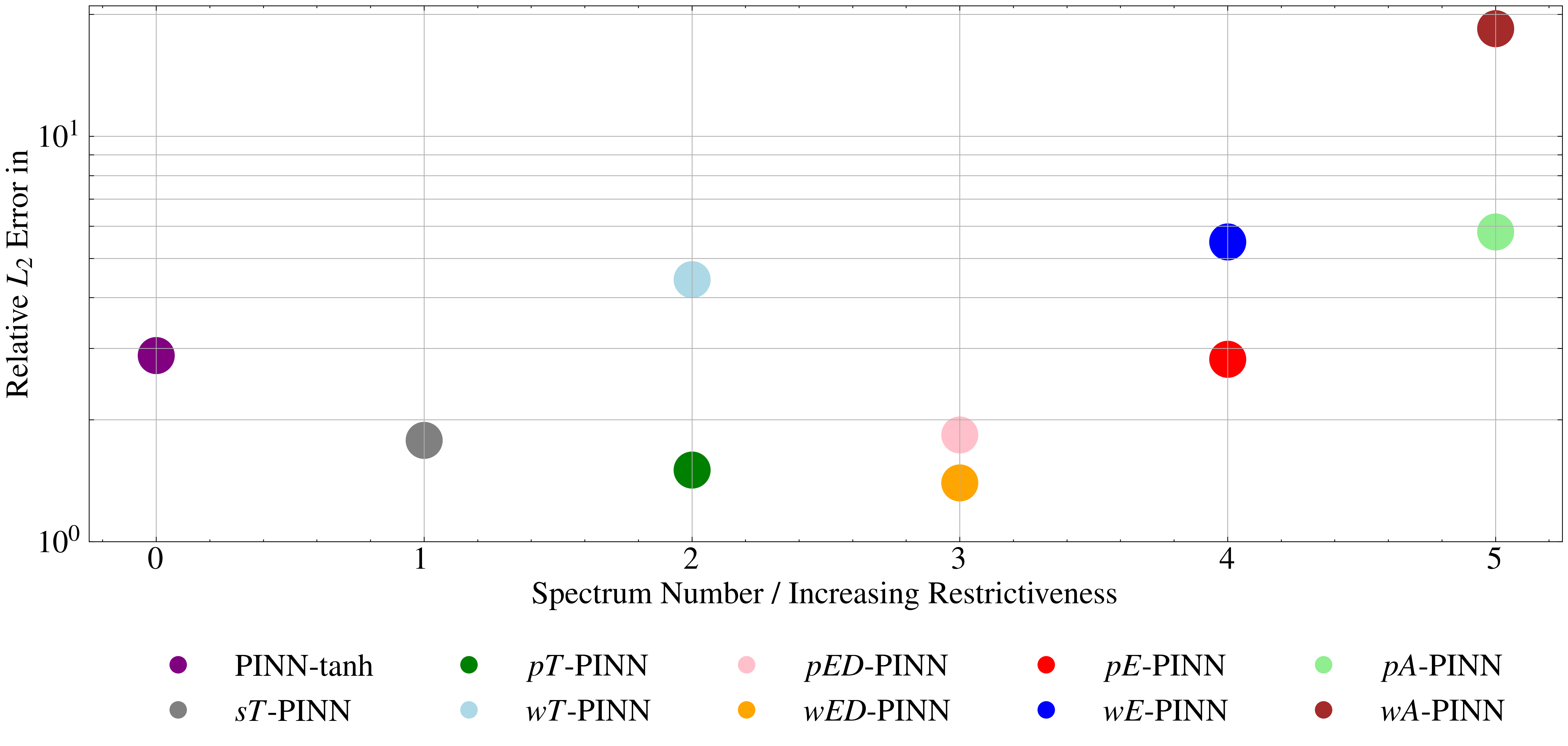}
  \caption{Mixture model Lam{\'e} parameters}
  \label{fig:sub2}
\end{subfigure}
% Third subfigure
\begin{subfigure}[b]{.8\textwidth}
  \includegraphics[width=\linewidth]{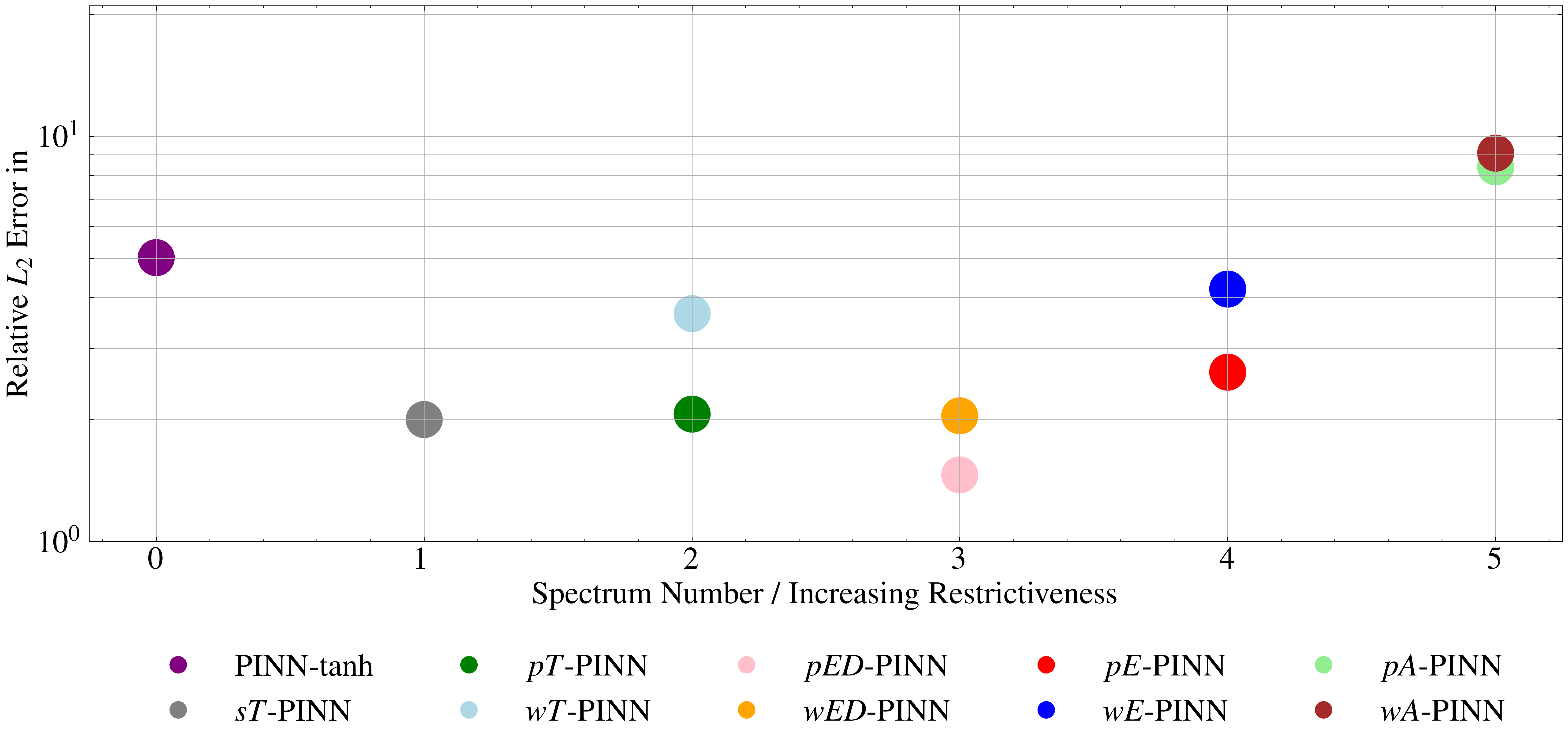}
  \caption{Layered model Lam{\'e} parameters}
  \label{fig:sub3}
\end{subfigure}
\caption{Average relative $L_2$ error of the best configuration of each architecture type, organised according to their placement on the spectrum. From top to bottom, the experiments used the following underlying Lam{\'e} parameters: Constant, Mixture and Layered.}
\label{fig:combined}
\end{figure}

\begin{figure}
\begin{center}
  \includegraphics[width=1.0\linewidth]{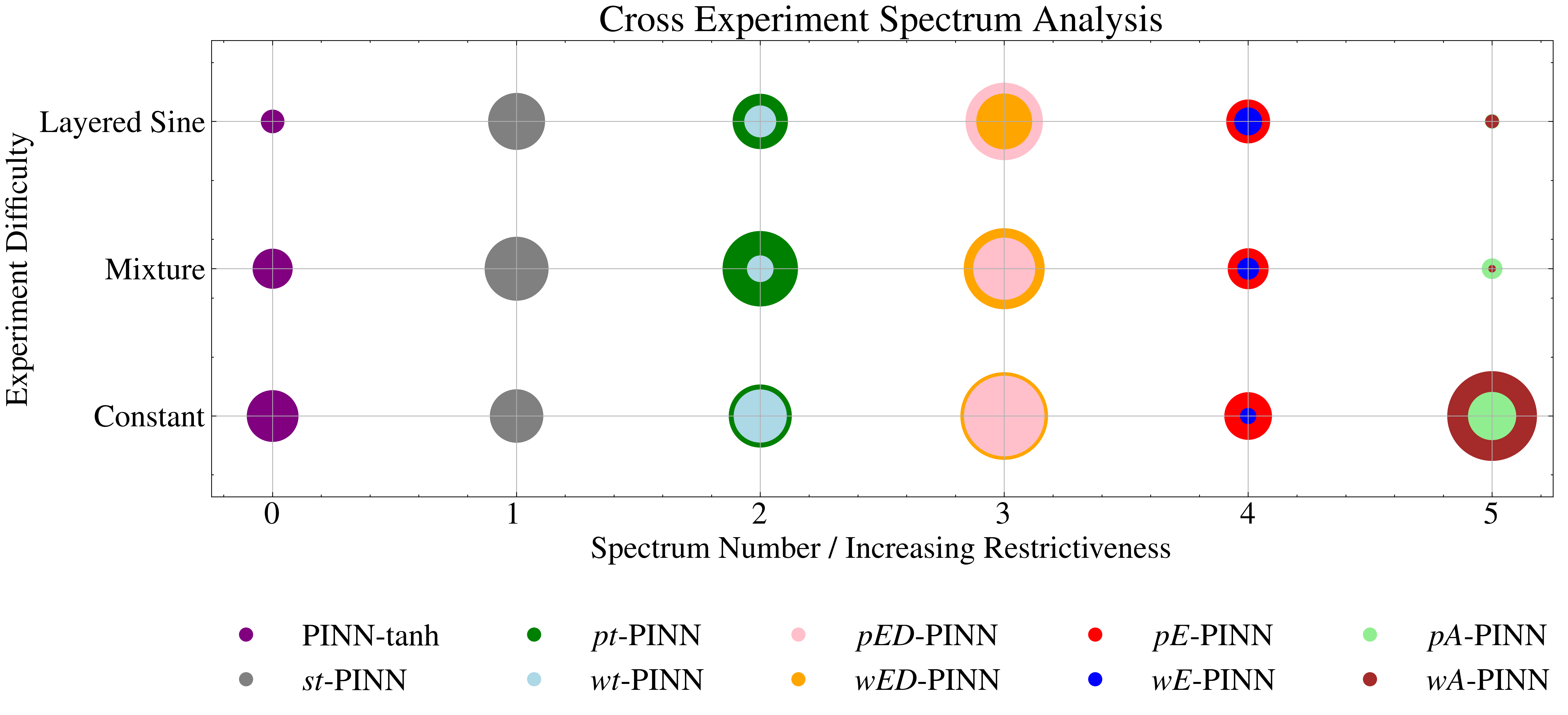}
  \caption{Accuracy analysis across architecture types with optimal hyperparameters, arranged on the y-axis by the extent of incorporated wave physics knowledge (rightward indicating more). It covers three experiments: Constant Lam{\'e} parameters, mixture, and layered models, with the y-axis reflecting experiment complexity.}
  \label{fig:bubble}
  \end{center}
\end{figure}

\paragraph{Findings}
Figure (\ref{fig:sub1}) illustrates the relative $L_2$ errors achieved by various network architectures using constant Lam{\'e} parameters ($\lambda=20, \mu=30$), arranged according to their positions on the spectrum defined in Section (\ref{spectrum_placement}). It is evident that, while performance across architectures is relatively similar, the \textit{wED}/\textit{pED}-PINNs and \textit{wA}-PINN exhibit the lowest relative $L_2$ errors. Figure (\ref{fig:sub2}) presents a comparison under the mixture model for Lam{\'e} parameters, where the \textit{wED}-PINN achieves the lowest average relative $L_2$ error. A trend is noticeable where the relative $L_2$ error decreases as we progress along the spectrum, up to the encoder-decoder type PINNs, after which it significantly increases. This pattern is similarly observed in Figure (\ref{fig:sub3}), with a decrease in relative $L_2$ error up to the encoder-decoder PINNs.
These experiments yield several insights:
\begin{itemize}
    \item Incorporating wave-physics knowledge or constraining the solution space improves PINN accuracy across all parameter settings. This is evidenced by the standard PINN-tanh architecture consistently underperforming compared to most other models, confirming that introducing prior wave physics knowledge enhances PINN model accuracy.
    \item The performance of purely encoder-based PINNs and amplitude-based PINNs varies, alternating in accuracy rankings. This suggests that overly restricting the solution space, as with encoder-type PINNs, can detrimentally affect performance. The \textit{wA}-PINN's fluctuating performance across different settings implies that model restrictiveness benefits simpler scenarios but complicates optimiser navigation in more complex cases.
    \item The observed trend suggests a decrease in $L_2$ error towards the right of the spectrum, culminating at the encoder-decoder PINNs, beyond which the error increases. This pattern highlights an optimal position within the spectrum where the integration of wave-physics knowledge significantly enhances solution accuracy. However, surpassing this optimal point introduces excessive constraints, detrimentally affecting PINN convergence.
    \item A novel architecture type, encoder-decoder PINNs with a custom wavelet/plane wave layer, outperforms all other tested models across various parameter settings. This highlights the advantage of incorporating prior physical knowledge through a flexible encoder-decoder framework.
\end{itemize}
Figure (\ref{fig:bubble}) further emphasises these findings by showing how average relative $L_2$ errors vary along the spectrum and with increasing experiment difficulty. With the bubble sizes representing the accuracy. Radii of bubbles are calculated as $r = (5e+3)\frac{1}{L_2^2}$. The \textit{wED} and \textit{pED}-PINNs maintain high acuracy across all experiments, demonstrating their superior convergence properties compared to other models that either consistently underperform or exhibit variable performance across experiments. Notably, input transformation type PINNs, particularly the \textit{pT}-PINN, also show improvement over the standard PINN, albeit not as significantly as encoder-decoder types, supporting the hypothesis of an optimal model placement along the defined spectrum. Models that failed to converge, such as MFF and MRB PINNs, are not included in these figures, as their training did not result in meaningful loss reduction.
Key observations are summarised as follows:
\begin{itemize}
\item Models lacking prior knowledge are disadvantaged, with the standard PINN consistently showing poorer convergence.
\item Overly restrictive models, such as encoder-PINNs and amplitude PINNs, yield inferior results compared to more balanced models on the spectrum.
\item Extremely restrictive models fail to converge, like MFF-PINNs and MRB-PINNs.
\item The \textit{wED} and \textit{pED}-PINNs are identified as the most accurate models across all tested parameter settings.
\end{itemize}
Addressing the impact of incorporating prior wave physics knowledge into network design on performance, readers are directed to Table (\ref{tab:results}). This table presents the average relative $L_2$ test error, compared to the FDM solution produced using DEVTIO, across various network architectures. The data reveals a reduction in average $L_2$ error, across the three experiments, from 3.34\% to 1.56\%. Additionally, by employing the appropriate encoder-decoder type PINN for each setting, a relative $L_2$ error lower than 1.46\% is consistently achievable.
\begin{table}[ht]
\centering
\begin{tabular}{lcccc}
\toprule
 & Constant & Mixture & Layered & Average \\
\midrule
PINN-tanh & 2.22\% & 2.88\% & 5.02\% & 3.37\% \\
\midrule
\textit{sT}-PINN & 2.14\% & 1.78\% & 2.00\% & 1.97\% \\
\midrule
\textit{pT}-PINN & 1.81\% & 1.51\% & 2.07\% & 1.80\% \\
\textit{wT}-PINN & 2.15\% & 4.44\% & 3.66\% & 3.42\% \\
\midrule
\textit{pED}-PINN & 1.40\% & 1.83\% & \textbf{1.46\%} & \textbf{1.56\%} \\
\textit{wED}-PINN & 1.28\% & \textbf{1.39\%} & 2.03\% & 1.57\% \\
\midrule
\textit{pE}-PINN & 2.38\% & 2.82\% & 2.62\% & 2.61\% \\
\textit{wE}-PINN & 7.54\% & 5.49\% & 4.20\% & 5.74\% \\
\midrule
\textit{pA}-PINN & 2.38\% & 5.80\% & 8.41\% & 5.53\% \\
\textit{wA}-PINN & \textbf{1.26\%} & 18.43\% & 9.08\% & 9.59\% \\
\bottomrule
\end{tabular}
\caption{Relative $L_2$ errors of all presented network architectures for the constant, mixture and layered Lam{\'e} parameter models}
\label{tab:results}
\end{table}

\subsection{Encoder-Decoder-PINNs: Extended Validation}
The preceding section provided a detailed evaluation of various network architectures over three increasingly complex parameter settings, highlighting an optimally performing encoder-decoder network enhanced by either a plane wave or wavelet layer. Our focus now shifts to a detailed comparison between this leading model and the conventional PINN approach. 
By analyzing wavefield predictions from both the encoder-decoder-PINN and the PINN-tanh model under the three previously discussed Lam{\'e} parameter settings, we emphasize the superior performance of our novel methodology. Additionally, we explore new experimental conditions, such as a high-frequency layered scenario and two novel source functions, to assert the robustness and adaptability of our model. These experiments are designed to validate our model's effectiveness across diverse scenarios and its resilience to parameter variations, including changes in the $t_1$ parameter. Furthermore, we investigate the impact of reduced source size on model accuracy, extending the discussion from Section (\ref{strong_scaling}). The generalizability of our approach is further tested against a different underlying PDE---the acoustic wave equation. Despite its relative simplicity compared to the elastic wave equation, it is a stringent test case, underscoring our novel architecture's versatility.
Through these rigorous tests, we aim to establish our model not as a narrowly specialized solution but as a broadly applicable and superior choice for solving the elastic wave equation, consistently outperforming standard fully connected networks.

\begin{figure}
\begin{center}
\includegraphics[width=0.64\linewidth]{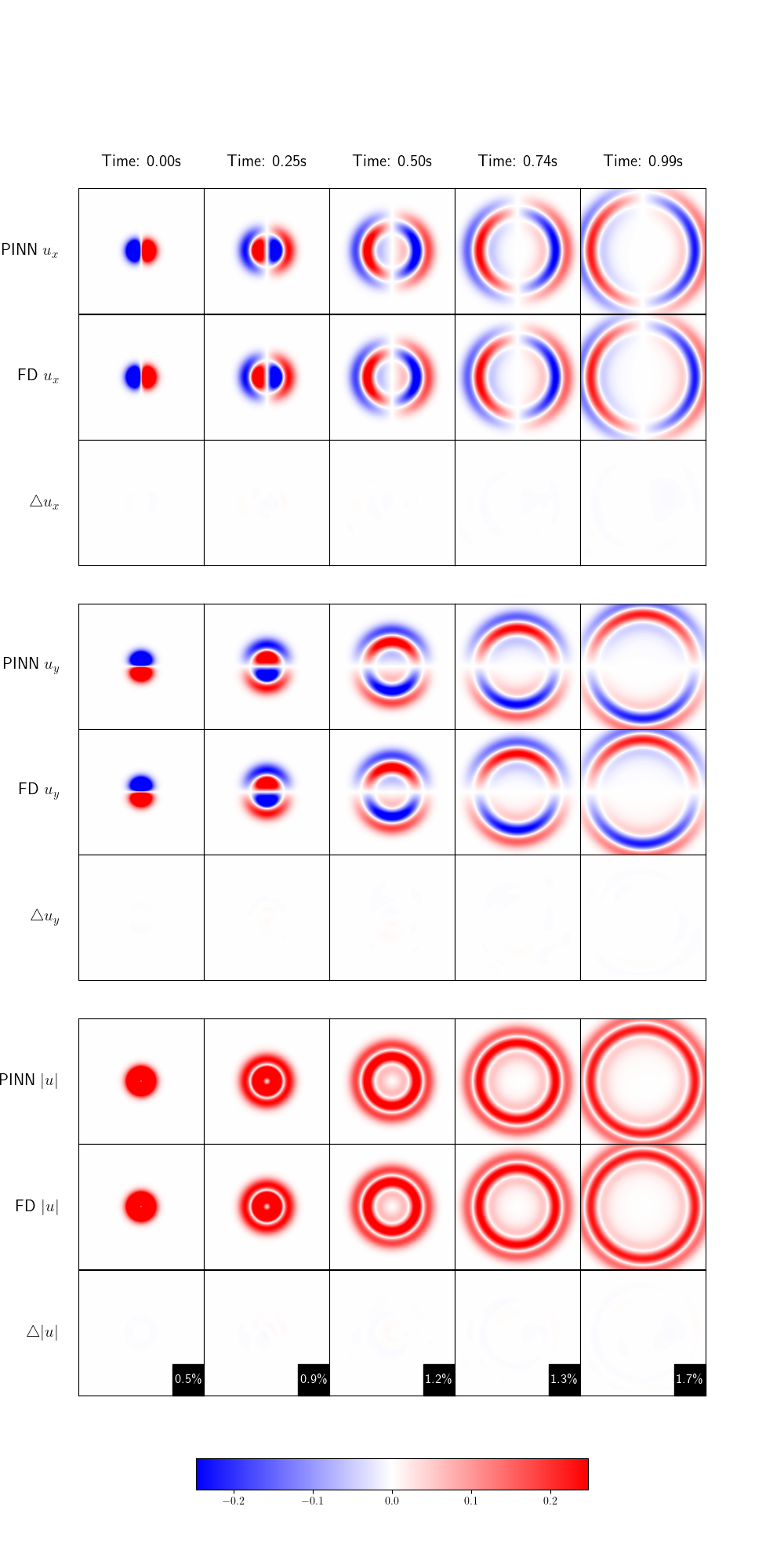}
\caption{Illustration of the outcomes achieved using optimal hyperparameter configurations with the \textit{wED}-PINN in the context of constant Lam{\'e} parameters. The data are organized into three groups of rows, each comparing the \textit{wED}-PINN predictions with our FDM reference solution for $u_x$, $u_y$, and the magnitude $|u|$, respectively}
\label{fig:novel_constant}
\end{center}
\end{figure}

\paragraph{Constant Parameters}
We initiate our comparative analysis with the constant parameter case, where the performance of the standard model has been previously detailed in Section (\ref{section:Elastic PINNs}) and illustrated in Figure (\ref{fig:vanilla_constant}). The results for the \textit{wED}-PINN are displayed in Figure (\ref{fig:novel_constant}). The hyperparameter configuration can be found in Table (\ref{tab:hyperparameters_constant}). The \textit{wED}-PINN achieved a relative $L_2$ error of $1.28\%$. Visually, the PINN approximation closely aligns with the FD solution, capturing the meeting point of the two outward-propagating wave fronts without noticeable boundary errors. Despite an increase in error alongside time and solution complexity, this increase remains marginal, with errors staying below $2\%$ at all time points. This suggests that the observed discrepancies might be attributed more to discretization and comparison errors rather than to the approximation capabilities of the PINN.

\paragraph{Variable Source Size Comparison}
In Section (\ref{strong_scaling}), the convergence behaviour of the PINN model was examined in relation to an increasing source size. We conducted the same analysis with our optimal model type the \textit{ED}-type-PINN, to compare the performance against the conventional PINN-tanh model.
\begin{figure}
\begin{center}
\includegraphics[width=0.8\linewidth]{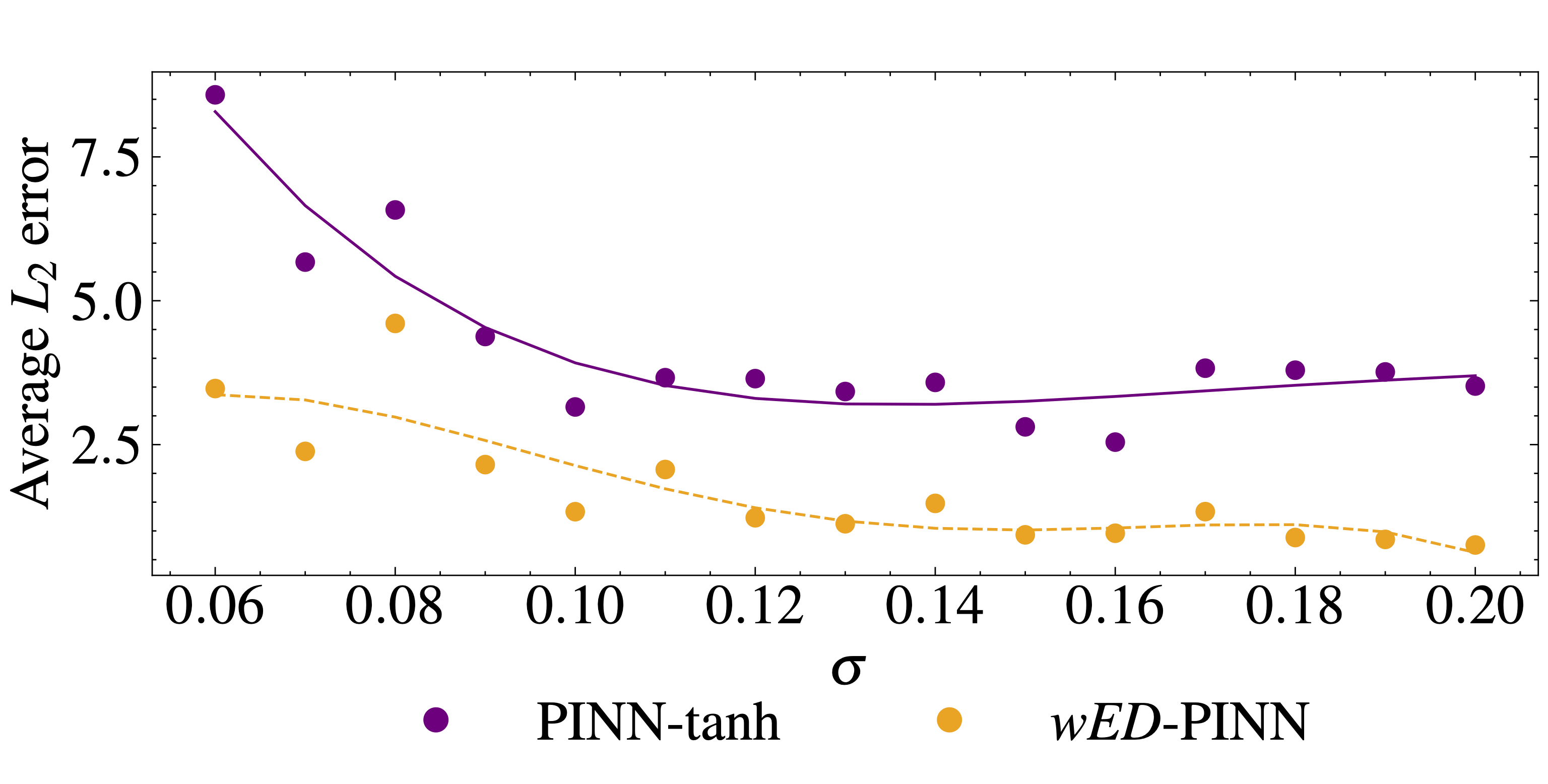}
\caption{Comparison of the performance of the standard PINN and the \textit{wED}-PINN with varying source size.}
\label{fig:strong_scaling_cross}
\end{center}
\end{figure}
Figure (\ref{fig:strong_scaling_cross}) showcases the performance comparison between the conventional PINN and our enhanced model, the \textit{wED}-PINN, across different source sizes in the constant parameter scenario. Identical experimental setups, as detailed in Section (\ref{section:exp_setup}), were employed to ensure a fair comparison. This comparative analysis yields several important insights. Firstly, the \textit{wED}-PINN consistently exhibits significantly lower $L_2$ errors for all source sizes tested, underscoring the effectiveness of encoder-decoder-PINNs. Secondly, the trend of $L_2$ error reduction with the \textit{wED}-PINN parallels findings discussed in Section (\ref{strong_scaling}): as source size increases, thereby diminishing high-frequency features in the solution, $L_2$ error correspondingly decreases. This reduction in $L_2$ error appears linear relative to the increase in $\sigma$, except for an anomaly at $\sigma = 0.08$. Notably, the \textit{wED}-PINN maintains its performance advantage without experiencing an increase in error during transitions towards solutions with very low-frequency characteristics, unlike the baseline PINN model, further demonstrating its robustness and efficacy under various conditions.

\begin{figure}
\begin{center}
\includegraphics[width=0.64\linewidth]{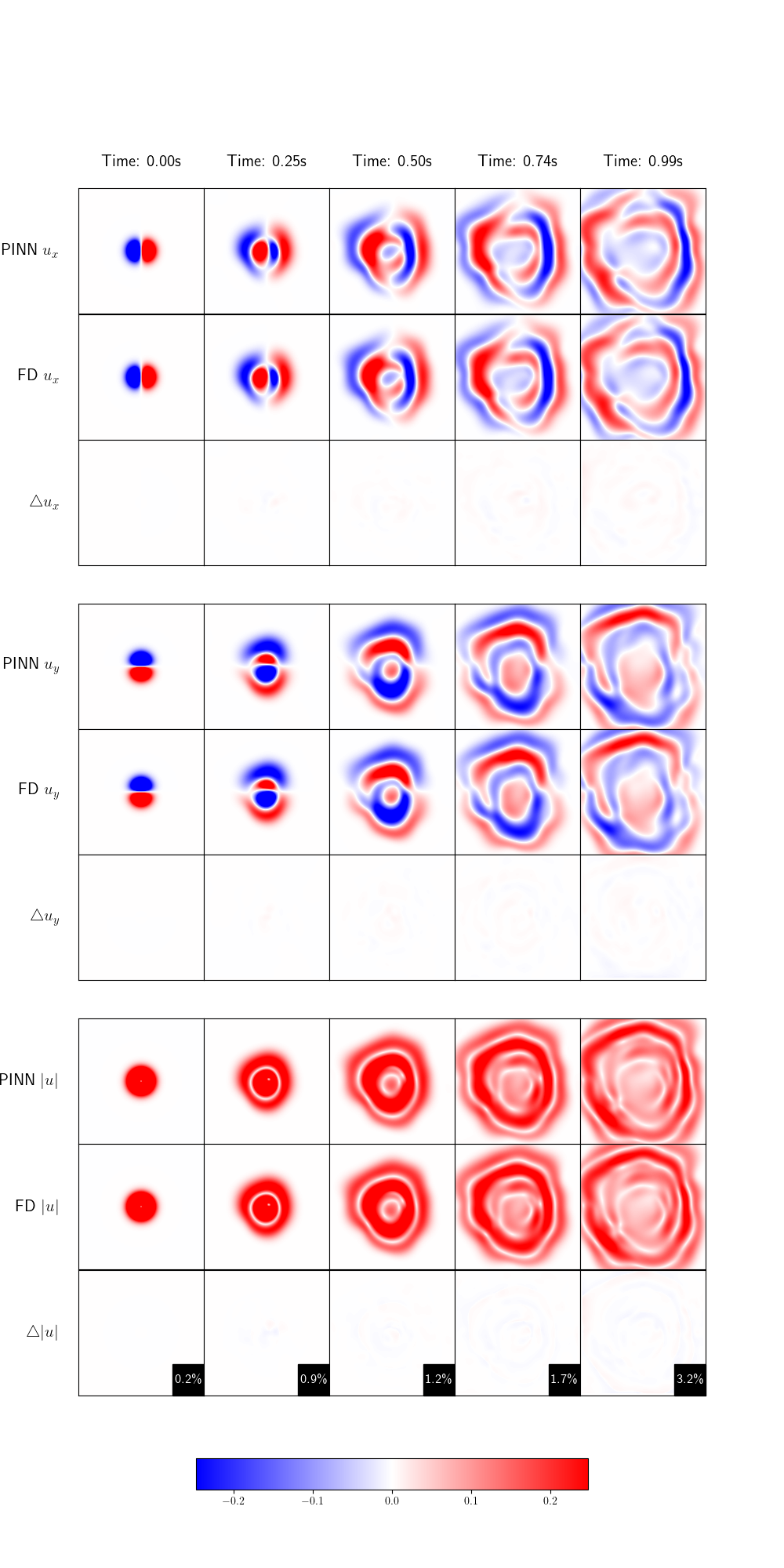}
\caption{Illustration of results obtained through optimal hyperparameter configurations for the \textit{wED}-PINN, with Lam{\'e} parameters defined by the mixture model. Data are grouped into three categories, each juxtaposing the \textit{wED}-PINN predictions against our FDM reference solution for $u_x$, $u_y$, and their magnitude $|u|$, respectively.}
\label{fig:novel_mixture}
\end{center}
\end{figure}

\paragraph{Mixture Model}
Next, we delve into the performance of the \textit{wED}-PINN in the context of the mixture model. The specific hyperparameter setup employed is detailed in Table (\ref{tab:hyperparameters_mixture}). The \textit{wED}-PINN achieved an average relative $L_2$ error of $1.39\%$. Visual comparisons, as depicted in Figure (\ref{fig:novel_mixture}), indicate minimal deviation from the FDM solution, with a slight discrepancy observable in the $u_x$ displacement field at $t=0.99$, which is considered to be of negligible impact. These findings further underscore the \textit{wED}-PINN's capacity for highly accurate and reliable simulations of complex wavefield behaviour.

\begin{figure}
\begin{center}
\includegraphics[width=0.64\linewidth]{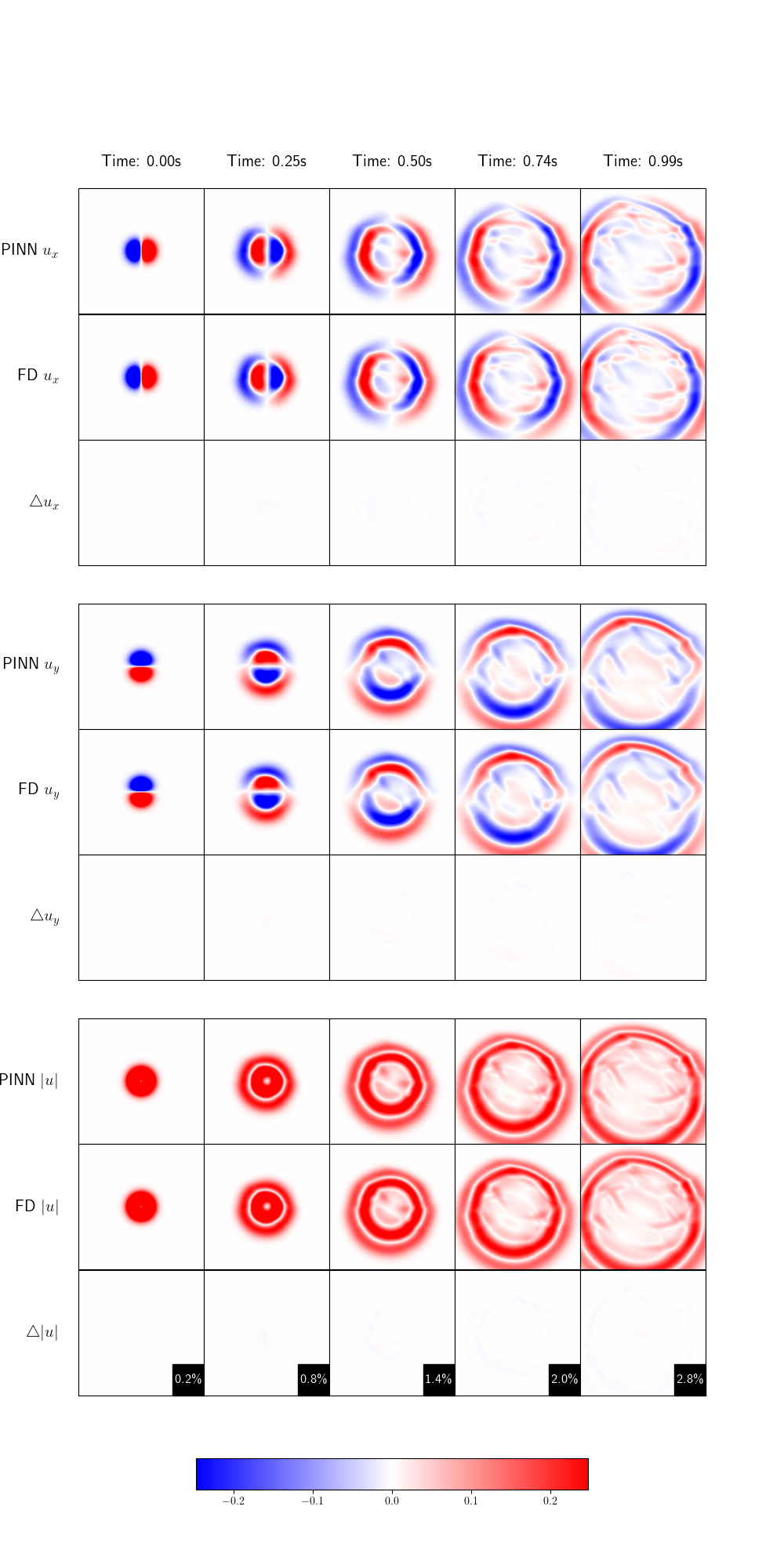}
\caption{Illustration of results obtained through optimal hyperparameter configurations for the \textit{pED}-PINN, with Lam{\'e} parameters defined by the layered model. Data are organized into three categories, each comparing the \textit{pED}-PINN predictions with our FDM reference solution for $u_x$, $u_y$, and their magnitude $|u|$, respectively.}
\label{fig:novel_layered}
\end{center}
\end{figure}

\paragraph{Layered Model}
The performance of the \textit{pED}-PINN within a layered Lam{\'e} parameter model is illustrated in Figure (\ref{fig:novel_layered}). The specific hyperparameter configuration utilized is detailed in Table (\ref{tab:hyperparameters_layered}). We report an average relative $L_2$ error of $1.46\%$, highlighting that despite its reduced size compared to the conventional PINN, the \textit{pED}-PINN attains a relative $L_2$ error that is approximately three and a half times lower than that of the standard model. The comparative analysis presented in Figure (\ref{fig:novel_layered}) reveals negligible differences between the FDM and \textit{pED}-PINN approximations, in stark contrast to the significant deviations observed with the standard PINN approach. This observation emphatically demonstrates the \textit{pED}-PINN's capability to accurately model the intricate dynamics of layered media, thereby affirming the enhanced effectiveness of an encoder-decoder network architecture equipped with specialized layers for diverse wave propagation scenarios.

\begin{figure} 
\begin{subfigure}{.45\textwidth} 
  \centering
  \includegraphics[width=\linewidth]{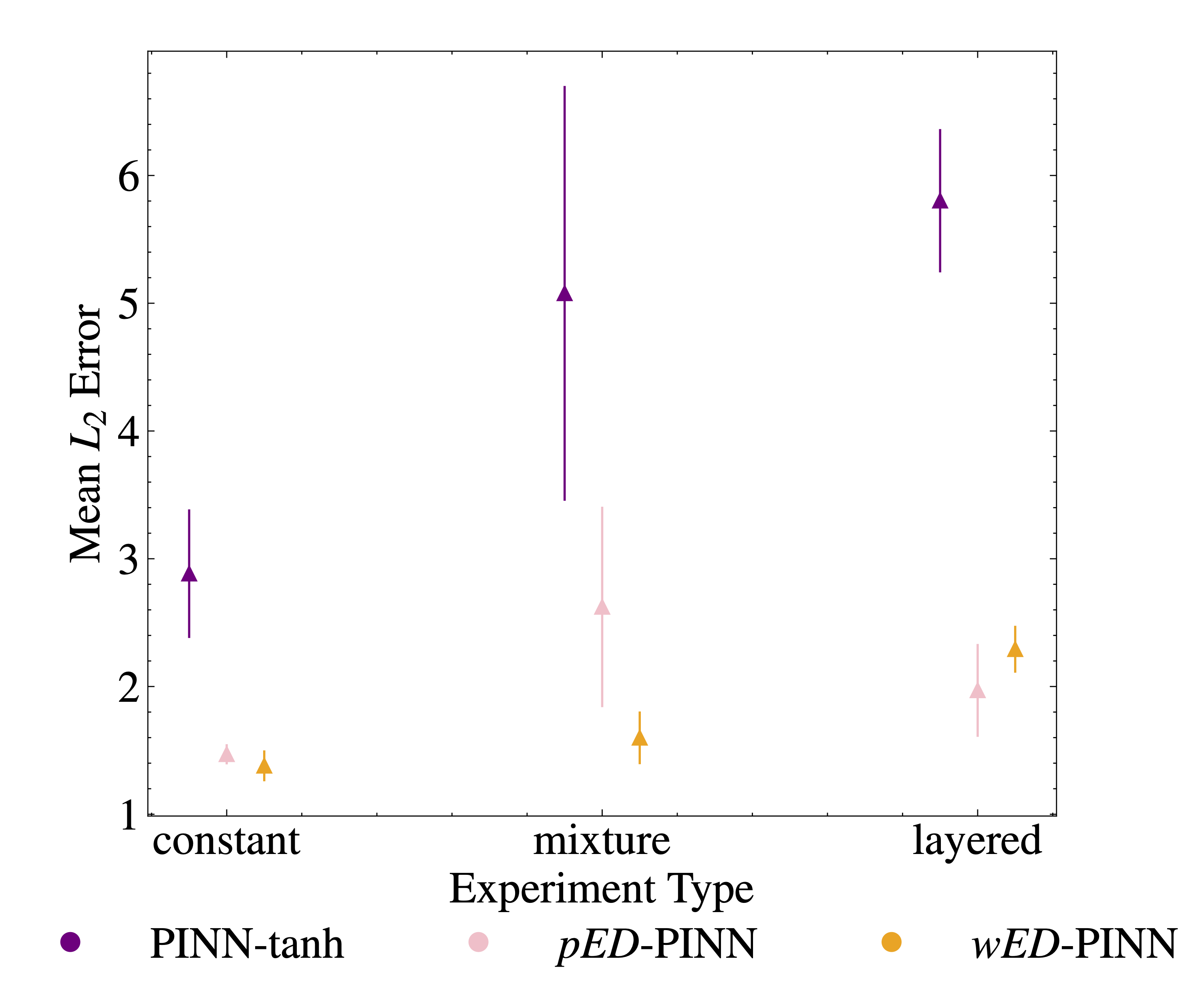}
  \caption{Mean (triangles) and variation of the average relative $L_2$ error. }
  \label{fig:t1_variance}
\end{subfigure}\hfill
\begin{subfigure}{.45\textwidth}
  \centering
  \includegraphics[width=\linewidth]{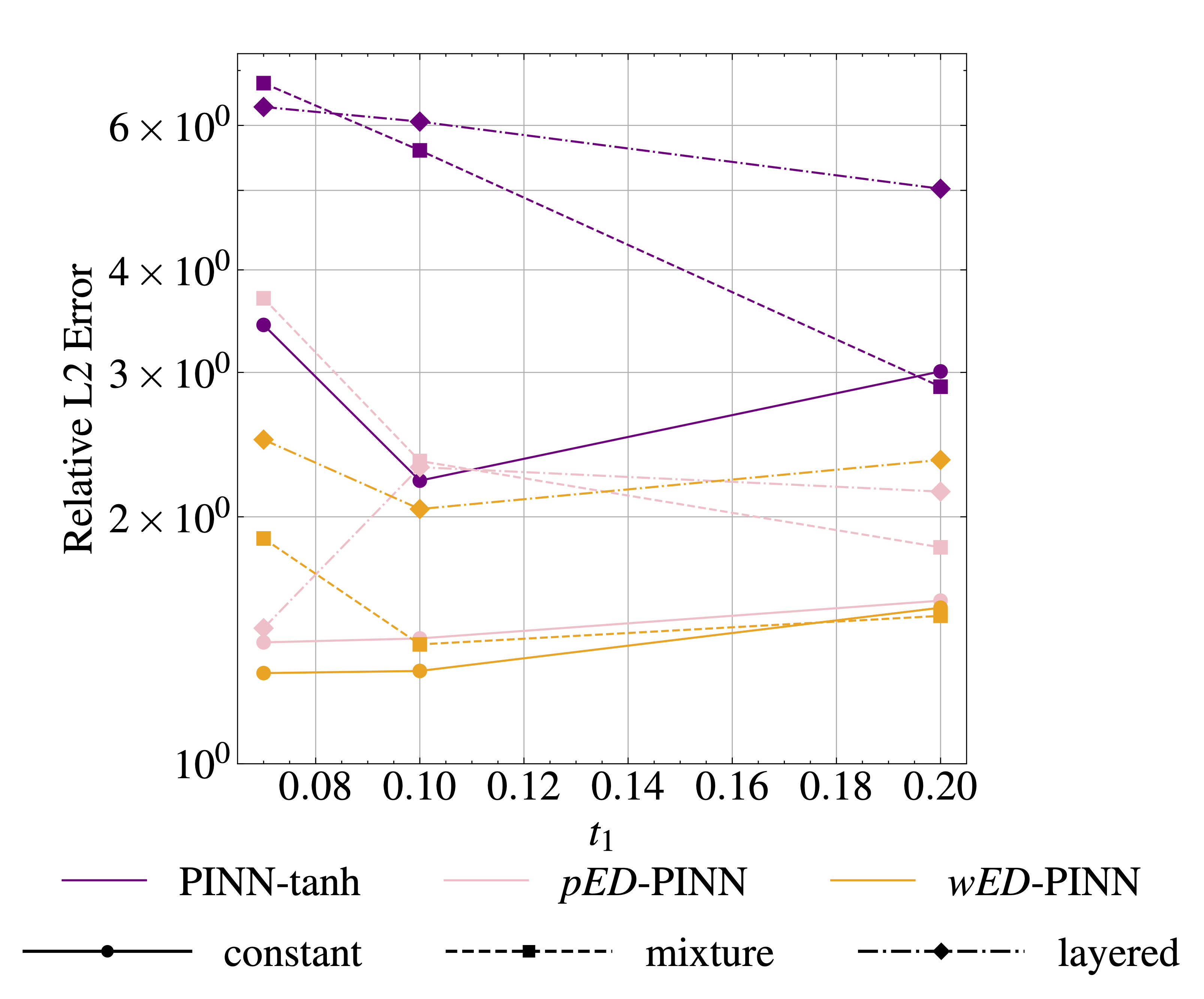}
  \caption{Average relative $L_2$ error evaluated at different $t_1$ values}
  \label{fig:function}
\end{subfigure}
\caption{Comparison of $t_1$ sensitivity of the PINN-tanh and encoder-decoder models across three different Lam{\'e} parameter models (constant, mixture, layered). Three $t_1$ values are considered: $t_1 \in[0.07,0.1,0.2]$.}
\label{fig:t1_both}
\end{figure}

\paragraph{$t_1$ sensitivity} 
The sensitivity of the PINN-tanh model to variations in the $t_1$ parameter was examined in Section (\ref{section:t1_vanilla}), with three distinct values tested: $t_1 \in [0.07, 0.1, 0.2]$. This analysis is extended to compare the $t_1$ sensitivity of encoder-decoder-PINNs against that of standard PINNs. A model's robustness is ideally demonstrated by its consistent performance across a diverse range of $t_1$ values. Figure (\ref{fig:t1_both}) showcases the comparative sensitivity of both standard PINNs and encoder-decoder PINNs to $t_1$ variations. Notably, ED-type PINNs show a markedly reduced sensitivity to changes in $t_1$, a trait that significantly bolsters their applicability and effectiveness in varied computational scenarios.
Crucially, the standard PINN's performance in response to $t_1$ adjustments does not exhibit a uniform pattern in terms of convergence. Notably, an increase in $t_1$ does not necessarily translate to improved convergence. For example, in the constant setting experiment, an optimal outcome was achieved at a $t_1$ of $0.1$, while in the layered and mixture model experiments, superior relative $L_2$ errors were observed at a $t_1$ of $0.2$. This variability highlights the importance of extensive experimentation to ascertain the most effective hyperparameters and pinpoint the optimal $t_1$ value. Although identifying an optimal $t_1$ is a demanding process, the utilization of an encoder-decoder-PINN can alleviate this burden. This novel network design consistently delivers strong performance across a spectrum of $t_1$ values, as evidenced by the reduced variance in relative $L_2$ errors shown in Figure (\ref{fig:t1_variance}), thereby streamlining the hyperparameter optimization process.

\paragraph{More Complex Seismic Sources}
In this set of experiments, we modify the characteristics of the seismic source by adjusting specific initial conditions. It is crucial to clarify that these modified sources, while not aiming to mimic realistic seismic events, are designed as conceptual tools to present a more challenging test environment for our models.

\begin{figure}
\begin{subfigure}{.5\textwidth}
  \centering
  \includegraphics[width=\linewidth]{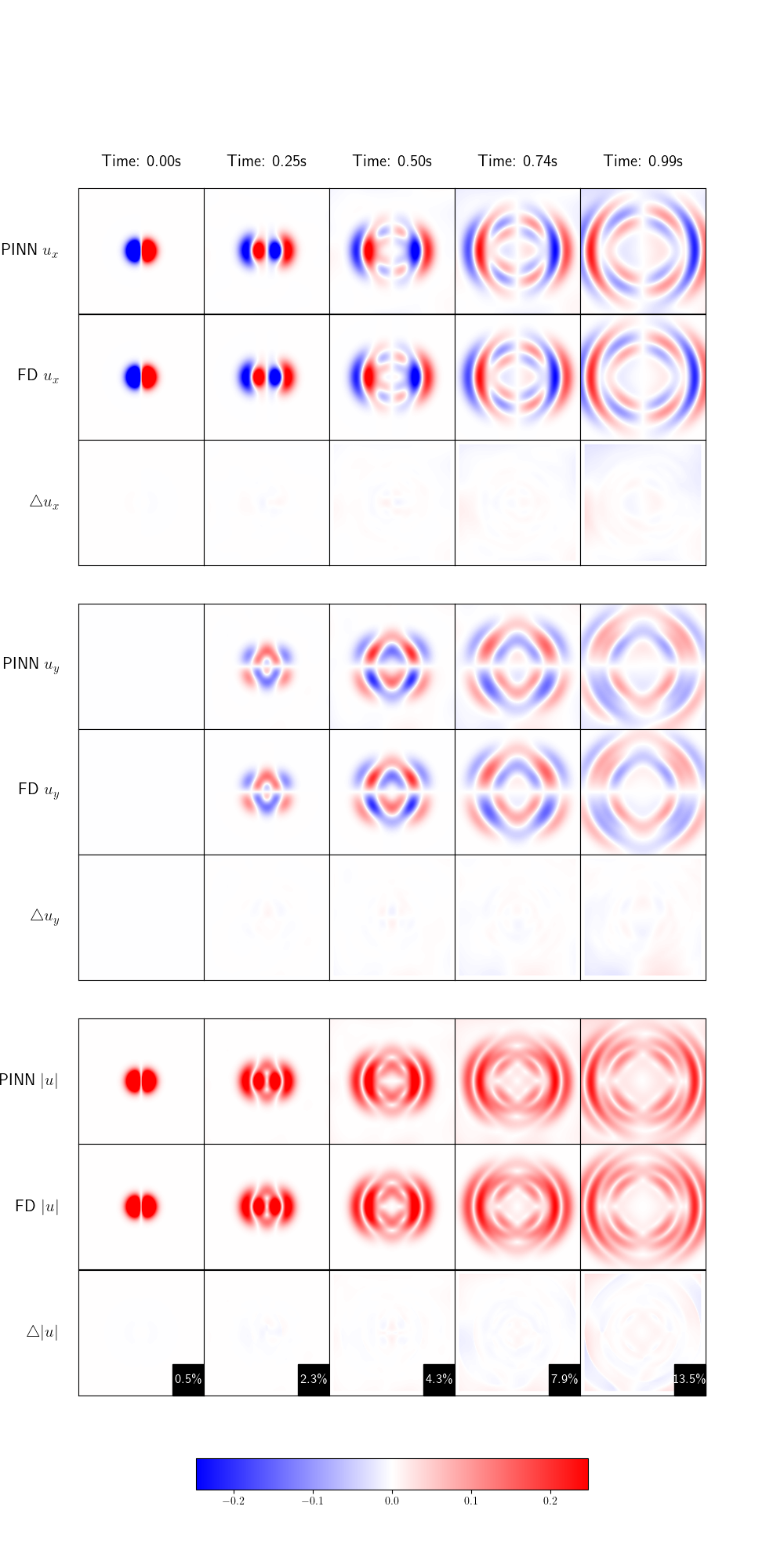}
  \caption{PINN-tanh model}
\end{subfigure}
\begin{subfigure}{.5\textwidth}
  \centering
  \includegraphics[width=\linewidth]{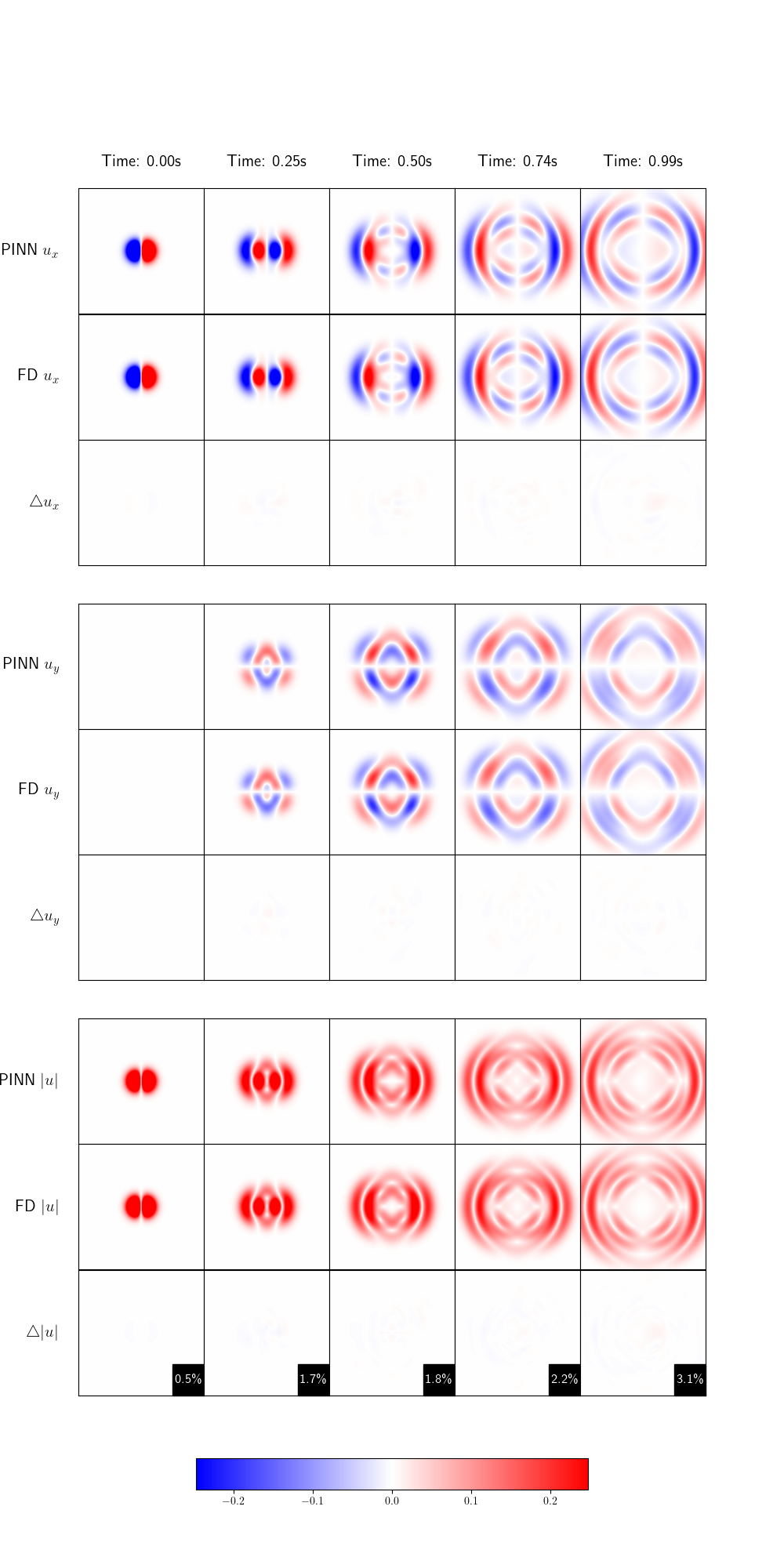}
  \caption{\textit{wED}-PINN model}
\end{subfigure}
\caption{Comparison of elastic wavefield predictions: The PINN-tanh model (left) and the \textit{wED}-PINN model (right), both utilizing constant underlying Lam{\'e} parameters and implementing the first novel seismic source, are juxtaposed with the FDM solution computed using DEVITO.}
\label{fig:new_source1}
\end{figure}

Two novel sources were introduced for evaluation. The first alteration involved maintaining the $u_x$ term constant while setting the initial displacement of the $u_y$ term to zero, effectively simulating a pure latitudinal shift. This adjustment led to the emergence of complex wavefield patterns, particularly facilitating the visual distinction between P-waves and S-waves, a differentiation more pronounced than that observed with traditional explosive sources. The experimental setup follows the protocol delineated in Section (\ref{section:exp_setup}). Both models were tested across a spectrum of three $t_1$ values: $t_1 \in [0.07, 0.1, 0.2]$. It was determined that the optimal $t_1$ value for both models is $0.1$. Beyond this, the hyperparameter configurations for the PINN-tanh and the \textit{wED}-PINN models are consistent with those outlined in Table (\ref{tab:hyperparameters_constant}).
Figure (\ref{fig:new_source1}) contrasts the outcomes from the standard PINN model, which incurred a relative $L_2$ error of $5.36\%$, with those from the \textit{wED}-PINN model, which demonstrated a relative $L_2$ error of $1.93\%$. This comparison highlights a substantial difference in accuracy, showcasing the limitations of the standard PINN approach in accurately capturing both P-wave and S-wave components across the displacement field, especially near spatial boundaries. Conversely, the \textit{wED}-PINN model exhibits remarkable precision, with only minor discrepancies in the $u_x$ displacement field at the final timestep. These minor errors, located at the domain's centre, are significantly less pronounced than those observed with the standard PINN method, affirming the \textit{wED}-PINN's superior performance and reliability.

\begin{figure}
\begin{subfigure}{.5\textwidth}
  \centering
  \includegraphics[width=\linewidth]{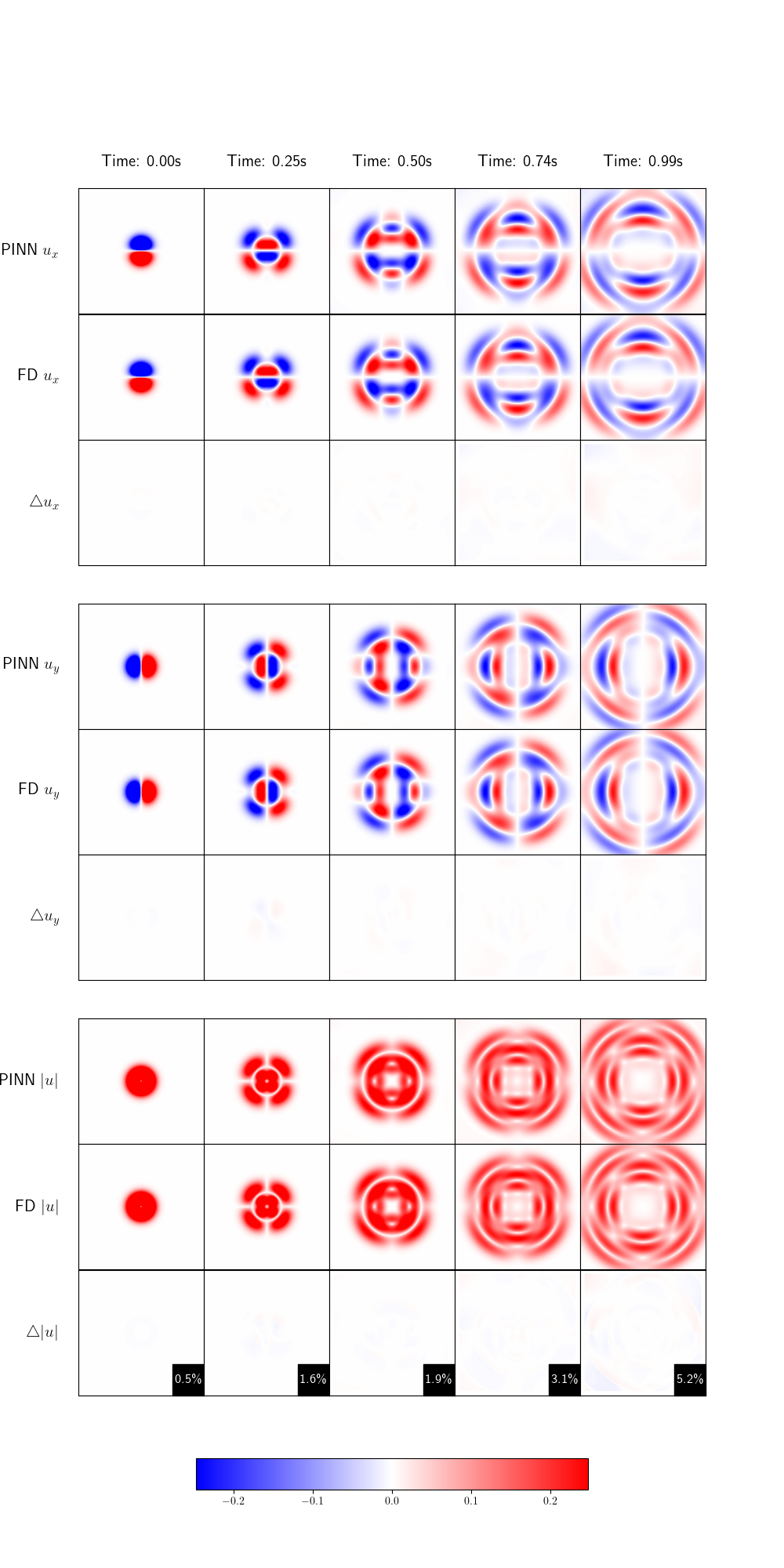}
  \caption{PINN-tanh model}
\end{subfigure}%
\begin{subfigure}{.5\textwidth}
  \centering
  \includegraphics[width=\linewidth]{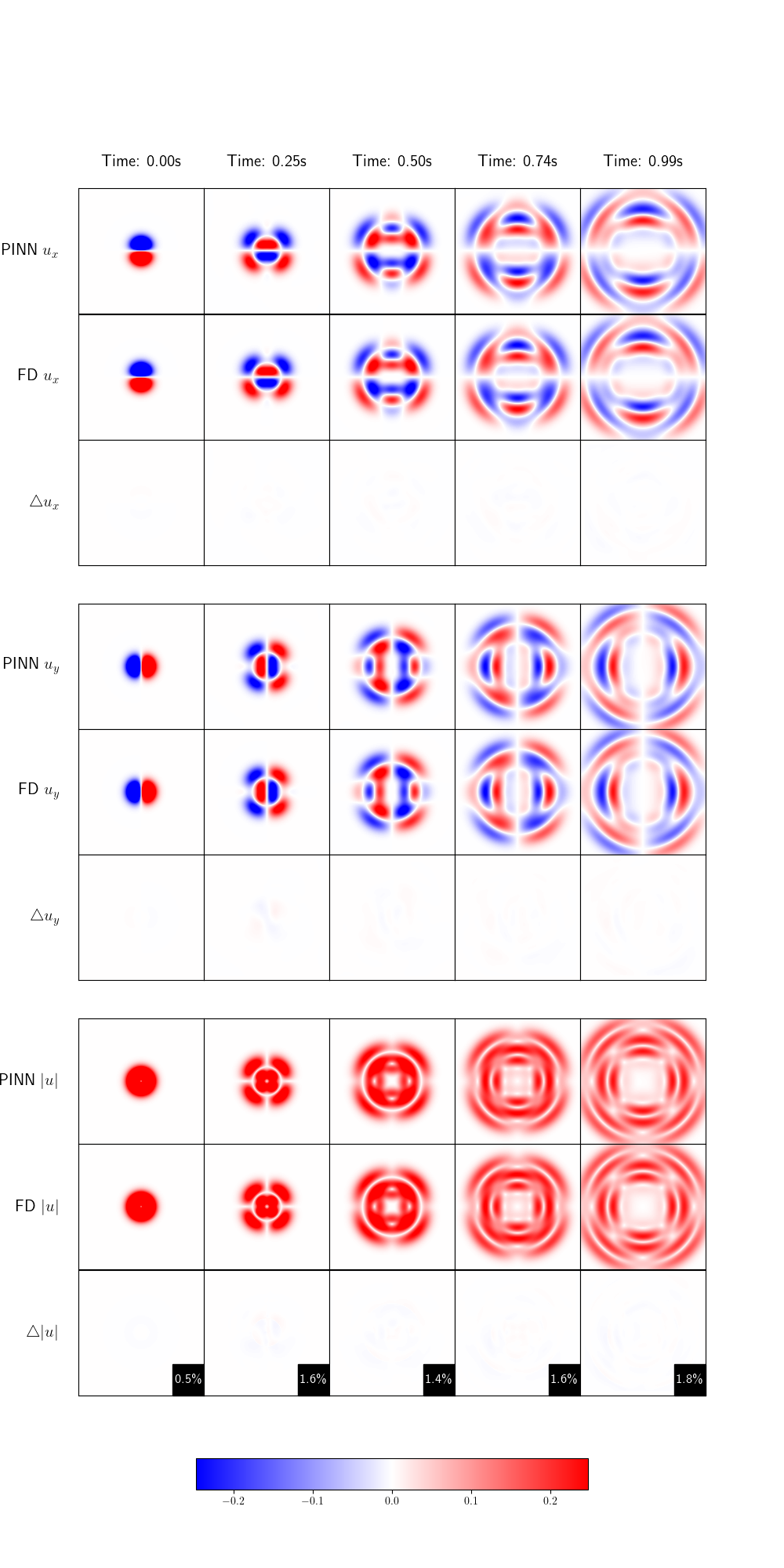}
  \caption{\textit{wED}-PINN model}
\end{subfigure}
\caption{Elastic wavefield predictions produced by the PINN-tanh model (left) and the \textit{wED}-PINN model (right), both incorporating constant underlying Lam{\'e} parameters and employing the second novel seismic source, are compared against the FDM solution computed using DEVITO.}
\label{fig:new_source2}
\end{figure}

A second modification involved swapping the initial displacement fields of $u_x$ and $u_y$. The comparative analysis of this adjustment, as depicted in Figure (\ref{fig:new_source2}), follows the experimental setup and hyperparameter configuration established for the first novel source, with the distinction that, for this second novel source, the optimal $t_1$ value for both models has been identified as $t_1 = 0.2$. While the accuracy discrepancy between the models for this second source variant is less marked, the \textit{wED}-PINN still outperforms the PINN-tanh model. The \textit{wED}-PINN achieved an average relative $L_2$ error of $1.4\%$, significantly enhancing accuracy over the PINN-tanh model, which recorded a relative $L_2$ error of $2.36\%$. These findings further validate the \textit{wED}-PINN's effectiveness in accommodating complex seismic source configurations.

\begin{figure}
\begin{subfigure}{.5\textwidth}
  \centering
  \includegraphics[width=\linewidth]{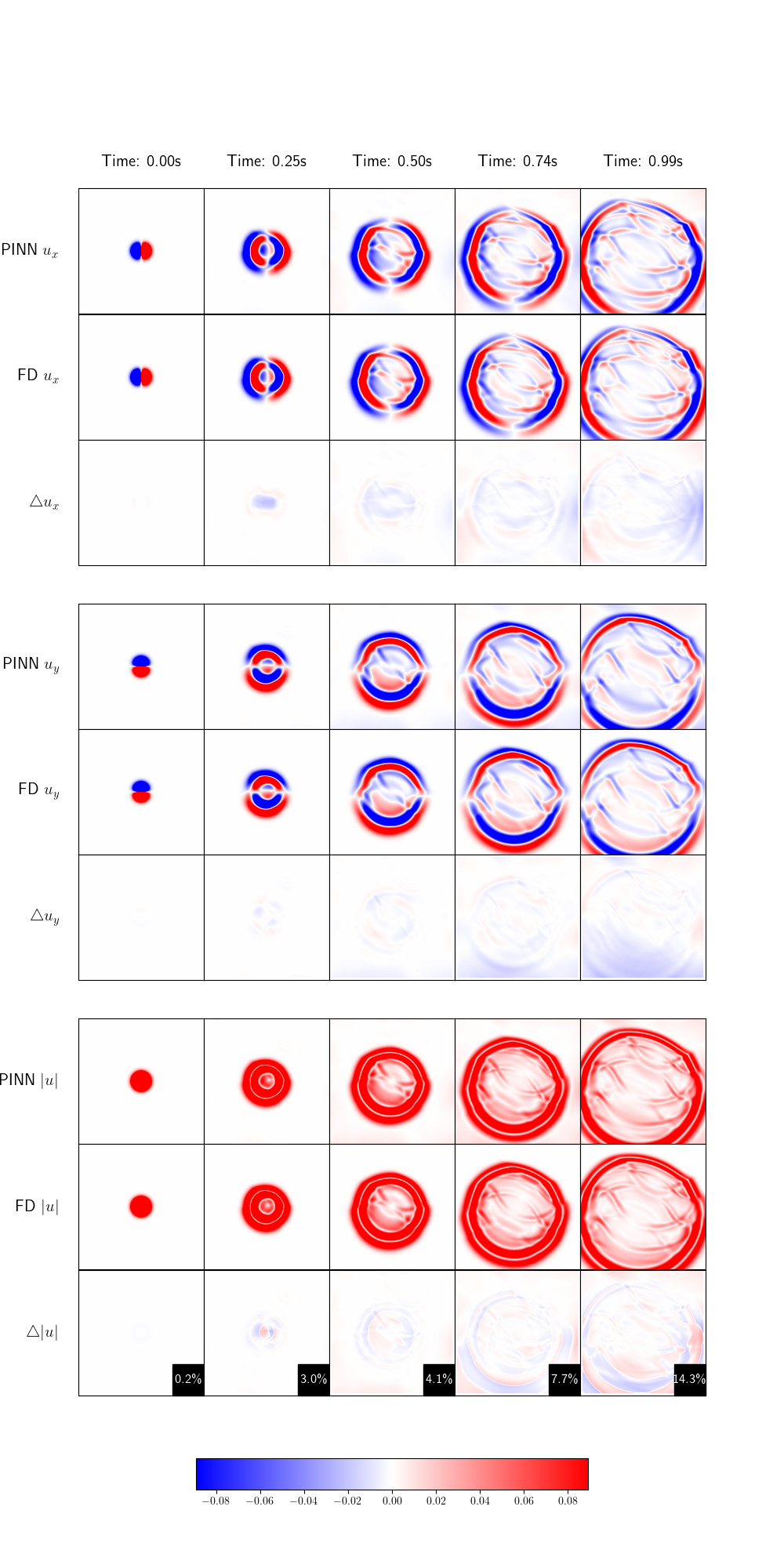}
  \caption{PINN-tanh model}
\end{subfigure}%
\begin{subfigure}{.5\textwidth}
  \centering
  \includegraphics[width=\linewidth]{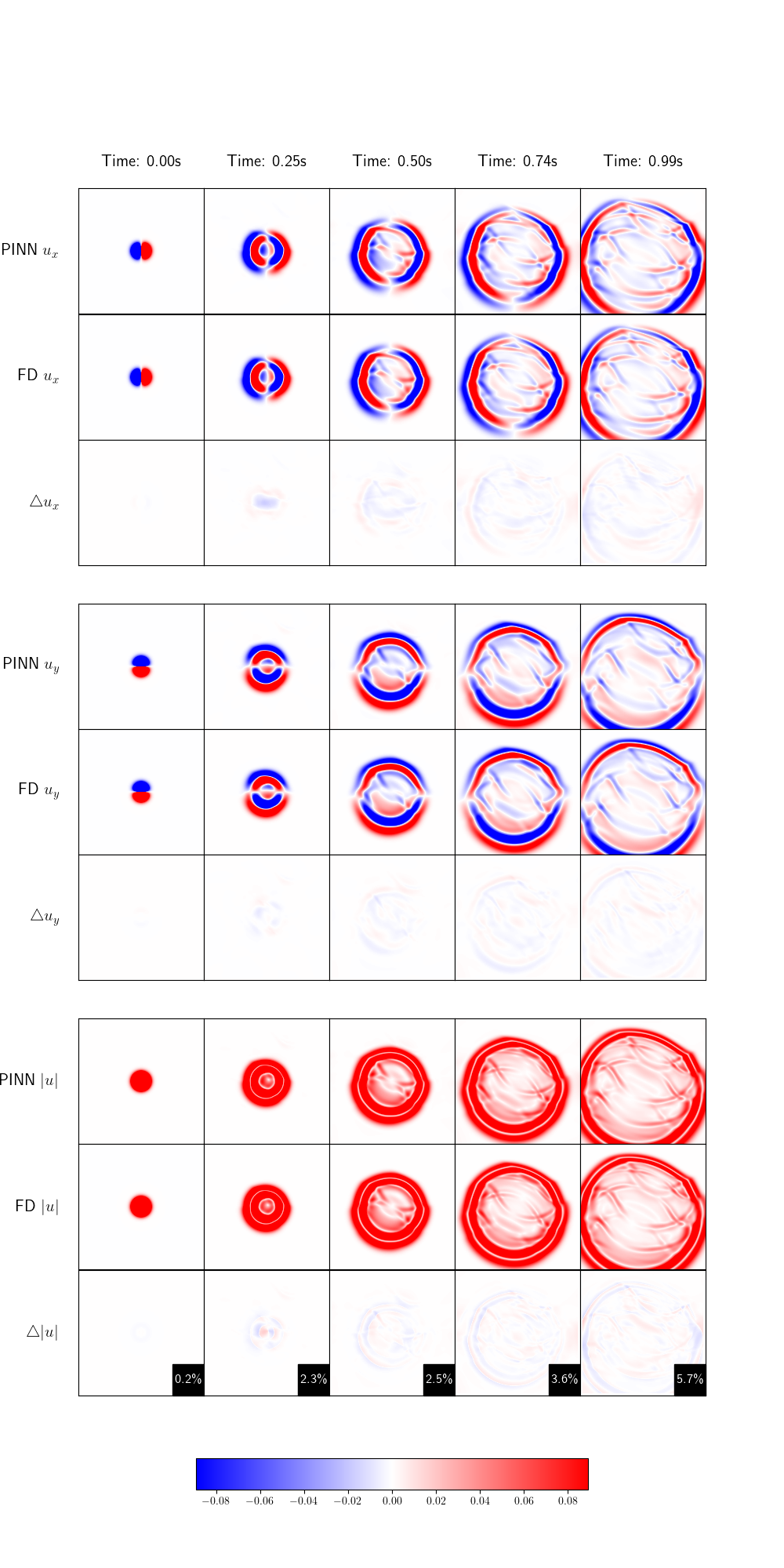}
  \caption{\textit{pED}-PINN model}
\end{subfigure}
\caption{Elastic wavefield predictions generated by the PINN-tanh model (left) and the \textit{pED}-PINN model (right), both employing a layered model for the underlying Lam{\'e} parameters, along with a smaller seismic source characterised by a standard deviation of $\sigma=0.06$, are contrasted with the FDM solution derived using DEVITO.}
\label{fig:high_frequency_layered}
\end{figure}

\paragraph{High Frequency Layered Model}
In this experiment, we revisit the layered model previously discussed, this time incorporating a seismic source characterized by a significantly smaller standard deviation, $\sigma = 0.06$, as opposed to the conventional $\sigma = 0.1$. This modification aims to evaluate the models' proficiency in handling the increase of high-frequency components alongside complex phenomena such as reflection and refraction. Figure (\ref{fig:high_frequency_layered}) presents a comparison of the outcomes from both models, unveiling several critical observations.

The standard PINN-tanh model exhibits notable deficiencies beyond its recurrent boundary issues. Specifically, the entire lower half of the $u_y$ displacement field at the final timestep shows a considerable deviation from the expected results, signifying a substantial drop in accuracy. This decrease is quantitatively evidenced by a relative $L_2$ error approaching $15\%$. Additionally, the model's inability to capture numerous details associated with reflections accentuates its limitations in complex scenarios.

In contrast, the \textit{pED}-PINN, despite registering an average relative $L_2$ error of $ 2.94\%$---higher than in other experimental contexts---does not manifest the severe deficiencies observed in the standard PINN-tanh model. The discrepancies between the FD solution and the \textit{pED}-PINN predictions are minor, demanding careful scrutiny to identify. This superior performance of the \textit{pED}-PINN highlights its advanced capability in accurately simulating high-frequency seismic wave propagation and interactions within intricate geological formations. Such proficiency affirms the \textit{pED}-PINN's reliability and its contribution to a deeper, more precise understanding of the involved physical phenomena.

\paragraph{Consolidated Results Summary}
Before transitioning to the acoustic wave equation, it is pertinent to succinctly recapitulate the findings from the series of experiments conducted on the elastic wave equation. 
Table (\ref{tab:summary}) provides a comprehensive overview of the outcomes from all conducted experiments, ranging from those with constant parameters to the layered model featuring a smaller seismic source. The data demonstrates the superior performance of our novel architecture over the standard PINN-tanh across every test scenario. Notably, the average $L_2$ error for the standard PINN remains relatively modest across various experiments, with only the most challenging setup---the layered model with a smaller source---exhibiting an $L_2$ error exceeding $2\%$. This consistency underscores the robustness of our novel architecture, highlighting its enhanced ability to accurately model complex seismic phenomena under various conditions, thereby establishing a new benchmark for performance in the field.

\begin{table}[ht]
\centering
\begin{tabular}{lcc}
\toprule
 & PINN-tanh & Encoder-decoder-PINN \\
\midrule
Constant & $2.22\%$ & $1.28\%^w$ \\
\midrule
Mixture & $2.88\%$ & $1.39\%^w$ \\
\midrule
Layered & $5.02\%$ & $1.46\%^p$ \\
\midrule
Constant - Source 2 & $5.36\%$ & $1.93\%^w$ \\
\midrule
Constant - Source 3 & $2.36\%$ & $1.40\%^w$ \\
\midrule
Layered - Source 2 & $5.59\%$ & $2.94\%^p$ \\
\bottomrule
\end{tabular}
\caption{Comparative summary of relative $L_2$ errors achieved by the baseline PINN (PINN-tanh) model and the encoder-decoder PINN model across different scenarios. The analysis spans from constant, mixture, and layered Lam{\'e} parameter cases to the first and second novel seismic sources, concluding with the high-frequency layered case. Within the encoder-decoder PINN category, the superscripts $^w$ and $^p$ indicate the use of \textit{wED}-PINNs and \textit{pED}-PINNs, respectively.}
\label{tab:summary}
\end{table}

It is noteworthy that \textit{wED}-PINNs generally outperform \textit{pED}-PINNs in scenarios without layers (constant and mixture model Lam{\'e} parameters), while the reverse holds true in layered contexts. This phenomenon suggests that plane waves are more capable of modelling reflections and refractions than wavelets.

\paragraph{Solving the Acoustic Wave Equation}
\label{section:arch_acoustic}
Previously, it was observed that encoder-decoder-PINNs can accommodate different source terms and tackle complex experiments. The objective now is to assess whether this approach is excessively complex and limited solely to the elastic wave equation or if it can be effectively applied to the acoustic wave equation, a special and simplified variant of the elastic wave equation. While solving the acoustic wave equation with PINNs has been 
successfully achieved before \cite{Ben} \cite{1Dacoustic} \cite{2DFWI} \cite{2Dacoustic_IC} \cite{SA-PINNS}, our goal is not to introduce a novel PINN model for the acoustic wave equation.
Instead, we aim to evaluate the performance of our novel encoder-decoder type PINN for this particular case. The rationale behind this is that our innovative model embeds prior knowledge of wave physics into the network architecture. Similar to the elastic wave equation, the acoustic wave equation deals with wavefields, though in this instance, the focus is on scalar pressure fields rather than on vectorial displacement fields. The properties of wavefield solutions discussed in Section (\ref{section:Arch_Motivation}) remain applicable to the acoustic wave equation (with certain exceptions, such as polarity), given that the acoustic wave equation can be considered a simplification or a specific instance of the elastic wave equation. The acoustic wave equation centres on the propagation of sound or pressure (P) waves, involving the compression and rarefaction of the medium, but crucially, without shear deformation. As such, the second Lame parameter, $\mu$, does not play a role in wave propagation. Under the assumptions of rotational motion \cite{acoustic_to_elastic} and isotropy, we derive the following scalar equation:
\begin{equation}
\frac{\partial^2\phi}{\partial t^2} = c^2\nabla^2\phi,
\end{equation}
known as the acoustic wave equation, where $c^2$ represents the variable speed of sound. It is important to emphasise that $\phi$, the pressure field, is a scalar field in contrast to the vectorial displacement field. The acoustic wave equation is sometimes preferred over the elastic wave equations for simulating seismic wave propagations due to its reduced complexity and lower computational cost, albeit with limited accuracy, as seismic waves, in reality, are not sound but elastic waves.

\begin{figure} 
\begin{subfigure}{.5\textwidth}
  \centering
  \includegraphics[width=\linewidth]{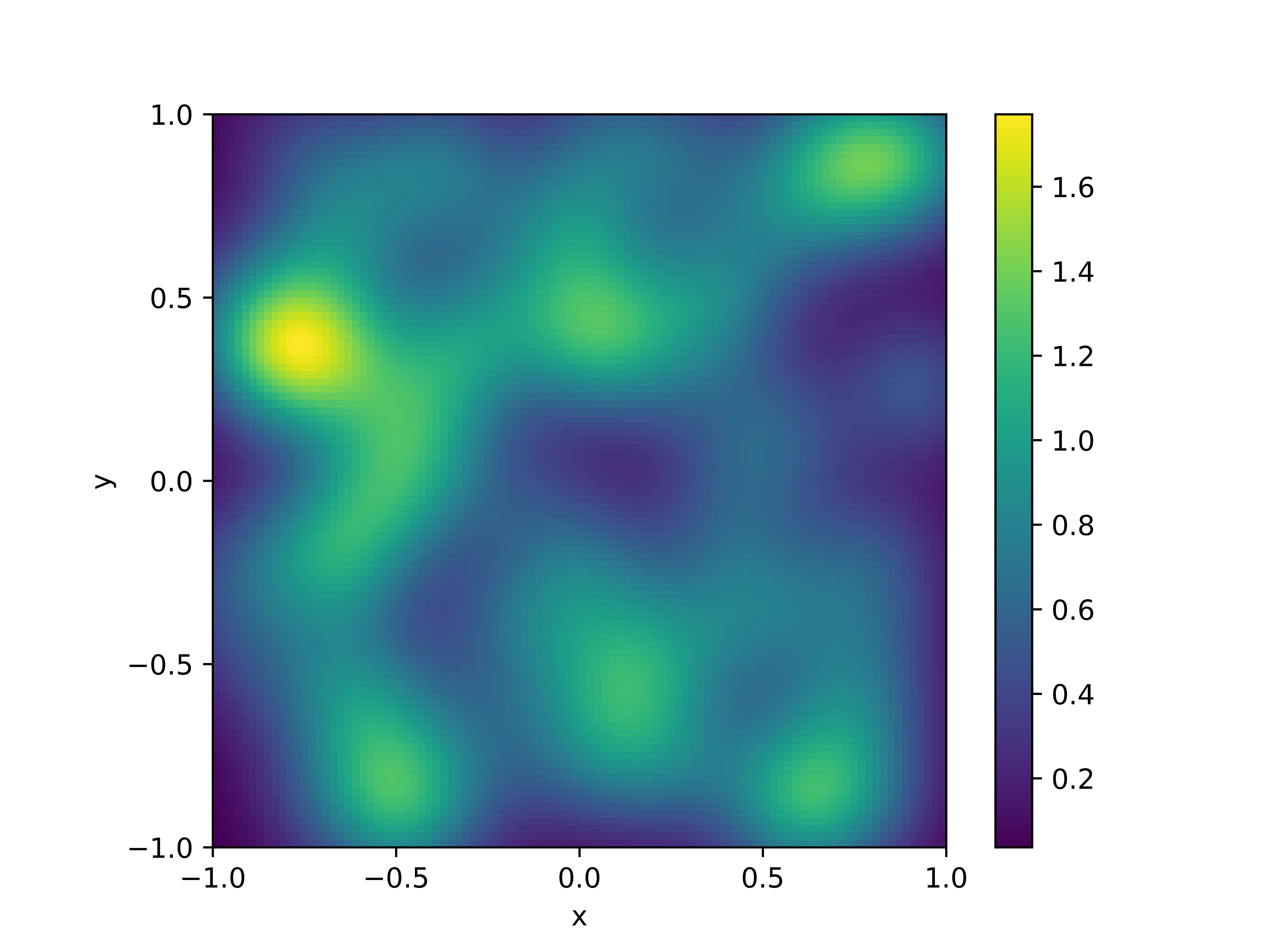}
  \caption{Mixture model}
  \label{fig:mixture_model_acc}
\end{subfigure}%
\begin{subfigure}{.5\textwidth}
  \centering
  \includegraphics[width=\linewidth]{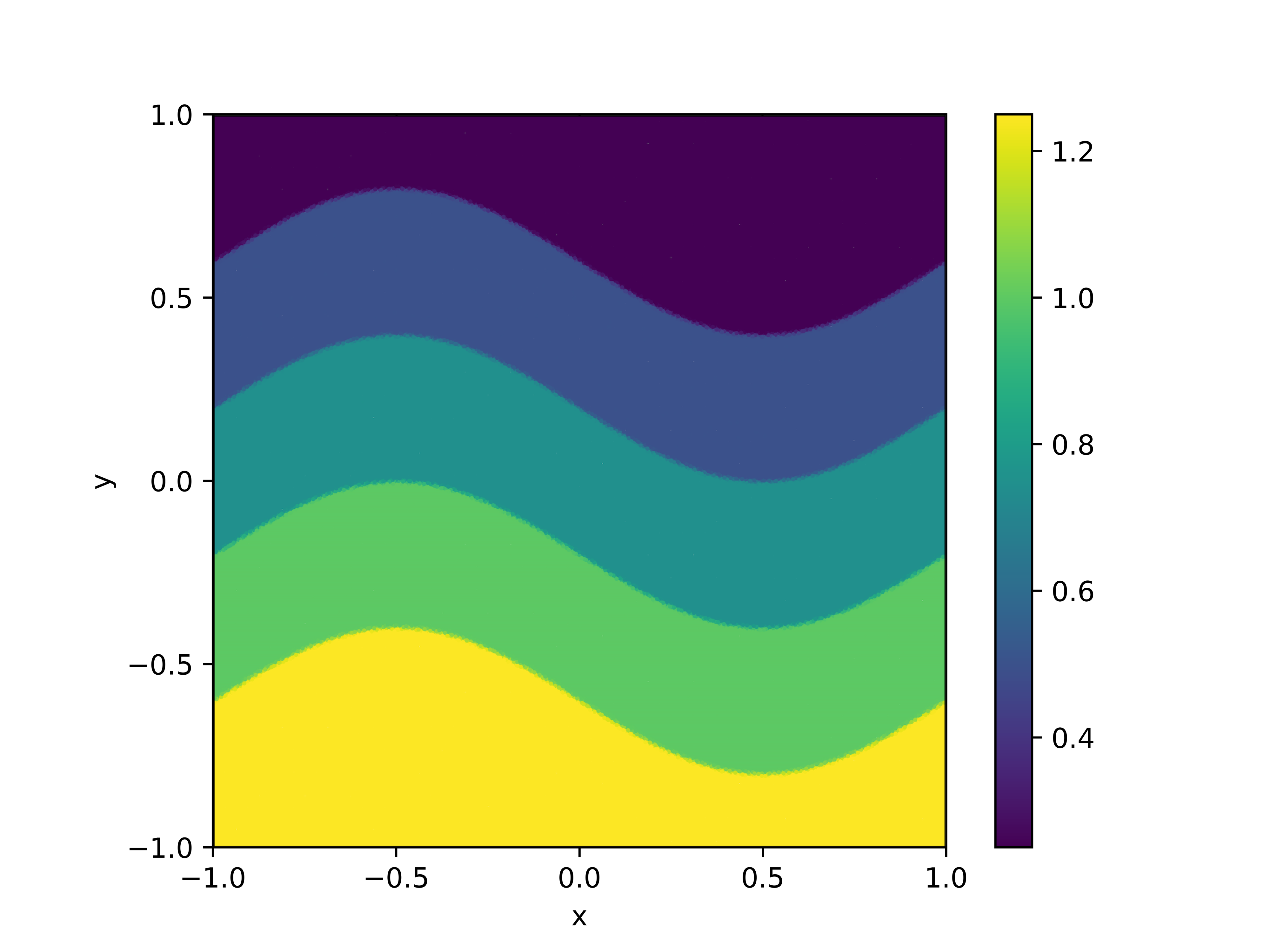}
  \caption{Layered model}
  \label{fig:layered_model_acc}
\end{subfigure}
\caption{Heterogeneous parameter models for the speed of sound $c$}
\label{fig:acoustic_setup}
\end{figure}

The evaluation of the acoustic wave equation was conducted across three scenarios: a constant velocity case ($c=1.0$), a mixture model adapted for the acoustic context similar to the approach described in Section (\ref{section:mixture_model_setup}), and a layered scenario, where the velocity parameter $c$ is derived from the methodology outlined in Section (\ref{section:layered_model_setup}), with adjustments to ensure the velocity values fall within a realistic range for sound propagation. Figure (\ref{fig:acoustic_setup}) provides a detailed overview of the specific ranges and configurations for these scenarios.

.

In contrast to the methodology applied to the elastic wave equation, an exhaustive search for hyperparameters was not undertaken for the acoustic wave equation analysis, as such a process demands substantial time and resources, diverging from the main objectives of this chapter.
Consequently, we focused on evaluating the performance of the most effective architecture and configuration previously identified. Specifically, the \textit{wED}-PINN model was employed, with hyperparameter settings aligned with those utilized in the elastic wave equation experiments, as detailed in Tables (\ref{tab:hyperparameters_constant})-(\ref{tab:hyperparameters_layered}). To facilitate a balanced comparison, the PINN-tanh model was allowed greater flexibility, with experimental setups featuring hidden layers ranging from five to seven, each comprising 128 neurons. A series of $t_1$ values between $[0.07,0.2]$ were explored to improve adaptability. For the PINN-tanh model, the optimal configurations identified were as follows: for the constant case, five hidden layers with a $t_1$ value of $0.1$ were optimal, for the mixture and layered scenarios, six hidden layers proved most effective, with $t_1$ values of $0.15$ and $0.12$, respectively. In contrast, the \textit{wED}-PINN maintained a consistent $t_1$ value of $0.15$ across all three scenarios.

\begin{figure}
    \centering
    % First Row
    \begin{subfigure}[b]{0.45\textwidth}
        \centering
        \includegraphics[width=\textwidth]{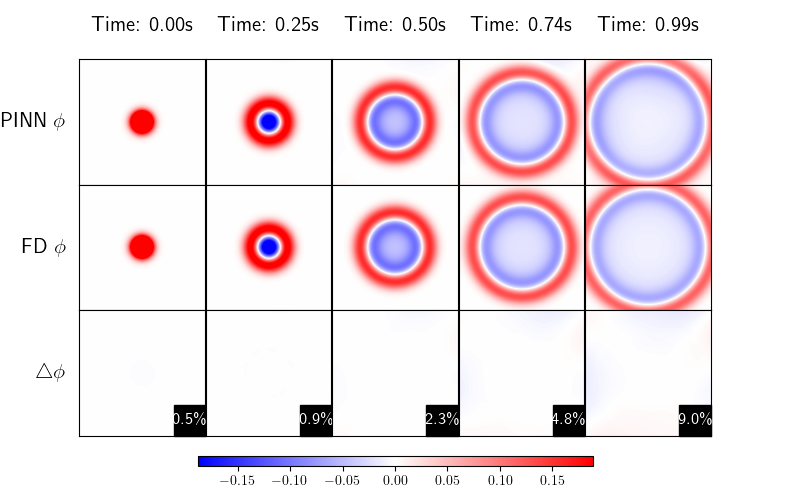}
        \caption{PINN-tanh - constant parameters}
        \label{fig:sub1a}
    \end{subfigure}
    \hfill % this will add a bit of spacing between the two subfigures
    \begin{subfigure}[b]{0.45\textwidth}
        \centering
        \includegraphics[width=\textwidth]{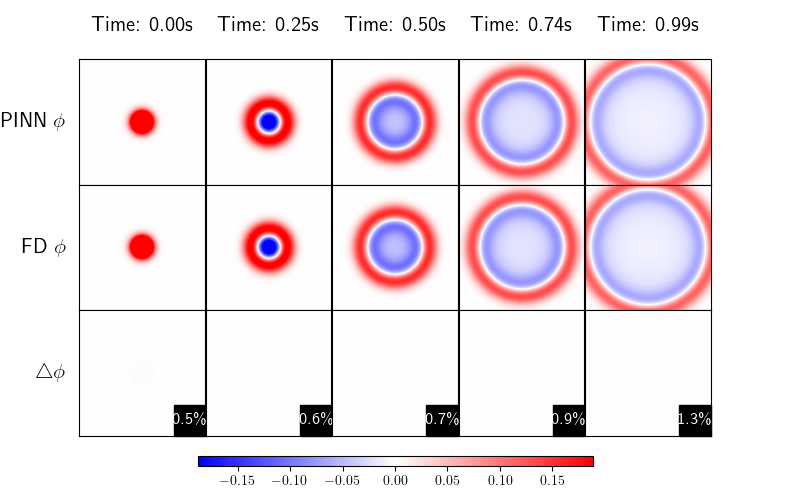}
        \caption{\textit{wED-PINN} - constant parameters}
        \label{fig:sub2a}
    \end{subfigure}
    % Second Row
    \begin{subfigure}[b]{0.45\textwidth}
        \centering
        \includegraphics[width=\textwidth]{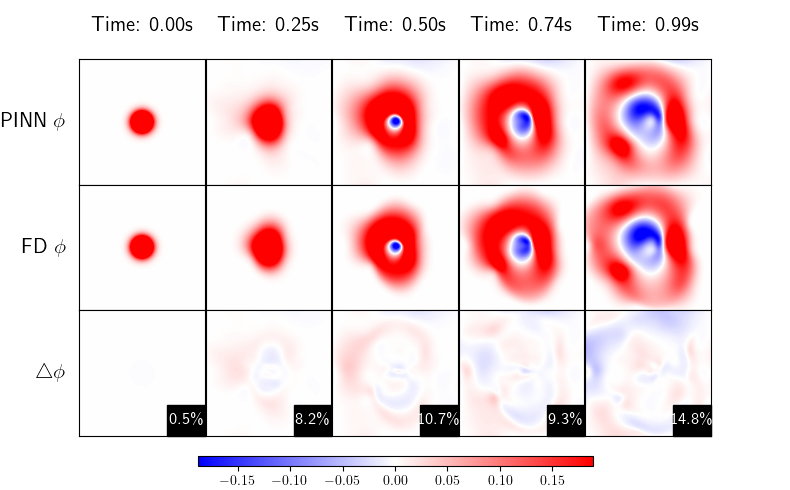}
        \caption{PINN-tanh - mixture model}
        \label{fig:sub3a}
    \end{subfigure}
    \hfill
    \begin{subfigure}[b]{0.45\textwidth}
        \centering
        \includegraphics[width=\textwidth]{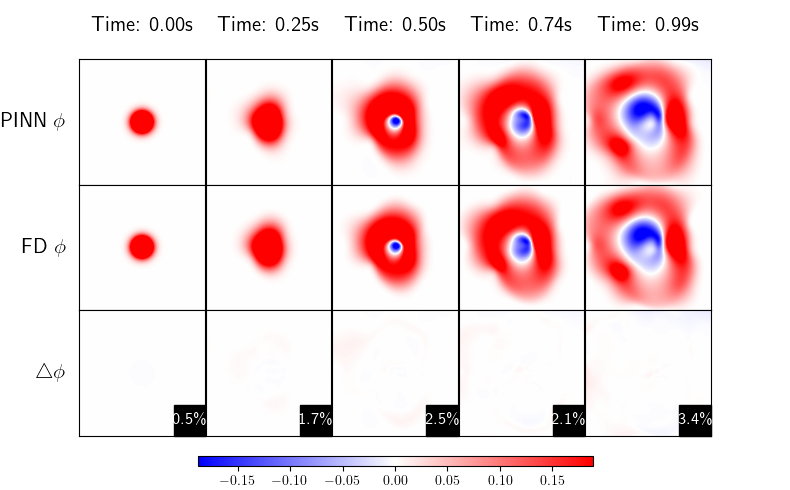}
        \caption{\textit{wED}-PINN - mixture model}
        \label{fig:sub4a}
    \end{subfigure}

    % Third Row
    \begin{subfigure}[b]{0.45\textwidth}
        \centering
        \includegraphics[width=\textwidth]{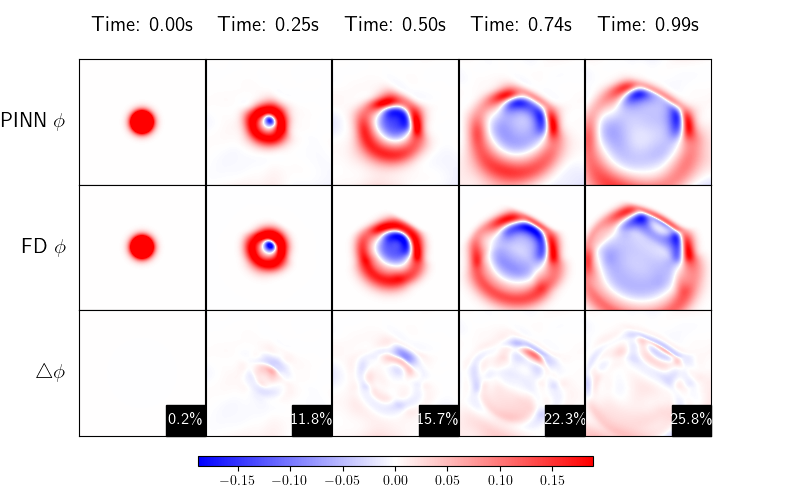}
        \caption{PINN-tanh - layered model}
        \label{fig:sub5}
    \end{subfigure}
    \hfill
    \begin{subfigure}[b]{0.45\textwidth}
        \centering
        \includegraphics[width=\textwidth]{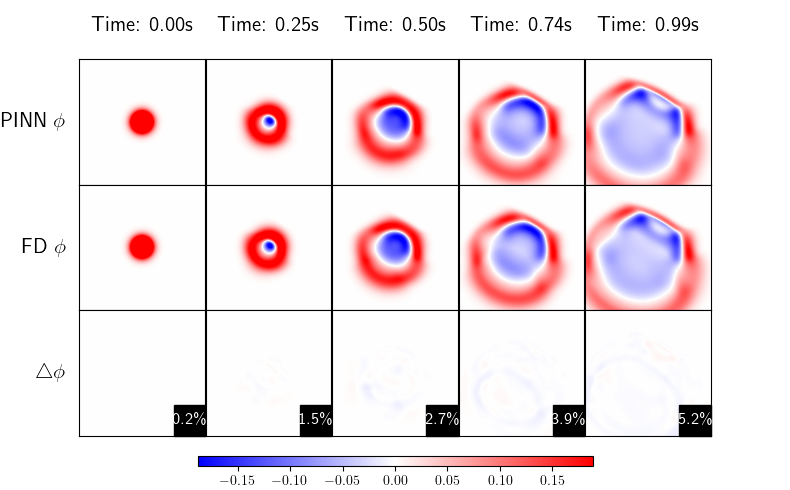}
        \caption{\textit{wED-PINN} - layered model}
        \label{fig:sub6}
    \end{subfigure}

    \caption{Pressure field prediction comparison between the PINN-tanh model and the \textit{wED}-PINN. Both models are evaluated against the FDM solution produced by DEVITO. The left group of results pertains to the PINN-tanh, while the right group illustrates the \textit{wED}-PINN results.}
    \label{fig:acoustic}
\end{figure}

Figure (\ref{fig:acoustic}) showcases the results of employing the PINN-tanh and \textit{wED}-PINN models to solve the acoustic wave equation under various conditions. The PINN-tanh model, while achieving acceptable performance in the constant velocity scenario, faces difficulties in converging satisfactorily in more complex cases. This observation is particularly notable given that prior research conducted by \cite{Ben} has reported promising results under similarly complex parameter conditions. It is critical to acknowledge, however, that these achievements are typically the result of thorough hyperparameter optimization explicitly tailored for the acoustic wave equation. Moreover, the mentioned study utilized advanced strategies, including curriculum-based learning, where training points are gradually introduced based on their temporal sequence to enhance model performance. Although a dedicated effort in hyperparameter tuning could potentially improve results, such an endeavour falls outside the purview of this thesis's aims. This situation underscores the need for precise and targeted adjustments when employing standard PINNs to model complex physical phenomena accurately.

On the other hand, the \textit{wED}-PINN model demonstrates remarkable robustness, achieving excellent convergence with errors below $1\%$ in the constant velocity scenario and maintaining commendable accuracy in both the mixture and layered configurations. This performance suggests that the \textit{wED}-PINN model is inherently adaptable and does not require extensive hyperparameter adjustments to achieve satisfactory accuracy across different, albeit related, PDE scenarios. Detailed statistics regarding the average relative $L_2$ errors for these experiments are provided in Table (\ref{tab:acoustic}), illustrating the \textit{wED}-PINN's superior adaptability and performance in handling the acoustic wave equation.

\begin{table}[ht]
\centering
\begin{tabular}{lcc}
\toprule
 & PINN-tanh & \textit{wED}-PINN \\
\midrule
Constant & $3.06\%$ & $0.76\%$ \\
Mixture & $8.92\%$ & $2.03\%$ \\
Layered & $15.48\%$ & $2.68\%$ \\
\bottomrule
\end{tabular}
\caption{Relative $L_2$ errors achieved by the baseline PINN (PINN-tanh) model and the \textit{wED}-PINN model when solving the acoustic wave equation across three different scenarios of the parameter speed of sound $c$.}
\label{tab:acoustic}
\end{table}

The conclusion that the \textit{wED}-PINN, or encoder-decoder type PINNs in general, exhibit superior generalization to the acoustic wave equation and represent a more accurate and robust alternative than the standard PINN is a reasonable assertion.

\section{Discussion}
The objective of this chapter was to explore the effects of incorporating prior knowledge of wave physics into the design of neural network architectures on their accuracy and convergence rates. To achieve this, we devised several innovative network architectures, each integrating varying degrees of wave physics knowledge. These architectures were then arrayed along a spectrum to assess their utilization of wave physics knowledge, spanning from the standard PINN model with a fully connected architecture to highly specialized architectures. Through extensive experimentation, we demonstrated the significance of incorporating physical insights into the architectural design of networks.

A key finding from our research is the clear benefit of embedding prior wave-physics knowledge into the neural network architecture, confirming the hypothesis that integrating prior wave-physics knowledge enhances model performance. Additionally, our investigation into various network designs revealed that models situated at either end of the spectrum---those with either minimal or excessive constraints on wave physics knowledge---exhibited poorer convergence rates compared to those positioned at a central point on the spectrum, thus identifying an optimal position. Specifically, encoder-decoder-PINNs, which strike a balance between the integration of physical knowledge and the maintenance of sufficient flexibility, avoiding the limitations of overly restrictive designs, occupy this optimal position. Encoder-decoder-PINNs consistently outperformed the standard PINN model across various problem settings, including those involving complex parameters and multiple seismic sources. Furthermore, our research extended to the acoustic wave equation, demonstrating the robust generalizability of the encoder-decoder architecture. The success of this architecture in these contexts, without the need for extensive hyperparameter optimization, highlights the versatility and efficiency of this approach.

Our findings indicate that instead of developing highly specialized solutions for each new problem scenario, investing in architectures that inherently align with physical principles offers a more scalable and flexible approach. Such models not only achieve higher accuracy but also improve efficiencies in resource utilization, notably reducing the necessity for intensive hyperparameter optimization efforts.

\section{Conclusion}
In this chapter, we have demonstrated that integrating prior wave physics knowledge into network design significantly enhances accuracy and robustness. Furthermore, it can, to a certain degree, mitigate the need for extensive hyperparameter search operations, as such models exhibit greater resilience to variations in certain hyperparameters. To contextualize these findings, we recall from the preceding chapter the identification of three primary shortcomings in conventional PINNs: accuracy, hyperparameter search intensity, and efficiency. The advancements presented in this chapter markedly improve upon the first of these issues---accuracy---and, to a lesser extent, the second---hyperparameter search. However, our novel, improved architectures do not substantially enhance the efficiency of PINNs. Although some proposed network architectures utilize more lightweight networks, this does not address the fundamental issue that, regardless of the architecture's robustness, PINNs must be retrained for each new problem setting. Consequently, the final challenge---efficiency---persists. Addressing this challenge will be the focus of our efforts in the subsequent chapter.
\newpage

\chapter{Conditioning}
\label{chap:conditioning}
\section{Introduction}
In the preceding chapter, we extensively discussed the significant enhancement in accuracy achieved by integrating physical knowledge into the network architecture, positioning PINNs as a potentially viable alternative to conventional numerical methods for solving seismic wavefield propagation with high accuracy. However, the improvement in accuracy, while necessary and desirable, does not alone make PINNs a formidable contender against traditional methodologies such as FDMs or SEMs. This is because, at their current stage, PINNs are generally less accurate or, in the best-case scenario, only match the accuracy of traditional numerical methods. Consequently, without substantial efficiency gains in computational speed over conventional methods, PINNs lack a compelling use case.

Currently, each new parameter configuration and source location necessitates retraining the PINN from scratch, which is a significant drawback. Although there are techniques to expedite PINN training---such as meticulous domain decomposition, as demonstrated in \cite{FBPINNS}---the training duration for PINNs remains considerably longer than the time required for simulating elastic wavefield propagation using traditional numerical methods, as exemplified in Table (\ref{tab:execution_times}). While PINNs offer advantages, such as being meshfree, these benefits alone do not establish them as a practical alternative.

To overcome this challenge, we propose conditioning the PINN on the source location, thereby eliminating the need to retrain the network for each new source location. Specifically, the PINN will learn a mapping $\Lambda: \mathbb{R}^5 \rightarrow \mathbb{R}^2$, incorporating the two-dimensional coordinates of the source location into the input space. By integrating our insights into the optimal network architecture discussed in the previous chapter, we aim to develop an optimally conditioned PINN model. This model seeks to provide a viable alternative to traditional FDMs in seismology, capable of accurately predicting seismic wavefields for numerous potential source locations without resorting to FDMs or retraining the PINN for each case. Such a capability is crucial, as scientists currently face prohibitive computational costs in conducting broad-band simulation, for instance, when creating a comprehensive statistical model of potential earthquake-induced surface responses across an entire region, considering uncertainties in seismic origin. 

\section{Contributions}
We propose a methodology to solve the elastic wave equation using PINNs conditioned on the source location. Our approach does not require any training data, yet it still aims to solve multiple instances of the elastic wave equation by incorporating the two-dimensional source location as part of the input. We implement an encoder-decoder type PINN, as discussed in Section (\ref{section:encoder-decoder}), alongside standard PINN methodologies, comparing their accuracy. Our work employs a non-constant parameter model, specifically the mixture model described in Section (\ref{section:mixture_model_setup}), to represent the underlying Lam{\'e} parameters.

We demonstrate that this conditioning strategy achieves reasonably low relative $L_2$ errors and significantly faster inference times than standard FDMs.

Furthermore, we highlight a clear seismological use case where this technique exhibits considerable potential by offering a promising alternative for broad-band simulations in computational seismology.

\section{Related Work}
\label{cond_related_work}
The concept of learning multiple instances of a function, as opposed to just a single instance, has emerged as a prominent research theme within sciML. A key category within this domain comprises Neural Operators, which further branches into two main approaches: Fourier neural operators (FNOs) and deep operator networks (DeepONet).

FNOs, as introduced by Li et al. \cite{FNO}, are designed to learn mappings between infinite-dimensional function spaces, diverging from the finite-dimensional function spaces typically utilised in sciML techniques, such as PINNs. They achieve this by parameterizing the integral kernel in the Fourier space. Zhang et al. extended FNOs to model the solution of the two-dimensional isotropic elastic wave equation using synthetic datasets \cite{FNO_elastic}. Despite showing promise, their findings indicated significant accuracy errors, particularly as simulation times increased.

DeepONets, conceptualised by Lu et al. \cite{DeepONets}, draw inspiration from the universal approximation theorem for operators \cite{UOAT}. This theorem posits that a pair of FCNs, whose outputs are merged via the dot product, can approximate any nonlinear operator with arbitrary precision, assuming a sufficiently large neuron count in the single hidden layer of the FCNs. Building on this theorem, DeepONet comprises two distinct sub-networks: the branch network, which encodes input functions at specific sensor points, and the trunk network, which encodes the locations of the output functions \cite{DeepONets}.
Both FNOs and DeepONets exhibit unique strengths and limitations. Neural Operators are generally categorised under supervised learning and rely heavily on labelled training data. Acquiring such data, particularly within seismology, can be challenging or sometimes entirely impractical.

%CHEKC WITH BEN CORRECTNESS OF PINOS
A novel hybrid approach, merging PINNs with Neural Operators, has been proposed by Li et al., called the physics-informed neural operator (PINO) \cite{PINO}. This approach combines the strengths of FNOs with PDE constraints, akin to PINNs. While promising, PINOs are not a universal solution, as they still largely depend on training data. It is feasible to devise unsupervised training schemes for PINOs, but their accuracy typically falls short of that achieved with labelled data \cite{PINO_self_improve}. Moreover, such unsupervised PINOs often use approximate gradients instead of exact ones, and their output becomes discretised, thus negating the meshfree advantage sought with PINNs. Rosofsky et al. explored applying PINO methods to the two-dimensional acoustic wave equation with spatially varying sound speed, $c(x,y)$, yet encountered substantial accuracy errors and the necessity for prior labelled training data \cite{PINO_acoustic}.
Unlike FNOs, DeepONets, and PINOs, our approach focuses on PINNs and does not require any labelled training data.

In the realm of PINNs, Moseley et al. \cite{Ben} explored conditioning a PINN on the source location for the two-dimensional acoustic wave equation, demonstrating accurate capture of the primary wavefront in complex parameter settings. However, beyond the primary wavefront, finer details of the wavefield were not predicted with high accuracy. Song et al. recently conditioned a PINN model on the source location for the scattered form of the frequency-domain elastic wave equation \cite{scattered}. Our approach diverges as it seeks to solve the elastic wave equation in the time domain rather than in the frequency domain, as discussed in Section (\ref{Section_ELASTic_relatedwork}).

\section{Methods}
The concept of augmenting PINNs to incorporate source location as part of their input is both theoretically straightforward and practically implementable. Previously, the approximation network was tasked with learning a mapping, $\textbf{u}: \mathbb{R}^3 \rightarrow \mathbb{R}^2$, which has now been extended to $\textbf{u}: \mathbb{R}^5 \rightarrow \mathbb{R}^2$. This extension involves including the $x$ and $y$ coordinates of the source locations, denoted as $s_x$ and $s_y$, into the input space, thus allowing the network to seamlessly adapt to this increased input dimensionality. This modification necessitates no alterations to the Loss functions. However, the initial condition requires a minor adjustment from its original formulation in Equation (\ref{eq:initial_condition}) to:
\begin{equation}
\begin{aligned}
g(\textbf{x}) = \frac{d f(\textbf{x}-\textbf{s})}{d\textbf{x}} / {\left\Vert \frac{d f(\textbf{x}-\textbf{s})}{d\textbf{x}}\right\Vert_{max}},
\end{aligned}
\end{equation}
where $\textbf{s}=(s_x,s_y)$. This ensures that the initial condition is consistently centred around the source location.

\subsection{Model Type}
We selected the mixture model for the underlying model of the Lam{\'e} parameters because of its highly heterogeneous nature. This characteristic is anticipated to produce significant variations in the wavefield for even minimal shifts in the source location.
\subsection{Sampling Strategies}
A pivotal aspect in enhancing the efficacy of conditioned PINNs is the methodology for sampling source locations. Two distinct strategies must be considered.

The first strategy entails designating the number of different source locations $N_s$ as an integer that is less than the total number of training points, $N$. This configuration divides the entire training set into $N_s$ hypothetical subsets, each of a size $N/N_s$ and encompassing a unique source location ($s_x$, $s_y$). These subsets are uniformly populated with spatial and temporal points generated via a Sobol sequence. This method is referred to as split-conditioning, alluding to the segmentation of the training dataset. It is crucial to note that, despite this segmentation, training still employs full batch optimisation techniques.

The alternative strategy, which we call full-conditioning, sets $N_s$ equal to $N$, resulting in each training sample having a distinct source location derived from a Sobol sequence. The choice between these strategies significantly impacts the network's training efficiency and its ability to generalise across various source locations.

Irrespective of the chosen approach, ensuring comprehensive coverage of source positions throughout the spatial training domain requires a considerable volume of training points. Split-conditioning offers the benefit of associating abundant spatial and temporal training points with each source location, provided the overall number of training points is reasonable. However, this strategy may lead the network to overfit specific source locations during training, potentially hindering its performance on the domain's new, unseen source locations. This risk is particularly pronounced in highly heterogeneous domains where minor shifts in source location can drastically alter the resulting wavefield.

Conversely, the full-conditioning strategy mitigates the risk of overfitting by vastly increasing $N_s$, albeit at the cost of associating only a single spatio-temporal training point with each source location. Therefore, sufficient collocation points are imperative to ensure a dense distribution of source locations under this approach.

Our empirical evidence suggests that split-conditioning, especially with $N_s < 100$, tends to trap the network in suboptimal, local minima due to overfitting. Even when $N_s$ values are higher, this approach still generally exhibits inferior convergence and interpolation capabilities compared to full-conditioning. Based on these findings, we have decided to proceed exclusively with the full-conditioning strategy.

It is also important to note that source locations were not sampled across the entire spatial domain $\Omega$. Instead, sampling was confined to a subdomain $\Omega_s = {(x,y) | -0.4\le x\le0.4,-0.4\le y\le0.4}$. Expanding the sampling to the entire domain $\Omega$ would significantly escalate the challenge, necessitating an increase in collocation points by a factor of $A_{\Omega}/A_{\Omega_s} = 6.25$ to maintain equivalent source location density. Given the constraints of our current GPU memory capacity, such an expansion is unfeasible. Moreover, the subdomain $\Omega_s$ already encompasses a considerable variance in the underlying Lam{\'e} parameters, justifying its selection for our experiments.

\subsection{Training Cost and Time}
Maximizing the number of collocation points has proven beneficial for the effectiveness of our network, prompting us to utilise as many points as our resources allow. Our experiments are constrained by the limitations of full batch training, which is necessitated due to the incompatibility of the LBFGS optimisation algorithm with batched training data. This limitation dictated that the maximum number of collocation points feasible within the memory constraints of our available hardware was approximately half a million. We conducted these computationally intensive experiments on shared NVIDIA A100 GPUs, each with 80GB of GPU memory. The bulk of this memory is consumed by the computation of second-order derivatives of the network parameters with respect to the input points. This computational intensity is also mirrored in the training duration: training a single conditioned PINN can require up to 30 hours to achieve convergence. 
\paragraph{Architecture Choice and Hyperparameters}
\label{section:arch_choice_hyper_conditioned}
Given the considerable computational demands and limited availability of high-end GPU resources, our exploration of network architectures and hyperparameters was inherently limited. We focused on two network types: encoder-decoder-PINNs (comprised of the \textit{wED}-PINN and \textit{pED}-PINN model) and the standard PINN model. They were chosen as the encoder-decoder-type PINNs have shown to be the most robust and best-performing architecture type (see Section (\ref{section:Arch_res})), and the PINN-tanh model is used as a benchmark as it represented the default model. This strategic choice allowed us to conduct a more targeted and efficient evaluation of the performances of these architectures in our experiments.

A hyperparameter search was conducted for both network types, employing the same methodology as described in Section (\ref{section:exp_setup}), to determine the optimal configurations for each. The outcomes of this search identified the following set of best-performing hyperparameters for each network:

\begin{itemize}
\item For the standard PINN model utilizing the tanh activation function (PINN-tanh), the optimal configuration was identified as comprising five hidden layers, each containing 128 neurons. Additionally, a $t_1$ value of $0.2$ was determined to be ideal.
\item For the encoder-decoder PINNs, the most effective configuration was found to be the (\textit{pED}-PINN) architecture, with four encoder layers, each with $64$ neurons, followed by a plane wave layer consisting of $135$ neurons and two decoder layers with $16$ neurons each. A $t_1$ value of $0.2$ was identified as optimal.
\end{itemize}

Both models were trained using the LBFGS optimiser with a starting learning rate of $1.0$, spanning 400 epochs and allowing for a maximum of 200 evaluations and iterations. This training regimen was applied using half a million collocation points.

\section{Results}
This section delineates the results achieved using the conditioned PINN models. The primary focus is on the elastic wave equation; however, the acoustic wave equation is also examined to assess the capability for generalization.
\subsection{Elastic Conditioning}
\label{ELASTIC_CONDITIONING_RESULTS}
To evaluate the post-training accuracy of the two network architectures under consideration, we distributed $1600$ source locations uniformly across the subdomain $\Omega_s$. The performance of each network and each source location was benchmarked against FD solutions generated using DEVITO. These reference solutions were computed on a $512 \times 512$ spatial grid, with the temporal dimension discretised into $100$ evenly spaced time steps. This rigorous comparison framework ensures a comprehensive assessment of the networks' ability to accurately simulate the wavefield for many different source locations.

\begin{figure}
\centering
\begin{subfigure}{\textwidth} % Use full text width for each subfigure
  \centering
  \includegraphics[width=\linewidth]{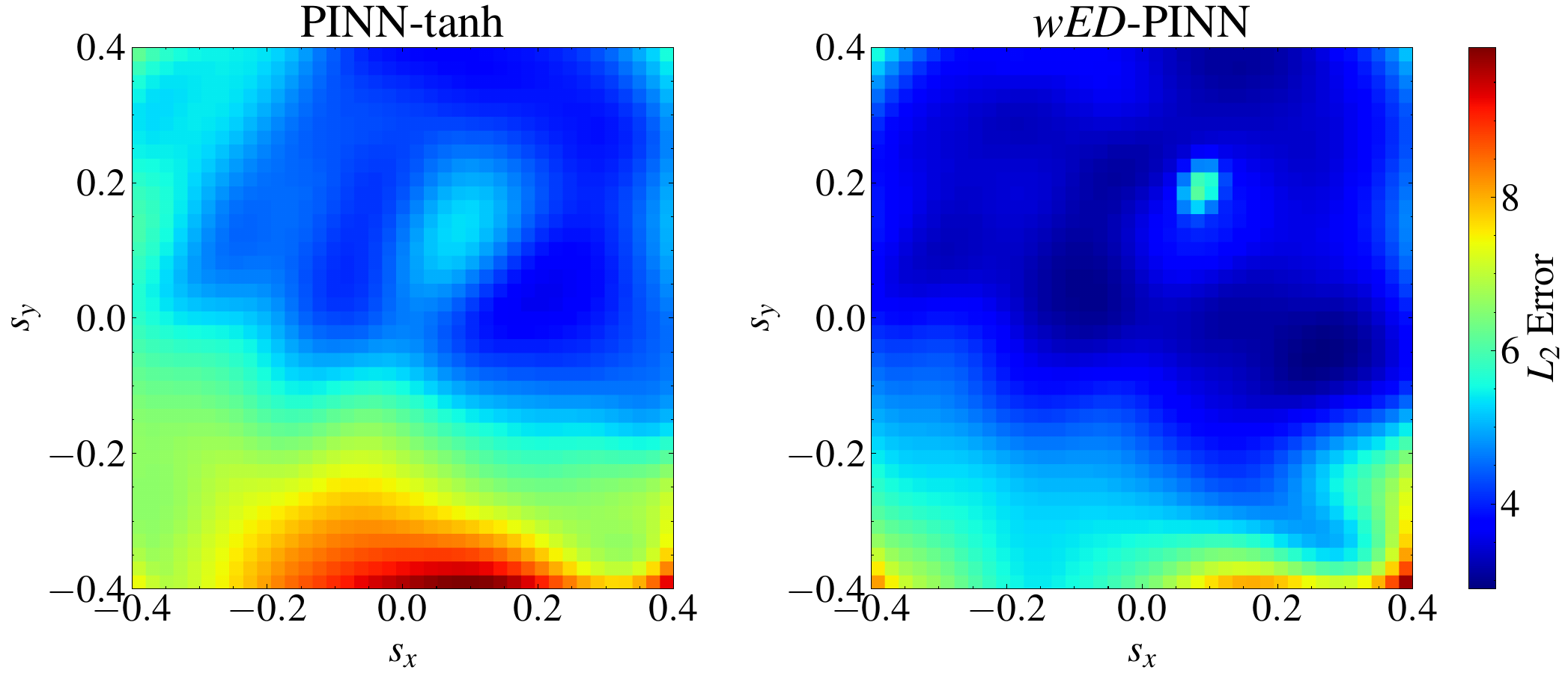} % Adjust the width as needed
  \caption{Error distribution in $\Omega_s$}
  \label{fig:04}
\end{subfigure}\\ % Use "\\" to stack the subfigures vertically

\begin{subfigure}{\textwidth} % Use full text width for the second subfigure
  \centering
  \includegraphics[width=\linewidth]{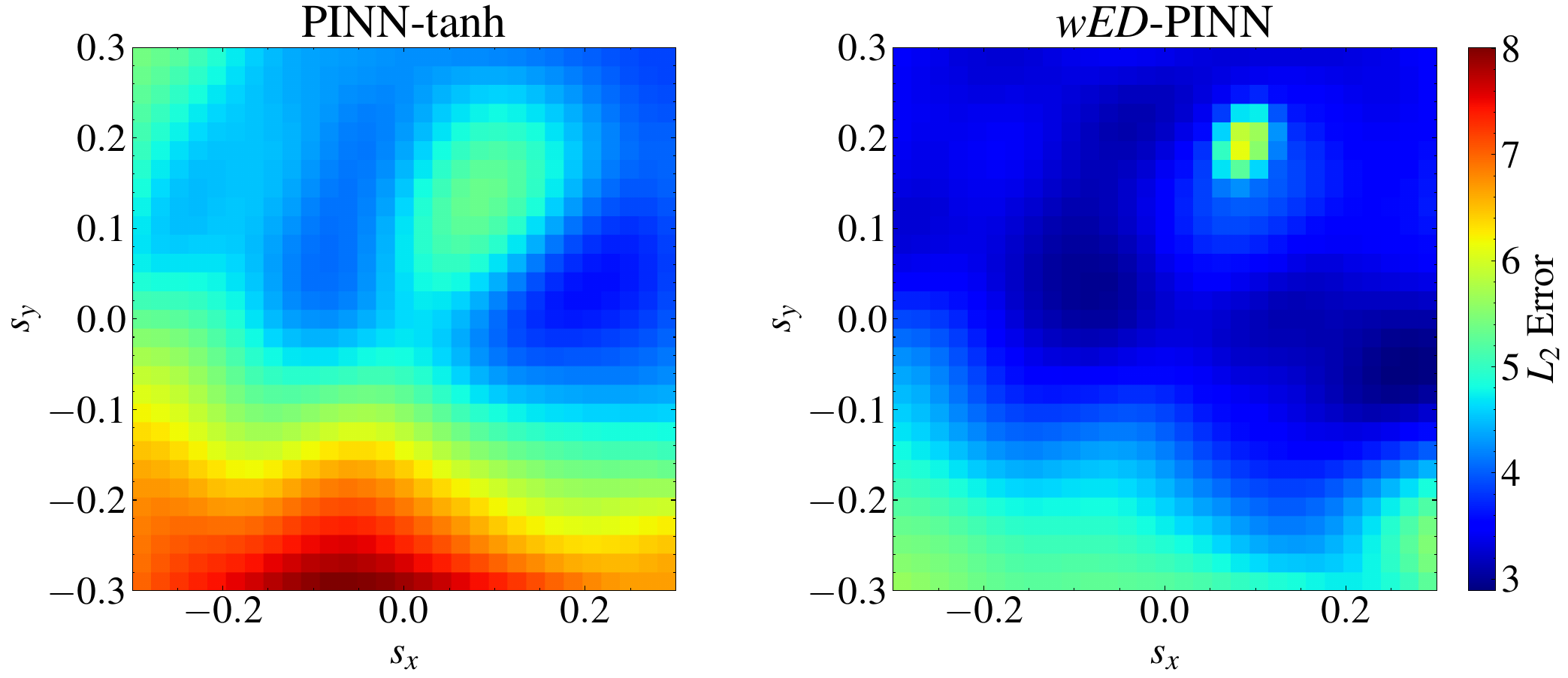} % Adjust the width as needed
  \caption{Error distribution in $\Omega_{\hat{s}}$}
  \label{fig:03}
\end{subfigure}
\caption{Comparison of the spatial variance in average relative $L_2$ errors between PINN-tanh and \textit{pED}-PINN models for solving the conditioned elastic wave equation. The errors are evaluated within the subdomains $\Omega_s = \{(x,y) \,|\, -0.4 \le x \le 0.4, -0.4 \le y \le 0.4\}$ and $\Omega_{\hat{s}} = \{(x,y) \,|\, -0.3 \le x \le 0.3, -0.3 \le y \le 0.3\}$, respectively, and tested against our FDM elastic wavefield solution generated with DEVITO.}
\label{fig:error_maps}
\end{figure}

\begin{figure}
\centering
\includegraphics[width=0.4\linewidth]{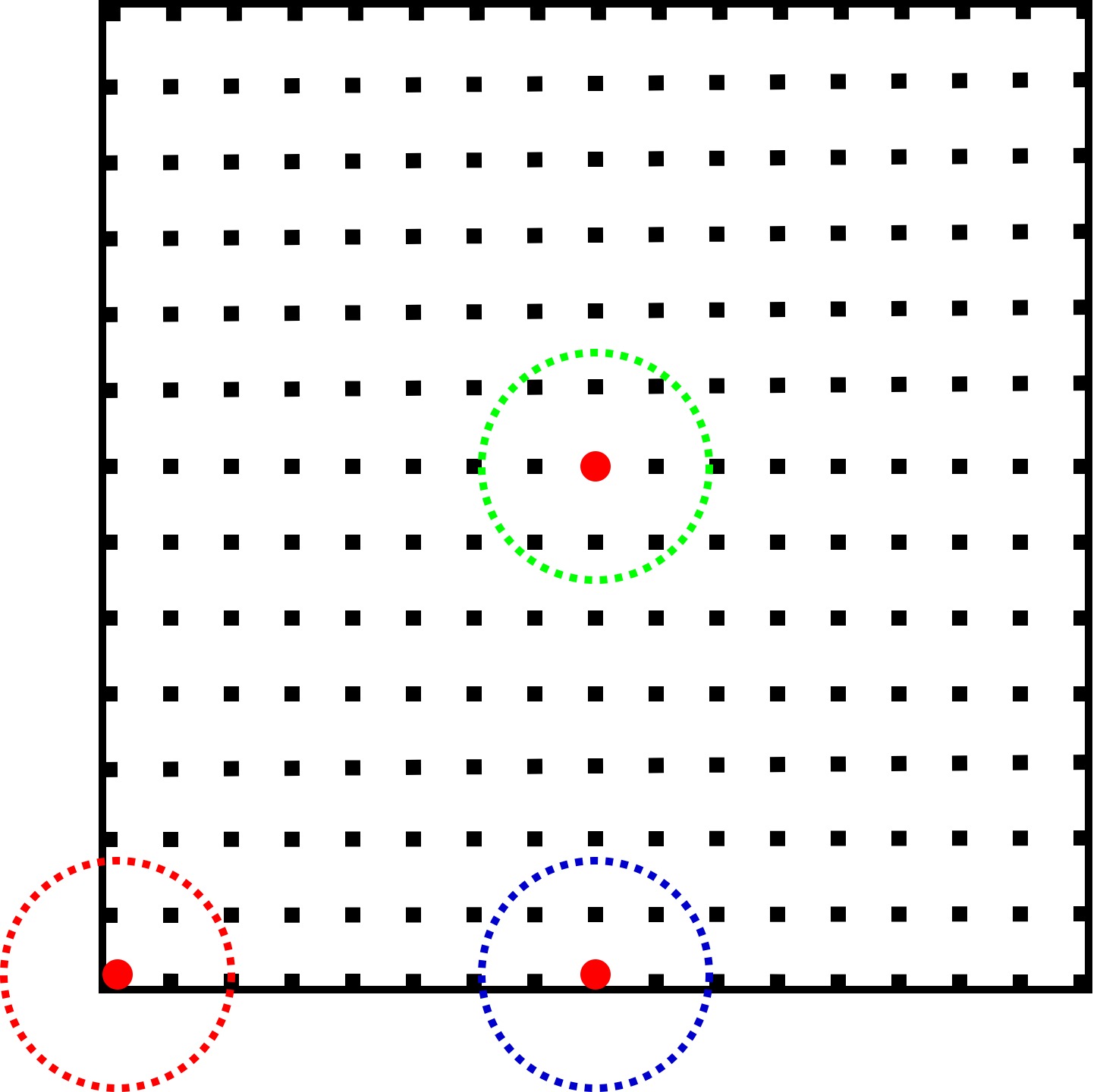}
\caption{Simplified sketch of the sampling densities around three select sources (red dots). The green circle indicates a source with high density, the blue circle indicates one with medium density, and the red circle indicates a source with low density.}
\label{fig:sampling_density}
\end{figure}

\begin{figure}
\begin{subfigure}{.48\textwidth}
  \centering
  \includegraphics[width=\linewidth]{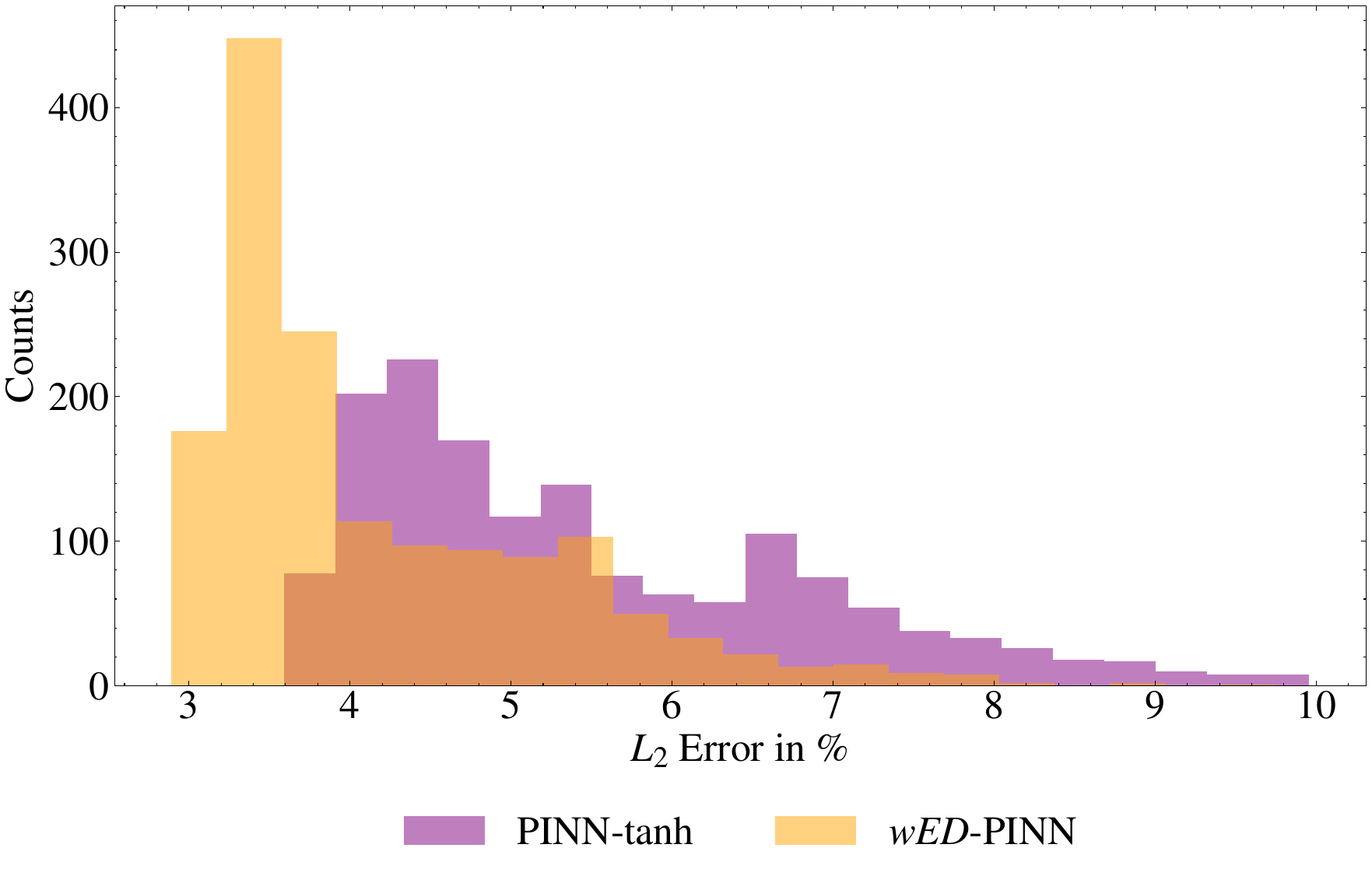}
  \caption{Histograms of the average relative $L_2$ errors in $\Omega_s$}
  \label{fig:04h}
\end{subfigure}\hfill
\begin{subfigure}{.48\textwidth}
  \centering
  \includegraphics[width=\linewidth]{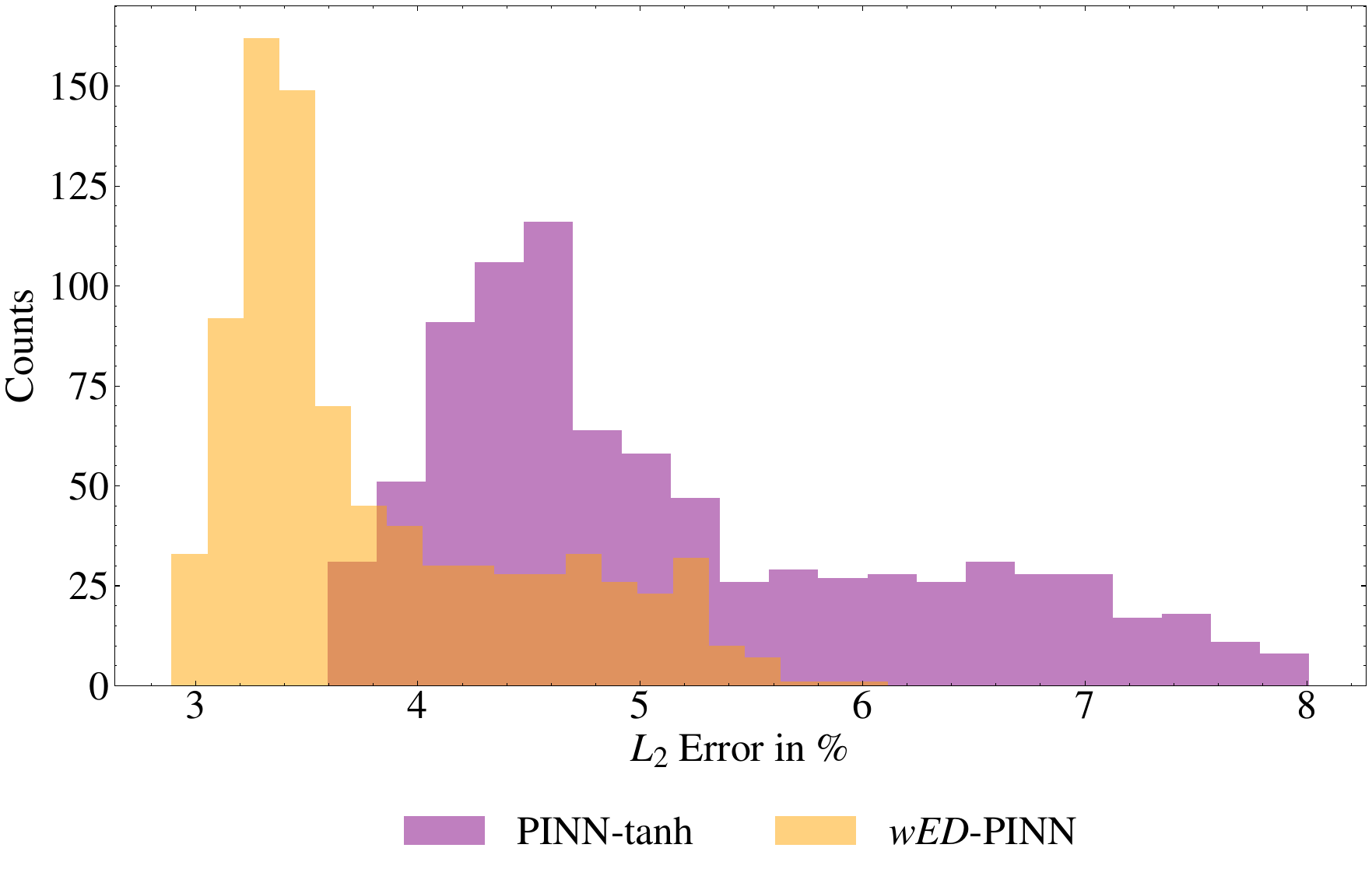}
  \caption{Histograms of the average relative $L_2$ errors in $\Omega_{\hat{s}}$}
  \label{fig:03h}
\end{subfigure}
\caption{Histograms of the average relative $L_2$ errors for the conditioned PINN-tanh and \textit{pED}-PINN models on the subdomains $\Omega_s = \{(x,y) | -0.4\le x\le0.4,-0.4\le y\le0.4\}$ (left) and $\Omega_{\hat{s}} = \{(x,y) | -0.3\le x\le0.3,-0.3\le y\le0.3\}$ (right), tested against the FDM elastic wavefield solution generated with DEVITO}
\label{fig:error_histograms}
\end{figure}

Figure (\ref{fig:03}) elucidates the variation in the average relative $L_2$ error between the two evaluated PINN models---\textit{pED}-model and PINN-tanh---based on the spatial positioning of the source locations. The \textit{pED}-model exhibits superior accuracy, with a mean\footnote{The mean relative $L_2$ error is calculated by averaging the average relative $L_2$ errors, where each error is an average over 100 time steps, across 1600 different sources.} relative $L_2$ error of $4.08\%$, in contrast to the $5.45\%$ observed for the PINN-tanh model, as detailed in Figure (\ref{fig:error_histograms}). A pattern emerges from the analysis, revealing that errors tend to peak near the subdomain's bottom and proximal to its corners. This phenomenon can be attributed to two primary factors: firstly, these regions might inherently produce more complex wavefield solutions; secondly, source locations near boundaries---and especially corners---exhibit a reduced density of neighbouring source locations, as visually represented in Figure (\ref{fig:sampling_density}).

Given the adoption of the full-conditioning approach, where each source location is paired with a singular tempo-spatial training point, the network's ability to achieve satisfactory convergence for a given source location is contingent upon a high density of nearby source locations and corresponding training points. This requisite is not met for locations proximate to the boundary of $\Omega_s$, and for sources near corners. To further dissect the behaviour of the relative $L_2$ error in regions less prone to such density discrepancies, an additional analysis is presented in Figure (\ref{fig:04}). This analysis focuses on the spatial distribution of the averaged relative $L_2$ error within a more central subdomain, $\Omega_{\hat{s}} = \{(x,y)|-0.3\le x\le0.3,-0.3\le y\le0.3\}$. Within this confined space, the \textit{pED}-model achieves a mean relative $L_2$ error of $3.79\%$, compared to the $5.15\%$ recorded for the PINN-tanh model, underscoring the \textit{pED}-model's enhanced accuracy in more uniformly sampled regions.

Figure (\ref{fig:error_histograms}) showcases the distribution of average relative $L_2$ errors for the two models under scrutiny across the subdomains $\Omega_s$ and $\Omega_{\hat{s}}$. This visualization further underscores the superior accuracy of the \textit{pED}-PINN over traditional PINN methodologies. An open, central question of this discussion revolves around whether the observed mean relative $L_2$ errors of $4.08\%$ and $3.79\%$, for $\Omega_s$ and $\Omega_{\hat{s}}$ respectively, are considered satisfactory. Given the challenging nature of the task, these error rates are considered reasonably low. Nonetheless, for PINNs to be recognised as a feasible alternative to conventional FD methods in terms of accuracy, these errors would need to be further reduced. The decision to opt for PINNs over traditional FD methods hinges significantly on the specific requirements of the use case, a topic that will be elaborated upon in Section (\ref{Section:Case_study}).

\begin{figure}
\begin{subfigure}{.5\textwidth}
  \centering
  \includegraphics[width=\linewidth]{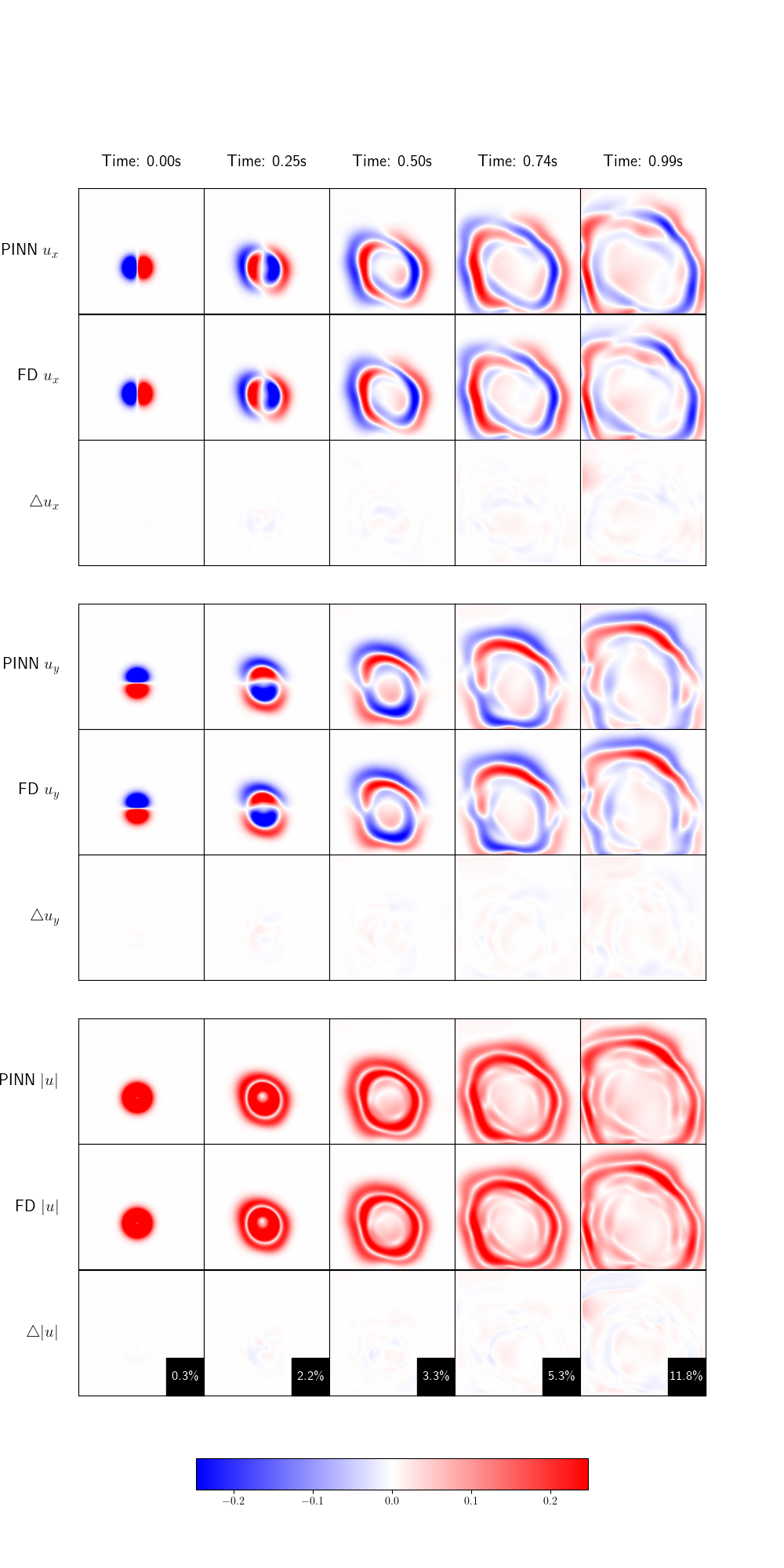}
  \caption{PINN-tanh model}
  \label{fig:tanh_good}
\end{subfigure}%
\begin{subfigure}{.5\textwidth}
  \centering
  \includegraphics[width=\linewidth]{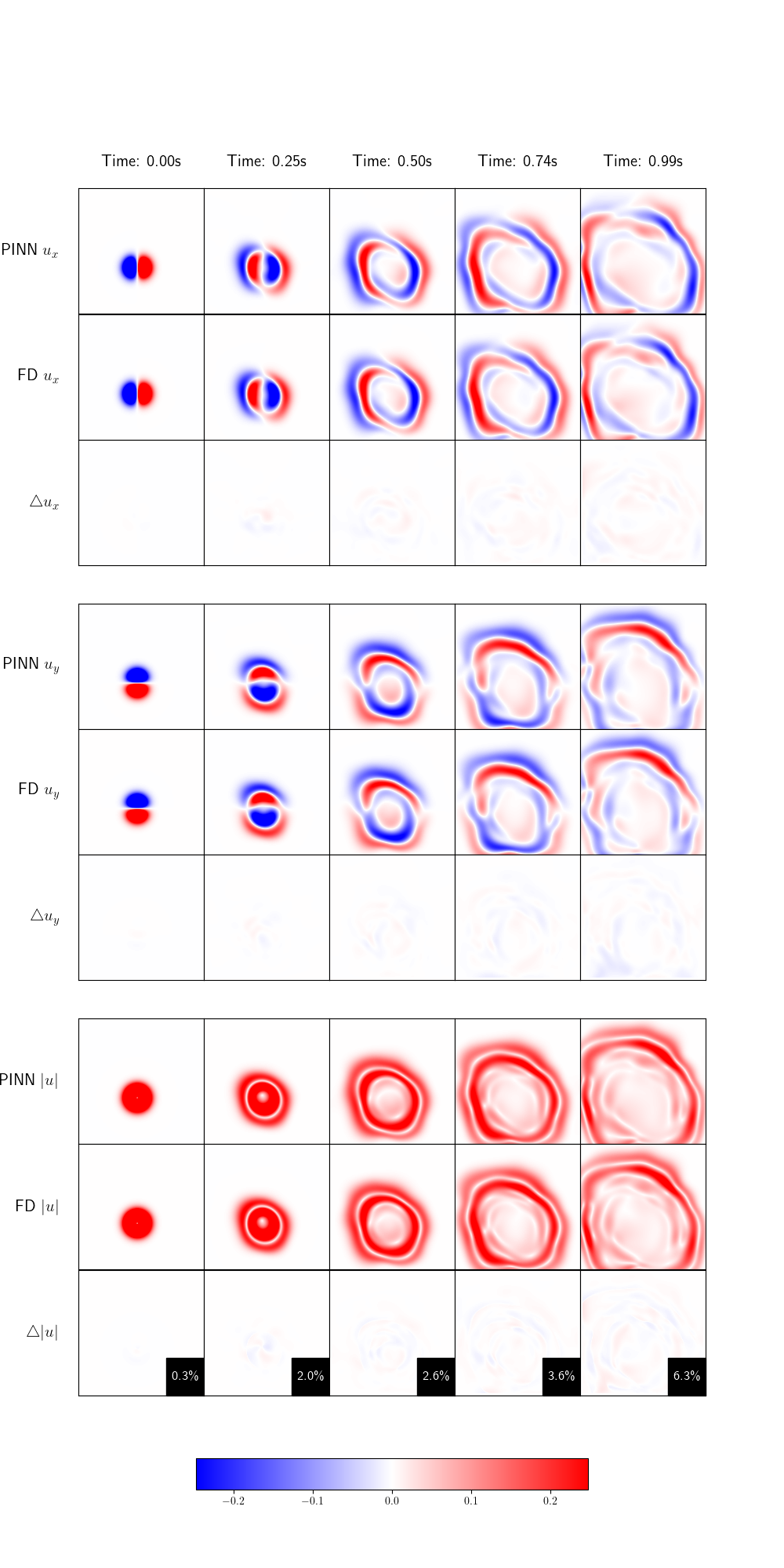}
  \caption{\textit{pED}-PINN model}
  \label{fig:planewave_good}
\end{subfigure}
\caption{Elastic wavefield solution generated with the conditioned PINN-tanh (left) and \textit{pED}-PINN (right) with $s = (0.26,-0.06)$, tested against the FDM elastic wavefield solution generated with DEVITO.}
\label{fig:conditioned_good}
\end{figure}

\begin{figure}
\begin{subfigure}{.5\textwidth}
  \centering
  \includegraphics[width=\linewidth]{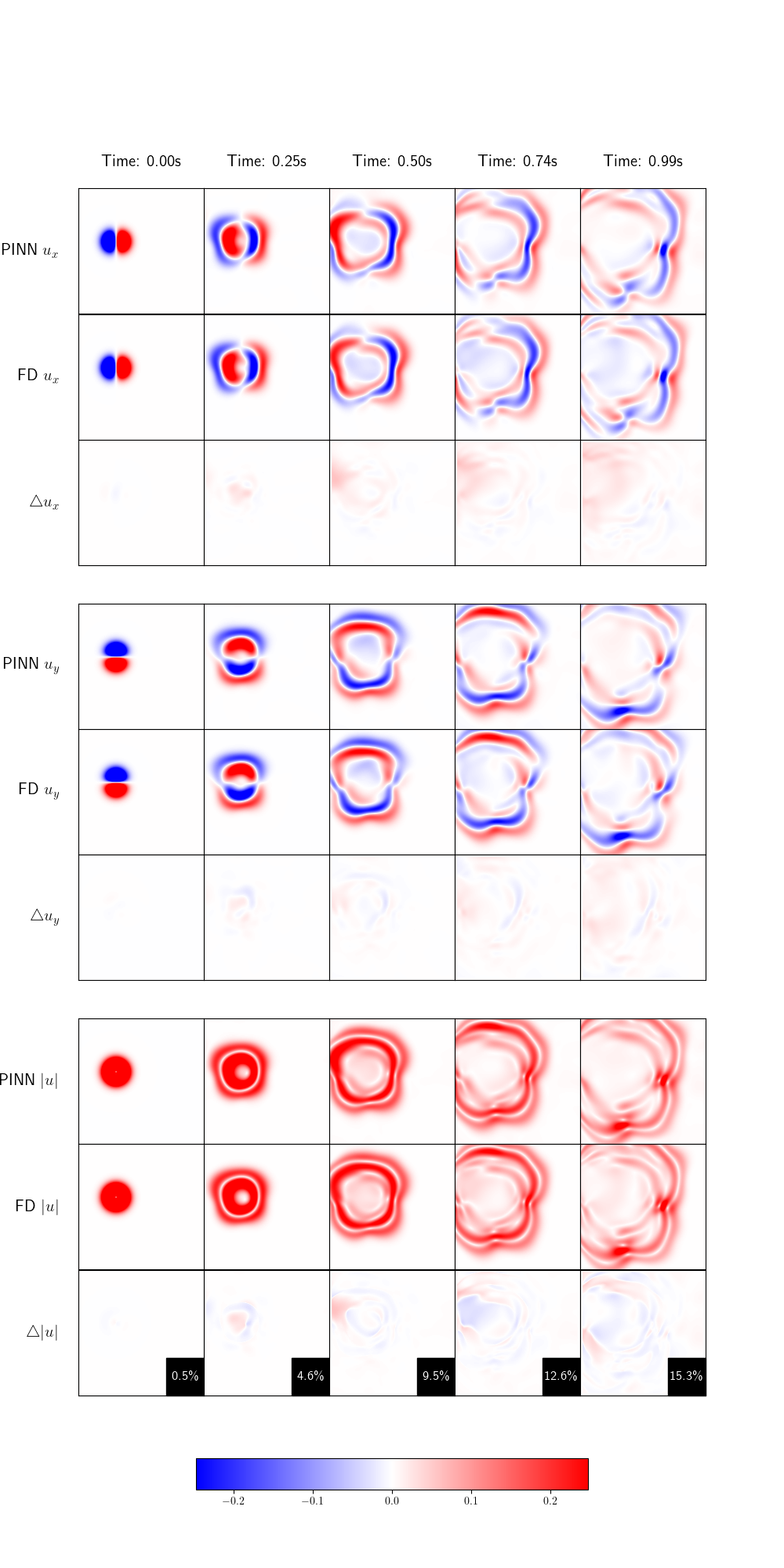}
  \caption{PINN-tanh model}
  \label{fig:tanh_bad}
\end{subfigure}%
\begin{subfigure}{.5\textwidth}
  \centering
  \includegraphics[width=\linewidth]{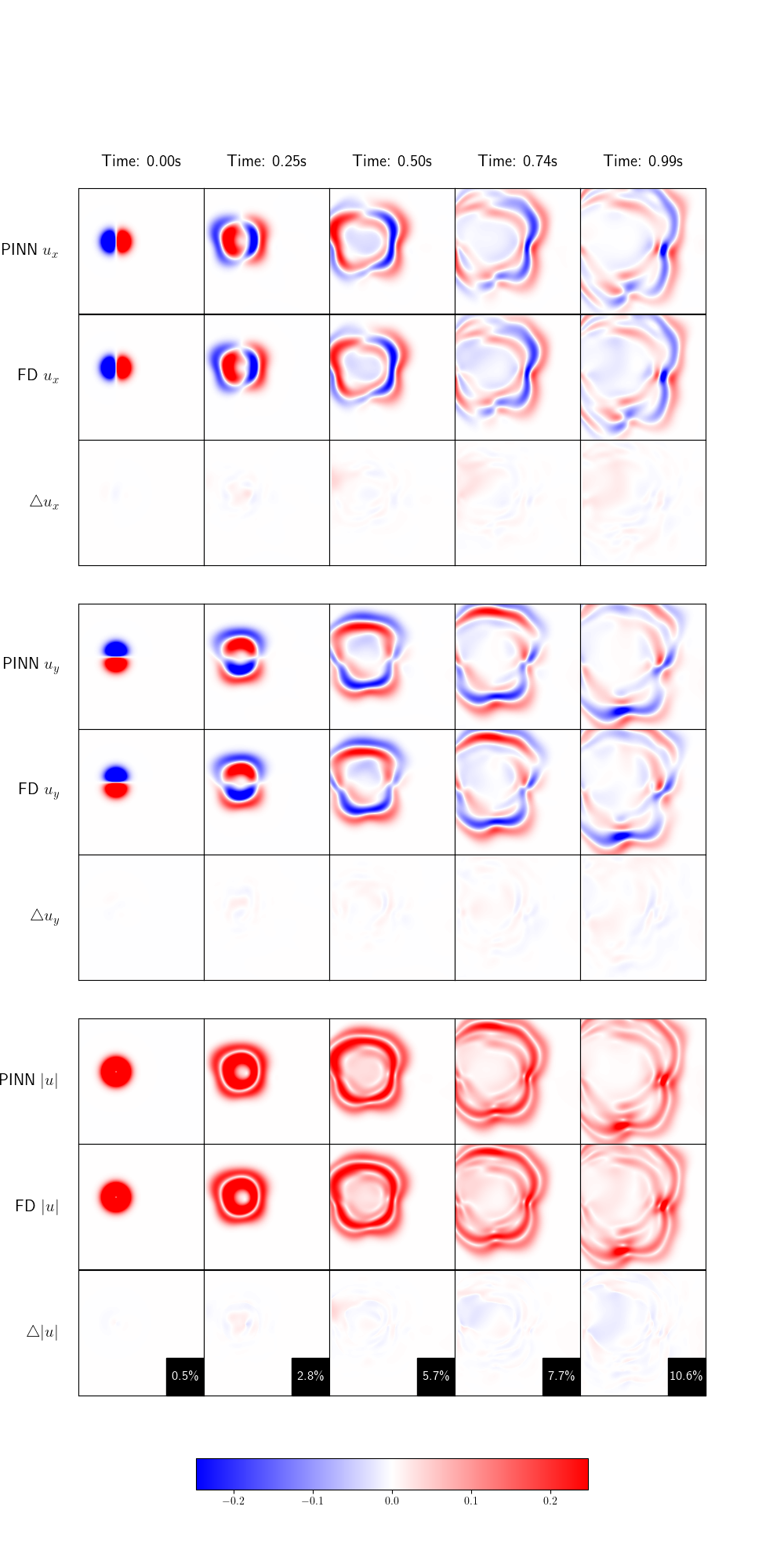}
  \caption{\textit{pED}-PINN model}
  \label{fig:planewave_bad_}
\end{subfigure}
\caption{Elastic wavefield solution generated with the conditioned PINN-tanh (left) and \textit{pED}-PINN (right) with $s = (-0.04,-0.4)$, tested against the FDM elastic wavefield solution generated with DEVITO.}
\label{fig:conditioned_bad}
\end{figure}

To conclude this analysis, we present wavefield simulations from two selected source locations, depicted in Figures (\ref{fig:conditioned_good}) and (\ref{fig:conditioned_bad}). The former illustrates the wave propagation for a source at $s = (0.26,-0.06)$, selected due to its association with lower $L_2$ errors, whereas the latter represents the outcome for a source at $s = (-0.04,-0.4)$, a location known to induce higher errors. Figure (\ref{fig:conditioned_good}) demonstrates that the \textit{pED}-PINN accurately captures primary wavefronts and even finer structural details, as exemplified by the $u_y$ displacement field at $t=0.99$, where such intricate structures are depicted with commendable precision. Conversely, the PINN-tanh model exhibits increased errors throughout the spatial domain, particularly near the boundaries and within the interior, where it fails to accurately represent finer detail structures at primary wavefronts. For the specified source, the \textit{pED}-PINN achieved an average relative $L_2$ error of $2.89\%$, compared to $4.20\%$ for the PINN-tanh model.

Figure (\ref{fig:conditioned_bad}) illustrates a scenario where both models exhibit high errors, with the PINN-tanh model displaying particularly low accuracy across the entire domain, especially in regions with fine-structured details. This is notably evident in the centre of the spatial domain at later timesteps, specifically at $t=0.74$ and $t=0.99$. In this instance, the \textit{pED}-PINN achieved an average relative $L_2$ error for the specified source of $5.43\%$, whereas the PINN-tanh model recorded an error of $8.70\%$.

%FOR ME: DONE UNTIL HERE
\subsection{Acoustic Conditioning}
\label{ACOUSTIC_CONDITIONING_RESULTS}
Building upon the rationale introduced in Section (\ref{section:arch_acoustic}), wherein we examined the PINN-tanh model and the encoder-decoder-PINN in the context of the acoustic wave equation, we extended our analysis to include their performance on the conditioned acoustic wave equation. This exploration was motivated by the objective to further validate the efficacy of our innovative encoder-decoder model (\textit{pED}-PINN) and to reinforce the assertion that it represents a versatile solution rather than one overly specialized for narrow scenarios of the elastic wave equation. Moreover, comparing the $L_2$ errors between the conditioned acoustic wave equation and its elastic counterpart provides valuable insights into their distinctions and parallels.

The training setup adheres to the one presented in Section (\ref{section:arch_choice_hyper_conditioned}).The network architecture and hyperparameters for the \textit{pED}-PINN remain unchanged. However, for the standard PINN model, an adjustment was necessary, increasing the number of layers from five to seven to obtain reasonable accuracy. This modification again underscores a significant limitation of standard FCN PINNs: their pronounced sensitivity to hyperparameter configurations.

\begin{figure}
\begin{subfigure}{\textwidth}
  \centering
  \includegraphics[width=\linewidth]{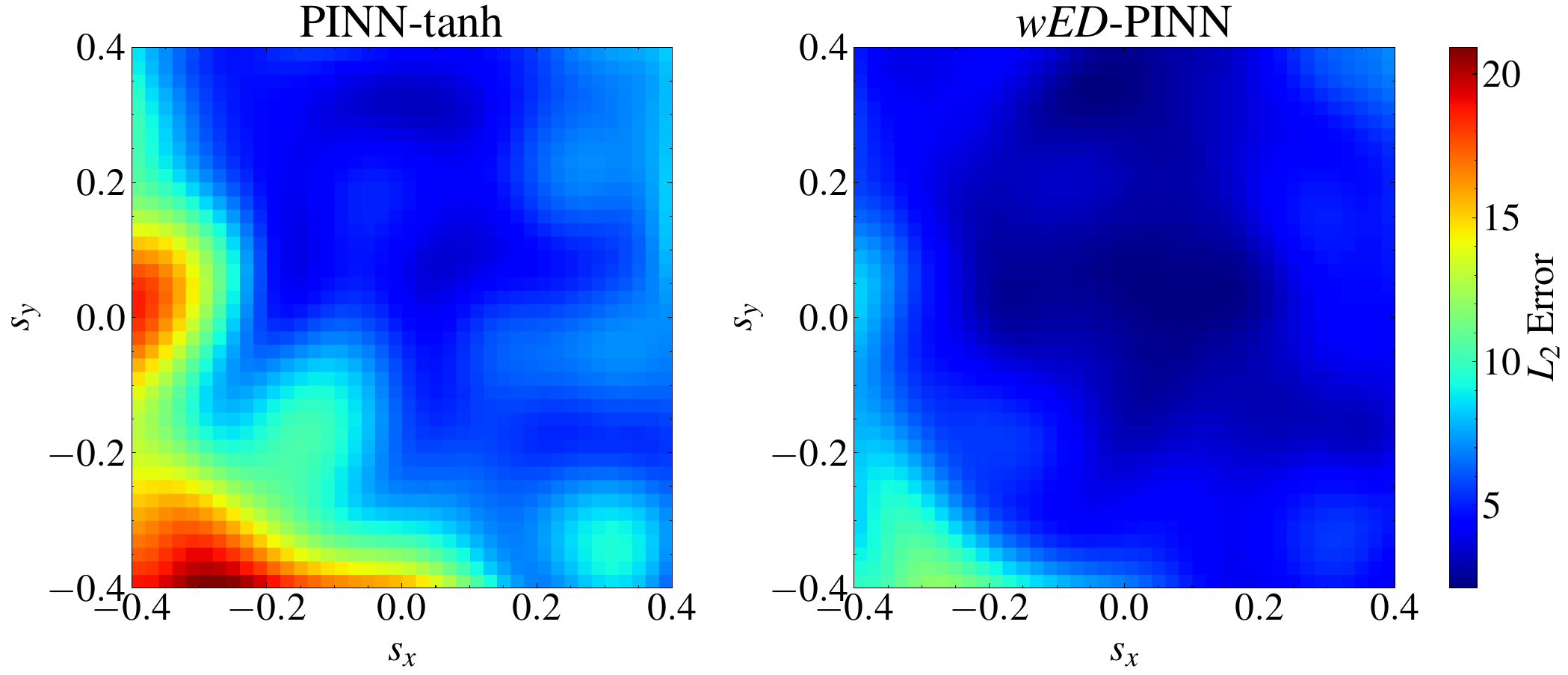}
  \caption{Error distribution in $\Omega_s$}
  \label{fig:04a}
\end{subfigure}\\
\begin{subfigure}{\textwidth}
  \centering
  \includegraphics[width=\linewidth]{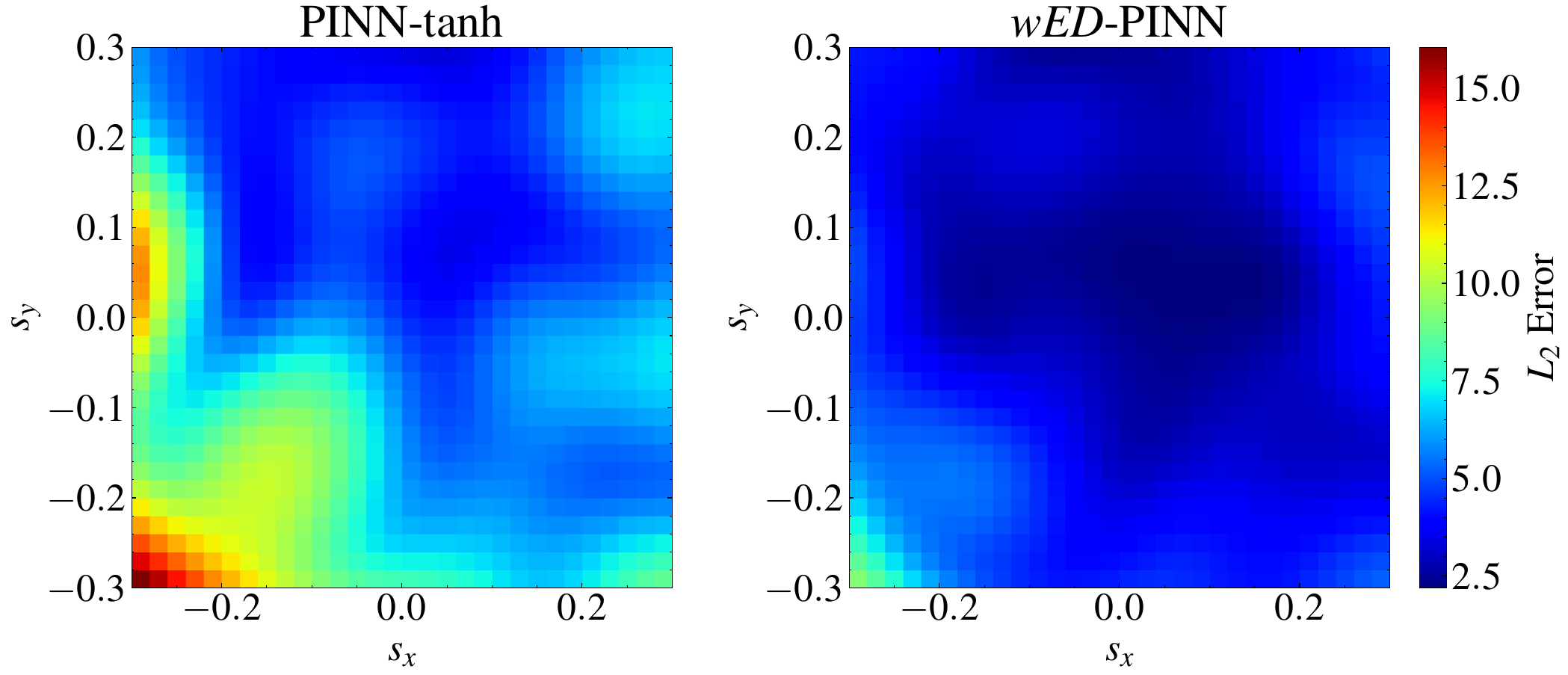}
  \caption{Error distribution in $\Omega_{\hat{s}}$}
  \label{fig:03a}
\end{subfigure}
\caption{Comparison of the spatial variance in average relative $L_2$ errors between PINN-tanh and \textit{pED}-PINN models for solving the conditioned acoustic wave equation. The errors are evaluated within the subdomains $\Omega_s = \{(x,y) \,|\, -0.4 \le x \le 0.4, -0.4 \le y \le 0.4\}$ and $\Omega_{\hat{s}} = \{(x,y) \,|\, -0.3 \le x \le 0.3, -0.3 \le y \le 0.3\}$, respectively, and tested the scalar pressure field generated with DEVITO.}
\label{fig:error_maps_accoustic}
\end{figure}

\begin{figure}
\begin{subfigure}{.48\textwidth}
  \centering
  \includegraphics[width=\linewidth]{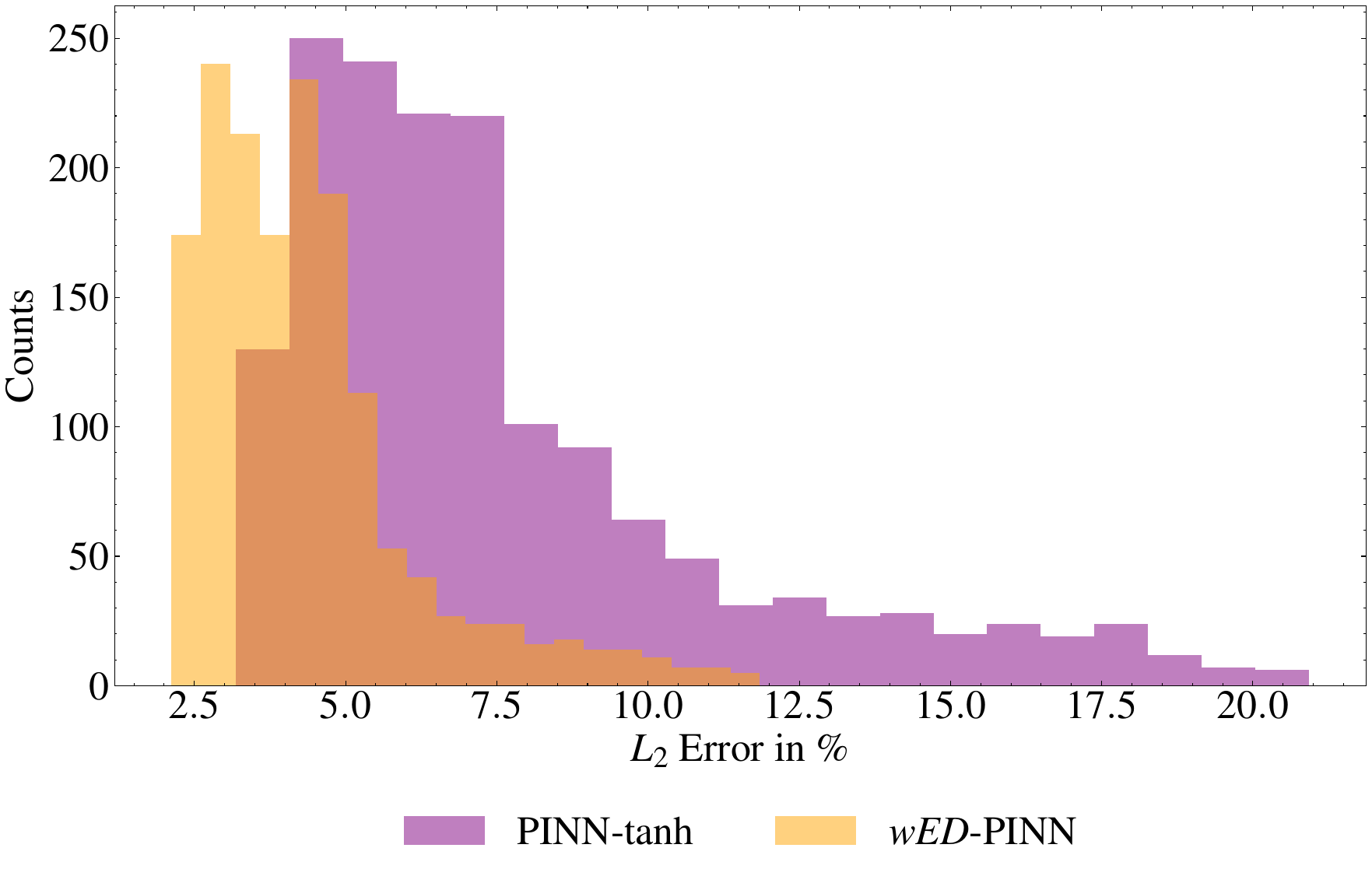}
  \caption{Histograms of the average relative $L_2$ errors in $\Omega_s$}
  \label{fig:04ah}
\end{subfigure}\hfill
\begin{subfigure}{.48\textwidth}
  \centering
  \includegraphics[width=\linewidth]{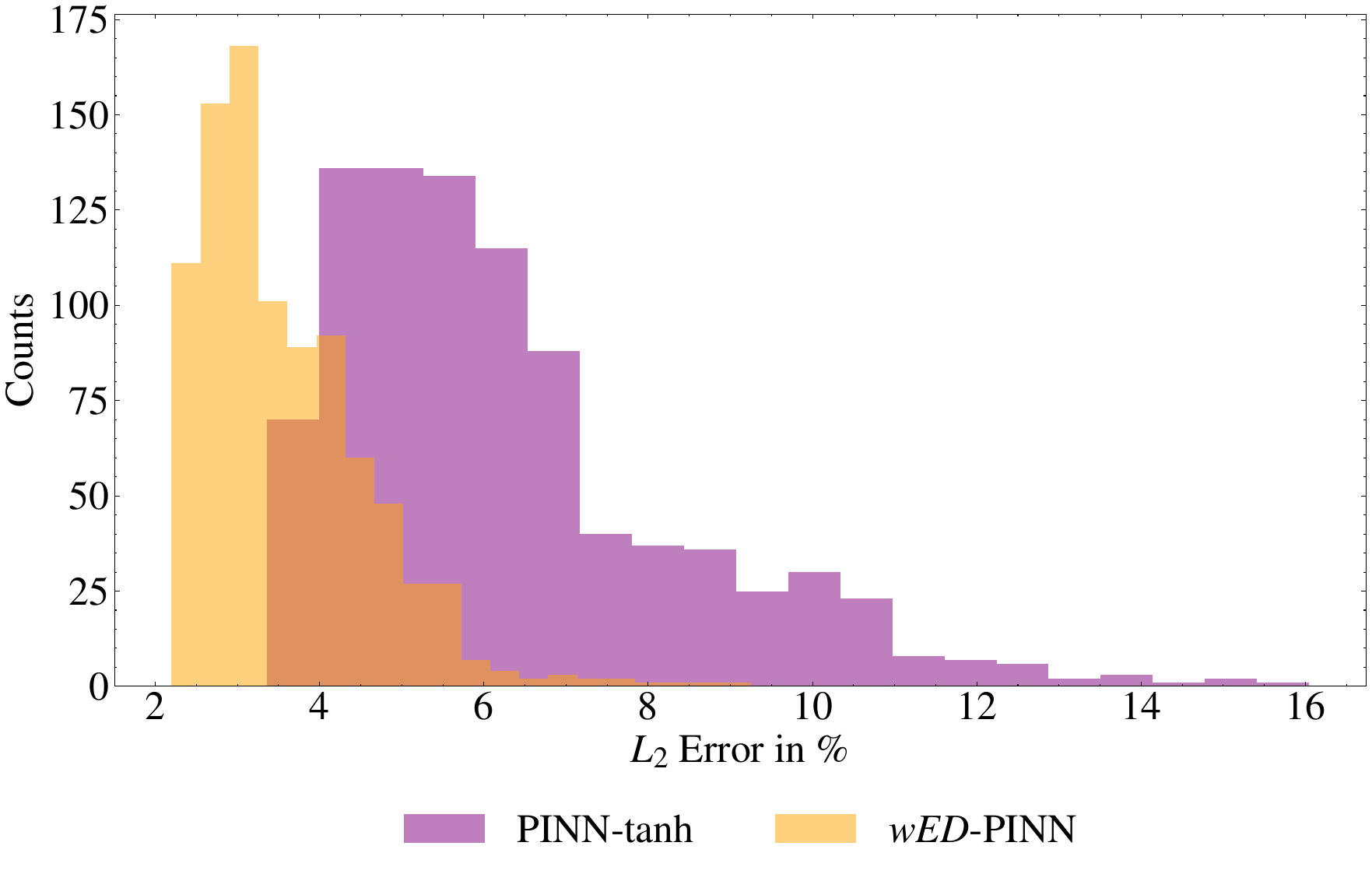}
  \caption{Histograms of the average relative $L_2$ errors in $\Omega_{\hat{s}}$}
  \label{fig:03ah}
\end{subfigure}
\caption{Histograms comparing the average relative $L_2$ errors of the conditioned PINN-tanh and the \textit{pED}-PINN models, applied to solve the acoustic wave equation within the subdomains $\Omega_s = \{(x,y)| -0.4 \le x \le 0.4, -0.4 \le y \le 0.4\}$ (left) and $\Omega_{\hat{s}} = \{(x,y)| -0.3 \le x \le 0.3, -0.3 \le y \le 0.3\}$ (right), tested against the FDM solution produced using DEVITO.}
\label{fig:histogram_acoustic}
\end{figure}

As with the elastic wave equation, we conducted inference on $1600$ source locations distributed on a regular $xy$-grid within the same subdomain $\Omega_s$ as previously described. Figure (\ref{fig:04a}) illustrates the spatial variation of the average relative $L_2$ error across different source locations. The analysis of these results closely mirrors the discussion presented for the elastic-conditioned experiment, with \textit{pED}-PINNs demonstrating superior accuracy over the standard PINN model when applied to the conditioned-acoustic wave equation.
For both models, errors tend to increase towards the corners and edges of the boundary, and notably, for the PINN-tanh model, towards the bottom half of the domain $\Omega_s$. In an effort to omit regions potentially affected by undersampling, as was undertaken for the elastic case, Figure (\ref{fig:03a}) highlights the error distribution within the subdomain $\Omega_{\hat{s}}$. Again, it is evident that \textit{pED}-PINNs offer a more accurate solution overall. The detailed error distributions for the two models across both $\Omega_s$ and $\Omega_{\hat{s}}$ are presented in Figure (\ref{fig:histogram_acoustic}).

\begin{figure}
\begin{subfigure}{.5\textwidth}
  \centering
  \includegraphics[width=\linewidth]
  {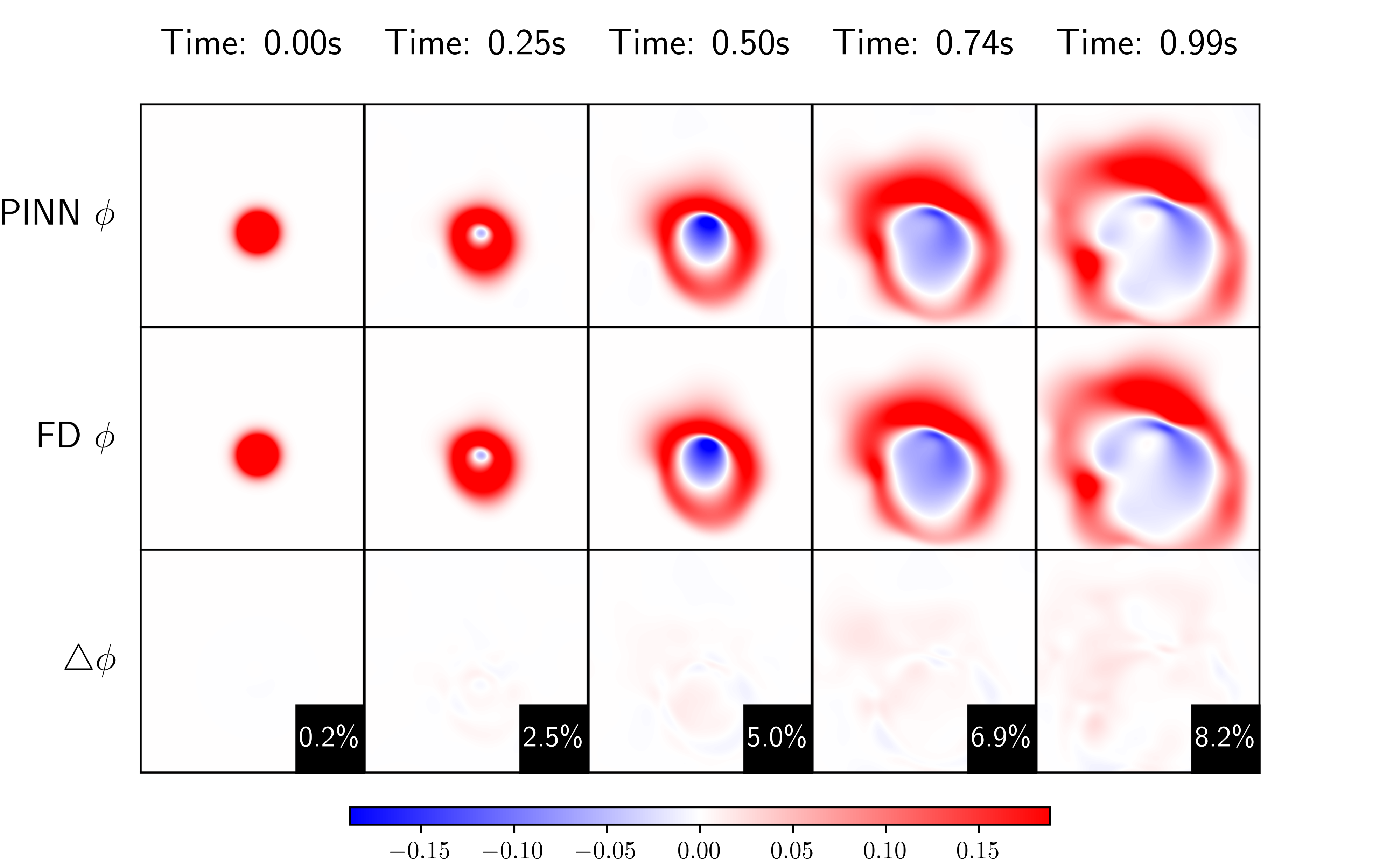}
  \caption{PINN-tanh model}
  \label{good_van_a}
\end{subfigure}%
\begin{subfigure}{.5\textwidth}
  \centering
  \includegraphics[width=\linewidth]
  {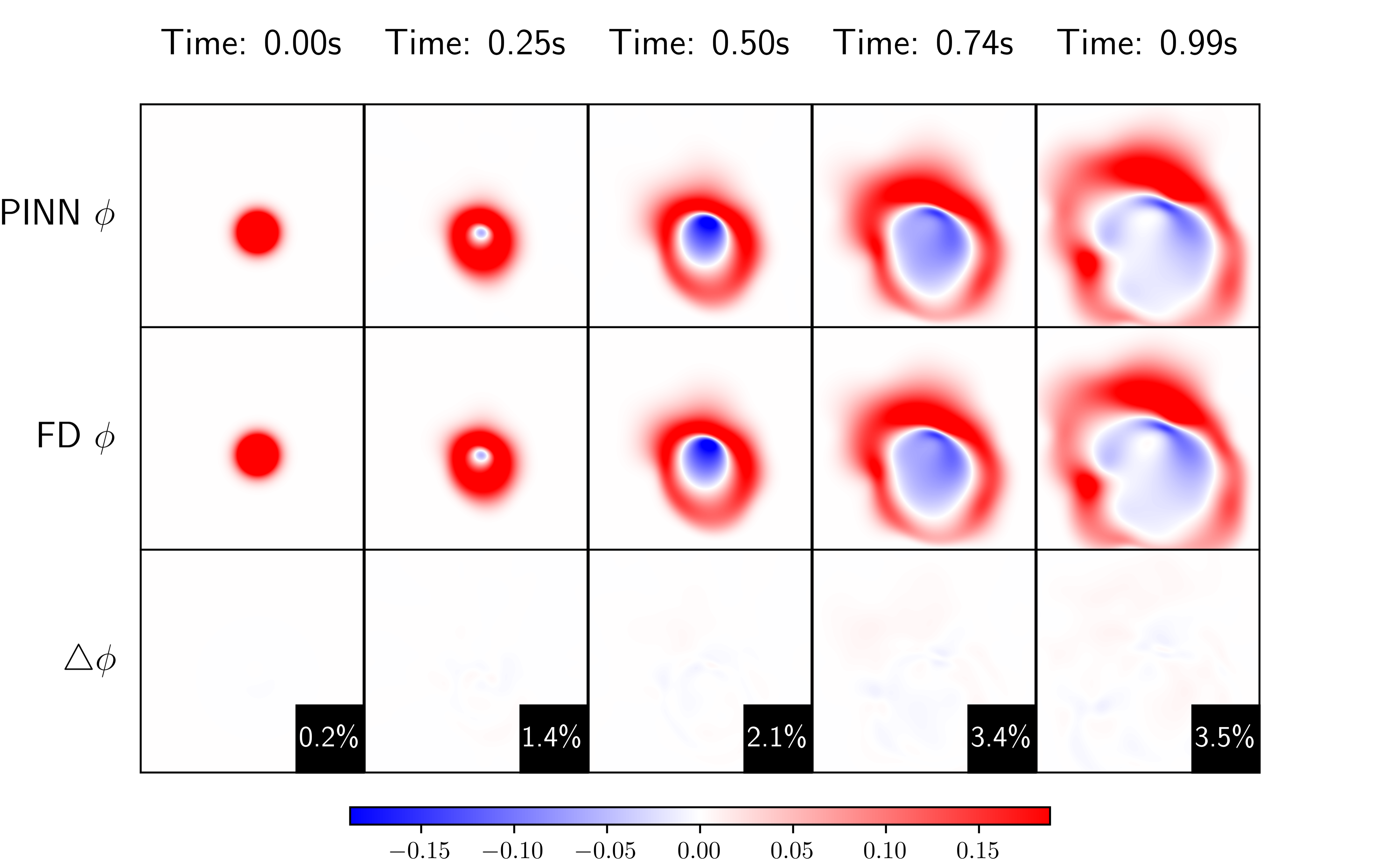}
  \caption{\textit{pED}-PINN model}
  \label{good_plane_a}
\end{subfigure}
\caption{Acoustic pressure field solution generated with the conditioned PINN-tanh (left) and \textit{pED}-PINN (right) with $s = (0.15,-0.05)$, tested against the FDM solution generated with DEVITO.}
\label{fig:example_conditioned_acoustic_good}
\end{figure}

\begin{figure}
\begin{subfigure}{.5\textwidth}
  \centering
  \includegraphics[width=\linewidth]
  {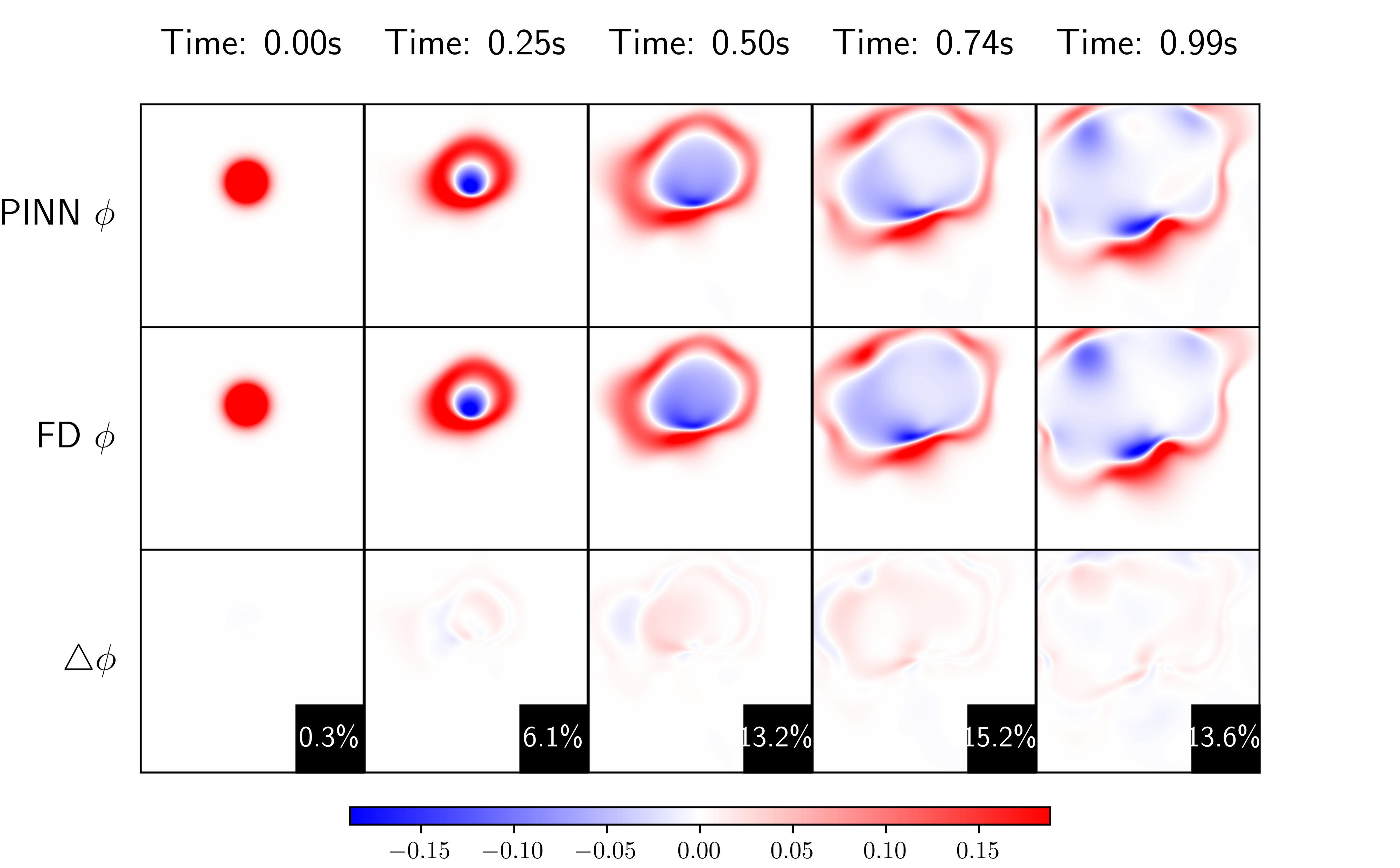}
  \caption{PINN-tanh model}
  \label{bad_van_a}
\end{subfigure}%
\begin{subfigure}{.5\textwidth}
  \centering
  \includegraphics[width=\linewidth]{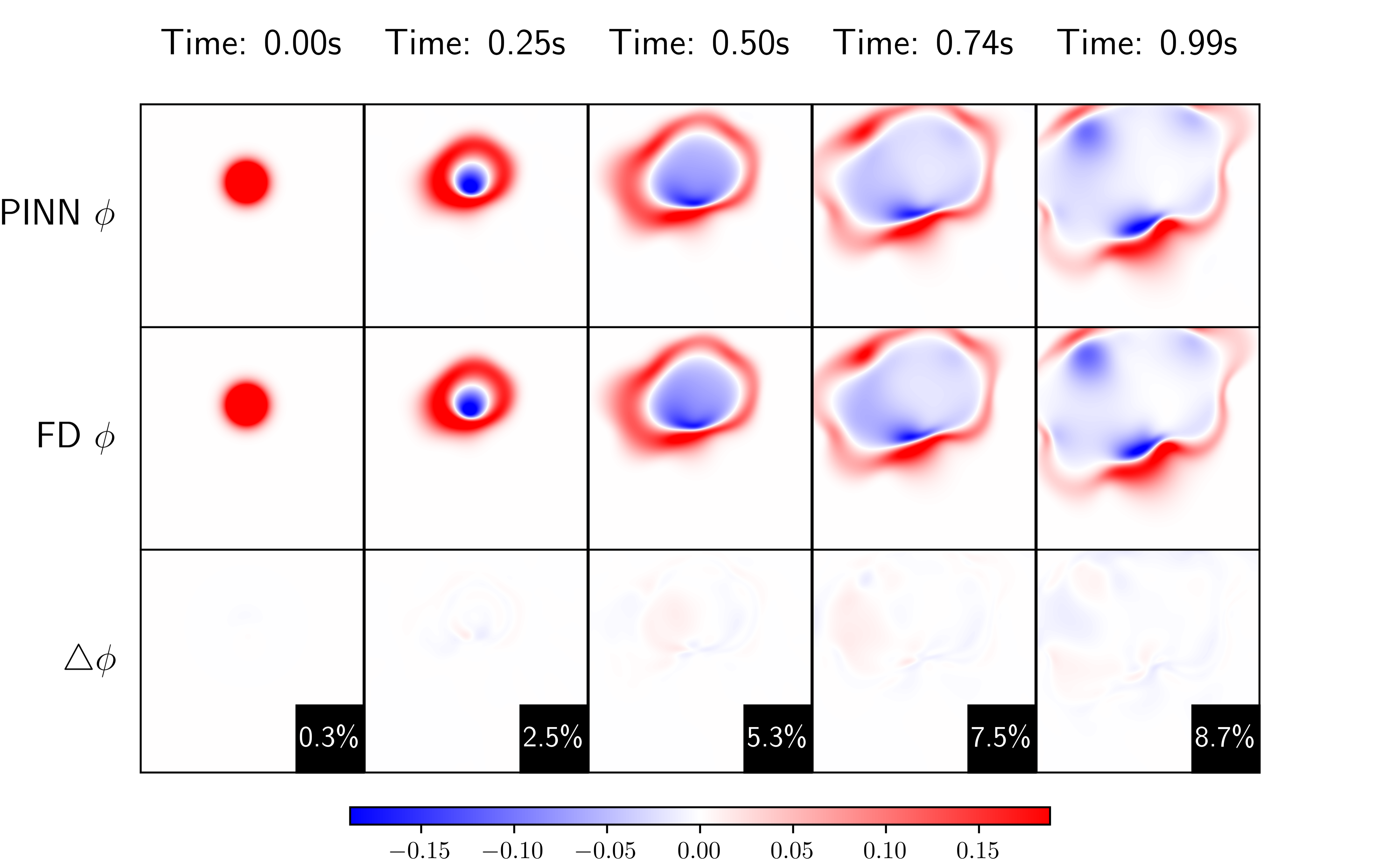}
  \caption{\textit{pED}-PINN model}
  \label{bad_plane_a}
\end{subfigure}
\caption{Acoustic pressure field solution generated with the conditioned PINN-tanh (left) and \textit{pED}-PINN (right) with $s = (-0.3,-0.05)$, tested against the FDM generated with DEVITO.}
\label{fig:example_conditioned_acoustic_bad}
\end{figure}

Figures (\ref{fig:example_conditioned_acoustic_good}) and (\ref{fig:example_conditioned_acoustic_bad}) display exemplary pressure fields generated using the source locations $s_1 = (0.15,-0.05)$ and $s_2 = (-0.3,-0.05)$, respectively. The source location $s_1$ is associated with an \textit{easier} solution, characterised by low-error outcomes, whereas $s_2$ corresponds to a \textit{harder} solution, indicative of higher-error results. These examples serve to illustrate the varying levels of challenge posed by different source locations within the conditioned-acoustic wave equation framework.

Both figures corroborate a consistent observation: the \textit{pED}-PINN model achieves superior accuracy and captures finer details more precisely than the standard PINN model, which exhibits significant discrepancies when contrasted with the FDM simulation results. Specifically, the \textit{pED}-PINN records an average relative $L_2$ error of $2.21\%$ for the pressure field generated from source location $s_1$, as illustrated in Figure (\ref{good_plane_a}), and $4.84\%$ for that originating from $s_2$, depicted in Figure (\ref{bad_plane_a}). In comparison, the PINN-tanh model exhibits higher average relative $L_2$ errors of $4.70\%$ and $10.33\%$ for the respective source locations, as shown in Figures (\ref{good_van_a}) and (\ref{bad_van_a}).

\subsection{Consolidated Results}
To encapsulate the findings from Sections (\ref{ELASTIC_CONDITIONING_RESULTS}) and (\ref{ACOUSTIC_CONDITIONING_RESULTS}), Table (\ref{tab:conditioned}) summarises the mean average relative $L_2$ errors for both subdomains and network types, covering both the elastic and acoustic wave equations. The table also includes the respective standard deviations of the error distributions, offering a comprehensive overview of the comparative performance and accuracy of the \textit{pED}-PINN and standard PINN models across different wave equations.

\begin{table}[ht]
\centering
\begin{tabular}{llcc}
\toprule
 & & \textbf{PINN-tanh} & \textbf{\textit{pED}-PINN} \\
\midrule
\multirow{2}{*}{Elastic} & $\Omega_s$& $\mu=5.48,\sigma=1.38$& $\mu=4.17,\sigma=1.06$\\
 & $\Omega_{\hat{s}}$& $\mu=5.15,\sigma=1.08$& $\mu=3.80,\sigma=0.67$ \\
\cmidrule{1-4}
\multirow{2}{*}{Acoustic} & $\Omega_s$ & $\mu=7.58,\sigma=3.62$ & $\mu=4.35,\sigma=1.76$ \\
 & $\Omega_{\hat{s}}$ & $\mu=6.28,\sigma=2.13$ & $\mu=3.57,\sigma=1.02$\\
\bottomrule
\end{tabular}
\caption{Summary of the spatial mean ($\mu$) and standard deviation ($\sigma$) for the average relative $L_2$ errors associated with the PINN-tanh and \textit{pED}-PINN models in solving the conditioned elastic and acoustic wave equations}
\label{tab:conditioned}
\end{table}
The results in Table (\ref{tab:conditioned}) demonstrate that by employing a suitable network architecture, specifically our encoder-decoder model designed to incorporate prior wave physics knowledge, we can successfully condition the network on the source location for both the elastic and acoustic wave equations within a highly heterogeneous parameter environment. While there is room for improvement in the achieved relative $L_2$ errors, we are confident in considering these errors to be reasonably low. This confidence is supported by the network's ability to accurately simulate wave propagation in complex settings, indicating a promising direction for future enhancements and applications of PINNs in wave physics simulations.

\section{Applicability of PINNs: A Seismic Hazards Case Study}
\label{Section:Case_study}
\subsection{Introduction}
Beyond the precision of PINNs, our exploration now extends into their practical applicability in real-world scenarios, particularly within the realm of seismology. The question at hand is whether PINNs represent a viable alternative to traditional numerical methods. Our observations indicate that while PINNs can approach the accuracy of conventional methods in simpler scenarios, such as those involving a single source or exhibiting very low relative $L_2$ errors (approximately $1\%$), their performance in more complex situations, like the conditioned cases, tends to be less accurate. Therefore, the appeal of PINNs does not solely rest on their accuracy, which, for the time being, may not surpass that of traditional methods.
The true advantage of PINNs lies in their capacity to learn and generalise across a broad class of problems. This capability was illustrated through the conditioning over source location, showcasing an aspect of PINNs that transcends mere accuracy metrics. We aim to evaluate how this feature of PINNs compares with traditional numerical methods, specifically FDM simulations using DEVITO, in terms of performance or inference time.
One immediate efficiency gain with PINNs is their ability to directly infer the wavefield solution at a specific time point $t_n$ without sequentially computing solutions at every timestep up to $t_n$, as is necessary with traditional methods. This characteristic potentially offers a significant speedup for larger values of $t_n$ and smaller timestep intervals. However, any speedup is negated if the PINN requires retraining for each new scenario, such as a novel source location, due to the substantial time investment needed for training.
PINNs reveal their full potential when considering scenarios with multiple seismic sources. Traditional computational approaches necessitate separate simulations for each source, resulting in a linear increase in computational effort. Conversely, PINNs are capable of learning a function that integrates source location, spatial coordinates, and time, thereby facilitating the evaluation of multiple source locations concurrently.
To illustrate these advantages, we propose a simplified seismological case study aimed at highlighting the performance and efficiency of PINNs in comparison to traditional FD simulations. This case study seeks to demonstrate the practical implications of PINNs' unique capabilities, particularly their ability to handle multiple seismic sources efficiently, marking a significant step forward in their applicability to real-world seismological challenges.
\subsection{Setup}
\begin{figure}
\begin{center}
  \includegraphics[width=0.99\linewidth]{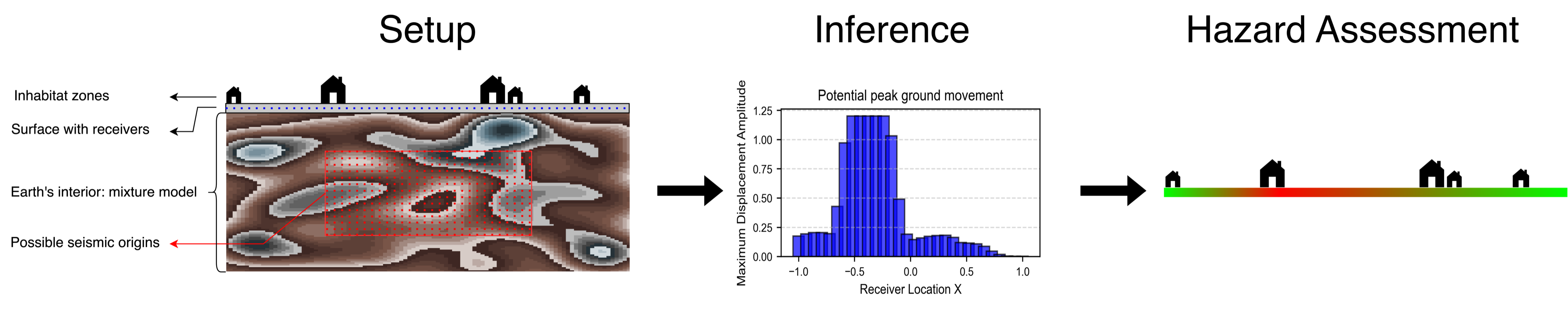}
  \caption{Visualization of the steps of a toy example seismic hazard assessment. Step A: Initially, the region of interest is delineated (red rectangle). Within this region, a number of possible seismic source locations are sampled (red dots).Step B: Utilizing the trained, conditioned PINN model, the maximum displacement field amplitudes at each receiver point (blue dots) are computed. Step C: Assessment of potential seismic hazard in the region of interest is conducted.}
  \label{fig:amplitude_experiment}
  \end{center}
\end{figure}

A pivotal aspect of seismology involves seismic hazard assessment, where the objective is to evaluate the seismic risk for a specific region statistically. To illustrate this, consider the following simplified scenario: an area with known seismic activity is analyzed, where the underlying structure and geometries of the Earth's crust are characterised by the first and second Lam{\'e} parameters, as described by the mixture model, alongside a constant density. Within this region, a subset is identified as a potential origin of earthquakes.
The methodology involves running simulations for each possible seismic origin within the identified subset and calculating the maximum amplitude of the displacement field at the surface, observed at predefined locations known as receiver positions. In real-world scenarios, these receiver positions are equipped with physical devices that record quantities such as the amplitude of incoming elastic waves. By conducting this analysis across all $N_s$ potential seismic sources, it is possible to determine the maximum amplitude measured at each receiver location, considering all considered seismic origins. This data then serves as a basis for assessing the seismic risk of the area in question.
Figure (\ref{fig:amplitude_experiment}) visually represents these procedural steps.

In the traditional setting, scientists are required to run an individual FD/FEM/SEM simulation for each seismic origin. This approach is computationally expensive, especially when considering a large number of potential seismic origins and the complexity of three-dimensional simulations. With conditioned PINNs, conducting the entire procedure in a single inference or forward pass is possible by designing a suitable input tensor.

The input tensor, denoted by \(\mathbf{I}\), encapsulates the simulation parameters across receiver locations, source locations, and time steps. It has dimensionality:
\[
\mathbf{I} \in \mathbb{R}^{(N_r \times N_t \times N_s) \times 5},
\]
and is constructed as follows:
\[
\mathbf{I}[i, j] = 
\begin{cases} 
t_{(i \mod N_t)} & \text{if } j = 0, \\
r_x\left[\left(\frac{i}{N_t \cdot N_s}\right) \mod N_r\right] & \text{if } j = 1, \\
0.99 & \text{if } j = 2, \\
s_x\left[\left(\frac{i}{N_t}\right) \mod N_s\right] & \text{if } j = 3, \\
s_y\left[\left(\frac{i}{N_t}\right) \mod N_s\right] & \text{if } j = 4.
\end{cases}
\]

\(N_s\) represents the number of source locations arranged in an arbitrary grid within the spatial domain. We specified the sources  with \(s_x \in [-0.4, -0.3]\) and \(s_y \in [-0.4, 0.3]\), defined by \((s_x[k], s_y[k])\) for each source \(k\), where \(k = 0, \ldots, N_s-1\). Typically, these source locations would be sampled along a pattern like a geological fault or line or randomly throughout the domain or subdomain. \(N_r=30\) is the total number of receivers, evenly distributed along the line with \(r_y = 0.99\), and \(r_x[k]\) being the \(x\)-coordinate of the \(k\)-th receiver, for \(k = 0, \ldots, N_r-1\). \(N_t\) is the number of time points, uniformly distributed  in $[0,1]$, repeated for each source and receiver pair, with \(N_t=10\) providing sufficient resolution. \(t_k\) is the \(k\)-th time step, with \(k = 0, \ldots, N_t-1\). Using these values we have \(N_s \times 300\) rows of input points, meaning that for each possible receiver and source pair, \(N_t\) time points are considered to evaluate the wavefield.

For example, consider an input tensor defined by $N_s, N_t, N_r = 2$ as follows:
\[
\mathbf{I} = \left[
\begin{array}{ccccc}
t_0 & r_{x_1} & 0.99 & s_{x_1} & s_{y_1} \\
t_1 & r_{x_1} & 0.99 & s_{x_1} & s_{y_1} \\
t_0 & r_{x_1} & 0.99 & s_{x_2} & s_{y_2} \\
t_1 & r_{x_1} & 0.99 & s_{x_2} & s_{y_2} \\
t_0 & r_{x_2} & 0.99 & s_{x_1} & s_{y_1} \\
t_1 & r_{x_2} & 0.99 & s_{x_1} & s_{y_1} \\
t_0 & r_{x_2} & 0.99 & s_{x_2} & s_{y_2} \\
t_1 & r_{x_2} & 0.99 & s_{x_2} & s_{y_2} \\
\end{array}
\right].
\]

Upon processing the input tensor with our PINN model, the resultant output tensor $\mathbf{U}$ exhibits the dimensionality
\[
\mathbf{U} \in \mathbb{R}^{(N_r \times N_t \times N_s) \times 2},
\]
The two dimensions correspond to the $x$ and $y$ components of the displacement field $\textbf{u}$, denoted as $u_x$ and $u_y$, respectively. The magnitude of displacement is calculated as $|u| = \sqrt{u_x^2 + u_y^2}$. This tensor is then partitioned into slices after every $N_s \times N_t$ rows, with each slice representing the wavefield at time steps $k = 0, \ldots, N_t-1$, for $N_s$ different source locations at a specific receiver location. The maximum value within each slice is identified to represent the peak displacement amplitude at the receiver location.

\subsection{Results}
The computational efficiency of this method was assessed for\newline $N_s \in[10,100,1000,10000,100000]$, comparing these runtimes with those required to execute the FD solution utilizing DEVITO for an equivalent count of source locations. This comparison was extended across three spatial grid resolutions employed in the FDM solution: $64\times64$, $256\times256$, and $512\times512$. As the grid becomes finer, the simulation duration increases, albeit yielding greater accuracy due to a reduced discretization error, which decreases with $\mathcal{O}(\triangle x^4)$, reflecting the fourth-order spatial scheme employed.

Upon comparing the FDM solutions at $64\times64$ and $256\times256$ resolutions against the pseudo-reference solution obtained with a $512\times512$ grid, it was found that the $64\times64$ FD solution exhibited an average relative $L_2$-error of approximately $7\%$ over multiple ($100$) source locations, indicating a higher error rate than that of the PINN solution, which recorded relative $L_2$ errors of $4.1\%$ and $3.8\%$, dependent on the chosen subdomain $\Omega_s$. Conversely, the $256\times256$ FD solution demonstrated an average relative $l_2$ error of around $1\%$ when benchmarked against the reference FD simulation.

Table (\ref{tab:runtimes}) details the runtime comparisons of the three FDM simulations with varying grid resolutions and the PINN simulation, all executed on the same single CPU. The PINN simulation was also conducted on a single NVIDIA TITAN RTX GPU, equipped with $20GB$ of GPU memory. It was noted that executing the FDM simulation on the GPU did not yield performance enhancements, as the FD simulation is not optimised for GPU utilization. Optimizing for GPU could potentially improve performance; however, such enhancements fall beyond the scope of this comparison. 

\begin{table}[ht]
\centering
\begin{tabular}{l|ccc||cc}
\toprule
 & \textbf{FD-64} & \textbf{FD-256} & \textbf{FD-512} & \textbf{PINN-CPU} & \textbf{PINN-GPU} \\
\midrule
$N_s:10$ & $4.2s$& $33.1s$& $2.3m$&$0.04s$ & $0.5s$\\
$N_s:100$ & $46.7s$& $6.1m$& $27.0m$&$0.1s$ & $0.5s$\\
$N_s:1000$ & $6.7m$& $>1h$& $>4h$&$2.2s$ & $0.5s$\\
$N_s:10000$ & $>1h$& $>10h$& $>1.8d$&$67.0s$ & $0.5s$ \\
$N_s:100000$ & $>10h$& $>4d$& $>10d$&$14.2m^\ast$ & $1.5s^\ast$\\
\bottomrule
\end{tabular}
\caption{Comparison of the FDM solution runtimes at resolutions ($64\times64, 256\times256, 512\times512$) against PINN inference times on CPU and GPU for varying numbers of sources ($N_s$). Times preceded by $>$ are minimum extrapolations from shorter runs. Runtimes marked with $\ast$ indicate computations performed over multiple CPU/GPU runs due to memory constraints.}
\label{tab:runtimes}
\end{table}
The results indicate that the PINN approach yields solutions which are magnitudes faster than the FDM, even when employing DEVITO, a highly efficient library. For instance, simulating $100,000$ source locations using an FDM approach could take days or weeks, depending on grid resolution. In contrast, PINNs reduce computation times to minutes on a CPU and seconds on a GPU. It is important to note that these comparisons are based on two-dimensional simulations and are conditioned solely on one parameter group: the source location. Should PINNs be successfully conditioned on additional parameters, such as the underlying parameter structure, or applied to solving the three-dimensional elastic wave equation, the speed advantage could increase exponentially.
However, the applicability of this comparison is highly dependent on the application's specific requirements. Traditional numerical methods may still be preferable for tasks that demand intricate dynamic behaviour and waveform analysis for a single seismic source. PINNs are particularly advantageous in situations that require quick, extensive evaluations, such as assessing amplitude responses across various potential earthquake sources. PINNs effectively bypass the traditional simulation process by learning the underlying physical functions. Nonetheless, the training duration for these conditioned networks should not be overlooked; our findings demonstrate that training for the most complex cases can take up to $30$ hours. Consequently, when the number of seismic sources to be evaluated does not exceed $100000$, and the resolution is capped at $256\times256$, the efficiency of PINNs aligns with that of traditional numerical methods for individual source location analyses. This highlights the context-dependent superiority of PINNs, which can surpass traditional numerical methods in function approximation when appropriately conditioned on a set of problems.

\section{Discussion}
In this chapter, our focus has been on expanding the capabilities of PINNs to solve the elastic wave equation for any source location within a specified subdomain $\Omega_s$, moving beyond the limitation of a single source location. We have successfully demonstrated this capability within a highly heterogeneous parameter setting and have further underscored the improvement in accuracy achieved by utilizing the encoder-decoder type PINN, specifically the \textit{pED}-PINN, in contrast to the standard PINN model. We argue that the level of accuracy attained is sufficient for the PINN solutions to be deemed reliable for practical seismological applications. However, there is potential for further enhancement in accuracy.
 
The most direct approach to improve accuracy involves the use of additional training/collocation points. Despite utilizing $500000$ collocation points, which may seem substantial, our full-conditioning sampling methodology allocates a single spatiotemporal training point per individual source location. An increase in the number of collocation points would enhance the density of source sampling, which is anticipated to improve accuracy, as our observations suggest a positive correlation between the number of collocation points and accuracy outcomes. Given that the available 80GB of GPU memory has been fully utilised, adopting this strategy would necessitate a transition to mini-batch training from full-batch training, consequently requiring the adoption of optimisers compatible with batch training, such as ADAM or ADAGRAD, due to LBFGS's incompatibility with such training modalities. Achieving similar convergence levels with these alternative optimisers, particularly for the elastic wave equation, remains a challenge beyond the scope of our current project timeline and would necessitate further engineering efforts.

Exploring more sophisticated network architectures represents another avenue for increasing accuracy. Notably, we employed the \textit{pED}-PINN for cases conditioned on the elastic wave equation with an underlying mixture model, whereas in the unconditioned scenario, the \textit{wED}-PINN exhibited superior performance compared to the \textit{pED}-PINN when applied to the mixture model. Although the difference is marginal, leveraging the \textit{wED}-PINN in the conditioned mixture case might enhance accuracy further. However, the \textit{wED}-PINN's performance was significantly inferior in the conditioned scenario compared to the \textit{pED}-PINN. The reason behind this discrepancy remains unclear and may be attributed to wavelets' inherent spatial localization characteristics compared to plane waves. Future work is necessary to develop an alternative conditioned \textit{wED}-PINN model that matches or surpasses the accuracy of the \textit{pED}-PINN. Additionally, exploring more innovative models, such as a \textit{wpED}-PINN that integrates both wavelets and plane waves, could yield further accuracy improvements.

In our case study, we demonstrated that although PINNs may not universally outperform traditional numerical solvers across all seismological analysis dimensions, they offer substantial advantages in scalability and flexibility, crucial for specific applications. Notably, PINNs' ability to simultaneously process multiple seismic sources in a single inference significantly enhances their utility in seismology. For instance, we highlighted PINNs' superiority over conventional FDMs in computational speed. This leap forward transforms seismic hazard assessment by reducing simulation times from days to mere minutes or seconds. This breakthrough underscores the potential for substantial progress, yet it also marks the beginning rather than the endpoint of research in this area.  Adhering to a no-labelled-data PINN setup, as opposed to using FNOs or PINOs, further conditioning on the underlying parameter model could significantly enhance the applicability of PINNs. However, this endeavour is not straightforward and necessitates carefully selecting network architecture and training setup alongside a significant degree of engineering ingenuity.

\section{Conclusion}
In this chapter, we explored how conditioning PINNs enhances efficiency by eliminating the need for retraining for every new source location. Additionally, selecting the appropriate network architecture improves accuracy in this context, providing an edge over traditional PINN approaches.
Chapter (\ref{section:Elastic PINNs}) outlined the main challenges with conventional PINNs: accuracy, hyperparameter search intensity, and efficiency. While the previous chapter addressed the first two challenges, the current chapter has leveraged insights from its predecessor to tackle the final hurdle---efficiency---through source location conditioning. This approach significantly boosts efficiency. Despite these advances, conditioned PINNs face challenges, such as reduced accuracy in some instances and lengthy training periods. Nonetheless, this strategy unveils new prospects for rapid seismic hazard assessment, showing notable speed advantages over traditional numerical methods in specific applications.
\newpage

%CHECK WITH BEN: THE NEW DISCUSSION SECTION
\chapter{Conclusion}
Throughout this thesis, we have delved deeply into the efficacy of PINNs for solving the elastic wave equation. In this concluding chapter, we aim to briefly summarize and reflect upon our discoveries, as outlined in Section (\ref{DISCUSSION}), and to articulate the current limitations alongside potential directions for future research, as will be discussed in Section (\ref{FUTURE_WORK}).
\section{Discussion}
\label{DISCUSSION}
This thesis investigates the application of PINNs as a novel approach that integrates physical laws with machine-learning techniques for solving the elastic wave equation. Traditional numerical methods, while precise, face challenges with high computational demands, particularly when simulating the elastic wave field under varied initial conditions within heterogeneous environments. Conventional machine learning, efficient in handling vast datasets, often falters in the face of sparse labelled data and generalization. Our approach circumvents these challenges by embedding the governing PDE of elastic wavefield propagation directly into the loss function of PINNs, negating the need for labelled data and offering a meshfree approach to the elastic wave equation.

While initial results appeared promising, an in-depth evaluation across a diverse range of parameter models and seismic source sizes unveiled accuracy limitations under complex scenarios. This realization spurred an extensive investigation into various network architectures designed to incorporate wave physics knowledge to varying extents. Our exploration, motivated by the hypothesis that such integration could refine PINN efficacy in solving the elastic wave equation, identified a spectrum of models with varying degrees of physical information integration. A nuanced balance emerged: optimal PINN performance was achieved at an intermediate level of physical knowledge integration that did not overly constrain the solution space.

A standout discovery was the encoder-decoder PINN architecture, augmented with wavelet or plane wave layers, which notably exceeded that of the conventional PINN model in terms of accuracy. This architecture demonstrated exceptional adaptability across various seismic scenarios and PDEs, consistently maintaining higher accuracy than the standard PINN model. This showcases the tangible advantages of incorporating physical principles into PINN architectures for seismological applications.

Although the increased accuracy represents a significant advancement for PINNs, the extended training durations pose a practical challenge, contrasting with the efficiency and accuracy of traditional methods. To counter this, we conditioned the neural network on the seismic source location, significantly improving efficiency. The conditioned encoder-decoder PINN architecture outperformed the conventional PINN method in accuracy and surpassed traditional numerical methods in inference speed for certain scenarios. Our case study on seismic hazard assessment illustrated the practical utility of the conditioned PINN, demonstrating its ability to produce hazard assessments for a vast number of potential seismic sources significantly faster than traditional solvers.

\section{Future Work}
\label{FUTURE_WORK}
Despite the promising results, we acknowledge several limitations in our study that necessitate further exploration to augment the method's accuracy and applicability on a broader scale.

A primary limitation encountered with our approach is the incompatibility of the LBFGS optimizer with batched training. This constraint limits the number of collocation points that can be used without exceeding the memory capacity of available GPUs. To address this issue, we propose the investigation of alternative optimizers, such as the ADAM optimizer, known for its compatibility with batched training. Implementing such optimizers could potentially allow for an increased number of collocation points, thereby enhancing the model's accuracy, especially in the conditioned case.

Exploring the integration of both plane waves and wavelets into an encoder-decoder PINN offers an intriguing research pathway. Our demonstration that each architecture excels under different conditions suggests that a combined architecture could lead to substantial accuracy improvements. Additionally, directly incorporating source location information into the wavelets or plane waves, thereby adapting their spatial localization explicitly, could further increase the accuracy of such models.

Our methodology currently conditions the networks solely on the source location. By augmenting this framework to incorporate additional parameters---such as the source size, type, and underlying parameter models---its applicability and competitiveness with existing numerical methods could be significantly enhanced. This expansion may require a shift towards operator learning, as exemplified by PINOs \cite{PINO}, to effectively manage the expanded parameter space with greater fidelity. Another promising direction involves foundational models, which fine-tune a large pre-trained model for each new problem setting. By adopting this strategy, one could pre-train a neural network on a comprehensive dataset encompassing multiple parameter settings, subsequently fine-tuning it on a task-specific dataset. However, this approach would steer the method towards a more data-driven technique.

The potential integration of our encoder-decoder architecture with existing methodologies, such as the efficient domain decomposition proposed in \cite{FBPINNS}, could lead to the development of a sophisticated PINN. This hybrid model would optimally adjust the wavelet or plane wave layers to their respective local subdomains, potentially shortening training times. Advanced models of this nature could significantly contribute to fast seismic hazard assessments, especially since our current models already demonstrate superior inference speed over traditional numerical methods in such applications.

Expanding these benefits to real-world geological domains presents a significant challenge and opportunity. Despite the complex challenges posed by our mixture and layered models, they are still simplified representations of the complex topologies found in natural geological settings. Testing our network architecture---or its subsequent iterations---against large-scale, real-world parameter models is crucial for determining the practical applicability of PINN methods in seismology. 

An intriguing yet unexplored area of research involves applying advanced PINN models to multi-physics domains, such as those combining elastic and acoustic domains prevalent at the Earth's ground-atmosphere interface. Given that our models have successfully solved the wave equation in these domains separately, evaluating their performance in the more complex setting of coupled domains could unveil new possibilities. Grasping the dynamics between seismic and acoustic waves in solid-fluid systems with PINNs could open the door to various applications, from enhancing planetary science research to refining terrestrial hazard predictions. 

Future enhancements in precision and expansions in the input parameter space of conditioned PINN models could position them as robust alternatives to traditional numerical methods in seismology, with a broad potential for advancing the field.

\newpage
\printbibliography
\newpage
\appendix
\chapter{Background}
\section{Python Implementation of DEVITO}
\label{DEVITO_IMPLEMENTATION}
\begin{lstlisting}[caption={Python implementation of our underlying numerical baseline built with DEVITO },captionpos=b,language=Python, label=lst:devito]
    
    u0x, u0y = initial_condition()

    ux.data[0] = u0x
    uy.data[0] = u0y
    ux.data[1] = u0x
    uy.data[1] = u0y

    div_stress_ux = (lambda_ + 2.0 * mu_) * ux.dx2 + 
    mu_ * ux.dy2 + lambda_ * uy.dy.dx + mu_ * uy.dx.dy

    div_stress_uy = (lambda_ + 2.0 * mu_) * uy.dy2 + 
    mu_ * uy.dx2 + lambda_ * ux.dx.dy + mu_ * ux.dy.dx

    pde_x = rho_solid * ux.dt2 - div_stress_ux 
    pde_y = rho_solid * uy.dt2 - div_stress_uy 
    
    stencil_x = Eq(ux.forward,solve(pde_x,ux.forward))
    stencil_y = Eq(uy.forward,solve(pde_y,uy.forward))
    
    op = Operator([stencil_x]+[stencil_y]+bc)

    op(time=end_time,dt=dt)
\end{lstlisting}

\chapter{Solving the Elastic Wave Equation with PINNs}
\label{APPENDIX:A}
\section{Pytorch Implementation of Physical Loss}
\label{Pytorch_LOSS}
\begin{lstlisting}[caption={Pytorch implementation of the physical loss used in our baseline PINN },captionpos=b,language=Python, label=lst:loss]
U = self.approximate_solution(input_s)
        
gradient_x = torch.autograd.grad(U[:, 0].unsqueeze(1).sum(), input_s, create_graph=True)[0]
gradient_y = torch.autograd.grad(U[:, 1].unsqueeze(1).sum(), input_s, create_graph=True)[0]

dt2_x = torch.autograd.grad(gradient_x[:, 0].sum(), input_s, create_graph=True)[0][:, 0]
dt2_y = torch.autograd.grad(gradient_y[:, 0].sum(), input_s, create_graph=True)[0][:, 0]

eps = 0.5 * torch.stack((torch.stack((2.0 * gradient_x[:, 1], gradient_x[:, 2] + gradient_y[:, 1])), torch.stack((gradient_x[:, 2] + gradient_y[:, 1], 2.0 * gradient_y[:, 2]))), dim=1)

stress_tensor_00 = self.lambda_m * (eps[0, 0] + eps[1, 1]) + 2.0 * self.mu_m * eps[0, 0]
stress_tensor_off_diag = 2.0 * self.mu_m * eps[0, 1]
stress_tensor_11 = self.lambda_m * (eps[0, 0] + eps[1, 1]) + 2.0 * self.mu_m * eps[1, 1]

off_diag_grad = torch.autograd.grad(stress_tensor_off_diag.sum(), input_s, create_graph=True)[0]
div_stress = torch.zeros(2, input_s.size(0), dtype=torch.float32, device=input_s.device)

div_stress[0, :] = torch.autograd.grad(stress_tensor_00.sum(), input_s, create_graph=True)[0][:, 1] + off_diag_grad[:, 2]
div_stress[1, :] = off_diag_grad[:, 1] + torch.autograd.grad(stress_tensor_11.sum(), input_s, create_graph=True)[0][:, 2]

residual_solid = self.rho_solid * torch.stack((dt2_x, dt2_y), dim=0) - div_stress
loss_solid = torch.mean(abs(residual_solid) ** 2)
\end{lstlisting}
\section[Minimum source width for Nyquist sampling]{Estimation of Minimum \texorpdfstring{$\sigma$}{sigma} to Comply with Nyquist--Shannon Sampling Theorem}
\label{Nyquist_approximation}
To ensure compliance with the Nyquist criterion, the spatial sampling interval, $\Delta x$, must meet the requirement:
\begin{equation}
\Delta x \leq \frac{\lambda_{min}}{2}.
\end{equation}
The relationship between $\Delta x$ and the total number of collocation points per dimension, $N_x = \sqrt[3]{N}$ (where $N$ is the total number of collocation points), is given by:
\begin{equation}
N_x = \frac{L}{\Delta x},
\end{equation}
with $L$ representing the domain's length.
The minimum resolvable wavelength, $\lambda_{min}$, is inversely related to the highest spatial frequency, $f_{max}$, determined by the Gaussian source's standard deviation, $\sigma$. Therefore, $\lambda_{min}$ can be expressed as:
\begin{equation}
\lambda_{min} = \frac{v_{max}}{f_{max}},
\end{equation}
where $v_{max}$ is the maximum wave speed. The formula for calculating $f_{max}$ is:
\begin{equation}
f_{max} = \frac{1}{\sigma \sqrt{2\pi}},
\end{equation}
based on an approximation from \cite{Fourier_approximation}. Given that $N = 100000$, $L=2$, and $v_{max} = 0.89$ (representing the p-wave velocity, $V_p$), we can rearrange the equation to estimate $\sigma$ as follows:
\begin{equation}
\sigma \geq \frac{2 \Delta x}{v_{max} \sqrt{2\pi}}.
\end{equation}
This calculation enables estimating the smallest source size that adheres to the Nyquist-Shannon theorem.
Although direct calculation suggests a threshold close to $\sigma = 0.045$, considering the approximations involved, a conservative approach adjusts this threshold to $\sigma = 0.06$.
\chapter{Neural Network Architecture Search}
\label{Appendix:Architecture}
\section{Modified Radial Bessel PINNs}
\label{APPENDIX:MRB}
The function $g(x,y)$ utilised in Equation (\ref{eq:bessel}) to generate a modulated radial-Gaussian envelope is defined as follows:
\begin{equation}
g(x,y) = \exp\left(-\sqrt{\alpha_{g}^2 + \varepsilon} \cdot (k \cdot \sqrt{\sqrt{\gamma^2 + \varepsilon} \cdot x^2 + \sqrt{\kappa^2 + \varepsilon} \cdot y^2 + \varepsilon} - \omega \cdot t)^2\right),
\end{equation}
where $\alpha_g$, $\kappa$, $\gamma$, and $\omega$ are trainable parameters.
The approximation of the Bessel function of the first kind is given by:
\begin{equation}
\begin{aligned}
&\tilde{j0}(x) =  \frac{1}{\sqrt[4]{1 + (\lambda_0^4 \cdot x^2)} \cdot (1 + q \cdot x^2)} \cdot 
 \Bigg[ \bigg(p_0 + p_1 \cdot x^2 + p_2 \cdot \sqrt{1 + (\lambda_{0}^{4} \cdot x^2)}\bigg) \\ &\cdot \cos(x) 
 + \bigg(p_{\tilde{0}} + p_{\tilde{1}} \cdot x^2\bigg) \cdot \sqrt{1 + \left(\lambda_0^4 \cdot x^2\right)} 
 + p_{\tilde{2}} \cdot x \cdot \sin(x) \Bigg].
\end{aligned}
\end{equation}
The constant parameters are defined as $\lambda_0 = 0.0815$, $p_1 = -0.2622$, $q = 0.5701$, $p_{\tilde{1}} = -0.3945$, $p0 = 0.5948$, $p_2 = -0.0493$, $p_{\tilde{0}} = 0.4544$, and $p_{\tilde{2}} = 0.0327$.

\section{Far Field PINNs}
Equations (\ref{eq:FF1}) - (\ref{eq:FFfin}) detail the computations for a single FarField neuron. Our approach closely follows \cite{quantitative_seismology}, albeit with adaptations from a three-dimensional to a two-dimensional case and focusing solely on the far field, omitting intermediate and near field equations for simplicity. Let
\begin{equation}\label{eq:FF1}
\hat{\phi}_x = -\hat{r}_y, \quad \hat{\phi}_y = \hat{r}_x.
\end{equation}
where $\hat{r}_x$ and $\hat{r}_y$ denote the unit vectors originating at the seismic source location.
\begin{equation}
\hat{r}_x = \frac{x - s_x}{r}, \quad \hat{r}_y = \frac{y - s_y}{r}.
\end{equation}
The derivative of the moment tensors $M$ is analytically defined as:
\begin{equation}
\dot{M}(t) = \frac{M_0}{T^2} \left(t - \frac{3T}{2}\right) \exp\left(-\frac{\left(t - \frac{3T}{2}\right)^2}{T^2}\right),
\end{equation}
which essentially represents a modified derivative of the exponential function. The characteristic time $T$ and the seismic moment $M_0$ are trainable parameters. The far-field radiation patterns for P and S waves are described by:
\begin{equation}
\begin{aligned}
A^{FP}_x &= \sin(\theta_1) \cos(\theta_3) \hat{r}_x, \\ 
A^{FS}_x &= -\cos(\theta_2) \sin(\theta_4) \hat{\phi}_x, \\
A^{FP}_y &= \sin(\theta_1) \cos(\theta_3) \hat{r}_y, \\
A^{FS}_y &= -\cos(\theta_2) \sin(\theta_4) \hat{\phi}_y,
\end{aligned}
\end{equation}
where [$\theta_1$, $\theta_2$, $\theta_3$, $\theta_4$] are trainable parameters. The x and y components of the displacement field are calculated as follows:
\begin{equation}
\begin{aligned}
u^{F}_x = \left(\frac{1}{4\pi v_{p}^{3}}\right) A^{FP}_x \left(\frac{1}{r}\right) \dot{M}([t + 1] + \delta - \frac{r}{v_p}) \\+ \left(\frac{1}{4\pi v_{s}^{3}}\right) A^{FS}_x \left(\frac{1}{r}\right) \dot{M}([t + 1] + \delta - \frac{r}{v_s}), \\
u^{F}_y = \left(\frac{1}{4\pi v_{p}^{3}}\right) A^{FP}_y \left(\frac{1}{r}\right) \dot{M}([t + 1] + \delta - \frac{r}{v_p}) \\+ \left(\frac{1}{4\pi v_{s}^{3}}\right) A^{FS}_y \left(\frac{1}{r}\right) \dot{M}([t + 1] + \delta - \frac{r}{v_s}).
\end{aligned}
\end{equation}
The delay $\delta$ is an additional trainable parameter. The speeds of P and S waves, $v_p$ and $v_s$, are calculated as described in Equation (\ref{eq:lame})
Lastly, the outputs are normalised, i.e.,
\begin{equation} \label{eq:FFfin}
u^{F}_x = \frac{u^{F}_x}{\lvert u^{F}x \rvert\_{max}}, \ u^{F}_y = \frac{u^{F}_y} {\lvert u^{F}y \rvert_{\max}}.
\end{equation}

\section{Model Hyperparameters}
Tables (\ref{tab:hyperparameters_constant}) -(\ref{tab:hyperparameters_layered}) outline each model configuration utilised to generate the results presented in Table (\ref{tab:results}).

\begin{table}[ht]
\centering
\begin{adjustbox}{width=\textwidth,center}
\begin{tabular}{lcccccccc}
\toprule
 & $N_{Neurons}$ & $N_{Layers}$ & $N^E_{Neurons}$ & $N^E_{Layers}$ & $N^D_{Neurons}$ & $N^D_{Layers}$ & $N^C_{Neurons}$ & $t_1$ \\
\midrule
PINN-tanh & 128 & 5 & - & - & - & - & - & 0.1\\
\textit{sT}-PINN & 128 & 5 & - & - & - & - & 32 & 0.1\\
\textit{pT}-PINN & 128 & 5 & - & - & - & - & 64 & 0.2\\
\textit{wT}-PINN & 128 & 5 & - & - & - & - & 128 & 0.1\\
\textit{pED}-PINN & - & - & 32 & 4 & 8 & 1 & 128 & 0.07\\
\textit{wED}-PINN & - & - & 32 & 4 & 64 & 3 & 128 & 0.07\\
\textit{pE}-PINN & - & - & 64 & 5 & - & - & 128 & 0.07\\
\textit{wE}-PINN & - & - & 64 & 5 & - & - & 128 & 0.1\\
\textit{pA}-PINN & 128 & 5 & - & - & - & - & 32 & 0.07\\
\textit{wA}-PINN & 128 & 5 & - & - & - & - & 128 & 0.2\\
\bottomrule
\end{tabular}
\end{adjustbox}
\caption{Hyperparameters for Constant Lam{\'e} parameters}
\label{tab:hyperparameters_constant}
\end{table}

\begin{table}[ht]
\centering
\begin{adjustbox}{width=\textwidth,center}
\begin{tabular}{lcccccccc}
\toprule
 & $N_{Neurons}$ & $N_{Layers}$ & $N^E_{Neurons}$ & $N^E_{Layers}$ & $N^D_{Neurons}$ & $N^D_{Layers}$ & $N^C_{Neurons}$ & $t_1$ \\
\midrule
PINN-tanh & 128 & 6 & - & - & - & - & - & 0.2\\
\textit{sT}-PINN & 128 & 5 & - & - & - & - & 16 & 0.2\\
\textit{pT}-PINN & 128 & 5 & - & - & - & - & 128 & 0.2\\
\textit{wT}-PINN & 128 & 5 & - & - & - & - & 128 & 0.1\\
\textit{pED}-PINN & - & - & 32 & 4 & 8 & 1 & 128 & 0.2\\
\textit{wED}-PINN & - & - & 32 & 4 & 64 & 3 & 128 & 0.1\\
\textit{pE}-PINN & - & - & 64 & 5 & - & - & 128 & 0.2\\
\textit{wE}-PINN & - & - & 32 & 5 & - & - & 128 & 0.07\\
\textit{pA}-PINN & 128 & 5 & - & - & - & - & 32 & 0.1\\
\textit{wA}-PINN & 16 & 5 & - & - & - & - & 32 & 0.2\\
\bottomrule
\end{tabular}
\end{adjustbox}
\caption{Hyperparameters for Mixture model Lam{\'e} parameters}
\label{tab:hyperparameters_mixture}
\end{table}

\begin{table}[ht]
\centering
\begin{adjustbox}{width=\textwidth,center}
\begin{tabular}{lcccccccc}
\toprule
 & $N_{Neurons}$ & $N_{Layers}$ & $N^E_{Neurons}$ & $N^E_{Layers}$ & $N^D_{Neurons}$ & $N^D_{Layers}$ & $N^C_{Neurons}$ & $t_1$ \\
\midrule
PINN-tanh & 128 & 5 & - & - & - & - & - & 0.2\\
\textit{sT}-PINN & 128 & 5 & - & - & - & - & 16 & 0.2\\
\textit{pT}-PINN & 128 & 5 & - & - & - & - & 128 & 0.1\\
\textit{wT}-PINN & 128 & 5 & - & - & - & - & 128 & 0.1\\
\textit{pED}-PINN & - & - & 32 & 4 & 8 & 2 & 128 & 0.07\\
\textit{wED}-PINN & - & - & 32 & 4 & 64 & 3 & 128 & 0.1\\
\textit{pE}-PINN & - & - & 64 & 5 & - & - & 128 & 0.2\\
\textit{wE}-PINN & - & - & 32 & 5 & - & - & 128 & 0.1\\
\textit{pA}-PINN & 128 & 5 & - & - & - & - & 64 & 0.07\\
\textit{wA}-PINN & 16 & 5 & - & - & - & - & 128 & 0.1\\
\bottomrule
\end{tabular}
\end{adjustbox}
\caption{Hyperparameters for Layered model Lam{\'e} parameters}
\label{tab:hyperparameters_layered}
\end{table}
\end{document}

%% file: eth-template/extrapackages.tex
\usepackage[capitalise]{cleveref}

\usepackage[german=swiss]{csquotes}

\usepackage{datetime}

\usepackage{mathtools}

\usepackage[h]{esvect}

\usepackage{array}

\usepackage{listings}
\usepackage{booktabs}

%% file: eth-template/layoutsetup.tex
\usepackage{eth-template/ETHlogo}

\nonzeroparskip
\defaultlists

\makeatletter

\if@twoside
  \copypagestyle{chapter}{Ruled}
\else
  \copypagestyle{chapter}{ruled}
\fi
\makeoddhead{chapter}{}{}{}
\makeevenhead{chapter}{}{}{}
\makeheadrule{chapter}{\textwidth}{0pt}
\copypagestyle{abstract}{empty}

\setsecheadstyle{\Large\bfseries\sffamily}
\setsubsecheadstyle{\large\bfseries\sffamily}
\setsubsubsecheadstyle{\bfseries\sffamily}
\setparaheadstyle{\normalsize\bfseries\sffamily}
\setsubparaheadstyle{\normalsize\itshape\sffamily}
\setsubparaindent{0pt}

\captionnamefont{\sffamily\bfseries\footnotesize}
\captiontitlefont{\sffamily\footnotesize}
\setsecnumdepth{subsection}
\settocdepth{subsection}

\pretitle{\vspace{0pt plus 0.7fill}\begin{center}\HUGE\sffamily\bfseries}
\posttitle{\end{center}\par}
\preauthor{\par\begin{center}\let\and\\\Large\sffamily}
\postauthor{\end{center}}
\predate{\par\begin{center}\Large\sffamily}
\postdate{\end{center}}

\def\@advisors{}
\newcommand{\advisors}[1]{\def\@advisors{#1}}
\def\@department{}
\newcommand{\department}[1]{\def\@department{#1}}
\def\@thesistype{}
\newcommand{\thesistype}[1]{\def\@thesistype{#1}}

\renewcommand{\maketitlehookb}{\vspace{1in}%
  \par\begin{center}\Large\sffamily\@thesistype\end{center}}

\renewcommand{\maketitlehookd}{%
  \vfill\par
  \begin{flushright}
    \sffamily
    \@advisors\par
    \@department, ETH Z\"urich
  \end{flushright}
}

\checkandfixthelayout

%% file: eth-template/theoremsetup.tex
\numberwithin{equation}{chapter}

%% file: eth-template/macrosetup.tex
\renewcommand{\epsilon}{\ensuremath\varepsilon}

\renewcommand{\phi}{\ensuremath{\varphi}}